\pdfoutput=1
\documentclass{arxivtmpl}

\def\1{\bm{1}}

\DeclareMathAlphabet{\mathsfit}{\encodingdefault}{\sfdefault}{m}{sl}
\SetMathAlphabet{\mathsfit}{bold}{\encodingdefault}{\sfdefault}{bx}{n}

\tcbuselibrary{skins,breakable}
\newtcolorbox{findingbox}{enhanced,breakable,frame hidden,colback=gred!6,borderline west={3pt}{0pt}{gred},sharp corners,boxsep=0pt,left=12pt,right=10pt,top=8pt,bottom=8pt,before skip=12pt,after skip=12pt}
\newtcolorbox{notebox}{enhanced,breakable,frame hidden,colback=gblue!6,borderline west={3pt}{0pt}{gblue},sharp corners,boxsep=0pt,left=12pt,right=10pt,top=8pt,bottom=8pt,before skip=12pt,after skip=12pt}
\newtcolorbox{tipbox}{enhanced,breakable,frame hidden,colback=ggreen!6,borderline west={3pt}{0pt}{ggreen},sharp corners,boxsep=0pt,left=12pt,right=10pt,top=8pt,bottom=8pt,before skip=12pt,after skip=12pt}
\newtcolorbox{warnbox}{enhanced,breakable,frame hidden,colback=gyellow!6,borderline west={3pt}{0pt}{gyellow},sharp corners,boxsep=0pt,left=12pt,right=10pt,top=8pt,bottom=8pt,before skip=12pt,after skip=12pt}

\tcbuselibrary{listings}
\newcommand{\boxfont}{\fontsize{9}{11}\selectfont}
\newtcblisting{promptbox}[1][]{%
  enhanced, listing only, breakable, frame hidden, width=\linewidth,
  colback=gblue!6, borderline west={3pt}{0pt}{gblue},
  sharp corners, boxsep=0pt,
  left=12pt, right=10pt, top=8pt, bottom=8pt, before skip=12pt, after skip=12pt,
  before upper={\def\promptlabel{#1}\ifx\promptlabel\empty\else{\sffamily\bfseries\color{gblue}#1}\par\smallskip\fi}, %
  listing options={basicstyle=\ttfamily\boxfont, escapeinside={(*}{*)}},
}

\newtcolorbox{transcriptbox}{%
  enhanced, breakable, frame hidden,
  colback=ggreen!6, borderline west={3pt}{0pt}{ggreen},
  sharp corners, boxsep=0pt, fontupper=\ttfamily\boxfont, halign upper=flush left, %
  left=12pt, right=10pt, top=6pt, bottom=8pt, before skip=12pt, after skip=12pt,
}

\newcommand{\teaserfigure}[2]{%
  \begingroup\centering\captionsetup{type=figure}
  #1\par\vspace{0.4em}\caption{#2}%
  \par\endgroup\vspace{1em}
}

\colorlet{codebg}{gyellow!6}

\hypersetup{
    colorlinks=true,
    linkcolor=linkcol,
    citecolor=linkcol,
    urlcolor=linkcol,
    pdfborderstyle={},
    pdfborder={0 0 0}
}
\renewcommand\Affilfont{\sffamily\mdseries}

\colorlet{perfhue}{gblue}
\colorlet{perf0}{white}
\colorlet{perf10}{perfhue!12}
\colorlet{perf20}{perfhue!22}
\colorlet{perf30}{perfhue!33}
\colorlet{perf40}{perfhue!44}
\colorlet{perf50}{perfhue!55}
\colorlet{perf60}{perfhue!66}
\colorlet{perf70}{perfhue!77}
\colorlet{perf80}{perfhue!88}

\newcommand{\RN}[1]{%
	\textup{\lowercase\expandafter{\it \romannumeral#1}}%
}

\SetKwInput{KwInput}{Input}
\SetKwInput{KwOutput}{Output}
\DontPrintSemicolon

\SetKwComment{Comment}{\color{ggreen}\# }{}

\SetKwProg{Function}{def}{:}{}

\SetKwProg{For}{for}{:}{}
\SetKwProg{If}{if}{:}{}

\usepackage{url}
\usepackage{longtable}
\usepackage{array}
\usepackage{xltabular}
\usepackage{pdflscape}
\usetikzlibrary{shapes.geometric,arrows.meta,calc}
\usepackage{pgfplots}
\pgfplotsset{compat=1.18}
\definecolor{SkillBlue}{HTML}{011F5B}
\definecolor{SkillPaleBlue}{HTML}{E8ECF3}
\definecolor{SkillInk}{HTML}{1A1A1A}
\definecolor{navyblue}{HTML}{0071BC}
\makeatletter
\newcommand{\panellabel}[2]{\begingroup\protected@edef\@currentlabel{\thefigure#1}\label{#2}\endgroup}
\makeatother
\iftrue
\newcommand{\displaytodo}[1]{#1}
\else
\newcommand{\displaytodo}[1]{}
\fi

\title{Video-Index: A Curated Meta-Benchmark for Video Understanding}

\author[1]{Enxin Song}
\author[1]{Yinuo Xu}
\author[2]{Shusheng Yang}
\author[3]{Wenhao Chai}
\author[1]{Jiatao Gu}
\affil[1]{University of Pennsylvania}
\affil[2]{New York University}
\affil[3]{Princeton University}

\begin{document}

\maketitle

\thispagestyle{firstpagestyle}

\begin{abstract}
{\sffamily\fontseries{eb}\normalsize\selectfont Abstract}\par\vspace{0.15\baselineskip}
\small
A video benchmark should reward the capability it claims to measure, yet models can exploit answer options, question text, or partial visual evidence. We introduce the attack pyramid, five levels of shortcut attacks with increasing access to each item, and audit 115 video benchmarks with it. On 35 benchmarks, attackers that never see a frame approach full-video accuracy. On 51 benchmarks with temporal probes, shuffled frames keep a median 96\% of full-video accuracy. Near-duplicate questions make up at least half the items in 63 benchmarks. We screen 505,518 question--answer pairs from 112 of them into an audited pool. Agents turn evaluation requests into specifications, and a deterministic selector with a red-team gate composes reproducible benchmarks. We release Video-Index, the 210 hardest verified items under these attacks in each of four capability groups, 840 items from 76 sources. With the same fixed input, Claude Opus 5 outscores every open-source model by over 37 percentage points, and agent tools add about 20 more, yet all systems leave room to improve efficiency and accuracy. \href{https://www.enxinsong.com/blog/video-index/}{Blog}~$\cdot$ \href{https://github.com/Espere-1119-Song/Video-Index}{GitHub}~$\cdot$ \href{https://huggingface.co/datasets/Video-Index/Video-Index}{Hugging~Face}

\end{abstract}
\vspace{1.5em}

\teaserfigure{\includegraphics[width=\linewidth]{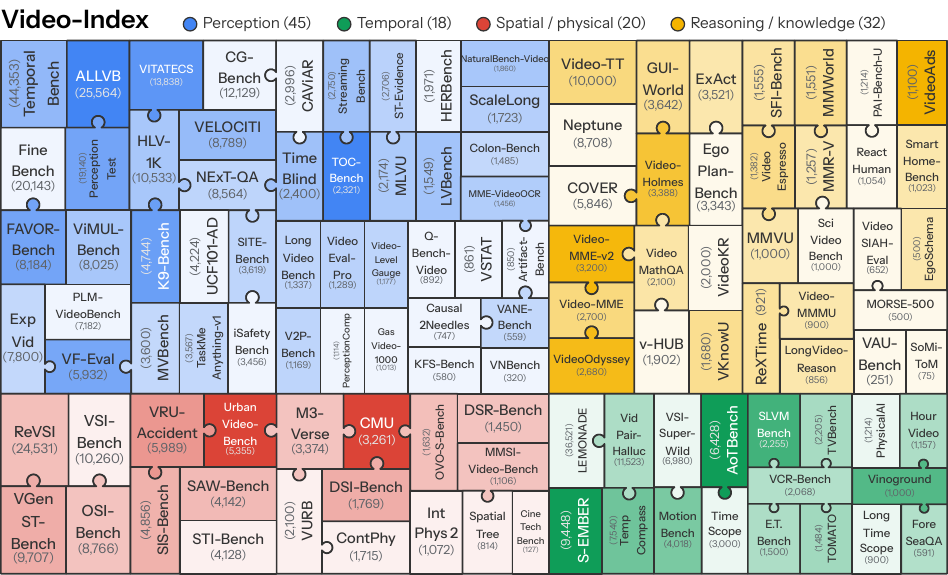}}{\textbf{Video-Index.} We sample 840 problems from the 505,518 in the 115 video benchmarks.\label{fig:poolsources}}

\clearpage
\section{Introduction}
\label{sec:intro}

A benchmark score is a certificate of a capability. A video model is trusted to follow a soccer match, an egocentric kitchen, or an hour-long lecture once it scores well on the matching benchmarks. These test temporal order \citep{cores2024tvbench, liu2024tempcompass}, long video \citep{chandrasegaran2024hourvideo, wang2024lvbench}, spatial reasoning \citep{yang2025vsibench}, egocentric memory \citep{mangalam2023egoschema}, knowledge \citep{hu2026video}, hallucination \citep{wang2024videohallucer}, streaming \citep{lin2024streamingbench}, and general perception \citep{li2024mvbench, fu2025video}. In the past five years, 605 video benchmarks have appeared, 342 of them with public data. The 115 multiple-choice benchmarks we evaluate hold 528,172 items over 123,281 videos, as many as the instruction-tuning sets behind the first video language models \citep{maaz2024video, zhang2024video}. Yet each benchmark was built against the models of its year. Stronger language priors, richer single frames, and retrieval over items already asked now let a model earn the certificate without watching the video.

We argue that a certificate is worth the difficulty of obtaining it without the capability, measured for every benchmark under one protocol. An adversary takes every item it can solve without the capability, and the benchmark keeps the rest. The attack pyramid organizes five attacker sets of growing access. The option level sees the option layout, the text level adds the question, and the pool level adds the other evaluated items. The frame level adds one frame or a caption, and the order level adds a shuffled or truncated video. The exploitability of a benchmark at a level is the largest margin over chance that its attackers reach. The benchmark breaks at the lowest level whose attackers approach the full protocol. Prior audits probe one benchmark with one shortcut \citep{buch2022revisiting, cores2024tvbench, shangguan2024tomato, yang2026cambrian, brown2025benchmark} and discard it when the shortcut works. Generators write new items with agents \citep{zhang2024taskmeanything, butt2024benchagents, li2025autobencher, shi2026harbor} and leave existing benchmarks unused. The pyramid instead tests every benchmark at every level and keeps what no tested attack solves.

\begin{figure}[h]
\centering
\captionsetup{skip=5pt}
\includegraphics[width=\linewidth]{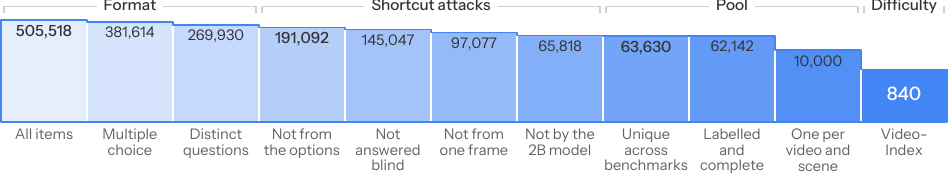}
\caption{\textbf{From pool to Video-Index.} We filter the 505,518 items in four stages and keep the 840 hardest.}
\label{fig:poolfunnel}
\vspace{-8pt}
\end{figure}

We run the pyramid over 115 video benchmarks. Newer benchmarks break earlier, each claim breaks at its own level, and the errors of most benchmarks concentrate in one or two capabilities. We keep the items that no attacker solves and let agents compose benchmarks from them behind a red-team gate. As shown in Figure~\ref{fig:poolfunnel}, screening keeps 840 of the 505,518 items in the pool for Video-Index, a meta-benchmark of the hardest items whose answers a judge verified against the frames. We evaluate open-source video models and proprietary agents on Video-Index.

\begin{itemize}
\item We introduce the attack pyramid, run it on 115 video benchmarks under one protocol, and find that newer benchmarks break earlier and that each claim breaks at its own level.
\item Our agent-assisted pipeline combines screening, specification generation, deterministic selection, and red-team validation for reproducible, on-demand benchmarks with lockfiles.
\item We release Video-Index, a meta-benchmark of the 840 hardest verified items of the pool, and evaluate open-source video models and proprietary systems on it.
\end{itemize}

\clearpage
\setcounter{tocdepth}{2}
\begingroup
\hypersetup{linkcolor=black}
\tableofcontents
\endgroup
\clearpage

\section{What scores certify}
\label{sec:threat}

We treat a score as evidence for the capability its benchmark claims to test. An attacker raises the score without that capability. An options-only model exploits answer wording, and a single-frame reader skips temporal order. Such shortcuts \citep{khandelwal2019generalization, buch2022revisiting, zheng2024large, balepur2024artifacts} motivate an adversarial test of what the score certifies.

Let $B$ be a benchmark, $c$ its chance accuracy, and $\mathcal{A}$ a declared set of attackers. Write $\mathrm{acc}_B(a)$ for the accuracy of attacker $a$ on $B$. The exploitability of $B$ is
\begin{equation}
\varepsilon_{\mathcal{A}}(B) = \max_{a \in \mathcal{A}} \mathrm{acc}_B(a) - c,
\end{equation}
the largest margin over chance that an attacker in the set achieves. The maximizing attacker can differ across benchmarks. For a model with accuracy $s$, the residual $s-c-\varepsilon_{\mathcal{A}}(B)$ measures the score above the strongest tested shortcut, and a positive residual rules out only the shortcuts tested.

We nest the attacker sets as $\mathcal{A}_0 \subseteq \mathcal{A}_1 \subseteq \cdots \subseteq \mathcal{A}_4$. The option level grants options alone, the text level adds the question, and the pool level adds other evaluation items. The frame level admits a degraded visual view, and the order level admits temporally perturbed video. No attacker sees the ground truth of the item under evaluation. We write $\varepsilon_\ell$ for the exploitability at level $\ell$, and the cumulative profile $\varepsilon_{\mathrm{opt}},\varepsilon_{\mathrm{text}},\varepsilon_{\mathrm{pool}},\varepsilon_{\mathrm{frame}},\varepsilon_{\mathrm{order}}$ is nondecreasing. Following \citet{finzi2026entropy}, each level represents an observer with bounded access and compute, as developed in the appendix.

A benchmark's threat model specifies the levels its capability claim must resist. General video understanding \citep{li2024mvbench, liu2024tempcompass, cai2024temporalbench, fu2025video} requires every level, and a knowledge claim \citep{he2024mmworld, hu2026video} could exempt the text level, yet we apply every measured level to every benchmark. Let the reference accuracy $s^{*}$ be the full-protocol accuracy of a reference model outside the attacker sets. A benchmark breaks when $\varepsilon_\ell \ge s^{*}-c-\delta$, with tolerance $\delta=5\%$, about the 95\% sampling margin of a 300-item accuracy, and its breaking level is the first such level. An unbroken benchmark resists every tested level within this tolerance.

\section{Auditing with shortcut attacks}
\label{sec:pyramid}
\label{sec:build}

\subsection{Benchmarks and protocol}
\label{sec:setup}\label{sec:screen}

We census 605 video benchmarks released in the past five years, 342 of them with public data, and collect 721,324 question--answer pairs across 161,580 videos. We retain 115 English multiple-choice benchmarks that do not require audio. We sample $n=300$ questions per benchmark, preserving subtask proportions and keeping every question when fewer are available. Videos are stored at a short side of 720 pixels, reduced to 480 for clips longer than 30 minutes without on-screen text. Frames reach the models with a long side of at most 768 pixels. Gemini 3.5 Flash \citep{google2026gemini35flash} supplies the reference accuracy $s^{*}$ at 32 frames. On long-video benchmarks where it stays within the tolerance $\delta$ of chance, its 128-frame probe supplies the reference instead. \textbf{Models in this audit answer once per item, without tools, retries, or an agent scaffold.} We label items with 18 fine capability categories in four groups, namely perception, temporal, spatial and physical, and reasoning and knowledge. Each benchmark takes the group that most of its items require. Before attacks, we score each answer by its option letter, then by an answer-line regular expression, then by its option text. Within each benchmark, we keep one item per near-duplicate group, which removes 1.1\% of sampled items, refill to $n=300$ where possible, and screen the replacements.

\subsection{Options without questions}
\label{sec:t0}

\begin{figure}[t]
\centering
\includegraphics[width=\linewidth]{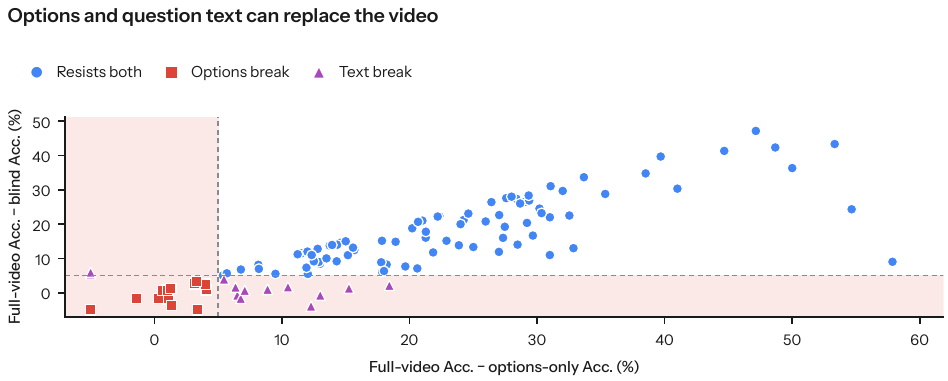} %
\caption{\textbf{Options and question text replace the video on 30 benchmarks.} Each benchmark is placed by the full video's gain over options alone and over the question text, and shaded bands mark breaks.}
\label{fig:t01}
\end{figure}
We test whether the answer options alone reveal the answer. Chance accuracy is the reciprocal of the option count of each item. A fixed-position attacker picks the most frequent answer position. An options-only attacker, Claude Opus 5 \citep{anthropic2026opus5}, sees only the choices in a call that shares no context with its other roles. The larger margin over chance defines $\varepsilon_{\mathrm{opt}}$. As shown in Figure~\ref{fig:t01}, options raise accuracy by up to 63\% over chance, lift it significantly on 78 benchmarks, and break 17. Across 70 complete runs, eight permutations of the option order give Qwen3-VL-8B \citep{bai2025qwen3} a median accuracy range of 5\%, significant on 15 benchmarks. The selected answer changes on a median 37\% of items and on 82\% of DSR-Bench \citep{zhou2025dsrbench}.

\subsection{Questions without video}
\label{sec:t1}

We test whether the question reveals the answer without the video. Gemini 3.5 Flash \citep{google2026gemini35flash} and Claude Opus 5 \citep{anthropic2026opus5} read the question and options independently, and their larger margin over chance defines $\varepsilon_{\mathrm{text}}$. Video dependence $m$ is reference accuracy minus the accuracy of the stronger blind reader, and a larger $m$ marks stronger resistance to text shortcuts. As shown in Figure~\ref{fig:t01}, question text raises accuracy by up to 71\% over chance, against 63\% for options alone. Among the 98 benchmarks that pass the option level, video dependence spans 51\%. The blind reader exceeds reference accuracy on 4 of them and breaks 13.

\subsection{Learning from evaluation items}
\label{sec:t2}

\begin{figure}[t]
\centering
\includegraphics[width=\linewidth]{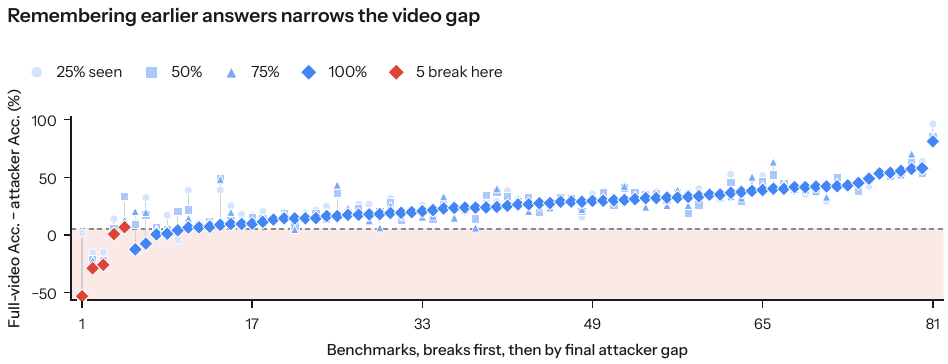} %
\caption{\textbf{Remembering earlier answers narrows the video gap.} Each column is a benchmark, with its gap to the strongest pool attacker after 25 to 100\% of the items.}
\label{fig:t2}
\end{figure}
We test whether the answers to other items help solve a question. A copy attacker copies the answer of the nearest earlier question when their BGE-large \citep{xiao2024c} cosine similarity exceeds 0.9. A retrieval attacker gives Qwen3-8B \citep{yang2025qwen3} the eight nearest question--answer pairs. Following the predictive auditing of \citet{brown2025benchmark}, an online logistic regression on BGE-large embeddings predicts answer letters out of sample over five rotating splits. As shown in Figure~\ref{fig:t2}, the gap to the strongest attacker narrows as more items enter the sequence, and 5 benchmarks break.

\subsection{Single frames and captions}
\label{sec:t3}

\begin{figure}[t]
\centering
\includegraphics[width=\linewidth]{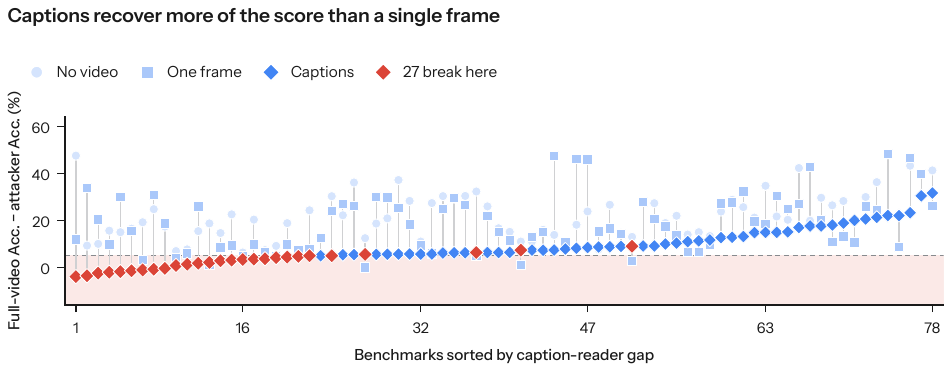} %
\caption{\textbf{Captions recover more of the video score than a single frame.} Each column is a benchmark, sorted by its caption gap, with gaps under no video, one frame, and captions.}
\label{fig:t3strips}
\end{figure}
We replace the video with a single frame or a text description. Gemini 3.5 Flash \citep{google2026gemini35flash} reads one frame, and Claude Opus 5 \citep{anthropic2026opus5} reads captions of the 32 sampled frames, each written independently by AuroraCap-7B-IMG \citep{chai2025auroracap} without the question or options. The larger margin over chance defines $\varepsilon_{\mathrm{frame}}$. As shown in Figure~\ref{fig:t3strips}, captions recover more reference accuracy than a single frame. Among the 78 benchmarks with all three probes at this level, one frame comes within the tolerance $\delta$ of the reference on 8. It beats the blind reader by a median 3.3\%. Captions come within $\delta$ on 23 and exceed the reference on 9, and 27 benchmarks break at this level.

\subsection{Order and temporal coverage}
\label{sec:t4}
\begin{figure}[t]
\centering
\includegraphics[width=\linewidth]{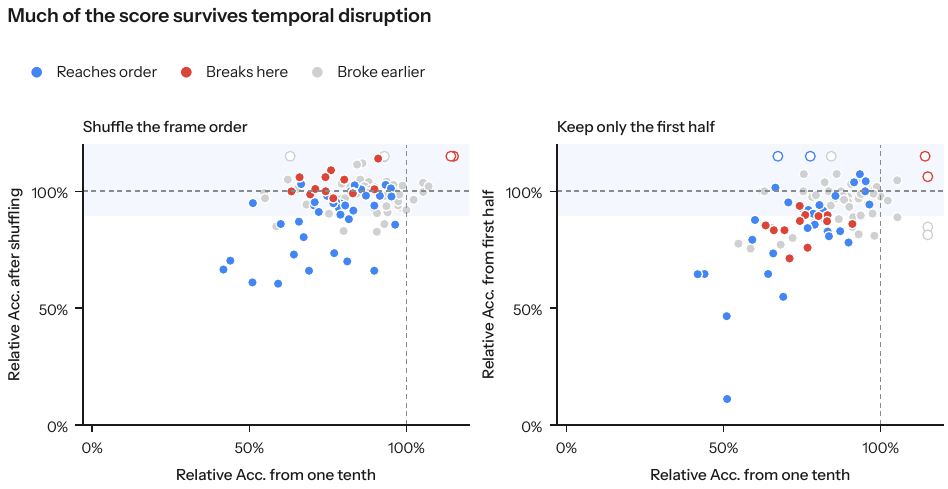} %

\caption{\textbf{Much of the score survives temporal perturbation.} Each dot is a benchmark, placed by the share of full-video accuracy that each perturbation retains.}
\label{fig:t4scatter}\panellabel{a}{fig:t4a}\panellabel{b}{fig:t4b}
\end{figure}
We test whether accuracy depends on frame order and on coverage of the full clip. Gemini 3.5 Flash \citep{google2026gemini35flash} and Qwen3-VL-8B \citep{bai2025qwen3} read 32 shuffled frames and, separately, a contiguous tenth of the video, and their largest margin over chance defines $\varepsilon_{\mathrm{order}}$. As shown in Figure~\ref{fig:t4scatter}, much of the score survives both perturbations. Among the 51 benchmarks with both probes at this level, shuffling retains a median 96\% of the same model's full-video accuracy and one tenth retains 78\%. Across all 96 profiled benchmarks, either half retains a median 92\%, and shifting the sampling grid leaves the median unchanged. Of 15 benchmarks whose papers claim temporal reasoning, 6 retain over 90\% under shuffling, yet 6 of the 7 reaching this level lose at least a quarter. In total, 38 benchmarks remain unbroken.

\section{What the audit finds}
\label{sec:findings}

\begin{figure}[t]
\centering
\includegraphics[width=\linewidth]{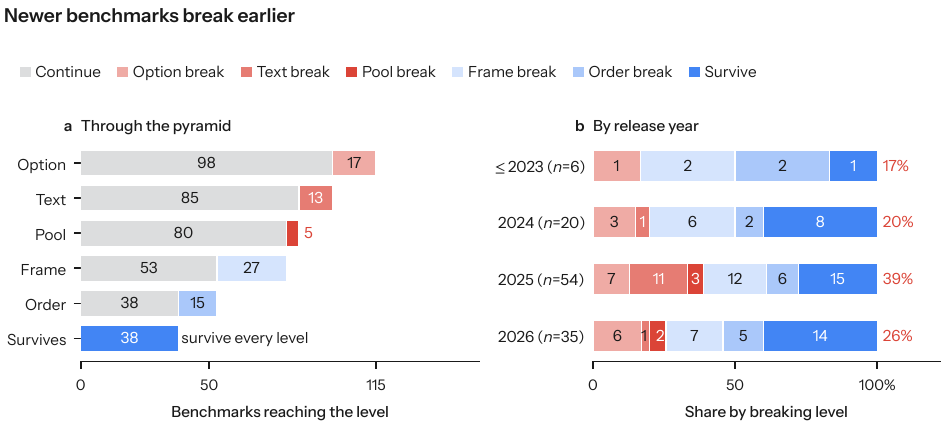} %

\caption{\textbf{Newer benchmarks break earlier.} \textbf{a}, Benchmarks that reach and break at each level. \textbf{b}, Breaking levels by release year, with the share breaking before visual input in red.}
\label{fig:funnel}\panellabel{a}{fig:funnela}\panellabel{b}{fig:funnelb}
\end{figure}
\subsection{Newer benchmarks break earlier}
\label{sec:f0}
We group the breaking levels by release year. As shown in Figure~\ref{fig:funnel}, newer benchmarks are easier to attack without video. Of the 115 audited benchmarks, 38 survive every level and 35 break before any visual input. The share breaking before visual input rises from 19\% among releases through 2024 to 34\% among releases in 2025 and 2026. The rise comes from the text and pool levels, where releases from 2025 and 2026 hold 17 of the 18 breaks. Blind readers break 12 of the 89 newer benchmarks against 1 of the 26 older ones. On most of them the blind reader comes within the tolerance $\delta$ of a reference that scores far above chance. The option level instead breaks a similar share in both periods, 4 older and 13 newer benchmarks, especially where reference accuracy approaches chance. The rise persists under attacks matched to each claim, as the appendix details. Recent benchmarks use LLMs to generate or refine questions and answers \citep{liu2024tempcompass, hu2026video}, which can make the answer recoverable from text alone. This may also leave recurring patterns in answer options, creating shortcuts even when questions resist blind reading.

\subsection{Each claim breaks differently}
\label{sec:f3}
\begin{figure}[t]
\centering
\includegraphics[width=\linewidth]{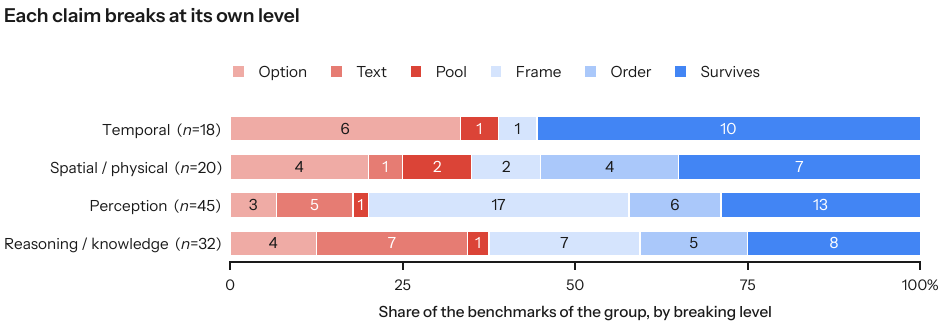} %
\caption{\textbf{Each capability group breaks at its own levels.} Bars split the benchmarks of each group by breaking level, and segment labels give counts.}
\label{fig:f3claims}
\end{figure}
As shown in Figure~\ref{fig:f3claims}, the four capability groups break at different levels. Reasoning and knowledge benchmarks break on text or captions that already carry the facts. Perception benchmarks break most often at the frame level, and spatial and physical ones on options and under temporal perturbation. Temporal benchmarks split sharply, with 6 of 18 breaking on options and 10 surviving every level. Audits should match their checks to the claimed capability and still screen options for temporal claims. The appendix provides the finer breakdown within each broad capability group.

\subsection{Frames or pixels}
\label{sec:f9fp}
\begin{figure}[t]
\centering
\includegraphics[width=\linewidth]{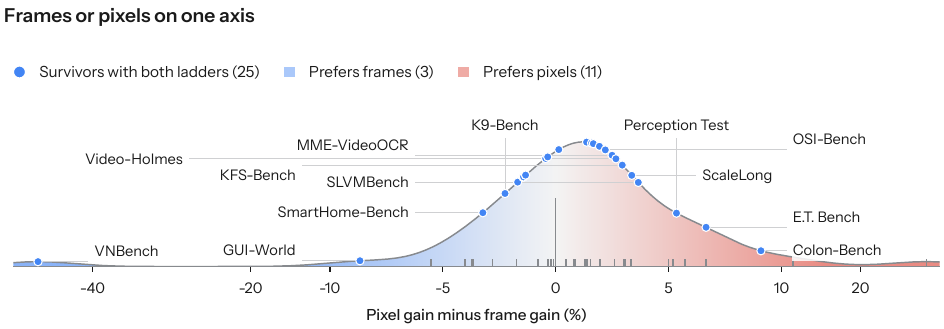} %
\caption{\textbf{Eleven benchmarks gain more from pixels and three from frames.} The curve is the density of pixel gain minus frame gain, and dots mark survivors with both ladders.}
\label{fig:f9fp}
\end{figure}
Additional visual input can supply finer spatial detail or denser temporal coverage, and we ask which helps more. On the benchmarks that pass the pool level, Qwen3-VL-8B \citep{bai2025qwen3} reads 32 frames at short sides of 168, 224, 336, and 448 pixels and at the stored resolution. It then reads 224-pixel frames sampled at 0.25 to 2 frames per second within its context. As shown in Figure~\ref{fig:f9fp}, 11 benchmarks favor pixels, 3 favor frames, and the two gains differ by less than $\delta$ on the rest. Resolution recovers small details in slides, endoscopy, and synthetic artifacts, and denser sampling catches brief events that sparse sampling misses.

\begin{figure}[t]
\centering
\includegraphics[width=\linewidth]{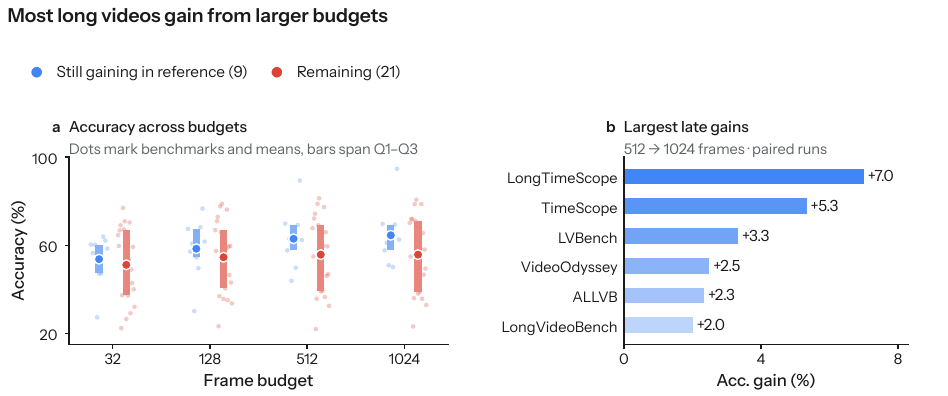} %
\caption{\textbf{Most long videos gain from larger budgets.} \textbf{a}, Scores per budget with means and quartiles, blue if the reference still gains at 600 frames. \textbf{b}, The six largest gains, 512 to 1024.}
\label{fig:f5long}
\end{figure}
We also test whether long-video questions need evidence that sparse sampling misses, with frame budgets from 32 to 1024. As shown in Figure~\ref{fig:f5long}, most long-video benchmarks gain from larger budgets. Among the 30 benchmarks whose videos average at least 300 seconds, 18 gain at least $\delta$ in the reference sweep and 9 still gain at its 600-frame limit. These gains indicate that sparse inputs can omit evidence useful for answering. In a diagnostic sweep at 224 pixels, only LongTimeScope \citep{zohar2025apollo2} and TimeScope \citep{zohar2025timescope} keep improving from 512 to 1024 frames. Since the observed gains also depend on the reference model's long-context ability, other benchmarks may still place high demands on processing events across long time spans.

\subsection{Errors cluster in few capabilities}
\label{sec:f10ability}
\begin{figure}[t]
\centering
\includegraphics[width=\linewidth]{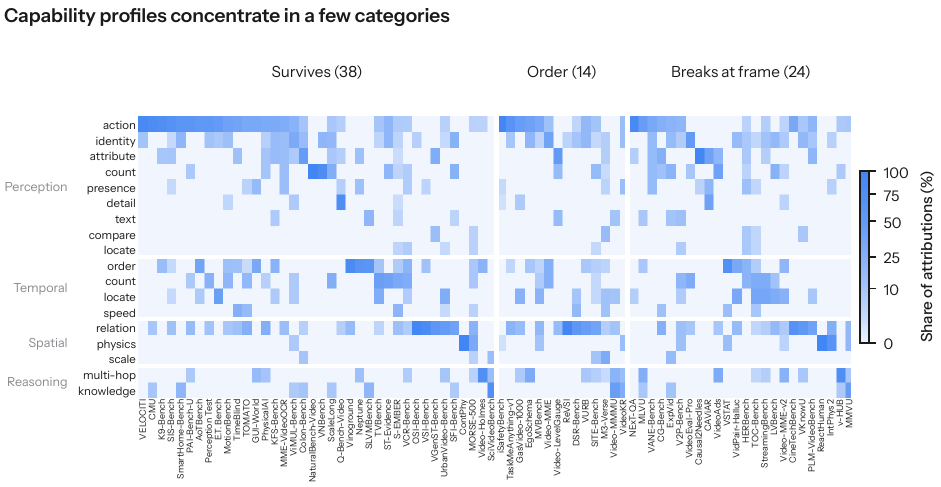} %
\caption{\textbf{Errors concentrate in few capabilities.} Columns are the 76 pool-level survivors with 20 or more agreed attributions, grouped by their first breaking level, each summing to 100\%.}
\label{fig:f10ability}
\end{figure}
We attribute the 32-frame errors of the reference model to the 18 fine categories of Section~\ref{sec:setup}, on the 76 pool-level survivors with at least 20 agreed attributions. Claude Opus 5 \citep{anthropic2026opus5} receives the 32 frames, question, options, gold answer, and incorrect answer with its reasoning, and only attributions confirmed by two independent judgments count. As shown in Figure~\ref{fig:f10ability}, errors concentrate in few capabilities. The two most frequent categories of a benchmark hold a median 71\% of its attributions, and profiles overlap weakly, with mean cosine similarity 0.29. Fine-grained action leads on 27 benchmarks, spatial relation on 13, and temporal order and action counting on 7 each, counting ties. Covering the observed capability gaps therefore takes several sources.

\subsection{Duplicates reduce effective size}
\label{sec:f7size}
\begin{figure}[!t]
\centering
\includegraphics[width=\linewidth]{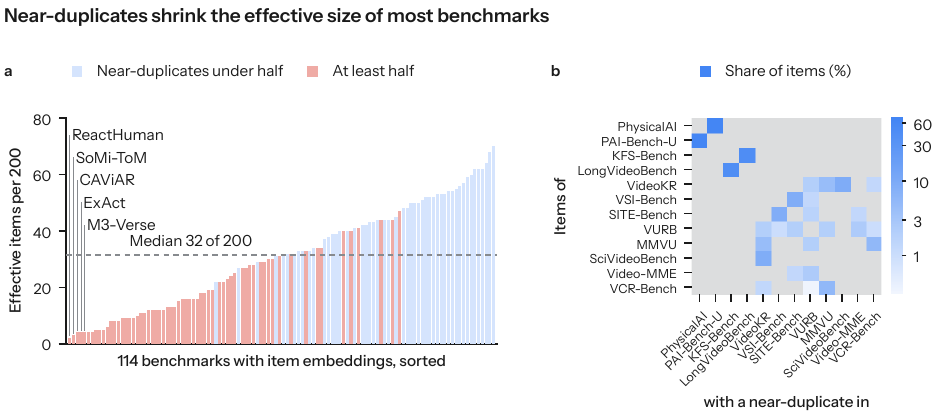} %
\caption{\textbf{Near-duplicates shrink effective size.} \textbf{a}, Effective items per 200, red where at least half are near-duplicates. \textbf{b}, The largest near-duplicate shares across benchmarks.}
\label{fig:f7size}
\end{figure}
We count distinct items, since item count can overstate diversity \citep{friedman2022vendi}. Let $\mathbf{v}_i$ be the unit-normalized mean SigLIP2 \citep{tschannen2025siglip} embedding of the 32 video frames of item $i$, and let $\mathbf{t}_i$ be its unit-normalized BGE-large \citep{xiao2024c} question embedding. For a sample of $n$ items, the joint similarity matrix $K$ and its Vendi score are
\begin{equation}
K_{ij} = (\mathbf{v}_i^{\top}\mathbf{v}_j)(\mathbf{t}_i^{\top}\mathbf{t}_j), \qquad
\mathrm{VS}(K) = \exp\Big(-\sum_{k=1}^{n} \lambda_k \log \lambda_k\Big),
\label{eq:vendi}
\end{equation}
\noindent where $\lambda_k$ is the $k$th eigenvalue of $K/n$. The Vendi score counts how many distinct items carry the observed diversity, and with $n=200$ we call it the effective size $n_{\mathrm{eff}}$. As shown in Figure~\ref{fig:f7size}, a 200-item draw holds a median of 32 effective items, and 39 benchmarks hold 20 or fewer. Near-duplicate questions make up at least half the items of 63 benchmarks, and weighting items by inverse duplicate-cluster size moves accuracy by more than 1\% on 93 benchmarks. Across sources, a tenth of items use videos that appear in another benchmark. PAI-Bench-U \citep{zhou2026paibench} and PhysicalAI \citep{zhou2026paibench} share 191 questions and yield nearly identical scores across conditions.

\section{Rebuilding video evaluation}
\label{sec:compose}

\subsection{Screening every item}
\label{sec:pool}
We screen all 381,614 multiple-choice items from 112 of the 115 audited benchmarks, leaving out the three added most recently. Qwen3-VL-8B \citep{bai2025qwen3} removes items it answers from text alone, options alone, or a single frame, and Qwen3-VL-2B \citep{bai2025qwen3} removes items it answers from 32 frames in order or shuffled. We then drop near-duplicate questions across sources and restrict repeated videos during selection. A labeling agent assigns task and video-content labels, and we record provenance, duration, attack margins, and question and middle-frame embeddings. As shown in Figure~\ref{fig:poolfunnel}, attacks remove 76\% of distinct questions, and 62,142 remain with complete screening and labeled videos. From these, the selector draws a 10,000-item coverage composition over 105 benchmarks, four capability groups, and 19 content categories.

\subsection{Composition from a request}
\label{sec:pipeline}
A benchmark request specifies task and event coverage, an item budget $k$, and a cap per source. Following agent-assisted benchmark construction \citep{butt2024benchagents, li2025autobencher}, a specification agent translates the request into a query over the pool and returns it for confirmation. A request can, for example, select household videos longer than 60 seconds with planning or memory questions. For the coverage composition, the selector draws items round-robin across duration groups, event types, and sources. Its initial caps allow one item per video and visual scene, five per question template, and 5\% of the budget per source, and scarce coverage cells trigger recorded relaxations. Within a source, the selector prefers less templated questions and then weaker worst-case attacker scores. The composition holds 9,688 videos from 105 benchmarks across seven duration groups. The appendix details this composition and its evaluation.

After selection, a red-team gate applies adversarial filtering \citep{le2020adversarial} to check the composition for residual shortcuts. It tests selected items with Qwen3-VL-8B \citep{bai2025qwen3} at 32 frames and the caption attacker, then reruns all five levels. It drops solved items and refills for a bounded number of rounds. A registrar randomizes option order, reports residual exploitability, and logs the pool snapshot, query, selector version, and seed in a lockfile whose hash names the release.

\subsection{Video-Index}
\label{sec:meta}
\begin{figure}[t]
\centering
\includegraphics[width=\linewidth]{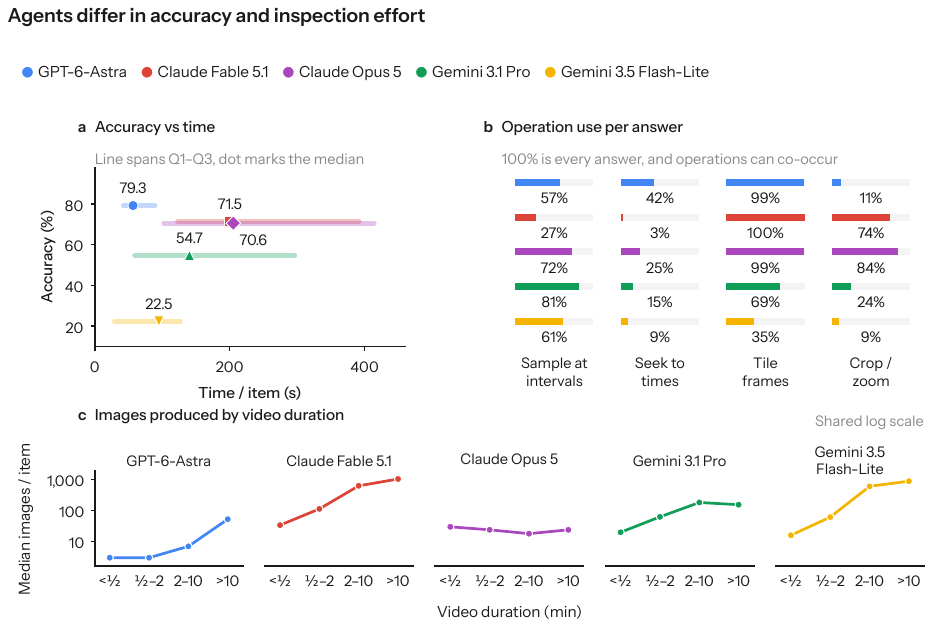}
\caption{\textbf{Agents differ in accuracy, runtime, and inspection.} \textbf{a}, Accuracy and median runtime, Q1--Q3. \textbf{b}, Operations per answer. \textbf{c}, Image medians by duration on shared log axes.}
\label{fig:agentcomparison}
\end{figure}
\nocite{bai2025qwen3,li2026videochat3,yang2026cambrian,wang2025internvideo2,zhang2025videollama,qin2025video,an2026llava,wang2025internvl3,clark2026molmo2,openai2026gpt6astra,anthropic2026fable51,anthropic2026opus5,google2026gemini31pro,google2026gemini35flashlite}
\begin{table}[t]
\centering
\caption{\textbf{Agents lead open models on Video-Index.} Accuracy in \%, fixed input then agent tools. Blind sees text only and Gain is Video minus Blind. Highest in bold. Protocols in the appendix.}
\label{tab:apex}
\begingroup
\scriptsize %
\begin{tabularx}{\textwidth}{>{\raggedright\arraybackslash}X|ccc|cccc}
\toprule
Model & Video & Blind & Gain & Percep. & Temp. & Spatial & Reason. \\
\midrule

VideoChat3-4B \citep{li2026videochat3} & 9.4 & 3.8 & 5.7 & 9.5 & 12.4 & 6.2 & 9.5 \\
Qwen3-VL-8B \citep{bai2025qwen3} & 9.5 & 1.4 & 8.1 & 9.0 & 11.4 & 6.7 & 11.0 \\
Cambrian-S-7B \citep{yang2026cambrian} & 11.3 & 5.3 & 6.0 & 12.4 & 10.0 & 3.8 & 19.0 \\
InternVideo2.5-8B \citep{wang2025internvideo2} & 12.4 & 7.5 & 4.9 & 12.9 & 11.0 & 8.1 & 17.6 \\
VideoLLaMA3-7B \citep{zhang2025videollama} & 12.4 & 5.2 & 7.1 & 11.9 & 18.6 & 8.6 & 10.5 \\
Video-XL-2 \citep{qin2025video} & 12.4 & 7.3 & 5.1 & 12.4 & 15.7 & 9.0 & 12.4 \\
LLaVA-OneVision-2-8B \citep{an2026llava} & 14.2 & 1.8 & 12.4 & 11.4 & 21.0 & 11.9 & 12.4 \\
InternVL3.5-8B \citep{wang2025internvl3} & 15.2 & 4.4 & 10.8 & 13.3 & 20.0 & 8.1 & 19.5 \\
Molmo2-8B \citep{clark2026molmo2} & 19.0 & 4.8 & 14.2 & 20.0 & 16.2 & 17.1 & 22.9 \\
Claude Opus 5 \citep{anthropic2026opus5} & 56.8 & 21.3 & 35.5 & 62.4 & 56.2 & 46.7 & 61.9 \\
\midrule
\multicolumn{8}{l}{\textit{Agentic evaluation}}\\
Gemini 3.5 Flash-Lite \citep{google2026gemini35flashlite} & 22.5 & 17.5 & 5.0 & 20.0 & 30.0 & 10.0 & 30.0 \\
Gemini 3.1 Pro \citep{google2026gemini31pro} & 54.7 & 23.9 & 30.8 & 58.2 & 59.1 & 60.6 & 40.9 \\
Claude Opus 5 \citep{anthropic2026opus5} & 70.6 & \textbf{24.9} & 45.7 & 68.4 & 77.2 & 69.1 & 67.6 \\
Claude Fable 5.1 \citep{anthropic2026fable51} & 71.5 & 20.7 & 50.8 & 68.4 & 77.7 & 71.1 & 68.7 \\
GPT-6-Astra \citep{openai2026gpt6astra} & \textbf{79.3} & 18.7 & \textbf{60.6} & \textbf{75.4} & \textbf{83.7} & \textbf{80.4} & \textbf{77.1} \\
\midrule
Human & 55.0 & 17.5 & 37.5 & 56.0 & 68.0 & 56.0 & 40.0 \\
\bottomrule
\end{tabularx}
\endgroup
\end{table}
Video-Index stress-tests models on the 840 hardest verified questions from 76 sources, 210 per capability group and one per video, ranked by their maximum percentile over five attacker margins. Claude Opus 5 \citep{anthropic2026opus5} labels the 3,000 hardest candidates by fine capability and verifies their answers against the frames, and selection ignores reference-model accuracy. We evaluate nine open video models, namely Qwen3-VL-8B \citep{bai2025qwen3}, VideoChat3-4B \citep{li2026videochat3}, Cambrian-S-7B \citep{yang2026cambrian}, InternVideo2.5-8B \citep{wang2025internvideo2}, VideoLLaMA3-7B \citep{zhang2025videollama}, Video-XL-2 \citep{qin2025video}, LLaVA-OneVision-2-8B \citep{an2026llava}, InternVL3.5-8B \citep{wang2025internvl3}, and Molmo2-8B \citep{clark2026molmo2}. These models and a fixed-input Claude Opus 5 receive at most 512 frames at 1 fps with a 224-pixel short side. Five agents use local tools on the same 2-fps videos, namely GPT-6-Astra \citep{openai2026gpt6astra}, Claude Fable 5.1 \citep{anthropic2026fable51}, Claude Opus 5 \citep{anthropic2026opus5}, Gemini 3.1 Pro \citep{google2026gemini31pro}, and Gemini 3.5 Flash-Lite \citep{google2026gemini35flashlite}. Blind scores average four option orders. As shown in Table~\ref{tab:apex}, fixed-input Claude Opus 5 outscores every open model by more than 37\%. On matched items, tools raise the same model by 19.8\%. Human volunteers reach 55.0\%, below the three leading agents and above every open model. At a matched item budget, the gate holds the strongest attacker to 13\%, against 19\% for ungated selections. Blind baselines \citep{chen2024we} give a mean text-only accuracy of 6\% and a video gain of 11\% over ten models, with video above blind for all ten. Item-bootstrap signal-to-noise analysis \citep{heineman2026signal} gives a noise floor of 2.0\% and a signal-to-noise ratio of 10.8, against 3.7 for the median source. The appendix gives protocols, validation, and coverage-first compositions.

We analyze execution logs to see how agents gather evidence. As shown in Figure~\ref{fig:agentcomparison}, GPT-6-Astra \citep{openai2026gpt6astra}, Claude Fable 5.1 \citep{anthropic2026fable51}, and Claude Opus 5 \citep{anthropic2026opus5} all tile frames into grids on nearly every answer and then diverge. Any two of them give the same answer on about four in five items but choose the same sampling strategy on fewer than half. \mbox{GPT-6-Astra} seeks timestamps more often, while Claude Fable 5.1 and Claude Opus 5 favor cropping and zooming. Claude Opus 5 extracts similar image counts across durations. GPT-6-Astra decodes all stored frames on short clips and combines methods on longer videos, whereas Claude Fable 5.1 decodes exhaustively and extracts far more images without reaching GPT-6-Astra's accuracy. All three score lower on videos past ten minutes, GPT-6-Astra least of all. Evaluation should separate the volume of extracted imagery from its relevance and reward knowing when to stop.

\section{Conclusion and Discussion}
\label{sec:conclusion}
The attack pyramid exposes shortcuts across 115 video benchmarks and yields Video-Index, 840 of the hardest verified questions. A score certifies a capability only above its strongest shortcut, and builders can report the breaking level beside the score. The screened pool supports further benchmarks with other capability and coverage goals through the same pipeline. Leading proprietary agents outperform open models but leave room to improve accuracy and evidence selection. Resistance holds only against tested attacks, and audits and benchmarks should evolve with new models and shortcuts. Progress requires accuracy, shortcut resistance, and efficient evidence gathering.

\paragraph{Acknowledgments.} We thank Saining Xie, Weili Xu, and Shengyu Hao for additional compute and API support throughout this project, and for their help in many other ways.

\clearpage

\bibliography{references,fig_tex/benchmarks}

\clearpage
\beginsupplement
\renewcommand{\floatpagefraction}{0.8}
\crefalias{section}{appendix} %

\phantomsection
\addcontentsline{toc}{section}{Appendix}
\section*{Appendix}
\noindent The appendix is organized as follows.
\begin{itemize}
\item Appendix~\ref{app:benchmarkreferences} lists the 115 audited benchmarks with their references.
\item Appendix~\ref{app:related} places the pyramid in the literature on benchmark validity, blind baselines, dataset artifacts, evidence grounding and evidence-seeking agents, data contamination, and benchmark composition and selection, and compares related audits in one table.
\item Appendix~\ref{app:limitations} states the limitations of the audit and of Video-Index.
\item Appendix~\ref{app:examples} shows one item for each shortcut and two Video-Index items.
\item Appendix~\ref{app:observers} reads the five levels as bounded observers and states what each level can and cannot certify.
\item Appendix~\ref{app:tierfigs} gives the per-benchmark figures of the option and text levels.
\item Appendix~\ref{app:settings} lists the settings of every stage of the audit, the pool, and Video-Index in one table.
\item Appendix~\ref{app:screen} details the measurement screen, the scoring rule, the error-attribution protocol, and the duplicate refill.
\item Appendix~\ref{app:t0} details the option level, with answer-position entropy, option counts against declared chance, and permutation sensitivity.
\item Appendix~\ref{app:t1} details the text level, with the blind readers, video dependence, and the paired McNemar test.
\item Appendix~\ref{app:t2} details the pool level, with shared videos and questions across benchmarks and the copy, retrieval, and learned attackers.
\item Appendix~\ref{app:t3} details the frame level, with the single-frame and caption attackers and the diagnostic profiles.
\item Appendix~\ref{app:t4} details the order level, with the shuffle, window, and half probes and the frame-budget ladder.
\item Appendix~\ref{app:findings} resolves the capability groups into fine categories and reports breaking levels by fine capability.
\item Appendix~\ref{app:claimmatched} recounts the breaking levels with only the attacks that match the claim of each benchmark.
\item Appendix~\ref{app:tolerance} recounts the breaking levels at tolerances of 3\% and 8\% and with Claude Opus 5 as the reference, and compares the order in which benchmarks break under these settings.
\item Appendix~\ref{app:framespixels} compares resolution against frame rate and reports the long-video frame-budget results.
\item Appendix~\ref{sec:f2} shows where adding video lowers a score that the blind reader had right.
\item Appendix~\ref{sec:f5} attributes the reference errors on surviving benchmarks to fine capability categories.
\item Appendix~\ref{app:whichbench} recommends benchmarks for each capability group.
\item Appendix~\ref{app:pool} details the pool construction, the watcher that keeps it current, and the Video-Index selection.
\item Appendix~\ref{sec:agentbehaviour} summarizes how the agents inspect video on Video-Index.
\item Appendix~\ref{app:fulltables} lists the taxonomy and the master table.
\item Appendix~\ref{app:cards} collects the report card of every audited benchmark.
\end{itemize}

\section{Audited benchmarks}
\label{app:benchmarkreferences}
We evaluate the following 115 benchmarks. References are collected here so that the appendix tables can use benchmark names without repeated citations, namely ALLVB \citep{tan2025allvb}, AoTBench \citep{xue2025aotbench}, Artifact-Bench \citep{tang2026artifactbench}, Causal2Needles \citep{li2025causal2needles}, CAViAR \citep{garg2026caviar}, CG-Bench \citep{chen2025cgbench}, CineTechBench \citep{wang2025cinetechbench}, CMU \citep{tan2026acam}, Colon-Bench \citep{hamdi2026colonbench}, ContPhy \citep{zheng2024contphy}, COVER \citep{zhou2025cover}, DSI-Bench \citep{zhang2025dsibench}, DSR-Bench \citep{zhou2025dsrbench}, E.T. Bench \citep{liu2024etbench}, EgoPlan-Bench \citep{chen2025egoplan}, EgoSchema \citep{mangalam2023egoschema}, ExAct \citep{yi2025exact}, ExpVid \citep{xu2025expvid}, FAVOR-Bench \citep{tu2025favorbench}, FineBench \citep{faure2026finebench}, ForeSeaQA \citep{park2026foresea}, GasVideo-1000 \citep{tang2026gasvideo}, GUI-World \citep{chen2025guiworld}, HERBench \citep{benami2026herbench}, HLV-1K \citep{zou2025hlv1k}, HourVideo \citep{chandrasegaran2024hourvideo}, IntPhys 2 \citep{bordes2025intphys2}, iSafetyBench \citep{abdullah2025isafetybench}, K9-Bench \citep{attarde2026k9bench}, KFS-Bench \citep{li2026kfsbench}, LEMONADE \citep{bonnetto2025lemonade}, LongTimeScope \citep{zohar2025apollo2}, LongVideo-Reason \citep{chen2025longvideoreason}, LongVideoBench \citep{wu2024longvideobench}, LVBench \citep{wang2024lvbench}, M3-Verse \citep{wei2025m3verse}, MLVU \citep{zhou2025mlvu}, MME-VideoOCR \citep{shi2025mmevideoocr}, MMR-V \citep{zhu2026mmrv}, MMSI-Video-Bench \citep{lin2025mmsivideo}, MMVU \citep{zhao2025mmvu}, MMWorld \citep{he2024mmworld}, MORSE-500 \citep{cai2025morse500}, MotionBench \citep{hong2025motionbench}, MVBench \citep{li2024mvbench}, NaturalBench-Video \citep{li2025naturalbenchvideo}, Neptune \citep{nagrani2024neptune}, NExT-QA \citep{xiao2021nextqa}, OSI-Bench \citep{wu2026osibench}, OVO-S-Bench \citep{li2026ovosbench}, PAI-Bench-U \citep{zhou2026paibench}, Perception Test \citep{patraucean2023perceptiontest}, PerceptionComp \citep{li2026perceptioncomp}, PhysicalAI \citep{zhou2026paibench}, PLM-VideoBench \citep{cho2025perceptionlm}, Q-Bench-Video \citep{zhang2025qbenchvideo}, ReactHuman \citep{reacthuman2026benchmark}, ReVSI \citep{zhang2026revsi}, ReXTime \citep{chen2024rextime}, S-EMBER \citep{wang2026sember}, SAW-Bench \citep{li2026sawbench}, ScaleLong \citep{ma2025scalelong}, SciVideoBench \citep{deng2025scivideobench}, SFI-Bench \citep{zhang2026sfibench}, SIS-Bench \citep{zou2026sisbench}, SITE-Bench \citep{wang2025site}, SLVMBench \citep{yang2026slvmbench}, SmartHome-Bench \citep{zhao2025smarthomebench}, SoMi-ToM \citep{fan2025somitom}, SpatialTree \citep{xiao2026spatialtree}, ST-Evidence \citep{wang2026stevidence}, STI-Bench \citep{li2025stibench}, StreamingBench \citep{lin2024streamingbench}, TaskMeAnything-v1 \citep{zhang2024taskmeanything}, TempCompass \citep{liu2024tempcompass}, TemporalBench \citep{cai2024temporalbench}, TimeBlind \citep{li2026timeblind}, TimeScope \citep{zohar2025timescope}, TOC-Bench \citep{anonymous2026tocbench}, TOMATO \citep{shangguan2024tomato}, TVBench \citep{cores2024tvbench}, UCF101-AD \citep{abdullah2026ucf101ad}, UrbanVideo-Bench \citep{zhao2025urbanvideo}, v-HUB \citep{shi2026vhub}, V2P-Bench \citep{zhao2026v2pbench}, VANE-Bench \citep{bharadwaj2025vanebench}, VAU-Bench \citep{zhu2025vaur1}, VCR-Bench \citep{qi2025vcrbench}, VELOCITI \citep{saravanan2025velociti}, VF-Eval \citep{song2025vfeval}, VGenST-Bench \citep{park2026vgenstbench}, Video-Holmes \citep{cheng2025videoholmes}, Video-LevelGauge \citep{xia2026videolevelgauge}, Video-MME \citep{fu2025video}, Video-MME-v2 \citep{fu2026videommev2}, Video-MMMU \citep{hu2026video}, Video-TT \citep{zhang2025videott}, VideoAds \citep{zhang2025videoads}, VideoEspresso \citep{han2025videoespresso}, VideoEval-Pro \citep{ma2026videoevalpro}, VideoKR \citep{fu2026videokr}, VideoMathQA \citep{rasheed2025videomathqa}, VideoOdyssey \citep{he2026videoodyssey}, VideoSIAH-Eval \citep{yang2026longvt}, VidPair-Halluc \citep{huang2026vidpairhalluc}, ViMUL-Bench \citep{shafique2025vimul}, Vinoground \citep{zhang2024vinoground}, VITATECS \citep{li2024vitatecs}, VKnowU \citep{jiang2025vknowu}, VNBench \citep{zhao2025vnbench}, VRU-Accident \citep{kim2025vruaccident}, VSI-Bench \citep{yang2025vsibench}, VSI-Super-Wild \citep{gu2026vsisuperwild}, VSTAT \citep{yu2026vstat}, and VURB \citep{wei2026vurb}.

\section{Related work}
\label{app:related}
\paragraph{Video understanding benchmarks.}
ActivityNet-QA \citep{yu2019activitynetqa} and NExT-QA \citep{xiao2021nextqa} evaluate short clips. MVBench \citep{li2024mvbench}, Perception Test \citep{patraucean2023perceptiontest}, EgoSchema \citep{mangalam2023egoschema}, and Video-MME \citep{fu2025video} broaden tasks and durations. Later evaluations focus on particular capabilities. HourVideo \citep{chandrasegaran2024hourvideo}, LVBench \citep{wang2024lvbench}, LongVideoBench \citep{wu2024longvideobench}, and MLVU \citep{zhou2025mlvu} target long videos. TVBench \citep{cores2024tvbench}, TOMATO \citep{shangguan2024tomato}, TempCompass \citep{liu2024tempcompass}, and Vinoground \citep{zhang2024vinoground} test temporal order. VSI-Bench \citep{yang2025vsibench} evaluates spatial reasoning, VideoHallucer \citep{wang2024videohallucer} hallucination, and StreamingBench \citep{lin2024streamingbench} online understanding. Audits reveal shortcuts within these settings. \citet{buch2022revisiting} answer many earlier benchmark items from one selected frame. \citet{lei2023revealing} show that a single-frame model can outperform multi-frame methods on popular video-and-language datasets. TVBench and TOMATO show that shuffling can preserve temporal-benchmark scores. VideoEval-Pro \citep{ma2026videoevalpro} finds that guessing and priors inflate multiple-choice scores on long-video benchmarks and adopts open-ended short answers. Cambrian-S \citep{yang2026cambrian} tests blind and single-frame performance and constructs VSI-Super \citep{yang2026cambrian} to resist both. VBenchComp \citep{feng2025breaking} separates text-answerable, semantic, and temporal items. \citet{brown2025benchmark} train on question text with cross-validation and remove predictable items from four test sets. We extend shortcut auditing to five levels across 115 benchmarks and compose new evaluations from surviving items.

\paragraph{Benchmark construction and auditing.}
MMStar \citep{chen2024we} removes image-benchmark items answerable without images or exposed through training data. NaturalBench \citep{li2024naturalbench} pairs each question with two images that yield different answers, which blocks blind solutions by design. DatBench \citep{joshi2026datbench} selects items requiring images and separating models, while PerceptionBench \citep{lin2026perceptionbench} isolates atomic perception. BetterBench \citep{reuel2024betterbench} assesses evaluation practice. VLMEvalKit \citep{duan2024vlmevalkit} and LMMs-Eval \citep{zhang2025lmms} run many benchmarks under shared protocols. tinyBenchmarks \citep{polo2024tinybenchmarks} and metabench \citep{kipnis2025metabench} select items predictive of full scores. Fluid Benchmarking \citep{hofmann2025fluid} selects adaptively. Lifelong Benchmarks \citep{prabhu2024efficient}, LiveXiv \citep{shabtay2025livexiv}, and LiveBench \citep{white2025livebench} refresh evaluations over time. \citet{perlitz2024benchmark} validate new benchmarks through agreement with established ones. Task-Me-Anything \citep{zhang2024taskmeanything} generates items from taxonomies, AutoBencher \citep{li2025autobencher} searches for difficult topics, and BenchAgents \citep{butt2024benchagents} connects planning, generation, and verification. Harbor-Index \citep{shi2026harbor} distills 82 agentic tasks from 54 benchmarks. SWE-Bench+ \citep{aleithan2024swe} and BenchJack \citep{wang2026androids} audit leaked solutions and exploitable graders. Our selection instruments build on AFLite's removal of predictable items \citep{le2020adversarial}, confident learning's label-error checks \citep{northcutt2021confident}, and IFD's model-relative difficulty \citep{li2024quantity}. Coverage-centric coreset selection \citep{zheng2022coverage} and QDIT \citep{bukharin2024data} balance coverage, quality, and diversity. We apply these principles to an audited pool drawn from existing video benchmarks. The selector preserves coverage, the Vendi score measures diversity \citep{friedman2022vendi}, and a lockfile makes each composition reproducible.

\paragraph{Dataset artifacts and adversarial data.}
Language priors let visual question answering models answer without looking at the image, which motivated balanced image pairs in VQA v2 \citep{goyal2017making} and shifted answer priors in VQA-CP \citep{agrawal2018don}. Hypothesis-only baselines expose annotation artifacts in natural language inference \citep{gururangan2018annotation, poliak2018hypothesis}. Adversarial collection with humans and models in the loop builds harder data \citep{nie2020adversarial, kiela2021dynabench}, yet filtering against one model family can make the resulting set unfair to that family \citep{phang2021adversarially}. The attack pyramid extends the blind baseline to five levels of access to a video, and Qwen3-VL-8B \citep{bai2025qwen3}, which screens the pool, is excluded from the ranking-agreement statistics of Video-Index. As shown in Table~\ref{tab:relatedcompare}, prior audits apply one or two probes to a few benchmarks, while the pyramid applies five probes to 115 benchmarks and releases a screened pool for new compositions.

\paragraph{Evidence grounding and evidence seeking.}
Evidence-grounded benchmarks score whether a model finds the evidence behind its answer. EG-VQA \citep{huang2026eg} pairs open-ended answers with annotated temporal evidence, and scene-text grounding \citep{zhou2025scene} asks for the text regions that support each answer. ViTexQA \citep{guo2026vitexqa} requires reading text across several frames. These benchmarks check the evidence directly, while Video-Index keeps the multiple-choice format and removes the items that attackers answer without it, and the two checks complement each other. Evidence-seeking systems choose frames before answering. Deep Video Discovery \citep{zhang2026deep}, TimeSearch-R \citep{pan2026timesearch}, and T$^{*}$ \citep{ye2025re} search the timeline for evidence. HiMu \citep{ben2026himu} and ToolMerge \citep{shlapentokh2026decomposing} decompose the question and score frames with lightweight tools, and an evidence-seeking agent for the TimeLogic challenge \citep{xu2026timelogic} compares timestamped observations. Such searchers could serve as stronger attackers at the frame and order levels. UCS-Bench \citep{wang2026keep} and EgoMonth \citep{chen2026egomonth} test memory over hours to months of egocentric video. TGPO \citep{xu2026incentivizing} rewards reasoning that depends on frame order by contrasting ordered and shuffled frames, the perturbation of our order level.

\begin{table}[!htbp]
\centering
\caption{Related audits and benchmark builders. Data gives the modality, and Audit gives the number of existing benchmarks each work examines, where Several means standard benchmarks without a count in the abstract. Probes marks the shortcuts a work tests for or designs against, O for options alone, T for question text without the visual input, P for other evaluation items or training on the test set, F for single or few frames, and R for frame order.}
\label{tab:relatedcompare}
\begin{tabularx}{\linewidth}{@{}llll>{\raggedright\arraybackslash}X@{}}
\toprule
Work & Data & Audit & Probes & Release \\
\midrule
ATP \citep{buch2022revisiting} & Video & Several & F & Temporally hard subsets \\
\citet{lei2023revealing} & Video & Several & F & Two temporal retrieval tasks \\
TVBench \citep{cores2024tvbench} & Video & Several & T F & New benchmark \\
TOMATO \citep{shangguan2024tomato} & Video & Several & F R & New benchmark, human annotated \\
VideoEval-Pro \citep{ma2026videoevalpro} & Video & Several & T F & Open-ended benchmark \\
TsT \citep{brown2025benchmark} & Image, video & 4 & T P & Debiased VSI-Bench \\
MMStar \citep{chen2024we} & Image & 6 & T & Curated subset, human reviewed \\
NaturalBench \citep{li2024naturalbench} & Image & -- & T & Paired benchmark, human verified \\
tinyBenchmarks \citep{polo2024tinybenchmarks} & Text & 4 & -- & Small predictive subsets \\
AutoBencher \citep{li2025autobencher} & Text & -- & -- & Generated datasets \\
\midrule
This work & Video & 115 & O T P F R & Screened pool, compositions on request, Video-Index \\
\bottomrule
\end{tabularx}
\end{table}

\section{Limitations}
\label{app:limitations}
\paragraph{Multiple choice only.}
We audit only multiple-choice benchmarks. The protocol covers 115 English multiple-choice benchmarks, and open-ended and generative video benchmarks fall outside it. Video-Index keeps the multiple-choice format that the option level shows can leak answers, and its blind scores average four option orders.

\paragraph{No audio.}
We exclude benchmarks that require audio. No attacker receives speech, subtitles, or sound, and audio-visual claims remain untested.

\paragraph{Lower bounds.}
We bound exploitability from below. A benchmark that survives every level resists the tested attackers, and a stronger or newer attacker can raise its exploitability and lower its breaking level.

\section{Qualitative examples}
\label{app:examples}
\newcommand{\qualbox}[1]{\par\smallskip\noindent\fbox{\parbox{\dimexpr\linewidth-2\fboxsep-2\fboxrule\relax}{#1}}\par\smallskip}
We show one sampled item for each of five shortcuts and two items of Video-Index. Each box gives the question and the options, with the correct option in bold, and the text above it states what each attacker saw and chose. Frames appear where an attacker saw visual input or where the item belongs to Video-Index, and no attacker sees the correct answer.

\paragraph{Options alone.}
ExAct \citep{yi2025exact} breaks at the option level. The options-only attacker, Claude Opus 5 \citep{anthropic2026opus5}, sees the five options of item 3175 without the question or the video and picks D, the correct option. D is the only option that says participant instead of player, and the only one without a reason clause. On 240 of the 300 sampled ExAct items, this attacker picks the correct option.
\qualbox{\textbf{Question}\enspace Which expert commentary correctly assesses the skill execution shown in the video?\\
A.~The player should focus on jumping higher before releasing the ball to generate more backspin, improving the shot's precision and reducing bounce-off potential.\\
B.~The player should focus on holding the elbow away from the body to create a wider arc, which will make the shot more difficult for opposing players to contest.\\
C.~The player should keep the fingers stiff and straight while shooting to ensure a stronger shot and prevent the ball from being underpowered or easily blocked.\\
\textbf{D.~The participant should focus on snapping the wrist down and pointing the fingers down when shooting.}\\
E.~The player should concentrate on flicking the wrist upwards and extending the fingers straight out when shooting to achieve more height and distance on the shot.}

\paragraph{Question text.}
MMR-V \citep{zhu2026mmrv} breaks at the text level. Gemini 3.5 Flash \citep{google2026gemini35flash} and Claude Opus 5 each read the question and the eleven options of item 337 without the video, and both pick D. Only D names a specific work, while the other ten options are generic titles about ratings, technology, and the workplace. Both readers answer 124 of the 300 sampled MMR-V items correctly without the video.
\qualbox{\textbf{Question}\enspace Please suggest a suitable title for this video.\\
A.~A Digital Society Perspective\quad B.~Social Ratings and Personal Perceptions\quad C.~Online Influence on Personal Life\quad \textbf{D.~Black Mirror - Social Credit System}\quad E.~Living Under the Social Lens\quad F.~Empathy in the Age of Technology\quad G.~Corporate Culture vs.\ Personal Image\quad H.~The Intersection of Ratings and Reality\quad I.~Navigating Social Standards\quad J.~The Power of Community Approval\quad K.~Workplace Dynamics in a Digital World}

\paragraph{Other evaluation items.}
TimeScope \citep{zohar2025timescope} breaks at the pool level. Items 1104 and 1892 come from videos of 3,600 and 180 seconds, yet they ask the same question with the same six options and the same answer. Their question embeddings coincide, and the copy attacker, which copies the answer of the nearest earlier question when the BGE-large \citep{xiao2024c} cosine similarity exceeds 0.9, answers whichever item comes second.
\qualbox{\textbf{Question}\enspace What does the girl do after she get a cup of coffee?\\
A.~reading a book\quad B.~working with her computer\quad \textbf{C.~taking some notes}\quad D.~cleaning the kitchen\quad E.~preparing a breakfast\quad F.~washing the dishes}

\paragraph{Captions.}
NExT-QA \citep{xiao2021nextqa} breaks at the frame level. For item 7426, Claude Opus 5 reads the captions of 32 frames, each written independently by AuroraCap-7B-IMG \citep{chai2025auroracap}, and picks C, the correct option. The captions describe a man with a small drone in a garden and later a small object resting on a white platform, and none of the 32 captions mentions a baby, a lady, or a girl. The same model shown a single frame picks B.
\qualbox{\includegraphics[width=\linewidth,height=0.9in,keepaspectratio]{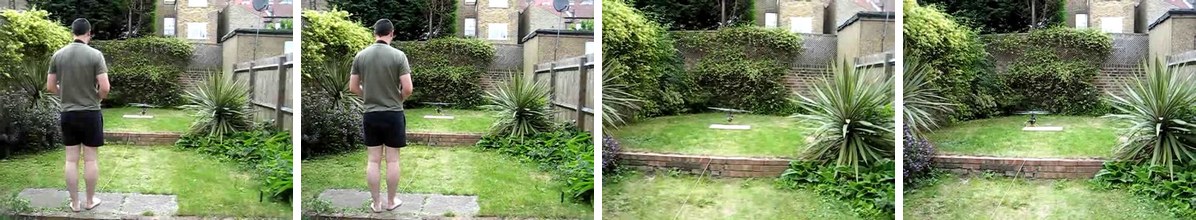}\\[2pt]
\textbf{Question}\enspace why did the blade stop moving at the end\\
A.~baby stopped it\quad B.~no battery\quad \textbf{C.~the man stopped it}\quad D.~lady pressed a button\quad E.~girl pulled the sockets\\[2pt]
\textbf{Captions}\enspace ``A man in a green t-shirt and black shorts stands barefoot on a lawn, facing a small, black, circular object on the ground, which appears to be a remote-controlled toy.'' ``The image shows a man standing in a garden, facing a drone that is in flight.'' ``The image captures a tranquil garden scene with a small, unmanned aircraft on the ground, a white rectangular object on the grass, and a variety of greenery.''}

\paragraph{Shuffled frames.}
MVBench \citep{li2024mvbench} breaks at the order level. For item action\_prediction\_88, Gemini 3.5 Flash reads the 32 frames in random order and still picks the correct next action, as it does with ordered frames. The question asks for the next action, yet the answer survives the loss of frame order.
\qualbox{\includegraphics[width=\linewidth,height=0.9in,keepaspectratio]{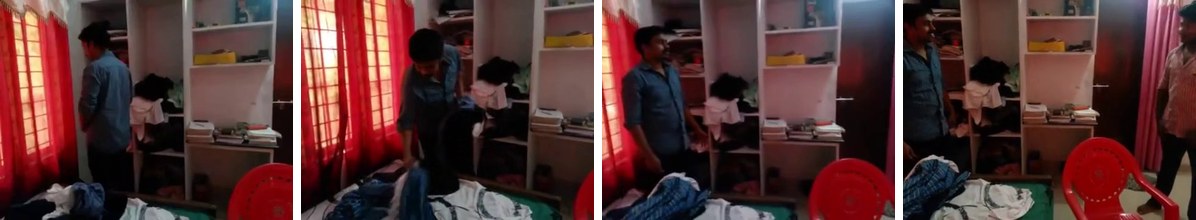}\\[2pt]
\textbf{Question}\enspace What will the person do next?\\
A.~Take the shoe.\quad B.~Lie on the floor.\quad C.~Close the box.\quad \textbf{D.~Tidy up the blanket.}}

\paragraph{A Video-Index item that agents solve.}
NaturalBench-Video \citep{li2025naturalbenchvideo} item 1481 plays an emergence in reverse, from a butterfly beside its chrysalis to the chrysalis alone. GPT-6-Astra \citep{openai2026gpt6astra}, Claude Fable 5.1 \citep{anthropic2026fable51}, and Claude Opus 5 working as agents pick B, the correct option. All nine open models and Claude Opus 5 with fixed input pick A, the forward transformation.
\qualbox{\includegraphics[width=\linewidth,height=0.9in,keepaspectratio]{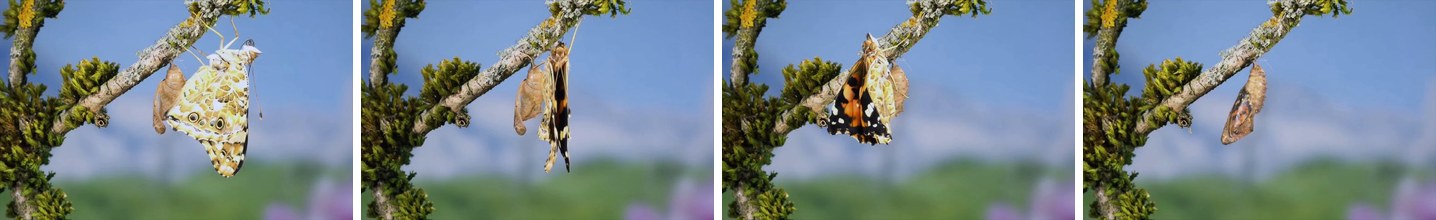}\\[2pt]
\textbf{Question}\enspace What transformation is occurring in the cocoon?\\
A.~The insect in the cocoon turns into a butterfly.\quad \textbf{B.~The butterfly turns back into an insect in the cocoon.}}

\paragraph{A Video-Index item that no system solves.}
S-EMBER \citep{wang2026sember} item 3823 follows the wearer of a camera through a furniture store for about five minutes and asks how many price tags the wearer touches after closing a nightstand drawer. The answer is four. The three agents pick one, Claude Opus 5 with fixed input picks three, and all nine open models also miss the answer.
\qualbox{\includegraphics[width=\linewidth,height=0.9in,keepaspectratio]{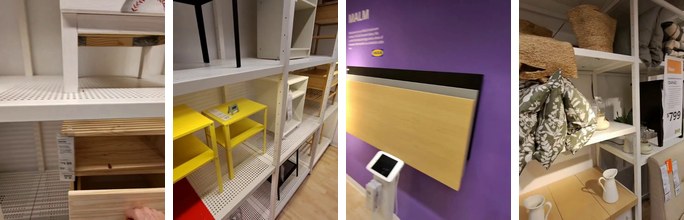}\\[2pt]
\textbf{Question}\enspace Since I closed the drawer on the nightstand, how many price tags have I touched?\\
A.~2 price tags.\quad B.~3 price tags.\quad C.~1 price tag.\quad D.~0 price tags.\quad \textbf{E.~4 price tags.}}

\section{Attack levels as bounded observers}
\label{app:observers}
\citet{finzi2026entropy} distinguish learnable structure from residual uncertainty relative to an observer's computational resources. Epiplexity denotes the structure available to a bounded observer. Time-bounded entropy denotes the remainder. This distinction motivates viewing each attacker set as an observer class. At the option level, the observer has only answer choices. Higher levels grant additional inputs. Exploitability $\varepsilon_\ell$ measures the score achieved by the declared class, and the breaking level identifies the first class approaching the reference. These operational measures differ from quantities such as answer-position entropy that depend only on the item distribution. A diverse-looking set can remain predictable to a shortcut model. At the pool level, sequential prediction measures how answer structure becomes learnable as earlier items accumulate.

Option permutations preserve the choices but change their positions. Frame shuffling preserves individual images but alters their order. Both interventions can change what a bounded model extracts from the input. The former tests sensitivity to answer layout. The latter tests dependence on temporal order. We interpret their score changes relative to the stated capability and the information retained by each intervention. Evaluation aims to separate a capability-bearing model from shortcut observers. An informative item is difficult for the declared attackers but solvable from the intended evidence. The residual score above the strongest shortcut quantifies this separation. Admissibility is therefore relative to a claim and its tested threat model.

\section{Per-benchmark option and text results}
\label{app:tierfigs}
\begin{figure}[!htbp]
\centering
\subfloat[Option attack margins above benchmark-specific chance across the audited multiple-choice sources.\label{fig:t0}]{\includegraphics[width=\linewidth]{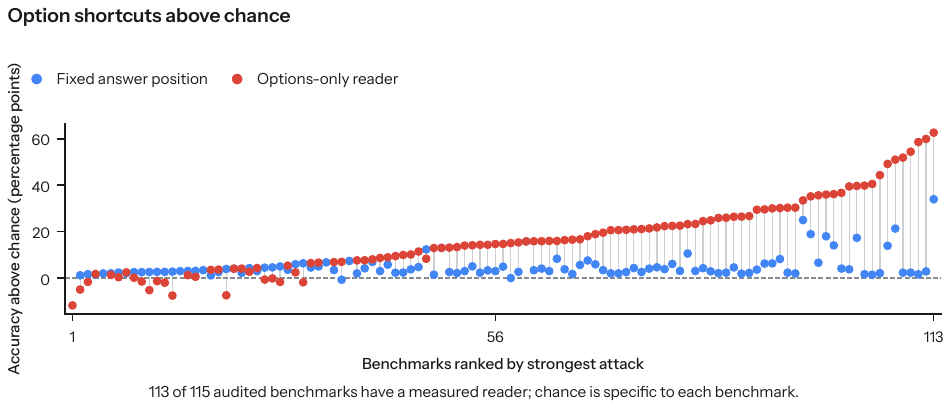}}

\subfloat[Blind-reader margins above benchmark-specific chance across the audited multiple-choice sources.\label{fig:t1m}]{\includegraphics[width=\linewidth]{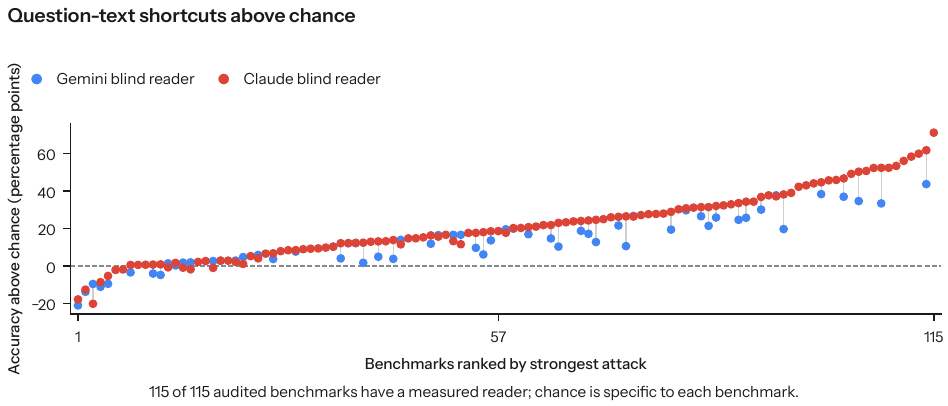}}
\caption{Measured option and blind-reader margins over chance within the 115-benchmark audit.}
\label{fig:tiers01}
\end{figure}

As shown in Figure~\ref{fig:tiers01}, option attacks and blind readers exceed each benchmark's own chance baseline on most audited benchmarks. Figure~\ref{fig:t0} compares fixed-position and options-only attacks, separating answer-position imbalance from predictability in the choices. Figure~\ref{fig:t1m} compares two blind readers given questions and options. Each panel ranks benchmarks by its own stronger attack, and a horizontal position marks a different benchmark in each panel. Vertical lines connect paired measurements, showing how much the attackers differ on each benchmark. Zero denotes chance performance. A break requires approaching the full-video reference, and above-chance accuracy alone does not meet that criterion. Each panel holds the benchmarks with that measurement.

\section{Experimental settings}
\label{app:settings}
Table~\ref{tab:settings} lists the settings of every stage of the audit, the screened pool, and Video-Index.
\begin{table}[!htbp]
\centering
\caption{Settings of the audit, the screened pool, and Video-Index. Each entry restates a setting given in the main text or the appendix.}
\label{tab:settings}
\small
\begin{tabularx}{\textwidth}{@{}>{\raggedright\arraybackslash}p{0.17\textwidth}>{\raggedright\arraybackslash}X@{}}
\toprule
Stage & Setting \\
\midrule
Audit sample & 300 questions per benchmark in a seed-42 draw stratified by subtask. One item per near-duplicate group is kept, and the draw refills within subtasks. \\
Media & Videos stored at a 720-pixel short side, or 480 for clips over 30 minutes without on-screen text. Frames reach the models at a long side of at most 768 pixels. \\
Reference & A reference model at 32 frames. Its 128-frame probe replaces it on long-video benchmarks where it stays within $\delta$ of chance. \\
Break rule & $\varepsilon_\ell \ge s^{*}-c-\delta$ with $\delta=5\%$, about the 95\% sampling margin of a 300-item accuracy. \\
Scoring & Option letter, then an answer-line regular expression, then the option text. \\
Option level & A fixed-position attacker picks the most frequent answer position, and Claude Opus 5 reads the options alone. Qwen3-VL-8B answers under eight option permutations with seed 42, 32 frames, and temperature 0. \\
Text level & Gemini 3.5 Flash and Claude Opus 5 read the question and options. \\
Pool level & A copy attacker at BGE-large cosine similarity above 0.9, Qwen3-8B reading the eight nearest question--answer pairs, and an online logistic regression on BGE-large embeddings over five rotating splits. \\
Frame level & Gemini 3.5 Flash reads one frame. Claude Opus 5 reads AuroraCap-7B-IMG captions of the 32 sampled frames, one caption per frame. \\
Order level & Gemini 3.5 Flash and Qwen3-VL-8B read 32 shuffled frames and, separately, a contiguous tenth of the video. \\
Duplicates & Near-duplicate questions at BGE-large cosine similarity of at least 0.9, confirmed with MPNet. Effective size is the Vendi score of 200 items under the product of SigLIP2 frame and BGE-large question similarities. \\
Pool screening & Qwen3-VL-8B answers from text alone, options alone, or one frame, and Qwen3-VL-2B from 32 frames in order or shuffled. \\
Composition & Round-robin selection over duration groups, event types, and sources with per-source caps. The gate reruns all five levels with Qwen3-VL-8B at 32 frames and the caption attacker and refills for a bounded number of rounds. A JSON lockfile records the snapshot hash, query, selector version, seed, item identifiers, option permutations, round count, and residual exploitability. \\
Video-Index & The 3,000 candidates with the lowest worst-case attacker percentile, one per video. Claude Opus 5 verifies answers at 32 frames, then at 512 frames, and 210 items fill each capability group. \\
Fixed-input evaluation & One frame per second, at most 512 frames, and a 224-pixel short side. Blind accuracy averages four option permutations with seed 42. \\
Agents & Local tools on videos stored at 2 fps and at most 1,024 frames. \\
\bottomrule
\end{tabularx}
\end{table}

\section{Scoring and screening protocol}
\label{app:screen}\label{app:tables}
\paragraph{Scoring and error verification.}
The scorer applies the three-step rule in Section~\ref{sec:setup}, checking a leading letter, a stated \texttt{Answer: C}, and option text. We verify errors through a reasoning-trace rerun on identical frames, rule-based pre-classification, two independent judges, and a dense-frame evidence check. Each judge, including Claude Opus 5 \citep{anthropic2026opus5}, receives the same 32 hash-verified frames, question, options, gold answer, and incorrect answer with its reasoning. The responding model's identity is hidden. The checks distinguish annotation issues, insufficient frame coverage, and capability failures. Dense verification uses $\min(n_{\mathrm{frames}},128)$ frames, where $n_{\mathrm{frames}}$ is the stored count. A self-check shifts the frame grid by half an interval on a tenth of confirmed failures. Judge disagreement marks a disputed case, an offset-induced answer flip an unstable case, and a missing dense check an unverified case. The remaining cases form the denominator for verified error shares.
\paragraph{Capability attribution.}
The profiles in Section~\ref{sec:f10ability} use the same 18 fine categories as the item taxonomy in Section~\ref{sec:setup}. These comprise nine perception categories, four temporal categories, three spatial and physical categories, and two reasoning and knowledge categories. Only cases with agreement on a fine label within this set contribute to Figure~\ref{fig:f10ability}. Each share divides that category's count by the total across the 18 categories within the benchmark. The detailed tables retain other diagnostic labels, unresolved cases, and their original denominators.
\paragraph{Deduplication and replacement.}
The screen removes within-source near-duplicates and refills the audit sample toward $n=300$. Duplicate questions share a video hash, options, and answer, with cosine similarity at least 0.90 under bge-large-en-v1.5 \citep{xiao2024c}. We retain one item per connected component. Refill continues the seed-42 stratified draw, replacing items within their subtasks before filling residual gaps in the original proportions. It excludes duplicates and non-multiple-choice items. A second screen checks scoring and duplication in the replacements.
\paragraph{Screening results.}
Table~\ref{tab:attrib} reports error shares, including protocol failures and judge disagreement. Table~\ref{tab:screen} lists screening counts by benchmark. Across 115 benchmarks and 34,274 items, the screen removes 368 near-duplicates. Refill adds 287 items to the 39 benchmarks that fall below 300 after the screen. The frame-offset self-check flips 108 of 1,103 failures, or 9.8\%, and the attribution excludes these as unstable. In Table~\ref{tab:screen}, $n$ sampled counts every sampled item of a listed benchmark, and pool remaining counts the never-sampled items of the full annotation pool after the audio-visual filter. Pool remaining counts items before the multiple-choice filter and the chance audit. Benchmarks without multiple-choice items are not listed. Every appendix table counts multiple-choice items only, and $n$ is that count.
\begingroup
\begingroup
\setlength{\LTcapwidth}{\linewidth}
\setlength{\tabcolsep}{4pt}
\begin{xltabular}{\linewidth}{>{\raggedright\arraybackslash}p{103.24bp}>{\raggedleft\arraybackslash}p{43.48bp}>{\raggedleft\arraybackslash}p{54.20bp}>{\raggedleft\arraybackslash}p{41.30bp}>{\raggedleft\arraybackslash}p{45.83bp}>{\raggedleft\arraybackslash}p{37.10bp}>{\raggedleft\arraybackslash}p{43.86bp}>{\raggedleft\arraybackslash}p{24.57bp}}
\caption{Question-level error shares (\%), including protocol failures and judge disagreement.}\label{tab:attrib}\\
\toprule
\rowcolor{SkillPaleBlue} Benchmark & Percep. & Temporal & Spatial & Reason. & Proto. & Disput. & $n$ \\
\midrule
\endfirsthead
\multicolumn{8}{l}{\tablename~\thetable{} (continued)}\\
\toprule
\rowcolor{SkillPaleBlue} Benchmark & Percep. & Temporal & Spatial & Reason. & Proto. & Disput. & $n$ \\
\midrule
\endhead
\bottomrule
\endfoot
\mbox{AoTBench} & 15.6 & 13.3 & 0.0 & 71.1 & 0.0 & 18.8 & 45 \\
\mbox{Artifact-Bench} & 46.6 & 0.0 & 6.9 & 46.6 & 2.0 & 20.1 & 149 \\
\mbox{CMU} & 35.2 & 0.0 & 5.6 & 59.3 & 1.8 & 25.0 & 55 \\
\mbox{COVER} & 26.9 & 3.9 & 3.9 & 65.4 & 30.7 & 27.3 & 75 \\
\mbox{DSI-Bench} & 7.7 & 0.0 & 65.4 & 26.9 & 0.9 & 2.5 & 105 \\
\mbox{DSR-Bench} & 9.2 & 20.2 & 29.4 & 41.2 & 4.0 & 22.7 & 124 \\
\mbox{EgoSchema} & 20.0 & 2.9 & 0.0 & 77.1 & 22.2 & 25.8 & 45 \\
\mbox{ExAct} & 26.2 & 0.0 & 0.0 & 73.8 & 6.2 & 17.2 & 65 \\
\mbox{ExpVid} & 56.4 & 9.1 & 0.0 & 34.5 & 75.2 & 11.0 & 222 \\
\mbox{FAVOR-Bench} & 40.8 & 12.7 & 0.0 & 46.5 & 10.1 & 19.0 & 79 \\
\mbox{GasVideo-1000} & 25.4 & 4.8 & 3.2 & 66.7 & 71.2 & 9.6 & 219 \\
\mbox{HourVideo} & 9.2 & 15.3 & 13.7 & 61.8 & 27.6 & 17.9 & 181 \\
\mbox{IntPhys 2} & 10.1 & 0.0 & 72.5 & 17.4 & 18.8 & 17.1 & 85 \\
\mbox{LEMONADE} & 0.0 & 78.9 & 16.9 & 4.2 & 68.9 & 11.3 & 228 \\
\mbox{LongTimeScope} & 50.0 & 0.0 & 0.0 & 50.0 & 87.6 & 2.1 & 194 \\
\mbox{LongVideoBench} & 10.5 & 15.8 & 5.3 & 68.4 & 58.7 & 8.1 & 46 \\
\mbox{LVBench} & 11.5 & 13.1 & 4.9 & 70.5 & 40.2 & 17.8 & 102 \\
\mbox{MLVU} & 26.9 & 3.9 & 0.0 & 69.2 & 59.4 & 10.7 & 64 \\
\mbox{MMR-V} & 2.8 & 1.4 & 0.0 & 95.8 & 25.0 & 26.3 & 96 \\
\mbox{MMSI-Video-Bench} & 7.0 & 3.1 & 49.2 & 40.6 & 14.1 & 9.2 & 149 \\
\mbox{MMVU} & 6.8 & 0.0 & 13.6 & 79.5 & 26.7 & 22.2 & 60 \\
\mbox{MMWorld} & 14.3 & 8.6 & 0.0 & 77.1 & 25.5 & 11.5 & 47 \\
\mbox{MORSE-500} & 19.4 & 29.6 & 22.4 & 28.6 & 15.5 & 24.8 & 116 \\
\mbox{MotionBench} & 30.0 & 16.7 & 5.0 & 48.3 & 9.1 & 9.8 & 66 \\
\mbox{MVBench} & 25.4 & 20.0 & 16.4 & 38.2 & 8.3 & 20.4 & 60 \\
\mbox{Neptune} & 33.3 & 33.3 & 0.0 & 33.3 & 72.7 & 15.4 & 11 \\
\mbox{OSI-Bench} & 1.5 & 7.5 & 70.7 & 20.3 & 3.6 & 7.4 & 138 \\
\mbox{Perception Test} & 31.9 & 8.5 & 4.3 & 55.3 & 4.1 & 28.1 & 49 \\
\mbox{PerceptionComp} & 27.1 & 0.7 & 39.4 & 32.9 & 16.7 & 17.7 & 186 \\
\mbox{Q-Bench-Video} & 47.6 & 0.0 & 0.0 & 52.4 & 38.2 & 23.3 & 34 \\
\mbox{ReVSI} & 16.9 & 0.0 & 62.1 & 21.1 & 11.7 & 0.0 & 188 \\
\mbox{SAW-Bench} & 2.1 & 0.0 & 58.7 & 39.2 & 0.0 & 6.5 & 143 \\
\mbox{SciVideoBench} & 0.0 & 57.1 & 0.0 & 42.9 & 84.3 & 21.5 & 223 \\
\mbox{SFI-Bench} & 25.0 & 0.0 & 16.7 & 58.3 & 3.2 & 19.4 & 62 \\
\mbox{SIS-Bench} & 79.3 & 0.9 & 4.3 & 15.5 & 55.4 & 7.3 & 260 \\
\mbox{SITE-Bench} & 23.7 & 6.8 & 30.5 & 39.0 & 9.2 & 10.2 & 65 \\
\mbox{SoMi-ToM} & 33.3 & 0.0 & 0.0 & 66.7 & 0.0 & 0.0 & 3 \\
\mbox{SpatialTree} & 14.6 & 0.0 & 52.1 & 33.3 & 15.8 & 6.5 & 57 \\
\mbox{ST-Evidence} & 27.3 & 9.1 & 4.5 & 59.1 & 15.4 & 23.5 & 26 \\
\mbox{STI-Bench} & 1.6 & 6.2 & 33.3 & 58.9 & 25.9 & 24.6 & 174 \\
\mbox{StreamingBench} & 16.3 & 30.2 & 2.3 & 51.2 & 30.6 & 21.1 & 62 \\
\mbox{TempCompass} & 17.0 & 5.1 & 3.4 & 74.6 & 56.6 & 8.0 & 136 \\
\mbox{TemporalBench} & 26.3 & 2.6 & 7.9 & 63.2 & 5.0 & 20.0 & 40 \\
\mbox{TimeBlind} & 28.1 & 12.5 & 6.2 & 53.1 & 5.9 & 19.1 & 34 \\
\mbox{TimeScope} & 30.0 & 0.0 & 0.0 & 70.0 & 82.5 & 5.7 & 114 \\
\mbox{TOMATO} & 22.2 & 22.2 & 13.3 & 42.2 & 10.0 & 29.5 & 100 \\
\mbox{TVBench} & 11.9 & 40.7 & 3.4 & 44.1 & 13.2 & 20.4 & 68 \\
\mbox{v-HUB} & 55.0 & 0.0 & 6.2 & 38.8 & 50.3 & 14.6 & 161 \\
\mbox{V2P-Bench} & 18.5 & 7.4 & 7.4 & 66.7 & 48.6 & 16.7 & 105 \\
\mbox{Video-Holmes} & 4.3 & 26.1 & 0.0 & 69.6 & 77.9 & 17.1 & 208 \\
\mbox{Video-MME} & 18.8 & 12.5 & 0.0 & 68.8 & 39.6 & 13.9 & 53 \\
\mbox{Video-MME-v2} & 7.6 & 15.2 & 3.8 & 73.4 & 28.2 & 15.7 & 110 \\
\mbox{Video-MMMU} & 10.0 & 3.3 & 3.3 & 83.3 & 37.5 & 25.6 & 48 \\
\mbox{Video-TT} & 16.7 & 25.0 & 0.0 & 58.3 & 14.3 & 22.2 & 14 \\
\mbox{VideoEval-Pro} & 13.4 & 6.0 & 1.5 & 79.1 & 46.8 & 6.3 & 126 \\
\mbox{VideoKR} & 12.8 & 0.0 & 5.1 & 82.0 & 51.8 & 30.5 & 81 \\
\mbox{VideoMathQA} & 16.7 & 0.0 & 12.5 & 70.8 & 42.9 & 11.8 & 42 \\
\mbox{VideoOdyssey} & 7.1 & 7.1 & 7.1 & 78.6 & 50.9 & 10.3 & 57 \\
\mbox{VideoSIAH-Eval} & 48.4 & 9.7 & 0.0 & 41.9 & 82.4 & 4.6 & 176 \\
\mbox{VidPair-Halluc} & 8.5 & 6.8 & 1.7 & 83.0 & 31.4 & 17.8 & 86 \\
\mbox{ViMUL-Bench} & 39.1 & 4.3 & 4.3 & 52.2 & 48.9 & 12.8 & 45 \\
\mbox{Vinoground} & 3.5 & 20.7 & 3.5 & 72.4 & 0.0 & 33.3 & 29 \\
\mbox{VITATECS} & 25.0 & 8.3 & 0.0 & 66.7 & 14.3 & 0.0 & 14 \\
\mbox{VKnowU} & 9.8 & 0.0 & 14.6 & 75.6 & 29.3 & 12.5 & 58 \\
\mbox{VSI-Bench} & 8.9 & 2.1 & 69.2 & 19.9 & 7.6 & 2.8 & 158 \\
\mbox{VSI-Super-Wild} & 20.2 & 24.6 & 0.0 & 55.3 & 14.9 & 16.5 & 134 \\
\mbox{VURB} & 19.4 & 13.9 & 16.7 & 50.0 & 41.9 & 22.6 & 62 \\
\end{xltabular}
\endgroup
\endgroup

\begingroup
\begingroup
\setlength{\LTcapwidth}{\linewidth}
\setlength{\tabcolsep}{4pt}
\begin{xltabular}{\linewidth}{>{\raggedright\arraybackslash}p{121.16bp}>{\raggedleft\arraybackslash}p{67.94bp}>{\raggedleft\arraybackslash}p{88.21bp}>{\raggedleft\arraybackslash}p{48.79bp}>{\raggedleft\arraybackslash}p{91.48bp}}
\caption{Audit items, near-duplicate removals, and replacement pools by benchmark.}\label{tab:screen}\\
\toprule
\rowcolor{SkillPaleBlue} Benchmark & $n$ sampled & Dup. removed & $n$ after & Pool remaining \\
\midrule
\endfirsthead
\multicolumn{5}{l}{\tablename~\thetable{} (continued)}\\
\toprule
\rowcolor{SkillPaleBlue} Benchmark & $n$ sampled & Dup. removed & $n$ after & Pool remaining \\
\midrule
\endhead
\bottomrule
\endfoot
\mbox{CMU} & 300 & 0 & 300 & 3,853 \\
\mbox{COVER} & 300 & 1 & 299 & 5,546 \\
\mbox{DSI-Bench} & 299 & 0 & 299 & 1,470 \\
\mbox{DSR-Bench} & 299 & 0 & 299 & 1,151 \\
\mbox{EgoSchema} & 300 & 0 & 300 & 200 \\
\mbox{GasVideo-1000} & 650 & 72 & 578 & 363 \\
\mbox{LEMONADE} & 300 & 5 & 295 & 36,221 \\
\mbox{LongTimeScope} & 300 & 53 & 247 & 600 \\
\mbox{LongVideoBench} & 300 & 0 & 300 & 632 \\
\mbox{LVBench} & 300 & 1 & 299 & 1,249 \\
\mbox{MMWorld} & 300 & 1 & 299 & 1,259 \\
\mbox{MORSE-500} & 299 & 1 & 298 & 201 \\
\mbox{Neptune} & 300 & 19 & 281 & 8,500 \\
\mbox{OSI-Bench} & 299 & 0 & 299 & 8,467 \\
\mbox{PAI-Bench-U} & 300 & 1 & 299 & 914 \\
\mbox{Perception Test} & 300 & 1 & 299 & 18,840 \\
\mbox{SAW-Bench} & 299 & 10 & 289 & 3,843 \\
\mbox{SciVideoBench} & 300 & 0 & 300 & 700 \\
\mbox{SITE-Bench} & 300 & 0 & 300 & 3,319 \\
\mbox{SoMi-ToM} & 75 & 7 & 68 & 0 \\
\mbox{ST-Evidence} & 300 & 0 & 300 & 3,704 \\
\mbox{STI-Bench} & 299 & 13 & 286 & 3,829 \\
\mbox{TempCompass} & 300 & 9 & 291 & 7,240 \\
\mbox{TemporalBench} & 300 & 8 & 292 & 44,053 \\
\mbox{TimeBlind} & 300 & 1 & 299 & 2,100 \\
\mbox{TimeScope} & 300 & 14 & 286 & 2,700 \\
\mbox{TOMATO} & 300 & 0 & 300 & 1,184 \\
\mbox{TVBench} & 300 & 2 & 298 & 2,225 \\
\mbox{VCR-Bench} & 299 & 15 & 284 & 1,769 \\
\mbox{Video-Holmes} & 300 & 0 & 300 & 3,088 \\
\mbox{Video-MME-v2} & 300 & 0 & 300 & 2,590 \\
\mbox{Video-MMMU} & 300 & 0 & 300 & 900 \\
\mbox{Video-TT} & 300 & 6 & 294 & 9,700 \\
\mbox{VideoEval-Pro} & 300 & 1 & 299 & 989 \\
\mbox{VideoMathQA} & 300 & 0 & 300 & 1,800 \\
\mbox{VideoOdyssey} & 320 & 1 & 319 & 1,478 \\
\mbox{VidPair-Halluc} & 300 & 1 & 299 & 11,223 \\
\mbox{VKnowU} & 300 & 1 & 299 & 1,380 \\
\mbox{VSI-Bench} & 300 & 4 & 296 & 9,960 \\
\mbox{VSI-Super-Wild} & 300 & 3 & 297 & 6,680 \\
\mbox{VSTAT} & 344 & 6 & 338 & 1,156 \\
\mbox{VURB} & 300 & 76 & 224 & 1,800 \\
\midrule
\multicolumn{5}{l}{\emph{Second run batch}}\\
\mbox{CG-Bench} & 120 & 0 & 120 & 12,009 \\
\mbox{CineTechBench} & 127 & 0 & 127 & 0 \\
\mbox{FineBench} & 300 & 22 & 278 & 19,843 \\
\mbox{ForeSeaQA} & 300 & 4 & 296 & 291 \\
\mbox{HERBench} & 300 & 1 & 299 & 1,671 \\
\mbox{iSafetyBench} & 299 & 0 & 299 & 3,157 \\
\mbox{K9-Bench} & 299 & 0 & 299 & 4,445 \\
\mbox{OVO-S-Bench} & 300 & 8 & 292 & 1,332 \\
\mbox{ReactHuman} & 297 & 0 & 297 & 757 \\
\mbox{S-EMBER} & 299 & 0 & 299 & 9,149 \\
\mbox{VAU-Bench} & 251 & 0 & 251 & 0 \\
\midrule
\textbf{Displayed rows total} & 15,674 & 368 & 15,306 & 271,530 \\
\end{xltabular}
\endgroup
\endgroup

\section{Option-level attacks}
\label{app:t0}
\paragraph{Experiments.}
The option-level audit measures answer-position entropy, fixed-position accuracy, options-only accuracy, actual option counts, and sensitivity to option permutations. The final measure is the diagnostic model's best-to-worst accuracy range across reorderings.
\paragraph{Settings.}
We normalize ground-truth position entropy by the logarithm of the modal option count. The fixed-position attacker chooses the modal answer position. Claude Opus 5 \citep{anthropic2026opus5} receives \texttt{You will see ONLY the answer options of a multiple-choice question about a video.} A probe counts after covering 95\% of items with options. Chance uses the reciprocal of the actual option count, then lettered choices in the question, the declared count, or a count inferred from answer letters. The audit reports disagreement with declared counts. Qwen3-VL-8B \citep{bai2025qwen3} evaluates eight permutations with seed 42, 32 frames, and temperature 0, and the original ordering is the ninth run. We flag accuracy ranges above $2.326\sqrt{c(1-c)/n}$, where $c$ is chance accuracy and $n$ the number of items. Item consistency is the fraction of runs choosing the same answer. Blind lift at or below $-10\%$ flags possible answer-format artifacts.
\paragraph{Results.}
Table~\ref{tab:chance} compares actual option counts and chance levels with their declared values.
\begingroup
\begingroup
\setlength{\LTcapwidth}{\linewidth}
\setlength{\tabcolsep}{4pt}
\begin{xltabular}{\linewidth}{>{\raggedright\arraybackslash}p{110.24bp}>{\raggedleft\arraybackslash}p{44.19bp}>{\raggedleft\arraybackslash}p{50.69bp}>{\raggedleft\arraybackslash}p{61.48bp}>{\raggedleft\arraybackslash}p{73.44bp}>{\raggedright\arraybackslash}p{69.53bp}}
\caption{Declared option counts and chance levels that disagree with sampled benchmark items.}\label{tab:chance}\\
\toprule
\rowcolor{SkillPaleBlue} Benchmark & Decl. $k$ & Modal $k$ & Deviant \% & Impl. chance & MCQ no ch. \\
\midrule
\endfirsthead
\multicolumn{6}{l}{\tablename~\thetable{} (continued)}\\
\toprule
\rowcolor{SkillPaleBlue} Benchmark & Decl. $k$ & Modal $k$ & Deviant \% & Impl. chance & MCQ no ch. \\
\midrule
\endhead
\bottomrule
\endfoot
\mbox{ALLVB} & -- & 5 & 0.0 & 0.200 & yes \\
\mbox{Artifact-Bench} & NA & 2 & 0.0 & 0.363 & yes \\
\mbox{Causal2Needles} & -- & 4 & 0.0 & 0.250 & yes \\
\mbox{CAViAR} & -- & 5 & 0.0 & 0.321 & yes \\
\mbox{CG-Bench} & -- & 7 & 0.0 & 0.146 & yes \\
\mbox{CineTechBench} & -- & 4 & 0.0 & 0.250 & yes \\
\mbox{CMU} & 4 & 2 & 56.3 & 0.367 & no \\
\mbox{ContPhy} & -- & 2 & 0.0 & 0.396 & yes \\
\mbox{COVER} & -- & 4 & 0.0 & 0.309 & yes \\
\mbox{E.T. Bench} & -- & 4 & 0.0 & 0.233 & yes \\
\mbox{EgoPlan-Bench} & 4 & 4 & 0.0 & 0.250 & yes \\
\mbox{EgoSchema} & 4 & 5 & 100.0 & 0.200 & no \\
\mbox{ExpVid} & NA & 4 & 0.0 & 0.243 & yes \\
\mbox{FAVOR-Bench} & -- & 5 & 0.0 & 0.200 & yes \\
\mbox{FineBench} & -- & 4 & 0.0 & 0.250 & yes \\
\mbox{ForeSeaQA} & -- & 4 & 0.0 & 0.334 & yes \\
\mbox{GasVideo-1000} & NA & 3 & 0.0 & 0.301 & yes \\
\mbox{GUI-World} & -- & 4 & 0.0 & 0.250 & yes \\
\mbox{HERBench} & -- & 5 & 0.0 & 0.200 & yes \\
\mbox{HLV-1K} & -- & 4 & 0.0 & 0.250 & yes \\
\mbox{HourVideo} & -- & 5 & 0.0 & 0.200 & yes \\
\mbox{IntPhys 2} & -- & 2 & 0.0 & 0.500 & yes \\
\mbox{iSafetyBench} & -- & 16 & 0.0 & 0.062 & yes \\
\mbox{K9-Bench} & -- & 5 & 0.0 & 0.200 & yes \\
\mbox{KFS-Bench} & -- & 5 & 0.0 & 0.211 & yes \\
\mbox{LongVideo-Reason} & -- & 4 & 0.0 & 0.268 & yes \\
\mbox{LongVideoBench} & 5 & 5 & 27.3 & 0.214 & yes \\
\mbox{M3-Verse} & -- & 4 & 0.0 & 0.236 & yes \\
\mbox{MLVU} & -- & 4 & 0.0 & 0.250 & yes \\
\mbox{MME-VideoOCR} & -- & 4 & 0.0 & 0.250 & yes \\
\mbox{MMR-V} & NA & 10 & 0.0 & 0.096 & yes \\
\mbox{MMSI-Video-Bench} & -- & 4 & 0.0 & 0.242 & yes \\
\mbox{MMVU} & -- & 5 & 0.0 & 0.200 & yes \\
\mbox{MVBench} & -- & 4 & 0.0 & 0.287 & yes \\
\mbox{OVO-S-Bench} & -- & 4 & 0.0 & 0.321 & yes \\
\mbox{Perception Test} & -- & 3 & 0.0 & 0.333 & yes \\
\mbox{PLM-VideoBench} & 2 & 2 & 0.0 & 0.500 & yes \\
\mbox{ReactHuman} & -- & 3 & 0.0 & 0.333 & yes \\
\mbox{ReXTime} & -- & 4 & 0.0 & 0.250 & yes \\
\mbox{S-EMBER} & -- & 5 & 0.0 & 0.200 & yes \\
\mbox{SAW-Bench} & 3 & 4 & 90.3 & 0.278 & no \\
\mbox{ScaleLong} & -- & 4 & 0.0 & 0.250 & yes \\
\mbox{SciVideoBench} & 5 & 10 & 100.0 & 0.100 & yes \\
\mbox{SIS-Bench} & -- & 4 & 0.0 & 0.250 & yes \\
\mbox{SITE-Bench} & 4 & 4 & 27.3 & 0.260 & yes \\
\mbox{SLVMBench} & -- & 5 & 0.0 & 0.200 & yes \\
\mbox{SmartHome-Bench} & -- & 2 & 0.0 & 0.500 & yes \\
\mbox{ST-Evidence} & 5 & 2 & 100.0 & 0.475 & yes \\
\mbox{STI-Bench} & -- & 5 & 0.0 & 0.200 & yes \\
\mbox{StreamingBench} & -- & 4 & 0.0 & 0.249 & yes \\
\mbox{TaskMeAnything-v1} & -- & 4 & 0.0 & 0.253 & yes \\
\mbox{TOC-Bench} & -- & 4 & 0.0 & 0.321 & yes \\
\mbox{TOMATO} & 4 & 6 & 83.0 & 0.199 & no \\
\mbox{TVBench} & -- & 4 & 0.0 & 0.334 & yes \\
\mbox{UCF101-AD} & -- & 11 & 0.0 & 0.091 & yes \\
\mbox{UrbanVideo-Bench} & -- & 5 & 0.0 & 0.210 & yes \\
\mbox{VANE-Bench} & -- & 4 & 0.0 & 0.250 & yes \\
\mbox{VAU-Bench} & -- & 4 & 0.0 & 0.250 & yes \\
\mbox{VELOCITI} & -- & 2 & 0.0 & 0.500 & yes \\
\mbox{VF-Eval} & -- & 4 & 0.0 & 0.250 & yes \\
\mbox{VGenST-Bench} & -- & 4 & 0.0 & 0.267 & yes \\
\mbox{Video-Holmes} & 5 & 6 & 100.0 & 0.167 & no \\
\mbox{Video-LevelGauge} & -- & 4 & 0.0 & 0.250 & yes \\
\mbox{Video-MME} & 4 & 4 & 0.0 & 0.250 & yes \\
\mbox{Video-MME-v2} & 5 & 8 & 100.0 & 0.127 & yes \\
\mbox{Video-MMMU} & 4 & 10 & 94.9 & 0.116 & no \\
\mbox{VideoAds} & -- & 4 & 0.0 & 0.250 & yes \\
\mbox{VideoEspresso} & -- & 4 & 0.0 & 0.250 & yes \\
\mbox{VideoEval-Pro} & -- & 4 & 0.0 & 0.246 & yes \\
\mbox{VideoKR} & NA & 10 & 0.0 & 0.140 & yes \\
\mbox{VideoMathQA} & 5 & 2 & 81.7 & 0.445 & yes \\
\mbox{VidPair-Halluc} & 2 & 3 & 100.0 & 0.333 & no \\
\mbox{ViMUL-Bench} & 4 & 4 & 8.1 & 0.257 & yes \\
\mbox{Vinoground} & -- & 2 & 0.0 & 0.500 & yes \\
\mbox{VITATECS} & -- & 2 & 0.0 & 0.500 & yes \\
\mbox{VKnowU} & NA & 2 & 0.0 & 0.378 & yes \\
\mbox{VNBench} & -- & 4 & 0.0 & 0.250 & yes \\
\mbox{VRU-Accident} & -- & 4 & 0.0 & 0.250 & yes \\
\mbox{VSI-Super-Wild} & NA & 4 & 0.0 & 0.250 & yes \\
\end{xltabular}
\endgroup
\endgroup

\section{Text-level attacks}
\label{app:t1}
\paragraph{Experiments.}
Gemini 3.5 Flash \citep{google2026gemini35flash} and Claude Opus 5 \citep{anthropic2026opus5} answer each question from text and options alone. Video dependence is the reference advantage over the stronger reader, $m=s^{*}-c-\varepsilon_{\mathrm{text}}$. We apply a paired McNemar test to items answered both blind and with 32 frames by the same model. Fewer than 25 discordant pairs use an exact two-sided binomial test. Larger counts use a chi-square test. Tables~\ref{tab:e1}--\ref{tab:e1_cols2} report condition accuracies, and Table~\ref{tab:mcnemar} gives the paired tests.
\paragraph{Settings.}
Both readers receive independent calls without shared context. The prompt is \texttt{You will answer a question about a video WITHOUT seeing the video. Guess the most likely answer from the question and options alone.} It then supplies the question, lettered options, and a request for the option letter. Items without options request the exact short answer. Extraction skips Gemini's thinking text and applies the scorer in the appendix.
\paragraph{Results.}
Coverage is the share of multiple-choice items with a scored answer under the reference model's 32-frame condition.
\begingroup

\begingroup
\setlength{\LTcapwidth}{\linewidth}
\setlength{\tabcolsep}{4pt}
\begin{xltabular}{\linewidth}{>{\raggedright\arraybackslash}p{100.70bp}>{\raggedleft\arraybackslash}p{87.76bp}>{\raggedleft\arraybackslash}p{87.77bp}>{\raggedleft\arraybackslash}p{87.77bp}}
\caption{Input-condition accuracy (\%) across audited multiple-choice benchmarks (1/2).}\label{tab:e1}\\
\toprule
\rowcolor{SkillPaleBlue} Benchmark & blind & 1f & 32f \\
\midrule
\endfirsthead
\multicolumn{4}{l}{\tablename~\thetable{} (continued)}\\
\toprule
\rowcolor{SkillPaleBlue} Benchmark & blind & 1f & 32f \\
\midrule
\endhead
\bottomrule
\endfoot
\mbox{ALLVB} & 73.3 & 70.3 & -- \\
\mbox{AoTBench} & 49.1 & 49.8 & 76.5 \\
\mbox{Artifact-Bench} & 8.1 & 13.8 & 8.9 \\
\mbox{Causal2Needles} & 57.3 & 58.7 & -- \\
\mbox{CAViAR} & 45.3 & 72.7 & -- \\
\mbox{CG-Bench} & 47.5 & 51.7 & -- \\
\mbox{CineTechBench} & 40.2 & 50.4 & -- \\
\mbox{CMU} & 45.8 & 58.7 & 66.9 \\
\mbox{Colon-Bench} & 54.3 & 63.7 & -- \\
\mbox{ContPhy} & 33.1 & 47.2 & -- \\
\mbox{COVER} & 57.2 & 56.8 & 59.2 \\
\mbox{DSI-Bench} & 51.2 & 53.5 & 52.7 \\
\mbox{DSR-Bench} & 24.1 & 30.8 & 23.9 \\
\mbox{E.T. Bench} & 28.7 & 32.7 & -- \\
\mbox{EgoPlan-Bench} & 33.0 & 28.0 & -- \\
\mbox{EgoSchema} & 56.7 & 62.7 & 77.0 \\
\mbox{ExAct} & 81.3 & 79.3 & 69.2 \\
\mbox{ExpVid} & 52.1 & 64.5 & -- \\
\mbox{FAVOR-Bench} & 55.0 & 50.0 & 55.3 \\
\mbox{FineBench} & 63.3 & 57.4 & -- \\
\mbox{ForeSeaQA} & 70.9 & 71.4 & -- \\
\mbox{GasVideo-1000} & 44.9 & 57.0 & 75.4 \\
\mbox{GUI-World} & 81.0 & 82.7 & -- \\
\mbox{HERBench} & 48.0 & 49.7 & -- \\
\mbox{HLV-1K} & 67.0 & 65.5 & -- \\
\mbox{HourVideo} & 22.7 & 20.3 & 21.2 \\
\mbox{IntPhys 2} & 48.7 & 47.7 & 57.0 \\
\mbox{iSafetyBench} & 14.7 & 36.1 & -- \\
\mbox{K9-Bench} & 65.5 & 63.9 & -- \\
\mbox{KFS-Bench} & 51.3 & 52.3 & -- \\
\mbox{LEMONADE} & 41.7 & 32.3 & -- \\
\mbox{LongTimeScope} & 29.7 & 36.3 & 15.7 \\
\mbox{LongVideo-Reason} & 76.3 & 73.2 & -- \\
\mbox{LongVideoBench} & 52.7 & 56.3 & 60.4 \\
\mbox{LVBench} & 37.3 & 38.3 & 45.6 \\
\mbox{M3-Verse} & 26.3 & 33.7 & -- \\
\mbox{MLVU} & 57.0 & 56.7 & 64.6 \\
\mbox{MME-VideoOCR} & 42.7 & 57.7 & -- \\
\mbox{MMR-V} & 56.0 & 53.0 & 55.6 \\
\mbox{MMSI-Video-Bench} & 28.7 & 30.7 & 28.3 \\
\mbox{MMVU} & 57.5 & 78.7 & 76.9 \\
\mbox{MMWorld} & 77.8 & 83.5 & 74.4 \\
\mbox{MORSE-500} & 0.0 & 12.0 & 41.0 \\
\mbox{MotionBench} & 50.3 & 59.0 & 64.9 \\
\mbox{MVBench} & 46.0 & 53.7 & 68.9 \\
\mbox{NaturalBench-Video} & 50.7 & 54.7 & -- \\
\mbox{Neptune} & 72.2 & 70.6 & -- \\
\mbox{NExT-QA} & 59.0 & 75.3 & -- \\
\mbox{OSI-Bench} & 38.5 & 36.3 & 47.8 \\
\mbox{OVO-S-Bench} & 40.4 & 41.8 & -- \\
\mbox{PAI-Bench-U} & 34.2 & 49.7 & -- \\
\mbox{Perception Test} & 45.0 & 62.0 & 74.7 \\
\mbox{PerceptionComp} & 7.4 & 8.4 & 13.5 \\
\mbox{PLM-VideoBench} & 71.7 & 82.0 & -- \\
\mbox{Q-Bench-Video} & 48.1 & 63.2 & 72.6 \\
\mbox{ReactHuman} & 31.3 & 40.4 & -- \\
\mbox{ReVSI} & 39.3 & 29.5 & 52.1 \\
\mbox{ReXTime} & 84.3 & 87.7 & -- \\
\mbox{S-EMBER} & 29.4 & 21.7 & -- \\
\mbox{SAW-Bench} & 28.0 & 25.4 & 38.1 \\
\mbox{ScaleLong} & 48.7 & 52.5 & -- \\
\mbox{SciVideoBench} & 54.0 & 52.0 & -- \\
\mbox{SFI-Bench} & 62.3 & 56.7 & 68.9 \\
\mbox{SIS-Bench} & 50.0 & 58.3 & -- \\
\mbox{SITE-Bench} & 50.2 & 45.8 & 65.3 \\
\mbox{SLVMBench} & 63.0 & 64.7 & -- \\
\mbox{SmartHome-Bench} & 63.0 & 71.3 & -- \\
\mbox{SoMi-ToM} & 46.7 & 50.0 & 93.3 \\
\mbox{SpatialTree} & 69.2 & 62.5 & 62.2 \\
\mbox{ST-Evidence} & 66.2 & 65.6 & 82.1 \\
\mbox{STI-Bench} & 47.2 & 49.2 & -- \\
\mbox{StreamingBench} & 56.0 & 54.3 & 61.9 \\
\mbox{TaskMeAnything-v1} & 32.0 & 41.3 & -- \\
\mbox{TempCompass} & 27.8 & 30.9 & 42.4 \\
\mbox{TemporalBench} & 76.4 & 75.0 & 77.0 \\
\mbox{TimeBlind} & 47.5 & 49.2 & 70.9 \\
\mbox{TimeScope} & 40.3 & 42.3 & 52.6 \\
\mbox{TOC-Bench} & 35.0 & 41.1 & -- \\
\mbox{TOMATO} & 19.0 & 22.3 & 43.4 \\
\mbox{TVBench} & 40.7 & 44.0 & 62.8 \\
\mbox{UCF101-AD} & 7.3 & 35.7 & -- \\
\mbox{UrbanVideo-Bench} & 41.3 & 41.3 & -- \\
\mbox{v-HUB} & 20.8 & 56.1 & 64.3 \\
\mbox{V2P-Bench} & 41.0 & 46.7 & 52.7 \\
\mbox{VANE-Bench} & 77.3 & 87.3 & -- \\
\mbox{VAU-Bench} & 96.0 & 94.8 & -- \\
\mbox{VCR-Bench} & 45.1 & 47.0 & -- \\
\mbox{VELOCITI} & 65.3 & 75.3 & -- \\
\mbox{VF-Eval} & 83.3 & 81.3 & -- \\
\mbox{VGenST-Bench} & 45.3 & 60.7 & -- \\
\mbox{Video-Holmes} & 54.3 & 58.0 & -- \\
\mbox{Video-LevelGauge} & 52.7 & 64.7 & -- \\
\mbox{Video-MME} & 56.7 & 56.7 & 70.2 \\
\mbox{Video-MME-v2} & 29.3 & 28.0 & 36.4 \\
\mbox{Video-MMMU} & 56.2 & 59.2 & 70.5 \\
\mbox{Video-TT} & 40.0 & 40.0 & 60.7 \\
\mbox{VideoAds} & 51.0 & 57.3 & -- \\
\mbox{VideoEspresso} & 75.0 & 79.3 & -- \\
\mbox{VideoEval-Pro} & 35.3 & 37.3 & 45.0 \\
\mbox{VideoKR} & 35.1 & 43.2 & 55.2 \\
\mbox{VideoMathQA} & 57.3 & 63.0 & 51.7 \\
\mbox{VideoOdyssey} & 44.7 & 45.3 & 34.8 \\
\mbox{VideoSIAH-Eval} & 6.9 & 8.7 & 12.8 \\
\mbox{VidPair-Halluc} & 29.8 & 41.3 & 42.7 \\
\mbox{ViMUL-Bench} & 43.4 & 62.4 & 80.1 \\
\mbox{Vinoground} & 50.0 & 54.3 & 82.2 \\
\mbox{VITATECS} & 78.3 & 89.3 & 92.0 \\
\mbox{VKnowU} & 63.3 & 60.6 & 71.5 \\
\mbox{VNBench} & 26.0 & 25.4 & -- \\
\mbox{VRU-Accident} & 48.3 & 67.3 & -- \\
\mbox{VSI-Bench} & 39.0 & 29.4 & 60.8 \\
\mbox{VSI-Super-Wild} & 25.4 & 31.4 & 32.5 \\
\mbox{VSTAT} & 43.8 & 42.2 & -- \\
\mbox{VURB} & 38.4 & 42.2 & 62.4 \\
\end{xltabular}
\endgroup

\begingroup
\setlength{\LTcapwidth}{\linewidth}
\setlength{\tabcolsep}{4pt}
\begin{xltabular}{\linewidth}{>{\raggedright\arraybackslash}p{128.20bp}>{\raggedleft\arraybackslash}p{60.85bp}>{\raggedleft\arraybackslash}p{95.80bp}>{\raggedleft\arraybackslash}p{82.18bp}>{\raggedleft\arraybackslash}p{50.55bp}}
\caption{Input-condition accuracy (\%) across audited multiple-choice benchmarks (2/2).}\label{tab:e1_cols2}\\
\toprule
\rowcolor{SkillPaleBlue} Benchmark & 32f & video cap. & Frame cap. & Cov. \\
\midrule
\endfirsthead
\multicolumn{5}{l}{\tablename~\thetable{} (continued)}\\
\toprule
\rowcolor{SkillPaleBlue} Benchmark & 32f & video cap. & Frame cap. & Cov. \\
\midrule
\endhead
\bottomrule
\endfoot
\mbox{ALLVB} & 72.7 & 71.0 & 72.0 & 1.00 \\
\mbox{AoTBench} & 80.0 & 72.2 & 70.5 & 0.98 \\
\mbox{Artifact-Bench} & 17.1 & 11.4 & 13.8 & 1.00 \\
\mbox{Causal2Needles} & 63.7 & 59.7 & 60.3 & 1.00 \\
\mbox{CAViAR} & 77.7 & 75.0 & 71.3 & 1.00 \\
\mbox{CG-Bench} & 56.7 & 47.5 & 52.5 & 1.00 \\
\mbox{CineTechBench} & 60.2 & 54.9 & 56.6 & 0.89 \\
\mbox{CMU} & 72.3 & 56.8 & 58.1 & 0.99 \\
\mbox{Colon-Bench} & 74.3 & 62.0 & 54.3 & 1.00 \\
\mbox{ContPhy} & 55.8 & 33.7 & 33.7 & 1.00 \\
\mbox{COVER} & 62.9 & 59.8 & 60.3 & 1.00 \\
\mbox{DSI-Bench} & 63.9 & 59.9 & 55.2 & 1.00 \\
\mbox{DSR-Bench} & 29.1 & 34.1 & 30.8 & 0.78 \\
\mbox{E.T. Bench} & 54.7 & 51.7 & 48.3 & 1.00 \\
\mbox{EgoPlan-Bench} & 29.3 & 30.0 & 32.7 & 1.00 \\
\mbox{EgoSchema} & 78.3 & 80.3 & 71.0 & 0.94 \\
\mbox{ExAct} & 83.7 & 79.0 & 82.0 & 0.96 \\
\mbox{ExpVid} & 64.5 & 62.7 & 59.0 & 1.00 \\
\mbox{FAVOR-Bench} & 67.0 & 54.3 & 52.3 & 0.98 \\
\mbox{FineBench} & 61.9 & 62.6 & 60.7 & 0.52 \\
\mbox{ForeSeaQA} & 73.4 & 68.8 & 69.3 & 1.00 \\
\mbox{GasVideo-1000} & 75.4 & 72.7 & 62.2 & 0.97 \\
\mbox{GUI-World} & 92.0 & 88.3 & 86.3 & 1.00 \\
\mbox{HERBench} & 55.7 & 54.7 & 54.3 & 1.00 \\
\mbox{HLV-1K} & 68.9 & 69.9 & 69.6 & 0.99 \\
\mbox{HourVideo} & 24.0 & 24.0 & 22.0 & 0.96 \\
\mbox{IntPhys 2} & 64.3 & 55.3 & 52.3 & 0.99 \\
\mbox{iSafetyBench} & 42.1 & 42.8 & 36.4 & 1.00 \\
\mbox{K9-Bench} & 74.8 & 68.7 & 67.0 & 0.98 \\
\mbox{KFS-Bench} & 67.3 & 61.7 & 60.0 & 1.00 \\
\mbox{LEMONADE} & 40.0 & 37.7 & 38.0 & 1.00 \\
\mbox{LongTimeScope} & 33.3 & 27.3 & 28.0 & 0.89 \\
\mbox{LongVideo-Reason} & 77.3 & 79.0 & 76.6 & 0.98 \\
\mbox{LongVideoBench} & 76.3 & 69.7 & 63.0 & 0.93 \\
\mbox{LVBench} & 58.7 & 47.0 & 46.7 & 0.95 \\
\mbox{M3-Verse} & 40.3 & 34.0 & 29.3 & 1.00 \\
\mbox{MLVU} & 74.7 & 68.7 & 66.3 & 0.93 \\
\mbox{MME-VideoOCR} & 84.0 & 67.7 & 52.3 & 1.00 \\
\mbox{MMR-V} & 61.0 & 58.4 & 57.0 & 0.92 \\
\mbox{MMSI-Video-Bench} & 35.3 & 35.0 & 27.0 & 0.92 \\
\mbox{MMVU} & 89.9 & 83.5 & 78.2 & 0.97 \\
\mbox{MMWorld} & 83.5 & 80.8 & 80.8 & 0.96 \\
\mbox{MORSE-500} & 38.0 & 24.1 & 10.2 & 0.88 \\
\mbox{MotionBench} & 69.3 & 63.7 & 56.3 & 0.99 \\
\mbox{MVBench} & 67.7 & 66.3 & 63.0 & 0.96 \\
\mbox{NaturalBench-Video} & 80.7 & 66.7 & 60.0 & 1.00 \\
\mbox{Neptune} & 81.3 & 75.6 & -- & -- \\
\mbox{NExT-QA} & 83.3 & 79.3 & 78.3 & -- \\
\mbox{OSI-Bench} & 46.2 & 41.8 & 37.4 & 0.99 \\
\mbox{OVO-S-Bench} & 46.9 & 48.0 & 43.6 & 1.00 \\
\mbox{PAI-Bench-U} & 60.7 & 56.1 & 42.6 & 1.00 \\
\mbox{Perception Test} & 77.3 & 71.3 & 65.7 & 0.99 \\
\mbox{PerceptionComp} & 16.4 & 9.4 & 8.0 & 0.89 \\
\mbox{PLM-VideoBench} & 84.8 & 79.8 & 75.8 & 0.92 \\
\mbox{Q-Bench-Video} & 59.4 & 46.2 & 53.8 & 1.00 \\
\mbox{ReactHuman} & 50.2 & 51.2 & 45.8 & 1.00 \\
\mbox{ReVSI} & 54.1 & 40.2 & 34.4 & 0.98 \\
\mbox{ReXTime} & 91.8 & 89.8 & 91.0 & 0.81 \\
\mbox{S-EMBER} & 37.1 & 24.1 & 28.4 & 1.00 \\
\mbox{SAW-Bench} & 32.7 & 32.4 & 32.7 & 0.98 \\
\mbox{ScaleLong} & 64.0 & 59.9 & 57.6 & 0.99 \\
\mbox{SciVideoBench} & 65.0 & 62.0 & 57.7 & 1.00 \\
\mbox{SFI-Bench} & 74.3 & 60.7 & 61.7 & 0.99 \\
\mbox{SIS-Bench} & 72.0 & 69.0 & 61.7 & 1.00 \\
\mbox{SITE-Bench} & 68.1 & 64.9 & 59.9 & 0.99 \\
\mbox{SLVMBench} & 79.7 & 73.0 & 73.3 & 1.00 \\
\mbox{SmartHome-Bench} & 78.0 & 71.3 & 66.7 & 1.00 \\
\mbox{SoMi-ToM} & 100.0 & 85.0 & 70.0 & 0.80 \\
\mbox{SpatialTree} & 70.3 & 59.4 & 64.1 & 0.92 \\
\mbox{ST-Evidence} & 84.0 & 80.8 & 77.2 & 0.95 \\
\mbox{STI-Bench} & 48.8 & 48.2 & 44.5 & 1.00 \\
\mbox{StreamingBench} & 69.0 & 62.0 & 59.7 & 0.97 \\
\mbox{TaskMeAnything-v1} & 61.7 & 49.7 & 44.0 & 1.00 \\
\mbox{TempCompass} & 39.7 & 39.7 & 34.1 & 0.99 \\
\mbox{TemporalBench} & 79.6 & 76.1 & 78.5 & 0.99 \\
\mbox{TimeBlind} & 82.3 & 66.3 & 60.8 & 0.61 \\
\mbox{TimeScope} & 64.0 & 57.7 & 50.7 & 0.95 \\
\mbox{TOC-Bench} & 49.6 & 46.8 & 46.8 & 1.00 \\
\mbox{TOMATO} & 47.0 & 29.8 & 29.3 & 0.98 \\
\mbox{TVBench} & 66.0 & 56.3 & 49.7 & 0.99 \\
\mbox{UCF101-AD} & 46.7 & 48.7 & 34.3 & 1.00 \\
\mbox{UrbanVideo-Bench} & 54.0 & 44.0 & 49.0 & 1.00 \\
\mbox{v-HUB} & 74.6 & 70.8 & 67.7 & 0.97 \\
\mbox{V2P-Bench} & 58.0 & 53.0 & 54.7 & 0.98 \\
\mbox{VANE-Bench} & 88.3 & 82.0 & 81.0 & 1.00 \\
\mbox{VAU-Bench} & 94.4 & 95.2 & 96.4 & 1.00 \\
\mbox{VCR-Bench} & 66.3 & 57.7 & 51.7 & 0.99 \\
\mbox{VELOCITI} & 93.7 & 93.0 & 88.0 & 1.00 \\
\mbox{VF-Eval} & 84.0 & 81.7 & 79.3 & 1.00 \\
\mbox{VGenST-Bench} & 87.7 & 80.7 & 70.7 & 1.00 \\
\mbox{Video-Holmes} & 67.7 & 62.7 & 56.0 & 1.00 \\
\mbox{Video-LevelGauge} & 89.0 & 74.0 & 67.7 & 1.00 \\
\mbox{Video-MME} & 80.0 & 70.7 & 65.0 & 0.96 \\
\mbox{Video-MME-v2} & 48.0 & 39.0 & 35.3 & 0.84 \\
\mbox{Video-MMMU} & 80.6 & 69.6 & 54.9 & 0.91 \\
\mbox{Video-TT} & 60.0 & 45.0 & 35.0 & 0.93 \\
\mbox{VideoAds} & 81.3 & 75.3 & 76.3 & 1.00 \\
\mbox{VideoEspresso} & 79.6 & 76.7 & 77.8 & 0.92 \\
\mbox{VideoEval-Pro} & 56.7 & 46.3 & 47.7 & 0.96 \\
\mbox{VideoKR} & 67.0 & 56.8 & 44.9 & 0.88 \\
\mbox{VideoMathQA} & 80.7 & 69.0 & 59.0 & 0.69 \\
\mbox{VideoOdyssey} & 50.0 & 49.4 & 47.8 & 0.93 \\
\mbox{VideoSIAH-Eval} & 12.1 & 10.8 & 12.1 & 0.94 \\
\mbox{VidPair-Halluc} & 46.2 & 42.3 & 41.3 & 0.94 \\
\mbox{ViMUL-Bench} & 82.1 & 74.6 & 66.5 & 0.93 \\
\mbox{Vinoground} & 86.3 & 78.7 & 76.0 & 0.98 \\
\mbox{VITATECS} & 93.3 & 88.7 & 88.7 & 0.96 \\
\mbox{VKnowU} & 76.1 & 77.3 & 72.7 & 0.97 \\
\mbox{VNBench} & 73.9 & 60.2 & 51.8 & 1.00 \\
\mbox{VRU-Accident} & 76.0 & 74.7 & 64.7 & 1.00 \\
\mbox{VSI-Bench} & 59.6 & 44.5 & 53.4 & 0.89 \\
\mbox{VSI-Super-Wild} & 32.9 & 31.4 & 25.0 & 0.76 \\
\mbox{VSTAT} & 48.6 & 49.6 & 48.9 & -- \\
\mbox{VURB} & 63.1 & 58.2 & 50.5 & 0.91 \\
\end{xltabular}
\endgroup

\endgroup

\begingroup
\begingroup
\setlength{\LTcapwidth}{\linewidth}
\setlength{\tabcolsep}{4pt}
\begin{xltabular}{\linewidth}{>{\raggedright\arraybackslash}p{143.55bp}>{\raggedleft\arraybackslash}p{77.71bp}>{\raggedleft\arraybackslash}p{60.89bp}>{\raggedleft\arraybackslash}p{60.89bp}>{\raggedleft\arraybackslash}p{74.53bp}}
\caption{Paired blind versus 32-frame McNemar tests, where $b$ and $c$ count discordant pairs.}\label{tab:mcnemar}\\
\toprule
\rowcolor{SkillPaleBlue} Benchmark & $n$ pairs & $b$ & $c$ & $p$ \\
\midrule
\endfirsthead
\multicolumn{5}{l}{\tablename~\thetable{} (continued)}\\
\toprule
\rowcolor{SkillPaleBlue} Benchmark & $n$ pairs & $b$ & $c$ & $p$ \\
\midrule
\endhead
\bottomrule
\endfoot
\mbox{ALLVB} & 300 & 30 & 28 & 0.8955 \\
\mbox{AoTBench} & 290 & 21 & 106 & 0.0000 \\
\mbox{Artifact-Bench} & 123 & 6 & 10 & 0.4545 \\
\mbox{Causal2Needles} & 300 & 35 & 54 & 0.0564 \\
\mbox{CAViAR} & 300 & 15 & 112 & 0.0000 \\
\mbox{CG-Bench} & 120 & 13 & 24 & 0.1002 \\
\mbox{CineTechBench} & 113 & 15 & 38 & 0.0025 \\
\mbox{CMU} & 154 & 18 & 61 & 0.0000 \\
\mbox{Colon-Bench} & 300 & 28 & 88 & 0.0000 \\
\mbox{ContPhy} & 163 & 19 & 56 & 0.0000 \\
\mbox{COVER} & 228 & 35 & 54 & 0.0564 \\
\mbox{DSI-Bench} & 298 & 56 & 59 & 0.8521 \\
\mbox{DSR-Bench} & 234 & 40 & 34 & 0.5611 \\
\mbox{E.T. Bench} & 300 & 31 & 109 & 0.0000 \\
\mbox{EgoPlan-Bench} & 100 & 25 & 20 & 0.5510 \\
\mbox{EgoSchema} & 282 & 15 & 120 & 0.0000 \\
\mbox{ExAct} & 289 & 26 & 42 & 0.0689 \\
\mbox{FAVOR-Bench} & 293 & 45 & 73 & 0.0129 \\
\mbox{FineBench} & 155 & 23 & 32 & 0.2807 \\
\mbox{ForeSeaQA} & 199 & 17 & 22 & 0.5218 \\
\mbox{GasVideo-1000} & 382 & 26 & 136 & 0.0000 \\
\mbox{GUI-World} & 300 & 6 & 39 & 0.0000 \\
\mbox{HERBench} & 300 & 31 & 54 & 0.0170 \\
\mbox{HLV-1K} & 296 & 33 & 38 & 0.6350 \\
\mbox{HourVideo} & 288 & 41 & 36 & 0.6485 \\
\mbox{IntPhys 2} & 298 & 103 & 118 & 0.3463 \\
\mbox{iSafetyBench} & 299 & 17 & 99 & 0.0000 \\
\mbox{K9-Bench} & 294 & 27 & 54 & 0.0039 \\
\mbox{KFS-Bench} & 300 & 29 & 77 & 0.0000 \\
\mbox{LongTimeScope} & 267 & 71 & 24 & 0.0000 \\
\mbox{LongVideo-Reason} & 295 & 24 & 28 & 0.6774 \\
\mbox{LongVideoBench} & 280 & 50 & 76 & 0.0259 \\
\mbox{LVBench} & 285 & 41 & 88 & 0.0001 \\
\mbox{M3-Verse} & 300 & 33 & 75 & 0.0001 \\
\mbox{MLVU} & 280 & 45 & 87 & 0.0004 \\
\mbox{MME-VideoOCR} & 300 & 11 & 135 & 0.0000 \\
\mbox{MMR-V} & 277 & 32 & 57 & 0.0110 \\
\mbox{MMSI-Video-Bench} & 276 & 57 & 52 & 0.7016 \\
\mbox{MMVU} & 182 & 19 & 54 & 0.0001 \\
\mbox{MMWorld} & 285 & 27 & 70 & 0.0000 \\
\mbox{MORSE-500} & 95 & 4 & 33 & 0.0000 \\
\mbox{MotionBench} & 296 & 30 & 109 & 0.0000 \\
\mbox{MVBench} & 289 & 26 & 114 & 0.0000 \\
\mbox{NaturalBench-Video} & 150 & 10 & 55 & 0.0000 \\
\mbox{OSI-Bench} & 90 & 7 & 24 & 0.0041 \\
\mbox{OVO-S-Bench} & 275 & 36 & 54 & 0.0731 \\
\mbox{Perception Test} & 296 & 25 & 106 & 0.0000 \\
\mbox{PerceptionComp} & 267 & 12 & 31 & 0.0061 \\
\mbox{PLM-VideoBench} & 277 & 16 & 52 & 0.0000 \\
\mbox{Q-Bench-Video} & 106 & 12 & 42 & 0.0001 \\
\mbox{ReactHuman} & 297 & 21 & 77 & 0.0000 \\
\mbox{ReVSI} & 118 & 12 & 36 & 0.0009 \\
\mbox{ReXTime} & 244 & 6 & 23 & 0.0030 \\
\mbox{S-EMBER} & 299 & 38 & 61 & 0.0270 \\
\mbox{SAW-Bench} & 270 & 45 & 70 & 0.0252 \\
\mbox{ScaleLong} & 297 & 27 & 72 & 0.0000 \\
\mbox{SFI-Bench} & 296 & 27 & 68 & 0.0000 \\
\mbox{SITE-Bench} & 297 & 44 & 110 & 0.0000 \\
\mbox{SLVMBench} & 300 & 18 & 68 & 0.0000 \\
\mbox{SmartHome-Bench} & 300 & 47 & 92 & 0.0002 \\
\mbox{SoMi-ToM} & 60 & 1 & 27 & 0.0000 \\
\mbox{SpatialTree} & 116 & 13 & 26 & 0.0547 \\
\mbox{ST-Evidence} & 268 & 21 & 77 & 0.0000 \\
\mbox{StreamingBench} & 291 & 43 & 88 & 0.0001 \\
\mbox{TaskMeAnything-v1} & 300 & 19 & 108 & 0.0000 \\
\mbox{TempCompass} & 125 & 2 & 27 & 0.0000 \\
\mbox{TemporalBench} & 282 & 28 & 43 & 0.0966 \\
\mbox{TimeBlind} & 110 & 16 & 36 & 0.0084 \\
\mbox{TimeScope} & 285 & 53 & 102 & 0.0001 \\
\mbox{TOC-Bench} & 280 & 22 & 63 & 0.0000 \\
\mbox{TOMATO} & 295 & 31 & 95 & 0.0000 \\
\mbox{TVBench} & 296 & 31 & 107 & 0.0000 \\
\mbox{UCF101-AD} & 300 & 11 & 129 & 0.0000 \\
\mbox{UrbanVideo-Bench} & 300 & 26 & 64 & 0.0001 \\
\mbox{v-HUB} & 126 & 3 & 63 & 0.0000 \\
\mbox{V2P-Bench} & 294 & 38 & 84 & 0.0000 \\
\mbox{VANE-Bench} & 300 & 8 & 41 & 0.0000 \\
\mbox{VAU-Bench} & 251 & 9 & 5 & 0.4239 \\
\mbox{VELOCITI} & 300 & 13 & 98 & 0.0000 \\
\mbox{VF-Eval} & 299 & 23 & 26 & 0.7751 \\
\mbox{VGenST-Bench} & 300 & 9 & 136 & 0.0000 \\
\mbox{Video-LevelGauge} & 300 & 10 & 119 & 0.0000 \\
\mbox{Video-MME} & 289 & 31 & 85 & 0.0000 \\
\mbox{Video-MME-v2} & 253 & 25 & 69 & 0.0000 \\
\mbox{Video-MMMU} & 269 & 22 & 78 & 0.0000 \\
\mbox{Video-TT} & 56 & 3 & 22 & 0.0003 \\
\mbox{VideoAds} & 300 & 18 & 109 & 0.0000 \\
\mbox{VideoEspresso} & 275 & 19 & 30 & 0.1531 \\
\mbox{VideoEval-Pro} & 289 & 47 & 76 & 0.0116 \\
\mbox{VideoKR} & 161 & 15 & 64 & 0.0000 \\
\mbox{VideoMathQA} & 207 & 49 & 38 & 0.2837 \\
\mbox{VideoOdyssey} & 296 & 73 & 43 & 0.0071 \\
\mbox{VideoSIAH-Eval} & 219 & 7 & 26 & 0.0017 \\
\mbox{VidPair-Halluc} & 96 & 6 & 24 & 0.0019 \\
\mbox{ViMUL-Bench} & 161 & 13 & 69 & 0.0000 \\
\mbox{Vinoground} & 293 & 15 & 121 & 0.0000 \\
\mbox{VITATECS} & 288 & 11 & 76 & 0.0000 \\
\mbox{VKnowU} & 255 & 21 & 84 & 0.0000 \\
\mbox{VNBench} & 299 & 18 & 162 & 0.0000 \\
\mbox{VRU-Accident} & 300 & 19 & 102 & 0.0000 \\
\mbox{VSI-Bench} & 130 & 15 & 46 & 0.0001 \\
\mbox{VSI-Super-Wild} & 191 & 27 & 32 & 0.6025 \\
\mbox{VURB} & 189 & 16 & 61 & 0.0000 \\
\end{xltabular}
\endgroup
\endgroup

\begingroup
\begingroup
\setlength{\LTcapwidth}{\linewidth}
\setlength{\tabcolsep}{4pt}
\begin{xltabular}{\linewidth}{>{\raggedright\arraybackslash}p{102.88bp}>{\raggedleft\arraybackslash}p{25.68bp}>{\raggedleft\arraybackslash}p{26.58bp}>{\raggedleft\arraybackslash}p{43.93bp}>{\raggedleft\arraybackslash}p{52.11bp}>{\raggedleft\arraybackslash}p{66.42bp}>{\raggedleft\arraybackslash}p{28.41bp}>{\raggedleft\arraybackslash}p{47.56bp}}
\caption{Text-exposure indicators, namely blind lift, min-20\%, zlib, and infini-gram matches.}\label{tab:contam}\\
\toprule
\rowcolor{SkillPaleBlue} Benchmark & Yr & Lift & mink 7B & mink 1.5B & mink Mistral & zlib & ngram \% \\
\midrule
\endfirsthead
\multicolumn{8}{l}{\tablename~\thetable{} (continued)}\\
\toprule
\rowcolor{SkillPaleBlue} Benchmark & Yr & Lift & mink 7B & mink 1.5B & mink Mistral & zlib & ngram \% \\
\midrule
\endhead
\bottomrule
\endfoot
\mbox{VNBench} & 2025 & 1.0 & 6.99 & 6.37 & 6.04 & 0.148 & 16.0 \\
\mbox{v-HUB} & 2025 & -3.1 & 9.82 & 8.45 & 7.78 & 0.186 & 21.7 \\
\mbox{SoMi-ToM} & 2025 & 17.3 & 9.69 & 8.85 & 8.56 & 0.195 & 0.0 \\
\mbox{VGenST-Bench} & 2026 & 18.7 & 8.01 & 7.56 & 6.94 & 0.156 & 10.7 \\
\mbox{MME-VideoOCR} & 2025 & 17.7 & 7.34 & 6.74 & 6.40 & 0.166 & 11.0 \\
\mbox{UCF101-AD} & 2026 & -1.8 & 6.37 & 5.82 & 5.35 & 0.167 & 0.0 \\
\mbox{EgoSchema} & 2023 & 19.3 & 6.73 & 5.99 & 5.58 & 0.179 & 0.7 \\
\mbox{Video-LevelGauge} & 2026 & 27.7 & 7.96 & 7.48 & 6.70 & 0.174 & 18.0 \\
\mbox{Vinoground} & 2024 & -4.0 & 7.53 & 6.69 & 6.35 & 0.177 & 0.0 \\
\mbox{ViMUL-Bench} & 2025 & 18.3 & 7.32 & 6.39 & 5.90 & 0.189 & 3.0 \\
\mbox{Video-TT} & 2025 & 3.3 & 7.68 & 6.76 & 6.36 & 0.173 & 2.3 \\
\mbox{CAViAR} & 2026 & 13.2 & 6.20 & 5.93 & 5.53 & 0.114 & 25.0 \\
\mbox{VideoKR} & 2026 & 9.1 & 8.03 & 7.48 & 7.03 & 0.169 & 7.7 \\
\mbox{MORSE-500} & 2025 & -9.8 & 7.44 & 7.29 & 6.88 & 0.162 & 100.0 \\
\mbox{VideoAds} & 2025 & 26.0 & 6.31 & 5.78 & 5.47 & 0.140 & 19.7 \\
\mbox{NaturalBench-Video} & 2025 & 0.7 & 8.35 & 7.54 & 7.46 & 0.149 & 13.7 \\
\mbox{MVBench} & 2023 & 10.3 & 9.15 & 7.82 & 7.52 & 0.177 & 27.3 \\
\mbox{TaskMeAnything-v1} & 2024 & 6.7 & 8.47 & 7.28 & 6.93 & 0.157 & 10.7 \\
\mbox{GasVideo-1000} & 2026 & 16.1 & 11.02 & 8.49 & 8.50 & 0.189 & 16.8 \\
\mbox{AoTBench} & 2025 & 0.7 & 8.77 & 7.79 & 6.66 & 0.191 & 4.7 \\
\mbox{VELOCITI} & 2025 & 15.3 & 7.19 & 6.42 & 5.97 & 0.181 & 0.0 \\
\mbox{Q-Bench-Video} & 2024 & 17.0 & 7.79 & 7.29 & 7.11 & 0.145 & 10.0 \\
\mbox{VRU-Accident} & 2025 & 23.3 & 7.85 & 6.58 & 6.34 & 0.145 & 14.7 \\
\mbox{Perception Test} & 2023 & 13.7 & 9.04 & 7.61 & 7.37 & 0.188 & 25.3 \\
\mbox{iSafetyBench} & 2025 & 8.5 & 6.80 & 6.22 & 5.71 & 0.168 & 3.0 \\
\mbox{MotionBench} & 2025 & 13.0 & 6.88 & 6.39 & 6.24 & 0.151 & 13.7 \\
\mbox{PAI-Bench-U} & 2026 & 9.0 & 9.12 & 8.38 & 8.11 & 0.170 & 1.3 \\
\mbox{E.T. Bench} & 2024 & 5.3 & 8.08 & 6.93 & 6.36 & 0.187 & 31.7 \\
\mbox{TVBench} & 2024 & 4.0 & 8.25 & 7.19 & 6.76 & 0.173 & 34.0 \\
\mbox{VKnowU} & 2025 & 9.9 & 6.93 & 6.25 & 6.12 & 0.156 & 17.3 \\
\mbox{VSI-Bench} & 2024 & 7.5 & 7.94 & 6.66 & 6.58 & 0.150 & 1.3 \\
\mbox{VITATECS} & 2023 & 18.5 & 7.12 & 6.26 & 6.27 & 0.155 & 5.3 \\
\mbox{ContPhy} & 2024 & 2.3 & 8.63 & 7.84 & 7.58 & 0.169 & 8.3 \\
\mbox{VURB} & 2026 & 13.2 & 8.02 & 7.15 & 6.72 & 0.162 & 10.0 \\
\mbox{SITE-Bench} & 2025 & 17.3 & 9.07 & 7.62 & 7.32 & 0.185 & 21.0 \\
\mbox{Video-MMMU} & 2025 & 36.6 & 6.55 & 6.01 & 5.55 & 0.195 & 14.3 \\
\mbox{SIS-Bench} & 2026 & 25.0 & 9.16 & 8.43 & 7.77 & 0.146 & 0.0 \\
\mbox{TimeBlind} & 2026 & -0.8 & 7.91 & 7.21 & 7.23 & 0.146 & 12.3 \\
\mbox{TOMATO} & 2024 & 2.1 & 9.36 & 8.18 & 7.00 & 0.189 & 8.0 \\
\mbox{VCR-Bench} & 2025 & 21.1 & 8.91 & 8.12 & 7.89 & 0.153 & 0.0 \\
\mbox{ReVSI} & 2026 & 4.3 & 8.23 & 7.18 & 7.15 & 0.158 & 4.0 \\
\mbox{ST-Evidence} & 2026 & 14.2 & 7.90 & 6.95 & 6.98 & 0.160 & 18.7 \\
\mbox{TempCompass} & 2024 & -4.5 & 8.57 & 7.54 & 7.03 & 0.162 & 5.3 \\
\mbox{CineTechBench} & 2025 & 15.2 & 8.40 & 7.82 & 6.72 & 0.185 & 29.9 \\
\mbox{MMVU} & 2025 & 36.9 & 7.28 & 6.99 & 6.66 & 0.137 & 12.7 \\
\mbox{Colon-Bench} & 2026 & 34.3 & 6.76 & 6.68 & 5.86 & 0.163 & 12.7 \\
\mbox{OSI-Bench} & 2026 & 3.6 & 8.17 & 7.60 & 7.04 & 0.181 & 0.0 \\
\mbox{ReactHuman} & 2026 & -2.0 & 8.41 & 7.06 & 6.71 & 0.151 & 100.0 \\
\mbox{Video-MME} & 2024 & 27.0 & 7.01 & 6.57 & 6.17 & 0.153 & 14.0 \\
\mbox{Video-MME-v2} & 2026 & 6.0 & 6.50 & 6.12 & 5.58 & 0.178 & 5.0 \\
\mbox{LVBench} & 2024 & 3.7 & 8.33 & 7.26 & 6.42 & 0.174 & 17.0 \\
\mbox{SLVMBench} & 2026 & 43.0 & 6.30 & 5.92 & 5.41 & 0.148 & 19.7 \\
\mbox{TimeScope} & 2025 & 19.3 & 9.71 & 7.57 & 6.71 & 0.199 & 57.7 \\
\mbox{MLVU} & 2024 & 23.5 & 7.79 & 7.04 & 6.29 & 0.152 & 16.7 \\
\mbox{KFS-Bench} & 2026 & 30.2 & 7.23 & 6.55 & 6.11 & 0.181 & 17.0 \\
\mbox{MMWorld} & 2024 & 32.7 & 7.58 & 7.00 & 6.40 & 0.167 & 9.0 \\
\mbox{VidPair-Halluc} & 2026 & -14.2 & 9.00 & 8.13 & 7.77 & 0.181 & 0.0 \\
\mbox{V2P-Bench} & 2025 & 11.9 & 9.72 & 7.82 & 6.93 & 0.196 & 17.3 \\
\mbox{StreamingBench} & 2024 & 21.4 & 7.09 & 6.70 & 6.13 & 0.157 & 15.0 \\
\mbox{ScaleLong} & 2025 & 23.7 & 6.54 & 6.16 & 5.98 & 0.144 & 17.0 \\
\mbox{SmartHome-Bench} & 2025 & 13.0 & 5.87 & 6.13 & 5.57 & 0.168 & 0.0 \\
\mbox{TOC-Bench} & 2026 & 2.9 & 8.72 & 7.95 & 7.54 & 0.182 & 15.0 \\
\mbox{SFI-Bench} & 2026 & 29.7 & 7.57 & 6.85 & 6.29 & 0.164 & 20.0 \\
\mbox{M3-Verse} & 2025 & 2.7 & 7.81 & 7.19 & 7.28 & 0.183 & 9.7 \\
\mbox{Video-Holmes} & 2025 & 37.7 & 8.49 & 8.02 & 7.27 & 0.171 & 0.3 \\
\mbox{PLM-VideoBench} & 2025 & 21.7 & 6.33 & 5.78 & 5.47 & 0.152 & 33.3 \\
\mbox{UrbanVideo-Bench} & 2025 & 20.3 & 7.42 & 6.58 & 6.24 & 0.199 & 7.0 \\
\mbox{ExpVid} & 2025 & 27.8 & 8.70 & 7.97 & 7.41 & 0.169 & 6.3 \\
\mbox{SpatialTree} & 2025 & 33.3 & 7.42 & 6.53 & 5.95 & 0.146 & 1.3 \\
\mbox{GUI-World} & 2025 & 56.0 & 7.03 & 6.55 & 6.09 & 0.145 & 8.7 \\
\mbox{VANE-Bench} & 2025 & 52.3 & 6.87 & 6.74 & 6.34 & 0.141 & 17.0 \\
\mbox{SciVideoBench} & 2025 & 44.0 & 7.83 & 7.12 & 6.78 & 0.177 & 0.0 \\
\mbox{LongVideoBench} & 2024 & 29.3 & 8.24 & 7.21 & 6.47 & 0.199 & 14.7 \\
\mbox{VideoEval-Pro} & 2025 & 10.7 & 7.44 & 6.89 & 6.54 & 0.154 & 18.3 \\
\mbox{SAW-Bench} & 2026 & 2.9 & 8.77 & 7.84 & 7.22 & 0.189 & 0.3 \\
\mbox{K9-Bench} & 2026 & 45.5 & 7.51 & 6.90 & 6.12 & 0.201 & 0.0 \\
\mbox{CG-Bench} & 2024 & 32.9 & 6.36 & 6.06 & 5.84 & 0.163 & 16.7 \\
\mbox{VideoSIAH-Eval} & 2025 & -21.1 & 10.54 & 9.37 & 8.42 & 0.195 & 15.0 \\
\mbox{FAVOR-Bench} & 2025 & 26.7 & 8.00 & 7.28 & 6.87 & 0.181 & 7.3 \\
\mbox{MMR-V} & 2025 & 37.4 & 7.50 & 6.81 & 5.86 & 0.165 & 7.0 \\
\mbox{COVER} & 2025 & 26.1 & 7.42 & 6.75 & 6.61 & 0.149 & 12.3 \\
\mbox{S-EMBER} & 2026 & 9.4 & 6.46 & 5.79 & 5.38 & 0.171 & 10.4 \\
\mbox{HERBench} & 2026 & 28.0 & 6.98 & 6.54 & 5.89 & 0.173 & 9.3 \\
\mbox{ReXTime} & 2024 & 59.3 & 6.96 & 6.40 & 5.96 & 0.160 & 5.0 \\
\mbox{PerceptionComp} & 2026 & -13.3 & 8.71 & 8.02 & 7.21 & 0.194 & 8.3 \\
\mbox{OVO-S-Bench} & 2026 & 9.9 & 7.32 & 7.40 & 7.10 & 0.196 & 13.7 \\
\mbox{Causal2Needles} & 2025 & 32.3 & 9.04 & 8.12 & 7.21 & 0.189 & 12.3 \\
\mbox{IntPhys 2} & 2025 & 1.7 & 10.61 & 8.41 & 8.65 & 0.160 & 0.0 \\
\mbox{ExAct} & 2025 & 44.0 & 6.45 & 6.36 & 5.80 & 0.175 & 0.0 \\
\mbox{TemporalBench} & 2024 & 21.8 & 6.71 & 6.15 & 5.48 & 0.215 & 0.0 \\
\mbox{VideoEspresso} & 2025 & 50.0 & 6.82 & 6.47 & 5.80 & 0.172 & 0.0 \\
\mbox{VSI-Super-Wild} & 2026 & 4.0 & 7.56 & 7.30 & 6.86 & 0.174 & 0.0 \\
\mbox{Artifact-Bench} & 2026 & -11.0 & 10.37 & 8.31 & 7.93 & 0.193 & 0.0 \\
\mbox{ForeSeaQA} & 2026 & 45.9 & 6.53 & 6.26 & 6.31 & 0.147 & 10.7 \\
\mbox{HLV-1K} & 2025 & 42.0 & 6.67 & 6.29 & 5.89 & 0.166 & 8.7 \\
\mbox{STI-Bench} & 2025 & 27.2 & 8.09 & 7.39 & 6.79 & 0.185 & 0.0 \\
\mbox{DSI-Bench} & 2025 & 26.5 & 6.71 & 6.41 & 6.25 & 0.154 & 0.0 \\
\mbox{LongVideo-Reason} & 2025 & 49.5 & 8.06 & 7.51 & 6.71 & 0.175 & 6.3 \\
\mbox{VF-Eval} & 2025 & 58.3 & 7.61 & 6.73 & 6.26 & 0.163 & 3.3 \\
\mbox{DSR-Bench} & 2025 & -0.9 & 6.43 & 5.74 & 5.50 & 0.222 & 0.7 \\
\mbox{ALLVB} & 2025 & 53.3 & 6.73 & 6.48 & 5.93 & 0.148 & 16.0 \\
\mbox{FineBench} & 2026 & 38.3 & 7.47 & 6.98 & 6.59 & 0.193 & 20.7 \\
\mbox{HourVideo} & 2024 & 2.7 & 6.97 & 6.31 & 5.90 & 0.177 & 6.0 \\
\mbox{VAU-Bench} & 2025 & 71.0 & 6.30 & 6.11 & 5.85 & 0.138 & 0.0 \\
\mbox{LEMONADE} & 2025 & 16.7 & 10.67 & 9.26 & 8.11 & 0.190 & 3.7 \\
\mbox{MMSI-Video-Bench} & 2025 & 5.9 & 6.95 & 6.42 & 6.26 & 0.166 & 8.0 \\
\mbox{EgoPlan-Bench} & 2023 & 8.0 & 9.04 & 7.49 & 7.60 & 0.189 & 33.3 \\
\mbox{VideoMathQA} & 2025 & 11.7 & 7.31 & 6.18 & 6.44 & 0.166 & 9.3 \\
\mbox{VideoOdyssey} & 2026 & 19.4 & 7.12 & 6.63 & 6.22 & 0.164 & 10.9 \\
\mbox{LongTimeScope} & 2025 & 17.0 & 9.67 & 7.50 & 6.65 & 0.198 & 56.7 \\
\mbox{NExT-QA} & 2021 & -- & 9.05 & 7.76 & 8.06 & 0.209 & 44.3 \\
\mbox{Neptune} & 2024 & -- & 8.79 & 7.79 & 6.96 & 0.159 & 0.7 \\
\mbox{VSTAT} & 2026 & -- & 8.66 & 7.72 & 7.10 & 0.171 & 4.1 \\
\end{xltabular}
\endgroup
\endgroup

Table~\ref{tab:contam} reports three text-exposure indicators, namely min-$k$\% scores under three open-weight models \citep{shi2024detecting}, zlib compression ratios, and exact question matches in Dolma \citep{liu2024infini, soldaini2024dolma}. Across 110 benchmarks with both indicators, blind lift correlates with Qwen2.5-7B \citep{qwen2024qwen25} min-$k$\% at Spearman $-0.38$. These indicators are indirect evidence of exposure. Median blind lift by release year is $+14$\% for 2023, $+8$\% for 2024, $+21$\% for 2025, and $+13$\% for 2026. A benchmark's release date need not date its source material, and a question match establishes public text rather than answer access. The text attack directly measures answerability without video.

\section{Pool-level attacks}
\label{app:t2}
\paragraph{Shared content and duplicates.}
Content hashes identify 923 shared videos and 2,893 cross-benchmark video pairs. Text comparisons find 412 near-duplicate and 345 verbatim question pairs. Near-duplicates require question cosine similarity of at least 0.9 under BGE-large \citep{xiao2024c}, confirmed with MPNet \citep{song2020mpnet}. Table~\ref{tab:dups} lists the largest cross-benchmark duplicate flows, and Table~\ref{tab:dupshare} reports their directional shares.
\paragraph{Pool attacks.}
Attackers draw on a within-benchmark store of previously seen items or a cross-benchmark store preloaded with other benchmarks. Copying transfers the nearest stored answer and matches option text when choices differ. Retrieval supplies the eight nearest examples to Qwen3-8B \citep{yang2025qwen3} at temperature 0. Online logistic regression predicts answer letters from BGE-large \citep{xiao2024c} embeddings using five rotating held-out splits. Table~\ref{tab:t2att} also summarizes a Laplace-smoothed position learner and two same-video sibling-item attackers. Full-scale runs use the reported item counts, and we track accuracy as evaluation proceeds.
\paragraph{Scoring and controls.}
We use 50 sequence permutations for copying and 3 for retrieval. Pool exploitability $\varepsilon_{\mathrm{pool}}$ is permutation-averaged accuracy over the final 20\% of the sequence, smoothed over 20 items, minus chance. We obtain 95\% intervals from 1,000 item-level bootstrap resamples. A control shuffles gold answers among items with equal option counts, preserving position priors while disrupting question-answer relations. Its excess accuracy is consistent with zero on 402 of 447 benchmark-attacker combinations, and fourteen retain a margin above 5\%. The baseline $c$ averages chance over the multiple-choice questions in each benchmark.
\paragraph{Reading the tables.}
Tables~\ref{tab:t2eps}--\ref{tab:t2eps_cols2} list the target Benchmark and its full-scale item count $n$, rather than a 300-item sample. Superscript w denotes a store of previously seen items from the same benchmark, and superscript c a store preloaded with other benchmarks. The reported $\varepsilon_{\mathrm{pool}}$ is the largest final-stage margin among the listed attackers, in \%. Att. identifies that attacker. L0 learns answer-position priors, L1 copies near-duplicates, L2-rule and L2-LLM use same-video sibling items through rules and a language model, and L3 retrieves eight examples. ALC in Table~\ref{tab:t2eps} is the selected attacker's exported normalized area under the learning curve, reported as a percentage. It summarizes the full sequence rather than only its final 20\%. Zero denotes chance-level performance, and 100\% denotes perfect accuracy throughout.
\paragraph{Diversity and robustness.}
Table~\ref{tab:diversity} reports diversity, template entropy, and effective item counts. The table and Figure~\ref{fig:f7size} cover 114 of the 115 audited benchmarks, since CG-Bench \citep{chen2025cgbench} has no item embeddings. We compute Vendi scores on $n=200$ items for a fixed-size comparison. The 8 smaller benchmarks use all available items and remain marked by their sample size. The primary embedding pairs SigLIP2 \citep{tschannen2025siglip} with BGE-large \citep{xiao2024c}, and an alternative pairs DINOv3 \citep{simeoni2025dinov3} with MPNet \citep{song2020mpnet}. Video embeddings average the same 32 frames, and text embeddings average token representations. Effective-size rankings agree across visual encoders at Spearman 0.88, across text encoders at 0.96, and across joint spaces at 0.90. Near-duplicate shares correlate at 0.90 across the two spaces. Diversity and judgment reliability capture different properties. Across 110 benchmarks, split-half reliability of the item-by-judge matrix correlates 0.04 with effective size and 0.00 with near-duplicate share.
\begingroup
\begingroup
\setlength{\LTcapwidth}{\linewidth}
\setlength{\tabcolsep}{4pt}
\begin{xltabular}{\linewidth}{>{\raggedright\arraybackslash}p{128.18bp}>{\raggedright\arraybackslash}p{137.54bp}>{\raggedleft\arraybackslash}p{78.93bp}>{\raggedleft\arraybackslash}p{80.93bp}}
\caption{Largest cross-benchmark near-duplicate flows, confirmed in both embedding spaces.}\label{tab:dups}\\
\toprule
\rowcolor{SkillPaleBlue} Benchmark A & Benchmark B & Pairs & Exact \\
\midrule
\endfirsthead
\multicolumn{4}{l}{\tablename~\thetable{} (continued)}\\
\toprule
\rowcolor{SkillPaleBlue} Benchmark A & Benchmark B & Pairs & Exact \\
\midrule
\endhead
\bottomrule
\endfoot
\mbox{PAI-Bench-U} & \mbox{PhysicalAI} & 191 & 190 \\
\mbox{KFS-Bench} & \mbox{LongVideoBench} & 131 & 123 \\
\mbox{SITE-Bench} & \mbox{VSI-Bench} & 30 & 9 \\
\mbox{SciVideoBench} & \mbox{VideoKR} & 27 & 23 \\
\mbox{VideoKR} & \mbox{Video-MMMU} & 21 & 21 \\
\mbox{MMVU} & \mbox{VCR-Bench} & 16 & 14 \\
\mbox{MMVU} & \mbox{VideoKR} & 13 & 13 \\
\mbox{VKnowU} & \mbox{VSI-Bench} & 13 & 3 \\
\mbox{SpatialTree} & \mbox{VSI-Bench} & 10 & 7 \\
\mbox{Video-MME} & \mbox{VideoEval-Pro} & 9 & 9 \\
\mbox{VURB} & \mbox{Video-MME} & 7 & 7 \\
\mbox{MMVU} & \mbox{VURB} & 6 & 6 \\
\mbox{VURB} & \mbox{VideoKR} & 6 & 6 \\
\mbox{SITE-Bench} & \mbox{VURB} & 6 & 3 \\
\mbox{GasVideo-1000} & \mbox{Video-MME} & 5 & 5 \\
\mbox{GasVideo-1000} & \mbox{MVBench} & 5 & 5 \\
\mbox{MVBench} & \mbox{TVBench} & 5 & 5 \\
\mbox{SpatialTree} & \mbox{VKnowU} & 5 & 3 \\
\mbox{SITE-Bench} & \mbox{SpatialTree} & 5 & 3 \\
\mbox{KFS-Bench} & \mbox{VideoEval-Pro} & 4 & 4 \\
\mbox{VCR-Bench} & \mbox{VideoKR} & 4 & 4 \\
\mbox{MVBench} & \mbox{VURB} & 4 & 4 \\
\mbox{VSI-Bench} & \mbox{VURB} & 4 & 1 \\
\mbox{SITE-Bench} & \mbox{Video-MME} & 4 & 4 \\
\mbox{LVBench} & \mbox{VideoEval-Pro} & 3 & 0 \\
\mbox{Perception Test} & \mbox{TVBench} & 3 & 1 \\
\mbox{TOMATO} & \mbox{TVBench} & 3 & 0 \\
\mbox{VCR-Bench} & \mbox{VURB} & 3 & 3 \\
\mbox{MVBench} & \mbox{SITE-Bench} & 2 & 2 \\
\mbox{MLVU} & \mbox{VideoEval-Pro} & 2 & 0 \\
\end{xltabular}
\endgroup
\endgroup

\begingroup
\setlength{\LTcapwidth}{\linewidth}
\setlength{\tabcolsep}{4pt}
\begin{xltabular}{\linewidth}{>{\raggedright\arraybackslash}p{87.92bp}>{\raggedleft\arraybackslash}p{24.24bp}>{\raggedleft\arraybackslash}p{24.24bp}>{\raggedleft\arraybackslash}p{24.24bp}>{\raggedleft\arraybackslash}p{24.24bp}>{\raggedleft\arraybackslash}p{20.94bp}>{\raggedleft\arraybackslash}p{18.79bp}>{\raggedleft\arraybackslash}p{18.79bp}>{\raggedleft\arraybackslash}p{21.51bp}>{\raggedleft\arraybackslash}p{25.79bp}>{\raggedleft\arraybackslash}p{18.79bp}>{\raggedleft\arraybackslash}p{23.35bp}>{\raggedleft\arraybackslash}p{20.76bp}}
\caption{Cross-benchmark near-duplicate shares (\%). Rows supply items matched to columns. Column codes in row order are PA PhysicalAI, PU PAI-Bench-U, KF KFS-Bench, LV LongVideoBench, VK VideoKR, VS VSI-Bench, SI SITE-Bench, VU VURB, MM MMVU, SV SciVideoBench, VM Video-MME, VC VCR-Bench.}\label{tab:dupshare}\\
\toprule
\rowcolor{SkillPaleBlue} Items of & PA & PU & KF & LV & VK & VS & SI & VU & MM & SV & VM & VC \\
\midrule
\endfirsthead
\multicolumn{13}{l}{\tablename~\thetable{} (continued)}\\
\toprule
\rowcolor{SkillPaleBlue} Items of & PA & PU & KF & LV & VK & VS & SI & VU & MM & SV & VM & VC \\
\midrule
\endhead
\bottomrule
\endfoot
\mbox{PhysicalAI} & -- & 63.7 & 0 & 0 & 0 & 0 & 0 & 0 & 0 & 0 & 0 & 0 \\
\mbox{PAI-Bench-U} & 63.3 & -- & 0 & 0 & 0 & 0 & 0 & 0 & 0 & 0 & 0 & 0 \\
\mbox{KFS-Bench} & 0 & 0 & -- & 43.0 & 0 & 0 & 0 & 0 & 0 & 0 & 0 & 0 \\
\mbox{LongVideoBench} & 0 & 0 & 42.7 & -- & 0 & 0 & 0 & 0 & 0 & 0 & 0 & 0 \\
\mbox{VideoKR} & 0 & 0 & 0 & 0 & -- & 0 & 0 & 2.0 & 4.3 & 9.0 & 0 & 1.3 \\
\mbox{VSI-Bench} & 0 & 0 & 0 & 0 & 0 & -- & 8.3 & 1.3 & 0 & 0 & 0 & 0 \\
\mbox{SITE-Bench} & 0 & 0 & 0 & 0 & 0 & 8.3 & -- & 1.7 & 0 & 0 & 1.3 & 0 \\
\mbox{VURB} & 0 & 0 & 0 & 0 & 2.0 & 1.0 & 2.0 & -- & 2.0 & 0 & 2.3 & 1.0 \\
\mbox{MMVU} & 0 & 0 & 0 & 0 & 4.3 & 0 & 0 & 2.0 & -- & 0 & 0 & 5.3 \\
\mbox{SciVideoBench} & 0 & 0 & 0 & 0 & 8.7 & 0 & 0 & 0 & 0 & -- & 0 & 0 \\
\mbox{Video-MME} & 0 & 0 & 0 & 0 & 0 & 0 & 1.3 & 2.3 & 0 & 0 & -- & 0 \\
\mbox{VCR-Bench} & 0 & 0 & 0 & 0 & 1.3 & 0 & 0 & 0.3 & 5.3 & 0 & 0 & -- \\
\end{xltabular}
\endgroup
\begingroup
\begingroup
\setlength{\LTcapwidth}{\linewidth}
\setlength{\tabcolsep}{4pt}
\begin{xltabular}{\linewidth}{>{\raggedright\arraybackslash}p{24.93bp}>{\raggedright\arraybackslash}p{163.64bp}>{\raggedright\arraybackslash}p{32.20bp}>{\raggedright\arraybackslash}p{91.12bp}>{\raggedleft\arraybackslash}p{63.30bp}>{\raggedright\arraybackslash}p{34.38bp}}
\caption{Pool attacks at full scale and on 300-item subsets, with learning-curve summaries.}\label{tab:t2att}\\
\toprule
\rowcolor{SkillPaleBlue} Scale & Attacker & Pool & Median $\varepsilon_{\mathrm{pool}}$ [IQR] & Median ALC & Expl. \\
\midrule
\endfirsthead
\multicolumn{6}{l}{\tablename~\thetable{} (continued)}\\
\toprule
\rowcolor{SkillPaleBlue} Scale & Attacker & Pool & Median $\varepsilon_{\mathrm{pool}}$ [IQR] & Median ALC & Expl. \\
\midrule
\endhead
\bottomrule
\endfoot
full & L0 position prior & within & 0.0 [-0.3, 1.9] & 0.0 & 16/113 \\
full & L1 near-duplicate copy & within & 5.7 [1.5, 20.3] & 7.1 & 46/113 \\
full & L2-rule sibling-item rule & within & 6.8 [2.0, 21.2] & 7.8 & 46/113 \\
full & L2-LLM sibling-item LLM & within & 18.5 [8.1, 28.7] & 17.1 & 72/113 \\
full & L3 top-8 RAG & within & 15.7 [7.9, 30.8] & 20.2 & 80/113 \\
full & any attacker (max) & within & 23.7 [11.7, 37.6] & 24.0 & 89/113 \\
full & L0 position prior & cross & 0.0 [-0.3, 0.9] & 0.0 & 14/113 \\
full & L1 near-duplicate copy & cross & 9.3 [2.3, 28.1] & 10.6 & 58/113 \\
full & L2-rule sibling-item rule & cross & 11.1 [2.7, 29.0] & 11.5 & 60/113 \\
full & L2-LLM sibling-item LLM & cross & 20.0 [8.8, 33.8] & 21.4 & 77/113 \\
full & L3 top-8 RAG & cross & 19.8 [7.8, 39.0] & 25.0 & 82/113 \\
full & any attacker (max) & cross & 28.5 [13.5, 44.7] & 30.0 & 91/113 \\
\midrule
s300 & L0 position prior & within & 0.1 [-0.5, 3.2] & 0.0 & 8/113 \\
s300 & L1 near-duplicate copy & within & 4.5 [0.9, 13.2] & 4.0 & 25/113 \\
s300 & L2-rule sibling-item rule & within & 5.4 [1.0, 14.9] & 4.5 & 24/113 \\
s300 & L2-LLM sibling-item LLM & within & 8.8 [3.6, 15.2] & 6.5 & 34/113 \\
s300 & L3 top-8 RAG & within & 14.9 [5.0, 24.9] & 16.5 & 50/113 \\
s300 & any attacker (max) & within & 15.4 [7.2, 27.6] & 16.4 & 55/113 \\
s300 & L0 position prior & cross & 0.0 [-0.9, 2.2] & 0.0 & 8/113 \\
s300 & L1 near-duplicate copy & cross & 6.6 [1.0, 13.1] & 6.2 & 25/113 \\
s300 & L2-rule sibling-item rule & cross & 6.7 [1.2, 15.7] & 6.6 & 25/113 \\
s300 & L2-LLM sibling-item LLM & cross & 8.8 [3.6, 16.3] & 7.6 & 38/113 \\
s300 & L3 top-8 RAG & cross & 14.0 [6.2, 25.9] & 18.0 & 51/113 \\
s300 & any attacker (max) & cross & 18.1 [8.7, 27.7] & 19.0 & 56/113 \\
\end{xltabular}
\endgroup
\endgroup

\begingroup
\begingroup
\setlength{\LTcapwidth}{\linewidth}
\setlength{\tabcolsep}{4pt}
\begin{xltabular}{\linewidth}{>{\raggedright\arraybackslash}p{138.85bp}>{\raggedleft\arraybackslash}p{67.10bp}>{\raggedleft\arraybackslash}p{62.55bp}>{\raggedright\arraybackslash}p{79.23bp}>{\raggedleft\arraybackslash}p{69.86bp}}
\caption{Within-pool and cross-pool exploitability across audited benchmarks (1/2).}\label{tab:t2eps}\\
\toprule
\rowcolor{SkillPaleBlue} Benchmark & $n$ & $\varepsilon_{\mathrm{pool}}^{\mathrm{w}}$ & Att.$^{\mathrm{w}}$ & ALC$^{\mathrm{w}}$ \\
\midrule
\endfirsthead
\multicolumn{5}{l}{\tablename~\thetable{} (continued)}\\
\toprule
\rowcolor{SkillPaleBlue} Benchmark & $n$ & $\varepsilon_{\mathrm{pool}}^{\mathrm{w}}$ & Att.$^{\mathrm{w}}$ & ALC$^{\mathrm{w}}$ \\
\midrule
\endhead
\bottomrule
\endfoot
\mbox{TimeScope} & 3000 & 82.4 & L2-LLM & 92.1 \\
\mbox{LongTimeScope} & 900 & 81.2 & L1 & 86.3 \\
\mbox{UCF101-AD} & 4224 & 76.0 & L3 & 81.6 \\
\mbox{Video-TT} & 5000 & 73.2 & L2-LLM & 78.8 \\
\mbox{STI-Bench} & 4128 & 71.1 & L2-rule & 54.2 \\
\mbox{FineBench} & 20143 & 69.2 & L1 & 90.5 \\
\mbox{SAW-Bench} & 4142 & 67.7 & L2-rule & 70.9 \\
\mbox{VF-Eval} & 5932 & 67.3 & L2-rule & 50.0 \\
\mbox{VURB} & 2100 & 64.3 & L1 & 71.2 \\
\mbox{VAU-Bench} & 251 & 64.2 & L3 & 88.1 \\
\mbox{VRU-Accident} & 5989 & 56.3 & L1 & 72.4 \\
\mbox{VideoMathQA} & 2100 & 55.1 & L2-rule & 79.3 \\
\mbox{VANE-Bench} & 559 & 53.3 & L2-LLM & 47.8 \\
\mbox{GUI-World} & 3642 & 46.9 & L3 & 60.6 \\
\mbox{Neptune} & 5000 & 45.9 & L1 & 40.4 \\
\mbox{TaskMeAnything-v1} & 3567 & 45.8 & L2-rule & 47.7 \\
\mbox{MMWorld} & 1559 & 42.1 & L3 & 52.5 \\
\mbox{ForeSeaQA} & 591 & 41.7 & L2-LLM & 48.9 \\
\mbox{ALLVB} & 25564 & 40.8 & L3 & 49.0 \\
\mbox{Video-Holmes} & 3388 & 40.4 & L2-LLM & 36.1 \\
\mbox{VELOCITI} & 8789 & 39.0 & L2-rule & 50.6 \\
\mbox{HLV-1K} & 10533 & 38.9 & L3 & 50.5 \\
\mbox{SLVMBench} & 2255 & 38.5 & L3 & 44.7 \\
\mbox{LongVideo-Reason} & 856 & 37.4 & L3 & 52.9 \\
\mbox{VGenST-Bench} & 9707 & 35.1 & L3 & 46.9 \\
\mbox{FAVOR-Bench} & 5000 & 34.8 & L2-LLM & 29.9 \\
\mbox{MMR-V} & 1256 & 34.8 & L2-LLM & 25.8 \\
\mbox{CMU} & 4153 & 33.5 & L2-LLM & 22.8 \\
\mbox{ViMUL-Bench} & 5000 & 31.0 & L2-LLM & 35.0 \\
\mbox{Colon-Bench} & 1485 & 30.8 & L3 & 34.3 \\
\mbox{ExAct} & 3521 & 29.3 & L3 & 33.7 \\
\mbox{ReXTime} & 921 & 28.7 & L3 & 26.4 \\
\mbox{StreamingBench} & 2500 & 28.5 & L2-LLM & 31.4 \\
\mbox{DSI-Bench} & 1769 & 27.9 & L1 & 35.5 \\
\mbox{MLVU} & 2174 & 27.0 & L3 & 31.1 \\
\mbox{Causal2Needles} & 747 & 26.3 & L3 & 30.1 \\
\mbox{Q-Bench-Video} & 892 & 25.9 & L3 & 49.0 \\
\mbox{HERBench} & 1971 & 25.9 & L2-LLM & 30.5 \\
\mbox{SFI-Bench} & 1555 & 25.7 & L3 & 30.2 \\
\mbox{Video-MMMU} & 900 & 25.6 & L2-LLM & 22.1 \\
\mbox{COVER} & 5000 & 25.5 & L3 & 33.0 \\
\mbox{CAViAR} & 2996 & 25.2 & L3 & 38.6 \\
\mbox{SIS-Bench} & 4856 & 23.6 & L2-LLM & 20.8 \\
\mbox{SpatialTree} & 814 & 23.0 & L3 & 21.8 \\
\mbox{UrbanVideo-Bench} & 5355 & 23.0 & L3 & 26.5 \\
\mbox{SciVideoBench} & 1000 & 22.6 & L2-LLM & 18.2 \\
\mbox{VideoEspresso} & 1382 & 20.1 & L2-LLM & 21.7 \\
\mbox{VCR-Bench} & 2068 & 19.6 & L1 & 24.6 \\
\mbox{EgoSchema} & 500 & 18.9 & L3 & 24.8 \\
\mbox{M3-Verse} & 3374 & 18.7 & L3 & 21.3 \\
\mbox{K9-Bench} & 4744 & 18.5 & L2-LLM & 16.4 \\
\mbox{Video-MME} & 2700 & 18.5 & L2-LLM & 17.2 \\
\mbox{MotionBench} & 4018 & 17.5 & L3 & 42.5 \\
\mbox{Video-LevelGauge} & 1177 & 17.3 & L2-LLM & 13.5 \\
\mbox{LongVideoBench} & 815 & 17.0 & L3 & 23.5 \\
\mbox{Perception Test} & 5000 & 15.9 & L2-rule & 21.8 \\
\mbox{ReVSI} & 5000 & 15.8 & L1 & 24.0 \\
\mbox{MMVU} & 1000 & 15.8 & L3 & 23.6 \\
\mbox{VideoAds} & 1100 & 15.4 & L2-LLM & 9.1 \\
\mbox{OVO-S-Bench} & 1632 & 15.0 & L3 & 14.0 \\
\mbox{Artifact-Bench} & 850 & 14.1 & L1 & -8.1 \\
\mbox{CineTechBench} & 127 & 13.5 & L3 & 9.5 \\
\mbox{SmartHome-Bench} & 1023 & 13.2 & L0 & 25.3 \\
\mbox{VITATECS} & 5000 & 13.1 & L3 & 20.5 \\
\mbox{iSafetyBench} & 3456 & 12.8 & L3 & 12.1 \\
\mbox{TVBench} & 2525 & 12.6 & L1 & 14.7 \\
\mbox{KFS-Bench} & 580 & 12.2 & L2-LLM & 7.7 \\
\mbox{SITE-Bench} & 3619 & 12.1 & L2-rule & 12.1 \\
\mbox{PhysicalAI} & 1214 & 11.8 & L3 & 36.8 \\
\mbox{ScaleLong} & 1723 & 11.7 & L2-LLM & 12.9 \\
\mbox{PAI-Bench-U} & 1214 & 11.7 & L3 & 36.8 \\
\mbox{Video-MME-v2} & 2865 & 11.7 & L2-LLM & 8.2 \\
\mbox{VKnowU} & 1680 & 11.2 & L3 & 10.9 \\
\mbox{VSI-Bench} & 5000 & 10.8 & L1 & 20.7 \\
\mbox{TOMATO} & 1484 & 10.8 & L2-LLM & 10.3 \\
\mbox{MME-VideoOCR} & 1456 & 10.2 & L3 & 11.0 \\
\mbox{TemporalBench} & 5000 & 10.2 & L2-LLM & 48.3 \\
\mbox{MVBench} & 4000 & 9.1 & L2-LLM & 9.7 \\
\mbox{VideoOdyssey} & 1618 & 8.4 & L3 & 9.0 \\
\mbox{ContPhy} & 1715 & 7.9 & L1 & 0.5 \\
\mbox{GasVideo-1000} & 1013 & 7.6 & L2-rule & 19.0 \\
\mbox{VideoEval-Pro} & 1289 & 7.1 & L2-LLM & 5.0 \\
\mbox{DSR-Bench} & 1450 & 7.1 & L2-rule & 8.6 \\
\mbox{ST-Evidence} & 4004 & 6.6 & L3 & 24.2 \\
\mbox{SoMi-ToM} & 75 & 6.4 & L3 & 43.8 \\
\mbox{ExpVid} & 7800 & 6.4 & L3 & 15.4 \\
\mbox{ReactHuman} & 1054 & 5.7 & L0 & 7.6 \\
\mbox{TOC-Bench} & 2321 & 5.6 & L3 & 7.9 \\
\mbox{V2P-Bench} & 1169 & 5.5 & L3 & 31.7 \\
\mbox{VSI-Super-Wild} & 5000 & 4.5 & L2-LLM & 7.8 \\
\mbox{VideoKR} & 2000 & 4.5 & L3 & 8.3 \\
\mbox{LVBench} & 1549 & 4.4 & L2-LLM & 27.7 \\
\mbox{v-HUB} & 1902 & 3.7 & L2-LLM & 15.9 \\
\mbox{MMSI-Video-Bench} & 1106 & 3.6 & L3 & 6.1 \\
\mbox{E.T. Bench} & 1500 & 3.6 & L3 & 3.6 \\
\mbox{EgoPlan-Bench} & 3343 & 3.5 & L2-LLM & 2.4 \\
\mbox{S-EMBER} & 9448 & 3.2 & L3 & 4.3 \\
\mbox{VNBench} & 320 & 3.1 & L3 & 0.8 \\
\mbox{HourVideo} & 1157 & 1.8 & L3 & 1.5 \\
\mbox{IntPhys 2} & 1072 & 1.1 & L2-rule & 0.4 \\
\mbox{VSTAT} & 1500 & 0.6 & L1 & 16.2 \\
\mbox{PLM-VideoBench} & 665 & 0.0 & L0 & 0.0 \\
\mbox{LEMONADE} & 5000 & -0.1 & L2-rule & 24.9 \\
\mbox{Vinoground} & 1000 & -3.8 & L0 & -3.3 \\
\mbox{VidPair-Halluc} & 5000 & -3.9 & L1 & 18.2 \\
\mbox{TempCompass} & 5000 & -5.3 & L1 & 17.1 \\
\mbox{AoTBench} & 5000 & -6.7 & L3 & 43.3 \\
\mbox{MORSE-500} & 500 & -10.2 & L3 & 8.1 \\
\mbox{OSI-Bench} & 5000 & -12.2 & L1 & 5.1 \\
\mbox{NaturalBench-Video} & 1860 & -15.8 & L2-LLM & 29.8 \\
\mbox{PerceptionComp} & 1114 & -16.6 & L2-LLM & 2.4 \\
\mbox{TimeBlind} & 2400 & -21.3 & L2-LLM & 24.5 \\
\mbox{VideoSIAH-Eval} & 652 & -24.8 & L2-LLM & 0.1 \\
\end{xltabular}
\endgroup

\begingroup
\setlength{\LTcapwidth}{\linewidth}
\setlength{\tabcolsep}{4pt}
\begin{xltabular}{\linewidth}{>{\raggedright\arraybackslash}p{158.31bp}>{\raggedleft\arraybackslash}p{86.56bp}>{\raggedleft\arraybackslash}p{82.01bp}>{\raggedright\arraybackslash}p{98.69bp}}
\caption{Within-pool and cross-pool exploitability across audited benchmarks (2/2).}\label{tab:t2eps_cols2}\\
\toprule
\rowcolor{SkillPaleBlue} Benchmark & $n$ & $\varepsilon_{\mathrm{pool}}^{\mathrm{c}}$ & Att.$^{\mathrm{c}}$ \\
\midrule
\endfirsthead
\multicolumn{4}{l}{\tablename~\thetable{} (continued)}\\
\toprule
\rowcolor{SkillPaleBlue} Benchmark & $n$ & $\varepsilon_{\mathrm{pool}}^{\mathrm{c}}$ & Att.$^{\mathrm{c}}$ \\
\midrule
\endhead
\bottomrule
\endfoot
\mbox{TimeScope} & 3000 & 82.5 & L2-LLM \\
\mbox{LongTimeScope} & 900 & 82.3 & L2-LLM \\
\mbox{UCF101-AD} & 4224 & 76.0 & L3 \\
\mbox{Video-TT} & 5000 & 73.2 & L2-LLM \\
\mbox{STI-Bench} & 4128 & 70.9 & L2-rule \\
\mbox{FineBench} & 20143 & 69.2 & L1 \\
\mbox{SAW-Bench} & 4142 & 67.7 & L2-rule \\
\mbox{VF-Eval} & 5932 & 67.5 & L2-rule \\
\mbox{VURB} & 2100 & 69.8 & L1 \\
\mbox{VAU-Bench} & 251 & 63.7 & L3 \\
\mbox{VRU-Accident} & 5989 & 56.6 & L1 \\
\mbox{VideoMathQA} & 2100 & 55.2 & L2-rule \\
\mbox{VANE-Bench} & 559 & 53.3 & L2-LLM \\
\mbox{GUI-World} & 3642 & 46.8 & L2-LLM \\
\mbox{Neptune} & 5000 & 45.9 & L1 \\
\mbox{TaskMeAnything-v1} & 3567 & 45.8 & L2-rule \\
\mbox{MMWorld} & 1559 & 40.8 & L3 \\
\mbox{ForeSeaQA} & 591 & 41.7 & L2-LLM \\
\mbox{ALLVB} & 25564 & 40.3 & L3 \\
\mbox{Video-Holmes} & 3388 & 40.4 & L2-LLM \\
\mbox{VELOCITI} & 8789 & 39.0 & L2-rule \\
\mbox{HLV-1K} & 10533 & 37.9 & L3 \\
\mbox{SLVMBench} & 2255 & 39.0 & L3 \\
\mbox{LongVideo-Reason} & 856 & 34.9 & L3 \\
\mbox{VGenST-Bench} & 9707 & 34.8 & L3 \\
\mbox{FAVOR-Bench} & 5000 & 34.8 & L2-LLM \\
\mbox{MMR-V} & 1256 & 34.2 & L2-LLM \\
\mbox{CMU} & 4153 & 34.7 & L2-LLM \\
\mbox{ViMUL-Bench} & 5000 & 31.0 & L2-LLM \\
\mbox{Colon-Bench} & 1485 & 31.1 & L3 \\
\mbox{ExAct} & 3521 & 29.3 & L3 \\
\mbox{ReXTime} & 921 & 44.7 & L3 \\
\mbox{StreamingBench} & 2500 & 28.5 & L2-LLM \\
\mbox{DSI-Bench} & 1769 & 28.1 & L1 \\
\mbox{MLVU} & 2174 & 34.2 & L3 \\
\mbox{Causal2Needles} & 747 & 26.6 & L3 \\
\mbox{Q-Bench-Video} & 892 & 24.3 & L3 \\
\mbox{HERBench} & 1971 & 25.9 & L2-LLM \\
\mbox{SFI-Bench} & 1555 & 24.4 & L2-LLM \\
\mbox{Video-MMMU} & 900 & 45.5 & L3 \\
\mbox{COVER} & 5000 & 24.3 & L3 \\
\mbox{CAViAR} & 2996 & 25.2 & L3 \\
\mbox{SIS-Bench} & 4856 & 23.3 & L2-LLM \\
\mbox{SpatialTree} & 814 & 33.0 & L3 \\
\mbox{UrbanVideo-Bench} & 5355 & 22.9 & L3 \\
\mbox{SciVideoBench} & 1000 & 58.6 & L3 \\
\mbox{VideoEspresso} & 1382 & 20.1 & L2-LLM \\
\mbox{VCR-Bench} & 2068 & 19.6 & L1 \\
\mbox{EgoSchema} & 500 & 19.8 & L3 \\
\mbox{M3-Verse} & 3374 & 18.7 & L3 \\
\mbox{K9-Bench} & 4744 & 18.5 & L2-LLM \\
\mbox{Video-MME} & 2700 & 31.1 & L3 \\
\mbox{MotionBench} & 4018 & 20.0 & L3 \\
\mbox{Video-LevelGauge} & 1177 & 18.4 & L3 \\
\mbox{LongVideoBench} & 815 & 18.5 & L3 \\
\mbox{Perception Test} & 5000 & 16.8 & L1 \\
\mbox{ReVSI} & 5000 & 15.8 & L1 \\
\mbox{MMVU} & 1000 & 30.3 & L1 \\
\mbox{VideoAds} & 1100 & 15.4 & L2-LLM \\
\mbox{OVO-S-Bench} & 1632 & 16.5 & L3 \\
\mbox{Artifact-Bench} & 850 & 13.7 & L1 \\
\mbox{CineTechBench} & 127 & 13.5 & L3 \\
\mbox{SmartHome-Bench} & 1023 & 13.2 & L0 \\
\mbox{VITATECS} & 5000 & 13.1 & L3 \\
\mbox{iSafetyBench} & 3456 & 12.8 & L3 \\
\mbox{TVBench} & 2525 & 19.6 & L3 \\
\mbox{KFS-Bench} & 580 & 47.5 & L1 \\
\mbox{SITE-Bench} & 3619 & 34.8 & L1 \\
\mbox{PhysicalAI} & 1214 & 57.2 & L1 \\
\mbox{ScaleLong} & 1723 & 11.7 & L2-LLM \\
\mbox{PAI-Bench-U} & 1214 & 57.0 & L1 \\
\mbox{Video-MME-v2} & 2865 & 11.8 & L2-LLM \\
\mbox{VKnowU} & 1680 & 16.6 & L3 \\
\mbox{VSI-Bench} & 5000 & 16.9 & L1 \\
\mbox{TOMATO} & 1484 & 11.1 & L2-LLM \\
\mbox{MME-VideoOCR} & 1456 & 10.1 & L3 \\
\mbox{TemporalBench} & 5000 & 10.7 & L3 \\
\mbox{MVBench} & 4000 & 22.3 & L2-LLM \\
\mbox{VideoOdyssey} & 1618 & 8.1 & L2-LLM \\
\mbox{ContPhy} & 1715 & 7.7 & L1 \\
\mbox{GasVideo-1000} & 1013 & 24.8 & L2-rule \\
\mbox{VideoEval-Pro} & 1289 & 68.4 & L3 \\
\mbox{DSR-Bench} & 1450 & 8.9 & L2-rule \\
\mbox{ST-Evidence} & 4004 & 7.2 & L3 \\
\mbox{SoMi-ToM} & 75 & 6.4 & L3 \\
\mbox{ExpVid} & 7800 & 6.6 & L3 \\
\mbox{ReactHuman} & 1054 & 5.7 & L1 \\
\mbox{TOC-Bench} & 2321 & 5.7 & L3 \\
\mbox{V2P-Bench} & 1169 & 32.4 & L3 \\
\mbox{VSI-Super-Wild} & 5000 & 4.5 & L2-LLM \\
\mbox{VideoKR} & 2000 & 29.7 & L2-rule \\
\mbox{LVBench} & 1549 & 38.2 & L3 \\
\mbox{v-HUB} & 1902 & 3.7 & L2-LLM \\
\mbox{MMSI-Video-Bench} & 1106 & 1.8 & L2-LLM \\
\mbox{E.T. Bench} & 1500 & 3.6 & L3 \\
\mbox{EgoPlan-Bench} & 3343 & 3.5 & L2-LLM \\
\mbox{S-EMBER} & 9448 & 3.0 & L3 \\
\mbox{VNBench} & 320 & 3.6 & L1 \\
\mbox{HourVideo} & 1157 & 1.6 & L1 \\
\mbox{IntPhys 2} & 1072 & 1.1 & L2-rule \\
\mbox{VSTAT} & 1500 & 0.6 & L1 \\
\mbox{PLM-VideoBench} & 665 & 0.0 & L0 \\
\mbox{LEMONADE} & 5000 & -0.1 & L2-rule \\
\mbox{Vinoground} & 1000 & -0.8 & L0 \\
\mbox{VidPair-Halluc} & 5000 & -3.8 & L1 \\
\mbox{TempCompass} & 5000 & -5.3 & L1 \\
\mbox{AoTBench} & 5000 & -2.1 & L3 \\
\mbox{MORSE-500} & 500 & -10.2 & L3 \\
\mbox{OSI-Bench} & 5000 & -12.2 & L1 \\
\mbox{NaturalBench-Video} & 1860 & -15.8 & L2-LLM \\
\mbox{PerceptionComp} & 1114 & -16.6 & L2-LLM \\
\mbox{TimeBlind} & 2400 & -21.3 & L2-LLM \\
\mbox{VideoSIAH-Eval} & 652 & -24.8 & L2-LLM \\
\end{xltabular}
\endgroup
\endgroup

\begingroup
\begingroup
\setlength{\LTcapwidth}{\linewidth}
\setlength{\tabcolsep}{4pt}
\begin{xltabular}{\linewidth}{>{\raggedright\arraybackslash}p{104.82bp}>{\raggedleft\arraybackslash}p{32.36bp}>{\raggedleft\arraybackslash}p{63.32bp}>{\raggedleft\arraybackslash}p{60.84bp}>{\raggedleft\arraybackslash}p{52.64bp}>{\raggedleft\arraybackslash}p{59.00bp}>{\raggedleft\arraybackslash}p{28.60bp}}
\caption{Content diversity, duplication, template entropy, and effective item counts by benchmark.}\label{tab:diversity}\\
\toprule
\rowcolor{SkillPaleBlue} Benchmark & VS/n & NearDup \% & SameVid \% & TemplEnt & AnsPosEnt & EffN \\
\midrule
\endfirsthead
\multicolumn{7}{l}{\tablename~\thetable{} (continued)}\\
\toprule
\rowcolor{SkillPaleBlue} Benchmark & VS/n & NearDup \% & SameVid \% & TemplEnt & AnsPosEnt & EffN \\
\midrule
\endhead
\bottomrule
\endfoot
\mbox{ALLVB} & 0.170 & 24.3 & 0.7 & 1.00 & 0.83 & 34 \\
\mbox{AoTBench} & 0.161 & 35.0 & 0.0 & 0.95 & -- & 32 \\
\mbox{Artifact-Bench} & 0.112 & 49.3 & 0.0 & 0.13 & 1.43 & 22 \\
\mbox{Causal2Needles} & 0.137 & 41.7 & 0.8 & 1.00 & 1.00 & 27 \\
\mbox{CAViAR} & 0.022 & 100.0 & 0.1 & 0.19 & 1.22 & 4 \\
\mbox{CineTechBench} & 0.067 & 87.6 & 0.0 & 0.00 & 0.99 & 8 \\
\mbox{CMU} & 0.163 & 28.3 & 0.1 & 0.59 & 1.66 & 33 \\
\mbox{Colon-Bench} & 0.042 & 99.8 & 0.2 & 1.00 & 1.00 & 8 \\
\mbox{ContPhy} & 0.046 & 100.0 & 0.1 & 0.94 & 1.56 & 9 \\
\mbox{COVER} & 0.258 & 7.8 & 0.0 & 0.95 & 0.85 & 52 \\
\mbox{DSI-Bench} & 0.092 & 67.3 & 0.1 & 0.25 & 1.00 & 18 \\
\mbox{DSR-Bench} & 0.155 & 49.4 & 0.1 & 0.84 & 0.99 & 31 \\
\mbox{E.T. Bench} & 0.062 & 97.1 & 0.0 & 0.68 & 1.11 & 12 \\
\mbox{EgoPlan-Bench} & 0.043 & 97.0 & 0.0 & 0.96 & -- & 9 \\
\mbox{EgoSchema} & 0.119 & 66.0 & 0.0 & 1.00 & 0.99 & 24 \\
\mbox{ExAct} & 0.022 & 100.0 & 0.0 & 0.00 & 0.99 & 4 \\
\mbox{ExpVid} & 0.109 & 69.1 & 0.0 & 0.85 & 1.09 & 22 \\
\mbox{FAVOR-Bench} & 0.195 & 25.7 & 0.1 & 1.00 & 0.99 & 39 \\
\mbox{FineBench} & 0.039 & 100.0 & 25.5 & 0.91 & 0.99 & 6 \\
\mbox{ForeSeaQA} & 0.134 & 84.3 & 4.6 & 0.97 & 0.96 & 27 \\
\mbox{GasVideo-1000} & 0.272 & 39.2 & 0.4 & 0.92 & 1.18 & 54 \\
\mbox{GUI-World} & 0.268 & 11.6 & 0.0 & 1.00 & 0.79 & 54 \\
\mbox{HERBench} & 0.059 & 99.2 & 1.8 & 0.81 & 0.98 & 12 \\
\mbox{HLV-1K} & 0.164 & 46.1 & 0.1 & 0.94 & 0.79 & 33 \\
\mbox{HourVideo} & 0.141 & 73.8 & 2.2 & 0.87 & 0.99 & 28 \\
\mbox{IntPhys 2} & 0.024 & 100.0 & 0.0 & 0.00 & 1.00 & 5 \\
\mbox{iSafetyBench} & 0.050 & 98.6 & 0.1 & 0.12 & 0.98 & 10 \\
\mbox{K9-Bench} & 0.146 & 41.0 & 0.1 & 1.00 & 1.00 & 29 \\
\mbox{KFS-Bench} & 0.255 & 26.8 & 0.3 & 0.98 & 0.99 & 51 \\
\mbox{LEMONADE} & 0.047 & 99.9 & 2.1 & 0.72 & 0.99 & 9 \\
\mbox{LongTimeScope} & 0.054 & 100.0 & 0.1 & 0.67 & -- & 11 \\
\mbox{LongVideo-Reason} & 0.214 & 10.7 & 0.0 & 1.00 & 0.83 & 43 \\
\mbox{LongVideoBench} & 0.303 & 15.3 & 0.1 & 0.96 & 0.99 & 55 \\
\mbox{LVBench} & 0.152 & 67.5 & 0.9 & 1.00 & -- & 30 \\
\mbox{M3-Verse} & 0.019 & 100.0 & 0.4 & 0.95 & 1.16 & 4 \\
\mbox{MLVU} & 0.108 & 76.7 & 0.1 & 0.98 & 0.99 & 22 \\
\mbox{MME-VideoOCR} & 0.347 & 9.5 & 0.0 & 0.99 & 0.99 & 70 \\
\mbox{MMR-V} & 0.202 & 24.2 & 0.4 & 1.00 & 1.00 & 40 \\
\mbox{MMSI-Video-Bench} & 0.158 & 47.9 & 0.1 & 1.00 & 1.09 & 32 \\
\mbox{MMVU} & 0.247 & 25.4 & 0.2 & 1.00 & 0.99 & 50 \\
\mbox{MMWorld} & 0.234 & 52.6 & 0.9 & 0.94 & 0.94 & 47 \\
\mbox{MORSE-500} & 0.025 & 100.0 & 0.0 & 0.00 & -- & 5 \\
\mbox{MotionBench} & 0.222 & 29.4 & 0.0 & 0.96 & -- & 44 \\
\mbox{MVBench} & 0.152 & 67.2 & 0.0 & 0.86 & 0.97 & 30 \\
\mbox{NaturalBench-Video} & 0.217 & 67.3 & 0.2 & 0.98 & -- & 44 \\
\mbox{Neptune} & 0.279 & 13.9 & 0.1 & 0.97 & -- & 56 \\
\mbox{NExT-QA} & 0.262 & 7.1 & 0.1 & 0.99 & -- & 52 \\
\mbox{OSI-Bench} & 0.083 & 91.2 & 0.1 & 0.79 & 0.97 & 16 \\
\mbox{OVO-S-Bench} & 0.082 & 91.3 & 1.0 & 0.93 & 0.99 & 16 \\
\mbox{PAI-Bench-U} & 0.221 & 46.4 & 0.0 & 0.96 & 1.46 & 44 \\
\mbox{Perception Test} & 0.160 & 77.4 & 0.0 & 0.76 & 1.00 & 32 \\
\mbox{PerceptionComp} & 0.168 & 66.6 & 0.5 & 1.00 & -- & 34 \\
\mbox{PhysicalAI} & 0.221 & 45.1 & 0.0 & 0.95 & -- & 44 \\
\mbox{PLM-VideoBench} & 0.242 & 19.3 & 0.0 & 0.95 & -- & 48 \\
\mbox{Q-Bench-Video} & 0.192 & 12.1 & 0.0 & 0.99 & -- & 38 \\
\mbox{ReactHuman} & 0.013 & 100.0 & 0.0 & 0.00 & 1.00 & 2 \\
\mbox{ReVSI} & 0.095 & 93.7 & 0.1 & 0.67 & 0.97 & 19 \\
\mbox{ReXTime} & 0.309 & 6.2 & 0.1 & 1.00 & 1.00 & 62 \\
\mbox{S-EMBER} & 0.216 & 12.0 & 0.0 & 1.00 & 0.99 & 43 \\
\mbox{SAW-Bench} & 0.040 & 97.3 & 0.2 & 0.34 & 0.99 & 8 \\
\mbox{ScaleLong} & 0.266 & 28.4 & 0.3 & 1.00 & 1.00 & 53 \\
\mbox{SciVideoBench} & 0.082 & 76.1 & 0.4 & 0.98 & 0.99 & 16 \\
\mbox{SFI-Bench} & 0.140 & 59.0 & 0.9 & 1.00 & 1.00 & 28 \\
\mbox{SIS-Bench} & 0.056 & 84.9 & 0.1 & 0.56 & 1.00 & 11 \\
\mbox{SITE-Bench} & 0.205 & 46.8 & 0.1 & 0.88 & 1.05 & 41 \\
\mbox{SLVMBench} & 0.201 & 51.3 & 0.0 & 0.99 & 0.99 & 40 \\
\mbox{SmartHome-Bench} & 0.022 & 98.6 & 0.0 & 0.00 & 0.95 & 4 \\
\mbox{SoMi-ToM} & 0.047 & 100.0 & 6.8 & 0.87 & -- & 3 \\
\mbox{SpatialTree} & 0.081 & 78.6 & 0.0 & 0.33 & 0.95 & 16 \\
\mbox{ST-Evidence} & 0.242 & 34.9 & 0.0 & 0.96 & 1.24 & 48 \\
\mbox{STI-Bench} & 0.095 & 87.5 & 0.3 & 0.83 & 0.98 & 19 \\
\mbox{StreamingBench} & 0.205 & 54.2 & 0.1 & 0.93 & 0.99 & 41 \\
\mbox{TaskMeAnything-v1} & 0.080 & 84.4 & 0.0 & 0.74 & 0.99 & 16 \\
\mbox{TempCompass} & 0.172 & 58.3 & 0.3 & 1.00 & -- & 34 \\
\mbox{TemporalBench} & 0.144 & 62.3 & 0.0 & 0.98 & -- & 29 \\
\mbox{TimeBlind} & 0.187 & 80.2 & 0.3 & 0.98 & -- & 33 \\
\mbox{TimeScope} & 0.066 & 99.8 & 0.0 & 0.69 & -- & 13 \\
\mbox{TOC-Bench} & 0.164 & 41.7 & 0.0 & 0.93 & 0.95 & 33 \\
\mbox{TOMATO} & 0.147 & 71.6 & 0.0 & 0.83 & 0.97 & 29 \\
\mbox{TVBench} & 0.093 & 90.3 & 0.0 & 0.61 & 0.95 & 18 \\
\mbox{UCF101-AD} & 0.064 & 92.1 & 0.0 & 0.00 & 0.99 & 13 \\
\mbox{UrbanVideo-Bench} & 0.067 & 94.3 & 0.1 & 0.98 & 1.08 & 13 \\
\mbox{v-HUB} & 0.211 & 22.0 & 0.1 & 1.00 & -- & 42 \\
\mbox{V2P-Bench} & 0.194 & 25.1 & 0.1 & 1.00 & -- & 39 \\
\mbox{VANE-Bench} & 0.156 & 72.1 & 0.2 & 0.82 & 0.80 & 31 \\
\mbox{VAU-Bench} & 0.116 & 69.8 & 0.0 & 0.98 & 0.66 & 23 \\
\mbox{VCR-Bench} & 0.339 & 9.0 & 0.1 & 1.00 & 1.07 & 53 \\
\mbox{VELOCITI} & 0.021 & 99.1 & 0.1 & 0.00 & 1.00 & 4 \\
\mbox{VF-Eval} & 0.184 & 30.1 & 0.1 & 0.80 & 1.00 & 37 \\
\mbox{VGenST-Bench} & 0.252 & 28.2 & 0.1 & 0.81 & 0.94 & 50 \\
\mbox{Video-Holmes} & 0.159 & 36.3 & 0.2 & 0.96 & 0.99 & 32 \\
\mbox{Video-LevelGauge} & 0.341 & 22.1 & 0.2 & 1.00 & 0.99 & 68 \\
\mbox{Video-MME} & 0.267 & 5.3 & 0.1 & 0.99 & 0.99 & 53 \\
\mbox{Video-MME-v2} & 0.309 & 15.0 & 0.1 & 1.00 & 0.99 & 62 \\
\mbox{Video-MMMU} & 0.299 & 22.9 & 0.2 & 1.00 & 0.94 & 60 \\
\mbox{Video-TT} & 0.308 & 12.7 & 0.1 & 1.00 & -- & 62 \\
\mbox{VideoAds} & 0.200 & 27.4 & 0.5 & 0.98 & 1.00 & 40 \\
\mbox{VideoEspresso} & 0.262 & 23.3 & 0.0 & 1.00 & 0.89 & 52 \\
\mbox{VideoEval-Pro} & 0.296 & 16.9 & 0.5 & 1.00 & 1.04 & 59 \\
\mbox{VideoKR} & 0.249 & 25.6 & 0.1 & 1.00 & 0.93 & 50 \\
\mbox{VideoMathQA} & 0.200 & 63.3 & 0.2 & 0.95 & 1.54 & 40 \\
\mbox{VideoOdyssey} & 0.303 & 27.1 & 1.4 & 1.00 & 0.99 & 42 \\
\mbox{VideoSIAH-Eval} & 0.319 & 8.1 & 0.3 & 1.00 & -- & 64 \\
\mbox{VidPair-Halluc} & 0.135 & 75.0 & 0.1 & 0.80 & 0.99 & 27 \\
\mbox{ViMUL-Bench} & 0.284 & 9.6 & 0.1 & 0.93 & 0.99 & 57 \\
\mbox{Vinoground} & 0.060 & 81.4 & 0.0 & 0.00 & 1.00 & 12 \\
\mbox{VITATECS} & 0.062 & 78.8 & 0.0 & 0.00 & 1.00 & 12 \\
\mbox{VKnowU} & 0.227 & 41.7 & 0.1 & 0.80 & 1.60 & 45 \\
\mbox{VNBench} & 0.125 & 57.7 & 0.0 & 0.38 & 1.00 & 25 \\
\mbox{VRU-Accident} & 0.023 & 99.9 & 0.1 & 0.34 & 1.00 & 5 \\
\mbox{VSI-Bench} & 0.075 & 97.8 & 0.6 & 0.49 & 0.99 & 15 \\
\mbox{VSI-Super-Wild} & 0.044 & 97.0 & 0.3 & 0.33 & 0.99 & 9 \\
\mbox{VSTAT} & 0.062 & 100.0 & 0.1 & 0.70 & 0.99 & 12 \\
\mbox{VURB} & 0.219 & 60.9 & 0.9 & 0.93 & 1.03 & 44 \\
\end{xltabular}
\endgroup
\endgroup

\section{Frame-level attacks}
\label{app:t3}
\paragraph{Frame-level protocol.}
The frame-level attackers replace video with a single image or independently generated frame captions. Gemini 3.5 Flash \citep{google2026gemini35flash} reads the normalized video's middle frame, and Claude Opus 5 \citep{anthropic2026opus5} reads the captions. We compare each attack with the reference on their commonly scored items and subtract chance accuracy. The larger margin defines $\varepsilon_{\mathrm{frame}}$. This tests whether a benchmark requires information beyond one image or a sequence of frame descriptions.
\paragraph{Caption construction.}
AuroraCap-7B-IMG \citep{chai2025auroracap} describes each of the 32 sampled frames independently. Each call receives one frame and a fixed instruction to describe the visible objects, people, actions, and setting in one short sentence without guessing what happened before or after. The captioner never receives the question, the options, or the other frames. The reader receives these descriptions in frame order, without timestamps, followed by the question and options. Its instruction begins \texttt{You are given a text description of a video INSTEAD of the video itself.} and requests an answer letter. Table~\ref{tab:e1} separately reports a whole-video caption generated by Qwen3-VL-8B \citep{bai2025qwen3} with the prompt \texttt{Describe this video in detail based on the frames provided, in temporal order.} This whole-video description enters no attack level, and the caption attack uses only the independently generated frame descriptions.
\paragraph{Diagnostic conditions.}
Tables~\ref{tab:p1acc} and~\ref{tab:profiles} provide a broader diagnostic sweep with Qwen3-VL-8B. In Table~\ref{tab:p1acc}, blind supplies no video, 1f uses position 16 of the 32-frame grid, and 32f and 128f give frame counts. The 448px and 224px columns vary resolution, shuf permutes frame order, half1 and half2 restrict input to each video half, and off1 and off2 shift the sampling grid. Table~\ref{tab:profiles} summarizes the derived capability profiles. These conditions help interpret sensitivity to visual detail and temporal coverage, and not all of them are frame-level attacks. In particular, the 224-pixel half-resolution probe remains outside this level. Dashes mark unavailable measurements or derived scores.
\paragraph{What the comparisons show.}
As shown in Figure~\ref{fig:t3strips}, frame captions recover more reference accuracy than a single image. Because the descriptions cover multiple moments in order, this result tests substitution by text with the temporal information intact. The diagnostic sweep also shows that single-frame sufficiency varies across benchmarks. On MMWorld \citep{he2024mmworld}, accuracy is 67\% with one frame and 68\% with 32. On EgoSchema \citep{mangalam2023egoschema}, it rises from 59\% to 78\%. These contrasts motivate separate checks for static visual cues, information spread across frames, and dependence on frame order.
\begingroup

\begingroup
\setlength{\LTcapwidth}{\linewidth}
\setlength{\tabcolsep}{4pt}
\begin{xltabular}{\linewidth}{>{\raggedright\arraybackslash}p{101.38bp}>{\raggedleft\arraybackslash}p{27.76bp}>{\raggedleft\arraybackslash}p{13.27bp}>{\raggedleft\arraybackslash}p{16.90bp}>{\raggedleft\arraybackslash}p{30.91bp}>{\raggedleft\arraybackslash}p{30.91bp}>{\raggedleft\arraybackslash}p{22.36bp}>{\raggedleft\arraybackslash}p{23.55bp}>{\raggedleft\arraybackslash}p{26.43bp}>{\raggedleft\arraybackslash}p{26.43bp}>{\raggedleft\arraybackslash}p{20.84bp}>{\raggedleft\arraybackslash}p{20.84bp}}
\caption{Qwen3-VL-8B condition accuracy (\%), with blank cells for unrun probes.}\label{tab:p1acc}\\
\toprule
\rowcolor{SkillPaleBlue} Benchmark & blind & 1f & 32f & 448px & 224px & 128f & shuf & half1 & half2 & off1 & off2 \\
\midrule
\endfirsthead
\multicolumn{12}{l}{\tablename~\thetable{} (continued)}\\
\toprule
\rowcolor{SkillPaleBlue} Benchmark & blind & 1f & 32f & 448px & 224px & 128f & shuf & half1 & half2 & off1 & off2 \\
\midrule
\endhead
\bottomrule
\endfoot
\mbox{AoTBench} & 49 & 48 & 65 & 65 & 61 & 66 & 51 & 60 & 58 & 63 & 64 \\
\mbox{Artifact-Bench} & 0 & 2 & 1 & 1 & 1 & -- & 1 & -- & -- & -- & -- \\
\mbox{COVER} & 45 & 51 & 59 & 59 & 56 & 52 & 54 & 50 & 46 & 56 & 56 \\
\mbox{DSI-Bench} & 47 & 49 & 59 & 60 & 55 & 75 & 0 & 50 & 50 & 75 & 75 \\
\mbox{DSR-Bench} & 32 & 32 & 32 & 32 & 32 & 34 & 33 & 34 & 33 & 32 & 33 \\
\mbox{ST-Evidence} & 57 & 64 & 79 & 79 & 77 & 69 & 70 & 58 & 69 & 68 & 71 \\
\mbox{EgoSchema} & 36 & 59 & 78 & 77 & 75 & 78 & 76 & 70 & 74 & 74 & 75 \\
\mbox{ExAct} & 38 & 41 & 42 & 42 & 41 & 30 & 33 & 34 & 30 & 33 & 33 \\
\mbox{ExpVid} & 33 & 43 & 46 & 48 & 44 & 43 & 42 & 46 & 40 & 44 & 42 \\
\mbox{FAVOR-Bench} & 40 & 35 & 53 & 51 & 52 & 51 & 41 & 40 & 44 & 51 & 52 \\
\mbox{GasVideo-1000} & 38 & 51 & 69 & 66 & 62 & 70 & 66 & 62 & 60 & 70 & 70 \\
\mbox{HourVideo} & 20 & 20 & 20 & 21 & 20 & 21 & 19 & 18 & 19 & 20 & 20 \\
\mbox{IntPhys 2} & 51 & 47 & 48 & 47 & 50 & -- & -- & -- & -- & -- & -- \\
\mbox{LVBench} & 21 & 24 & 37 & 36 & 38 & 48 & 39 & 33 & 36 & 38 & 38 \\
\mbox{LEMONADE} & 31 & 29 & 34 & 34 & 34 & 32 & 35 & 34 & 33 & 33 & 33 \\
\mbox{LongTimeScope} & 34 & 26 & 27 & 27 & 27 & 30 & 26 & 24 & 26 & 25 & 25 \\
\mbox{VideoSIAH-Eval} & 3 & 2 & 10 & 10 & 8 & 14 & 9 & 8 & 5 & 9 & 9 \\
\mbox{LongVideoBench} & 37 & 39 & 64 & 64 & 66 & 63 & 60 & 56 & 53 & 59 & 62 \\
\mbox{MLVU} & 37 & 40 & 70 & 68 & 66 & 76 & 65 & 54 & 58 & 66 & 65 \\
\mbox{MMR-V} & 38 & 39 & 44 & 45 & 43 & 51 & 46 & 43 & 42 & 46 & 46 \\
\mbox{MMSI-Video-Bench} & 25 & 26 & 27 & 28 & 29 & 28 & 29 & 29 & 28 & 28 & 28 \\
\mbox{MMVU} & 49 & 56 & 71 & 70 & 70 & 68 & 71 & 71 & 62 & 72 & 70 \\
\mbox{MMWorld} & 53 & 67 & 68 & 69 & 68 & 75 & 75 & 72 & 72 & 75 & 76 \\
\mbox{MORSE-500} & 0 & 3 & 27 & 28 & 31 & 23 & 21 & 3 & 26 & 20 & 25 \\
\mbox{MVBench} & 31 & 48 & 66 & 66 & 65 & 71 & 66 & 55 & 61 & 72 & 71 \\
\mbox{MotionBench} & 40 & 51 & 65 & 64 & 61 & 51 & 47 & 42 & 51 & 51 & 51 \\
\mbox{NExT-QA} & 46 & 72 & 81 & 82 & 80 & 85 & 84 & 79 & 80 & 84 & 83 \\
\mbox{Neptune} & 6 & 6 & 24 & 18 & 12 & 18 & 24 & 19 & 18 & 18 & 24 \\
\mbox{OSI-Bench} & 36 & 35 & 41 & 46 & 44 & 42 & 41 & 44 & 39 & 44 & 44 \\
\mbox{PAI-Bench-U} & 26 & 41 & 43 & 48 & 45 & 50 & 41 & 50 & 50 & 50 & 50 \\
\mbox{PerceptionComp} & 3 & 4 & 9 & 10 & 10 & 12 & 8 & 9 & 7 & 9 & 11 \\
\mbox{Perception Test} & 37 & 54 & 70 & 70 & 65 & 72 & 68 & 65 & 65 & 71 & 73 \\
\mbox{Q-Bench-Video} & 53 & 46 & 53 & 54 & 54 & 0 & 45 & 50 & 0 & 50 & 0 \\
\mbox{ReVSI} & 30 & 37 & 48 & 46 & 44 & 48 & 51 & 45 & 39 & 48 & 46 \\
\mbox{SAW-Bench} & 33 & 23 & 31 & 31 & 29 & 30 & 33 & 30 & 25 & 29 & 31 \\
\mbox{SFI-Bench} & 49 & 32 & 57 & 58 & 55 & 59 & 55 & 50 & 49 & 60 & 56 \\
\mbox{SIS-Bench} & 43 & 54 & 65 & 64 & 63 & 54 & 46 & 42 & 45 & 56 & 56 \\
\mbox{STI-Bench} & 38 & 38 & 43 & 43 & 38 & 42 & 40 & 42 & 42 & 43 & 42 \\
\mbox{SciVideoBench} & 18 & 21 & 27 & 27 & 26 & 29 & 27 & 27 & 25 & 29 & 27 \\
\mbox{SITE-Bench} & 27 & 37 & 63 & 62 & 61 & 61 & 60 & 55 & 55 & 63 & 64 \\
\mbox{SoMi-ToM} & 47 & 52 & 87 & 83 & 57 & 83 & 83 & 62 & 83 & 88 & 88 \\
\mbox{SpatialTree} & 37 & 30 & 38 & 38 & 40 & 31 & 28 & 31 & 34 & 28 & 31 \\
\mbox{StreamingBench} & 48 & 48 & 66 & 66 & 63 & 71 & 66 & 63 & 61 & 68 & 67 \\
\mbox{TOMATO} & 23 & 22 & 31 & 29 & 30 & 8 & 25 & 17 & 33 & 17 & 25 \\
\mbox{TVBench} & 37 & 37 & 50 & 51 & 48 & 53 & 37 & 46 & 40 & 50 & 52 \\
\mbox{TempCompass} & 6 & 16 & 24 & 26 & 26 & 38 & 42 & 33 & 42 & 38 & 38 \\
\mbox{TemporalBench} & 55 & 56 & 64 & 65 & 64 & 56 & 60 & 62 & 63 & 63 & 61 \\
\mbox{TimeBlind} & 39 & 37 & 58 & 55 & 55 & 55 & 36 & 27 & 64 & 55 & 55 \\
\mbox{TimeScope} & 40 & 33 & 58 & 59 & 61 & 73 & 57 & 45 & 39 & 59 & 58 \\
\mbox{V2P-Bench} & 34 & 34 & 51 & 52 & 51 & 60 & 48 & 50 & 45 & 54 & 53 \\
\mbox{VCR-Bench} & 38 & 36 & 52 & 53 & 50 & 54 & 48 & 47 & 46 & 53 & 50 \\
\mbox{VITATECS} & 64 & 79 & 87 & 87 & 86 & 89 & 89 & 86 & 86 & 89 & 89 \\
\mbox{VKnowU} & 41 & 45 & 66 & 67 & 66 & 66 & 65 & 62 & 64 & 67 & 65 \\
\mbox{VSI-Bench} & 40 & 34 & 60 & 60 & 60 & 64 & 51 & 55 & 47 & 64 & 56 \\
\mbox{VSI-Super-Wild} & 31 & 28 & 33 & 33 & 33 & 33 & 31 & 34 & 28 & 32 & 32 \\
\mbox{VSTAT} & 38 & 37 & 41 & 39 & 39 & 40 & 40 & 40 & 42 & 40 & 41 \\
\mbox{VURB} & 38 & 29 & 50 & 52 & 52 & 55 & 55 & 43 & 42 & 51 & 50 \\
\mbox{ViMUL-Bench} & 31 & 46 & 63 & 62 & 64 & 66 & 59 & 54 & 60 & 60 & 61 \\
\mbox{VidPair-Halluc} & 27 & 44 & 47 & 47 & 46 & -- & 43 & -- & -- & -- & -- \\
\mbox{Video-Holmes} & 38 & 35 & 49 & 49 & 49 & 54 & 49 & 43 & 46 & 49 & 51 \\
\mbox{Video-MME} & 40 & 42 & 66 & 66 & 63 & 69 & 64 & 59 & 55 & 65 & 64 \\
\mbox{Video-MME-v2} & 17 & 19 & 28 & 27 & 24 & 32 & 26 & 28 & 25 & 29 & 27 \\
\mbox{Video-TT} & 13 & 15 & 8 & 13 & 12 & 5 & 14 & 12 & 14 & 2 & 9 \\
\mbox{VideoEval-Pro} & 26 & 27 & 42 & 41 & 43 & 52 & 43 & 37 & 39 & 41 & 44 \\
\mbox{VideoKR} & 19 & 23 & 36 & 37 & 32 & 39 & 35 & 30 & 33 & 33 & 32 \\
\mbox{Video-MMMU} & 38 & 43 & 58 & 56 & 54 & 58 & 54 & 44 & 53 & 53 & 54 \\
\mbox{VideoMathQA} & 48 & 50 & 55 & 54 & 56 & 56 & 57 & 53 & 54 & 55 & 56 \\
\mbox{VideoOdyssey} & 32 & 26 & 36 & 36 & 32 & 41 & 36 & 32 & 31 & 37 & 40 \\
\mbox{Vinoground} & 49 & 53 & 73 & 73 & 71 & 71 & 29 & 57 & 29 & 71 & 71 \\
\mbox{v-HUB} & 20 & 49 & 59 & 58 & 58 & 52 & 48 & 48 & 41 & 52 & 52 \\
\mbox{ALLVB} & 59 & 56 & 62 & -- & 62 & 67 & 61 & 63 & 64 & 65 & 62 \\
\mbox{Causal2Needles} & 43 & 39 & 52 & -- & 51 & 66 & 53 & 54 & 45 & 54 & 56 \\
\mbox{CAViAR} & 26 & 73 & 71 & -- & 69 & 69 & 70 & 72 & 71 & 72 & 72 \\
\mbox{CG-Bench} & 20 & 26 & 43 & -- & 38 & -- & 42 & -- & -- & 39 & 40 \\
\mbox{CineTechBench} & 32 & 53 & 50 & -- & 50 & -- & 43 & -- & -- & -- & -- \\
\mbox{Colon-Bench} & 27 & 40 & 51 & -- & 47 & 48 & 48 & 48 & 49 & 48 & 50 \\
\mbox{ContPhy} & 18 & 21 & 30 & -- & 32 & -- & 28 & -- & -- & -- & -- \\
\mbox{E.T. Bench} & 22 & 32 & 53 & -- & 48 & 43 & 32 & 42 & 34 & 54 & 54 \\
\mbox{EgoPlan-Bench} & 26 & 26 & 27 & -- & 24 & -- & 24 & -- & -- & 27 & 26 \\
\mbox{FineBench} & 58 & 45 & 54 & -- & 56 & 58 & 55 & 55 & 56 & 54 & 54 \\
\mbox{ForeSeaQA} & 63 & 58 & 60 & -- & 59 & 66 & 61 & 56 & 58 & 64 & 62 \\
\mbox{GUI-World} & 74 & 78 & 89 & -- & 87 & 86 & 85 & 75 & 84 & 88 & 87 \\
\mbox{HERBench} & 47 & 50 & 58 & -- & 58 & 55 & 54 & 52 & 55 & 54 & 56 \\
\mbox{HLV-1K} & 60 & 46 & 64 & -- & 68 & 75 & 67 & 67 & 55 & 66 & 66 \\
\mbox{iSafetyBench} & 16 & 33 & 39 & -- & 39 & -- & 40 & -- & -- & -- & -- \\
\mbox{K9-Bench} & 37 & 39 & 51 & -- & 53 & 57 & 52 & 50 & 48 & 54 & 54 \\
\mbox{KFS-Bench} & 39 & 40 & 58 & -- & 55 & 56 & 53 & 48 & 46 & 54 & 56 \\
\mbox{LongVideo-Reason} & 69 & 65 & 74 & -- & 74 & 77 & 76 & 70 & 71 & 75 & 74 \\
\mbox{M3-Verse} & 34 & 35 & 39 & -- & 35 & 41 & 39 & 34 & 38 & 37 & 41 \\
\mbox{MME-VideoOCR} & 25 & 49 & 65 & -- & 55 & 77 & 67 & 66 & 59 & 67 & 71 \\
\mbox{NaturalBench-Video} & 51 & 55 & 71 & -- & 71 & 83 & 57 & 83 & 67 & 83 & 83 \\
\mbox{OVO-S-Bench} & 35 & 31 & 42 & -- & 42 & 44 & 40 & 38 & 34 & 41 & 40 \\
\mbox{PLM-VideoBench} & 59 & 71 & 74 & -- & 70 & -- & 67 & -- & -- & 67 & 64 \\
\mbox{ReactHuman} & 31 & 34 & 47 & -- & 46 & 78 & 53 & 72 & 39 & 56 & 56 \\
\mbox{ReXTime} & 42 & 66 & 82 & -- & 82 & 80 & 81 & 77 & 80 & 82 & 80 \\
\mbox{S-EMBER} & 16 & 11 & 26 & -- & 21 & 34 & 27 & 21 & 21 & 24 & 28 \\
\mbox{ScaleLong} & 36 & 40 & 52 & -- & 52 & 57 & 51 & 54 & 44 & 56 & 50 \\
\mbox{SLVMBench} & 38 & 43 & 63 & -- & 55 & 68 & 59 & 60 & 50 & 64 & 64 \\
\mbox{SmartHome-Bench} & 37 & 64 & 70 & -- & 70 & 71 & 68 & 73 & 68 & 69 & 69 \\
\mbox{TaskMeAnything-v1} & 26 & 37 & 50 & -- & 48 & 62 & 66 & 59 & 64 & 65 & 68 \\
\mbox{TOC-Bench} & 22 & 22 & 25 & -- & 26 & 26 & 24 & 24 & 29 & 27 & 26 \\
\mbox{UCF101-AD} & 84 & 50 & 47 & -- & 46 & -- & 48 & -- & -- & -- & -- \\
\mbox{UrbanVideo-Bench} & 32 & 37 & 41 & -- & 39 & 41 & 41 & 34 & 41 & 41 & 42 \\
\mbox{VANE-Bench} & 72 & 72 & 76 & -- & 77 & -- & 76 & -- & -- & -- & -- \\
\mbox{VAU-Bench} & 95 & 94 & 98 & -- & 97 & 96 & 96 & 97 & 96 & 96 & 97 \\
\mbox{VELOCITI} & 61 & 60 & 86 & -- & 84 & -- & 82 & -- & -- & -- & -- \\
\mbox{VF-Eval} & 46 & 53 & 59 & -- & 59 & -- & 58 & -- & -- & -- & -- \\
\mbox{VGenST-Bench} & 41 & 50 & 72 & -- & 70 & -- & 65 & -- & -- & -- & -- \\
\mbox{Video-LevelGauge} & 30 & 53 & 89 & -- & 83 & 89 & 89 & 76 & 78 & 90 & 90 \\
\mbox{VideoAds} & 27 & 33 & 64 & -- & 64 & 68 & 65 & 54 & 62 & 65 & 64 \\
\mbox{VideoEspresso} & 37 & 51 & 59 & -- & 60 & 58 & 61 & 58 & 57 & 56 & 59 \\
\mbox{VNBench} & 25 & 23 & 62 & -- & 63 & 79 & 41 & 40 & 24 & 57 & 49 \\
\mbox{VRU-Accident} & 50 & 60 & 68 & -- & 66 & 78 & 73 & 73 & 62 & 71 & 73 \\
\end{xltabular}
\endgroup

\endgroup

\begingroup
\begingroup
\setlength{\LTcapwidth}{\linewidth}
\setlength{\tabcolsep}{4pt}
\begin{xltabular}{\linewidth}{>{\raggedright\arraybackslash}p{103.39bp}>{\raggedleft\arraybackslash}p{32.54bp}>{\raggedleft\arraybackslash}p{32.54bp}>{\raggedleft\arraybackslash}p{32.54bp}>{\raggedleft\arraybackslash}p{32.54bp}>{\raggedleft\arraybackslash}p{28.91bp}>{\raggedleft\arraybackslash}p{41.46bp}>{\raggedleft\arraybackslash}p{28.91bp}>{\raggedright\arraybackslash}p{52.74bp}}
\caption{Qwen3-VL-8B diagnostic capability profiles across audited benchmarks}\label{tab:profiles}\\
\toprule
\rowcolor{SkillPaleBlue} Benchmark & $m$ & $S$ & $T_{coarse}$ & $O$ & $L$ & PosBias & $\Phi_{flip}$ & PriorDom \\
\midrule
\endfirsthead
\multicolumn{9}{l}{\tablename~\thetable{} (continued)}\\
\toprule
\rowcolor{SkillPaleBlue} Benchmark & $m$ & $S$ & $T_{coarse}$ & $O$ & $L$ & PosBias & $\Phi_{flip}$ & PriorDom \\
\midrule
\endhead
\bottomrule
\endfoot
\mbox{AoTBench} & 0.166 & 0.245 & 0.089 & 0.801 & 0.985 & 0.020 & 0.069 & no \\
\mbox{Artifact-Bench} & 0.008 & -- & -- & -- & -- & -- & -- & yes \\
\mbox{COVER} & 0.140 & 0.219 & -0.413 & 0.275 & 0.967 & 0.038 & 0.019 & no \\
\mbox{DSI-Bench} & 0.117 & 0.314 & -0.000 & 6.407 & 0.667 & 0.000 & 0.000 & no \\
\mbox{DSR-Bench} & -0.007 & -- & -- & -- & 0.979 & 0.013 & 0.053 & yes \\
\mbox{ST-Evidence} & 0.221 & -0.024 & -0.073 & 0.000 & 0.966 & -0.113 & 0.057 & no \\
\mbox{EgoSchema} & 0.420 & 0.071 & -0.000 & 0.032 & 0.987 & -0.033 & 0.117 & no \\
\mbox{ExAct} & 0.040 & -- & -- & -- & 1.000 & 0.038 & 0.048 & yes \\
\mbox{ExpVid} & 0.129 & -0.244 & 0.108 & 0.000 & 0.967 & 0.056 & 0.028 & no \\
\mbox{FAVOR-Bench} & 0.130 & 0.146 & 0.095 & 0.665 & 0.963 & -0.043 & 0.086 & no \\
\mbox{GasVideo-1000} & 0.306 & 0.117 & -0.012 & 0.143 & 0.953 & 0.015 & 0.080 & no \\
\mbox{HourVideo} & 0.003 & -- & -- & -- & 0.787 & -0.010 & 0.140 & yes \\
\mbox{IntPhys 2} & -0.037 & -- & -- & -- & -- & -- & -- & yes \\
\mbox{LVBench} & 0.167 & -0.020 & 0.620 & -0.080 & 0.812 & -0.023 & 0.370 & no \\
\mbox{LEMONADE} & 0.030 & -- & -- & -- & 0.932 & 0.013 & 0.150 & yes \\
\mbox{LongTimeScope} & -0.077 & -- & -- & -- & 0.825 & -0.013 & 0.127 & yes \\
\mbox{VideoSIAH-Eval} & 0.065 & -0.078 & 0.667 & 0.133 & 0.636 & 0.030 & 0.104 & no \\
\mbox{LongVideoBench} & 0.273 & 0.021 & 0.071 & 0.024 & 0.925 & 0.033 & 0.221 & no \\
\mbox{MLVU} & 0.330 & 0.057 & 0.182 & 0.131 & 0.871 & -0.047 & 0.233 & no \\
\mbox{MMR-V} & 0.067 & -0.052 & 0.684 & 0.105 & 0.916 & 0.007 & 0.131 & no \\
\mbox{MMSI-Video-Bench} & 0.023 & -- & -- & -- & 0.955 & 0.008 & 0.114 & yes \\
\mbox{MMVU} & 0.218 & 0.049 & -0.168 & 0.056 & 0.949 & 0.085 & 0.043 & no \\
\mbox{MMWorld} & 0.151 & -0.050 & -0.000 & -0.041 & 0.967 & -0.006 & 0.055 & no \\
\mbox{MORSE-500} & 0.269 & -0.138 & 0.122 & -0.061 & 0.917 & -0.230 & 0.115 & no \\
\mbox{MVBench} & 0.350 & 0.038 & -0.000 & 0.157 & 0.942 & -0.062 & 0.055 & no \\
\mbox{MotionBench} & 0.250 & 0.027 & -0.000 & 0.151 & 1.000 & -0.094 & 0.038 & no \\
\mbox{NExT-QA} & 0.353 & -0.019 & 0.055 & -0.033 & 0.986 & -0.012 & 0.062 & no \\
\mbox{Neptune} & 0.176 & 0.354 & -0.333 & 0.000 & 1.000 & 0.062 & 0.059 & no \\
\mbox{OSI-Bench} & 0.044 & -- & -- & -- & 0.909 & 0.050 & 0.037 & yes \\
\mbox{PAI-Bench-U} & 0.174 & -0.311 & -0.000 & 0.111 & 1.000 & 0.000 & 0.000 & no \\
\mbox{PerceptionComp} & 0.064 & -0.333 & 0.526 & 0.105 & 0.778 & 0.017 & 0.087 & no \\
\mbox{Perception Test} & 0.327 & -0.020 & 0.050 & 0.076 & 0.942 & 0.000 & 0.062 & no \\
\mbox{Q-Bench-Video} & 0.000 & -- & -- & -- & 1.000 & 0.500 & 0.500 & yes \\
\mbox{ReVSI} & 0.180 & 0.096 & -0.000 & -0.182 & 0.845 & 0.057 & 0.172 & no \\
\mbox{SAW-Bench} & -0.015 & -- & -- & -- & 0.859 & 0.048 & 0.051 & yes \\
\mbox{SFI-Bench} & 0.087 & -0.077 & 0.192 & 0.231 & 0.930 & 0.007 & 0.217 & no \\
\mbox{SIS-Bench} & 0.217 & 0.031 & -0.044 & 0.440 & 0.948 & -0.029 & 0.009 & no \\
\mbox{STI-Bench} & 0.043 & -- & -- & -- & 0.945 & 0.000 & 0.054 & yes \\
\mbox{SciVideoBench} & 0.090 & 0.049 & 0.222 & -0.037 & 0.864 & 0.017 & 0.187 & no \\
\mbox{SITE-Bench} & 0.354 & 0.033 & 0.013 & 0.038 & 0.925 & -0.004 & 0.196 & no \\
\mbox{SoMi-ToM} & 0.383 & 0.783 & -0.087 & 0.087 & 1.000 & -0.217 & 0.133 & no \\
\mbox{SpatialTree} & 0.016 & -- & -- & -- & 0.889 & -0.035 & 0.190 & yes \\
\mbox{StreamingBench} & 0.183 & 0.020 & 0.260 & 0.040 & 0.950 & 0.026 & 0.187 & no \\
\mbox{TOMATO} & 0.077 & 0.301 & -0.000 & -2.167 & 1.000 & -0.167 & 0.167 & no \\
\mbox{TVBench} & 0.133 & -0.049 & 0.362 & 0.797 & 0.899 & 0.063 & 0.111 & no \\
\mbox{TempCompass} & 0.179 & -0.143 & -0.000 & -0.232 & 1.000 & -0.083 & 0.000 & no \\
\mbox{TemporalBench} & 0.088 & -0.239 & -0.131 & -0.261 & 0.980 & -0.011 & 0.081 & no \\
\mbox{TimeBlind} & 0.227 & 0.121 & -0.000 & 0.800 & 1.000 & -0.364 & 0.000 & no \\
\mbox{TimeScope} & 0.183 & -0.051 & 0.800 & 0.054 & 0.840 & 0.063 & 0.193 & no \\
\mbox{V2P-Bench} & 0.167 & -0.286 & 0.528 & 0.192 & 0.915 & 0.044 & 0.228 & no \\
\mbox{VCR-Bench} & 0.138 & -0.083 & 0.128 & 0.255 & 0.912 & 0.007 & 0.165 & no \\
\mbox{VITATECS} & 0.237 & 0.042 & -0.000 & 0.000 & 1.000 & 0.000 & 0.000 & no \\
\mbox{VKnowU} & 0.246 & -0.118 & 0.038 & 0.000 & 0.971 & -0.019 & 0.100 & no \\
\mbox{VSI-Bench} & 0.199 & 0.035 & 0.172 & 0.483 & 0.886 & 0.082 & 0.294 & no \\
\mbox{VSI-Super-Wild} & 0.024 & -- & -- & -- & 0.868 & 0.059 & 0.194 & yes \\
\mbox{VSTAT} & 0.028 & -- & -- & -- & 0.923 & -0.023 & 0.061 & yes \\
\mbox{VURB} & 0.112 & -0.507 & 0.426 & -0.480 & 0.845 & 0.006 & 0.327 & no \\
\mbox{ViMUL-Bench} & 0.324 & -0.018 & 0.153 & 0.066 & 0.919 & -0.057 & 0.142 & no \\
\mbox{VidPair-Halluc} & 0.202 & 0.000 & -- & 0.191 & -- & -- & -- & no \\
\mbox{Video-Holmes} & 0.113 & 0.000 & 0.471 & 0.000 & 0.925 & -0.030 & 0.213 & no \\
\mbox{Video-MME} & 0.263 & 0.023 & 0.115 & 0.077 & 0.918 & 0.040 & 0.219 & no \\
\mbox{Video-MME-v2} & 0.105 & 0.069 & 0.414 & 0.172 & 0.870 & 0.029 & 0.174 & no \\
\mbox{Video-TT} & -0.050 & -- & -- & -- & 0.500 & -0.023 & 0.093 & yes \\
\mbox{VideoEval-Pro} & 0.163 & -0.020 & 0.592 & -0.020 & 0.858 & -0.017 & 0.323 & no \\
\mbox{VideoKR} & 0.168 & -0.048 & 0.132 & 0.099 & 0.924 & -0.028 & 0.116 & no \\
\mbox{Video-MMMU} & 0.201 & 0.102 & -0.017 & 0.200 & 0.948 & -0.087 & 0.120 & no \\
\mbox{VideoMathQA} & 0.063 & 0.053 & 0.112 & -0.337 & 0.929 & -0.007 & 0.110 & no \\
\mbox{VideoOdyssey} & 0.043 & -- & -- & -- & 0.824 & 0.007 & 0.286 & yes \\
\mbox{Vinoground} & 0.237 & 0.070 & -0.000 & 1.811 & 0.800 & 0.286 & 0.000 & no \\
\mbox{v-HUB} & 0.392 & 0.049 & -0.000 & 0.094 & 1.000 & 0.074 & 0.000 & no \\
\mbox{ALLVB} & 0.027 & -- & -- & -- & 0.941 & -0.007 & 0.217 & yes \\
\mbox{Causal2Needles} & 0.090 & 0.111 & 1.482 & -0.111 & 0.822 & 0.097 & 0.427 & no \\
\mbox{CAViAR} & 0.453 & 0.044 & -0.051 & 0.029 & 0.995 & 0.007 & 0.020 & no \\
\mbox{CG-Bench} & 0.233 & 0.214 & -- & 0.071 & -- & -- & 0.225 & no \\
\mbox{CineTechBench} & 0.212 & 0.042 & -- & 0.333 & -- & -- & -- & no \\
\mbox{Colon-Bench} & 0.243 & 0.164 & -0.062 & 0.062 & 0.969 & -0.011 & 0.095 & no \\
\mbox{ContPhy} & 0.123 & -0.150 & -- & 0.150 & -- & -- & -- & no \\
\mbox{E.T. Bench} & 0.307 & 0.152 & -0.301 & 0.670 & 0.791 & 0.079 & 0.151 & no \\
\mbox{EgoPlan-Bench} & 0.004 & -- & -- & -- & -- & -- & 0.015 & yes \\
\mbox{FineBench} & 0.019 & -- & -- & -- & 0.964 & -0.013 & 0.181 & yes \\
\mbox{ForeSeaQA} & -0.025 & -- & -- & -- & 0.967 & -0.025 & 0.126 & yes \\
\mbox{GUI-World} & 0.157 & 0.170 & -0.122 & 0.182 & 0.967 & -0.086 & 0.057 & no \\
\mbox{HERBench} & 0.107 & 0.000 & -0.219 & 0.344 & 0.925 & -0.033 & 0.150 & no \\
\mbox{HLV-1K} & 0.044 & -- & -- & -- & 0.905 & 0.115 & 0.270 & yes \\
\mbox{iSafetyBench} & 0.234 & 0.014 & -- & -0.014 & -- & -- & -- & no \\
\mbox{K9-Bench} & 0.146 & -0.140 & 0.364 & -0.024 & 0.959 & 0.018 & 0.152 & no \\
\mbox{KFS-Bench} & 0.193 & 0.155 & -0.000 & 0.166 & 0.892 & 0.018 & 0.232 & no \\
\mbox{LongVideo-Reason} & 0.051 & 0.000 & 0.533 & -0.400 & 0.977 & -0.014 & 0.125 & no \\
\mbox{M3-Verse} & 0.053 & 0.812 & 0.236 & 0.158 & 0.894 & -0.038 & 0.210 & no \\
\mbox{MME-VideoOCR} & 0.400 & 0.242 & 0.161 & 0.107 & 0.955 & 0.065 & 0.150 & no \\
\mbox{NaturalBench-Video} & 0.200 & 0.033 & -0.000 & 0.700 & 1.000 & 0.167 & 0.000 & no \\
\mbox{OVO-S-Bench} & 0.069 & 0.053 & 0.316 & 0.263 & 0.845 & 0.036 & 0.218 & no \\
\mbox{PLM-VideoBench} & 0.158 & 0.250 & -- & -0.175 & -- & -- & 0.028 & no \\
\mbox{ReactHuman} & 0.158 & 0.085 & -0.000 & -0.340 & 0.929 & 0.333 & 0.278 & no \\
\mbox{ReXTime} & 0.406 & 0.000 & -0.062 & 0.031 & 0.995 & -0.029 & 0.054 & no \\
\mbox{S-EMBER} & 0.107 & 0.531 & 0.719 & -0.062 & 0.810 & 0.000 & 0.221 & no \\
\mbox{ScaleLong} & 0.164 & 0.028 & 0.307 & 0.061 & 0.890 & 0.098 & 0.261 & no \\
\mbox{SLVMBench} & 0.253 & 0.303 & 0.198 & 0.158 & 0.942 & 0.104 & 0.171 & no \\
\mbox{SmartHome-Bench} & 0.330 & -0.010 & 0.054 & 0.013 & 1.000 & 0.049 & 0.022 & no \\
\mbox{TaskMeAnything-v1} & 0.243 & 0.082 & -0.350 & 0.175 & 0.924 & -0.053 & 0.117 & no \\
\mbox{TOC-Bench} & 0.036 & -- & -- & -- & 0.927 & -0.049 & 0.045 & yes \\
\mbox{UCF101-AD} & -0.367 & -- & -- & -- & -- & -- & -- & yes \\
\mbox{UrbanVideo-Bench} & 0.090 & 0.259 & 0.039 & 0.039 & 0.914 & -0.073 & 0.063 & no \\
\mbox{VANE-Bench} & 0.040 & -- & -- & -- & -- & -- & -- & yes \\
\mbox{VAU-Bench} & 0.024 & -- & -- & -- & 0.996 & 0.013 & 0.025 & yes \\
\mbox{VELOCITI} & 0.243 & 0.055 & -- & 0.164 & -- & -- & -- & no \\
\mbox{VF-Eval} & 0.123 & 0.000 & -- & 0.027 & -- & -- & -- & no \\
\mbox{VGenST-Bench} & 0.310 & 0.043 & -- & 0.204 & -- & -- & -- & no \\
\mbox{Video-LevelGauge} & 0.587 & 0.102 & -0.006 & 0.018 & 1.000 & -0.014 & 0.038 & no \\
\mbox{VideoAds} & 0.377 & 0.009 & 0.106 & -0.018 & 0.979 & -0.077 & 0.147 & no \\
\mbox{VideoEspresso} & 0.214 & -0.034 & 0.055 & -0.192 & 1.000 & 0.006 & 0.106 & no \\
\mbox{VNBench} & 0.368 & -0.018 & 0.754 & 0.286 & 0.759 & 0.158 & 0.330 & no \\
\mbox{VRU-Accident} & 0.177 & 0.094 & 0.377 & -0.126 & 0.969 & 0.111 & 0.044 & no \\
\end{xltabular}
\endgroup
\endgroup

\section{Order-level attacks}
\label{app:t4}
\paragraph{Experiments.}
Four probes perturb the video along time. The shuffle probe reads the 32 uniform frames in a random order. The window probe reads 32 frames from one tenth of the video. Two half probes of the diagnostic model read 16 frames from the first or the second half. A frame-budget ladder reads 8, 32, and 128 frames, extended to 512 and 600 frames by Claude Opus 5 \citep{anthropic2026opus5} on the long-video set. The shuffle and window runs of Gemini 3.5 Flash \citep{google2026gemini35flash} and Qwen3-VL-8B \citep{bai2025qwen3} form one attacker set, and $\varepsilon_{\mathrm{order}}$ is the largest margin among them. Table~\ref{tab:e1b} gives the probes per benchmark.
\paragraph{Settings.}
Gemini's shuffle uses \texttt{random.Random(42)} on 32 frames. Qwen3-VL-8B uses the per-item seed \texttt{42|qid}. The Gemini window starts uniformly in $[0,0.9]$ and covers a tenth of the clip. Qwen3-VL-8B uses the first tenth. Both sample 32 frames within the window, or all available frames when fewer exist. Half-video probes sample 16 frames, and offset probes shift the 32-cell grid by a random phase. Frame-budget probes sample uniformly over the complete clip and use all frames when the requested count exceeds availability. Retention divides perturbed accuracy by the same model's 32-frame accuracy on matched items. Tables retain ratios above one. Figures cap values at 1.15 and identify them with hollow markers.
\paragraph{Results.}
In Table~\ref{tab:e1b}, the window probe reads 32 frames inside one random contiguous tenth of the video, and retention divides probe accuracy by 32-frame accuracy on the same items. Table~\ref{tab:lvbudget} reports the Claude Opus 5 frame-budget sweep on the long-video set, the benchmarks whose sampled videos average at least 300 seconds, with the same items, prompt, and scorer and uniform frames over the whole video. Median frames is the median number of frames sent at the largest budget that ran, since a video holds at most 1,024 stored frames and short videos send every frame once. Saturation is the first budget within 5\% of the maximum over the budgets that ran. The request limit ends the reference sweep at 600 frames.
\begingroup
\begingroup
\setlength{\LTcapwidth}{\linewidth}
\setlength{\tabcolsep}{4pt}
\begin{xltabular}{\linewidth}{>{\raggedright\arraybackslash}p{124.51bp}>{\raggedleft\arraybackslash}p{36.39bp}>{\raggedleft\arraybackslash}p{40.03bp}>{\raggedleft\arraybackslash}p{45.48bp}>{\raggedleft\arraybackslash}p{85.39bp}>{\raggedleft\arraybackslash}p{77.78bp}}
\caption{Frame-budget accuracy (\%) and temporal-retention probes on 88 benchmarks}\label{tab:e1b}\\
\toprule
\rowcolor{SkillPaleBlue} Benchmark & 8f & 32f & 128f & Window ret. & Shuffle ret. \\
\midrule
\endfirsthead
\multicolumn{6}{l}{\tablename~\thetable{} (continued)}\\
\toprule
\rowcolor{SkillPaleBlue} Benchmark & 8f & 32f & 128f & Window ret. & Shuffle ret. \\
\midrule
\endhead
\bottomrule
\endfoot
\mbox{VNBench} & 41 & 74 & 90 & 0.37 & 0.71 \\
\mbox{v-HUB} & 70 & 64 & 69 & 1.70 & 2.15 \\
\mbox{SoMi-ToM} & 76 & 93 & 95 & 0.71 & 1.01 \\
\mbox{VGenST-Bench} & 87 & 88 & 87 & 0.76 & 0.84 \\
\mbox{MME-VideoOCR} & 79 & 84 & 83 & 0.67 & 0.94 \\
\mbox{UCF101-AD} & 45 & 47 & 46 & 0.94 & 0.97 \\
\mbox{Video-LevelGauge} & 85 & 89 & 89 & 0.80 & 0.98 \\
\mbox{Vinoground} & 80 & 82 & 78 & 0.72 & 0.66 \\
\mbox{ViMUL-Bench} & 82 & 80 & 76 & 0.79 & 0.90 \\
\mbox{Video-TT} & 51 & 61 & 64 & 0.55 & 0.97 \\
\mbox{CAViAR} & 77 & 78 & 78 & 0.96 & 0.99 \\
\mbox{VideoKR} & 47 & 55 & 66 & 0.66 & 1.06 \\
\mbox{VideoAds} & 74 & 81 & 85 & 0.78 & 0.95 \\
\mbox{NaturalBench-Video} & 83 & 81 & 82 & 0.74 & 0.66 \\
\mbox{MVBench} & 55 & 69 & 64 & 0.56 & 0.75 \\
\mbox{TaskMeAnything-v1} & 61 & 62 & 63 & 0.80 & 0.96 \\
\mbox{GasVideo-1000} & 64 & 75 & 84 & 0.76 & 1.09 \\
\mbox{AoTBench} & 68 & 77 & 69 & 0.51 & 0.70 \\
\mbox{VELOCITI} & 94 & 94 & 94 & 0.81 & 0.89 \\
\mbox{VRU-Accident} & 73 & 76 & 78 & 0.94 & 1.00 \\
\mbox{iSafetyBench} & 43 & 42 & 41 & 0.89 & 0.97 \\
\mbox{PAI-Bench-U} & 57 & 61 & 59 & -- & -- \\
\mbox{E.T. Bench} & 54 & 55 & 47 & 0.63 & 0.67 \\
\mbox{TVBench} & 51 & 63 & 58 & 0.55 & 0.48 \\
\mbox{VKnowU} & 69 & 71 & 70 & 0.66 & 0.97 \\
\mbox{VSI-Bench} & 43 & 61 & -- & 0.39 & 0.54 \\
\mbox{ContPhy} & 58 & 56 & 58 & 0.87 & 0.88 \\
\mbox{VURB} & 48 & 62 & 66 & 0.91 & 1.14 \\
\mbox{SITE-Bench} & 48 & 65 & 62 & 0.60 & 1.00 \\
\mbox{Video-MMMU} & 60 & 70 & 74 & 0.42 & 0.97 \\
\mbox{TimeBlind} & 77 & 71 & 67 & 0.51 & 0.61 \\
\mbox{TOMATO} & 38 & 43 & 39 & 0.69 & 0.66 \\
\mbox{VCR-Bench} & 55 & 66 & 69 & -- & -- \\
\mbox{TempCompass} & 29 & 42 & 34 & 0.80 & 0.83 \\
\mbox{CineTechBench} & 66 & 60 & 63 & 0.79 & 0.71 \\
\mbox{Colon-Bench} & 72 & 74 & 75 & 0.85 & 0.96 \\
\mbox{ReactHuman} & 52 & 50 & 51 & 0.94 & 1.05 \\
\mbox{Video-MME} & 60 & 70 & 81 & 0.75 & 1.05 \\
\mbox{Video-MME-v2} & 26 & 36 & 46 & 0.63 & 1.28 \\
\mbox{LVBench} & 30 & 46 & 58 & 0.72 & 0.91 \\
\mbox{SLVMBench} & 68 & 80 & 81 & 0.81 & 0.97 \\
\mbox{TimeScope} & 26 & 53 & 63 & 0.35 & 0.90 \\
\mbox{MLVU} & 42 & 65 & 75 & 0.53 & 0.98 \\
\mbox{KFS-Bench} & 60 & 67 & 70 & 0.77 & 0.90 \\
\mbox{MMWorld} & 74 & 74 & 88 & 0.93 & 1.14 \\
\mbox{V2P-Bench} & 52 & 53 & 65 & 0.88 & 1.04 \\
\mbox{StreamingBench} & 47 & 62 & 69 & 0.76 & 1.00 \\
\mbox{ScaleLong} & 54 & 64 & 68 & 0.85 & 0.96 \\
\mbox{SmartHome-Bench} & 78 & 78 & 77 & 0.88 & 1.01 \\
\mbox{TOC-Bench} & 49 & 50 & 49 & 0.77 & 0.87 \\
\mbox{SFI-Bench} & 41 & 69 & 69 & 0.60 & 0.86 \\
\mbox{M3-Verse} & 35 & 40 & 42 & 0.82 & 0.93 \\
\mbox{PLM-VideoBench} & 80 & 85 & 80 & 0.96 & 0.99 \\
\mbox{UrbanVideo-Bench} & 51 & 54 & 52 & 0.81 & 0.88 \\
\mbox{SpatialTree} & 61 & 62 & 73 & 0.93 & 1.36 \\
\mbox{GUI-World} & 90 & 92 & 91 & 0.87 & 0.97 \\
\mbox{VANE-Bench} & 87 & 88 & 89 & 0.99 & 0.98 \\
\mbox{LongVideoBench} & 50 & 60 & 67 & 0.61 & 0.83 \\
\mbox{SAW-Bench} & 33 & 38 & 28 & 0.88 & 0.73 \\
\mbox{K9-Bench} & 70 & 75 & 78 & 0.83 & 0.97 \\
\mbox{CG-Bench} & 55 & 57 & 57 & 0.87 & 1.06 \\
\mbox{VideoSIAH-Eval} & 7 & 13 & 21 & 0.44 & 0.91 \\
\mbox{FAVOR-Bench} & 48 & 55 & 62 & 0.46 & 0.85 \\
\mbox{MMR-V} & 42 & 56 & 52 & 1.00 & 0.92 \\
\mbox{COVER} & 58 & 59 & 61 & 0.91 & 0.98 \\
\mbox{S-EMBER} & 20 & 37 & 46 & 0.77 & 0.91 \\
\mbox{HERBench} & 53 & 56 & 64 & 0.95 & 1.06 \\
\mbox{ReXTime} & 92 & 92 & 88 & 0.95 & 1.00 \\
\mbox{PerceptionComp} & 10 & 13 & 15 & 0.31 & 1.05 \\
\mbox{OVO-S-Bench} & 49 & 47 & 55 & 0.93 & 1.05 \\
\mbox{Causal2Needles} & 56 & 64 & 70 & 0.92 & 1.04 \\
\mbox{IntPhys 2} & 56 & 57 & 48 & 0.81 & 1.02 \\
\mbox{ExAct} & 67 & 69 & 64 & 0.98 & 0.93 \\
\mbox{TemporalBench} & 74 & 77 & 74 & 0.93 & 0.98 \\
\mbox{VideoEspresso} & 79 & 80 & 80 & 0.97 & 0.99 \\
\mbox{VSI-Super-Wild} & 22 & 32 & 32 & 0.63 & 0.63 \\
\mbox{ForeSeaQA} & 69 & 73 & 74 & 0.94 & 1.01 \\
\mbox{HLV-1K} & 68 & 69 & 76 & 0.95 & 0.97 \\
\mbox{DSI-Bench} & 70 & 53 & 74 & 1.21 & 1.27 \\
\mbox{LongVideo-Reason} & 74 & 77 & 79 & 0.96 & 1.03 \\
\mbox{VF-Eval} & 85 & 84 & 85 & 0.98 & 1.02 \\
\mbox{DSR-Bench} & 26 & 24 & 41 & 1.61 & 1.37 \\
\mbox{ALLVB} & 69 & 73 & 72 & 1.00 & 1.01 \\
\mbox{FineBench} & 62 & 62 & 68 & 0.97 & 0.97 \\
\mbox{HourVideo} & 16 & 21 & 21 & 0.82 & 0.81 \\
\mbox{VAU-Bench} & 96 & 94 & 95 & 1.00 & 1.01 \\
\mbox{MMSI-Video-Bench} & 34 & 28 & 25 & 0.86 & 1.12 \\
\mbox{EgoPlan-Bench} & 26 & 29 & 26 & 0.89 & 0.96 \\
\mbox{VideoMathQA} & 34 & 52 & 67 & 0.39 & 1.00 \\
\mbox{VideoOdyssey} & 28 & 35 & 42 & 0.69 & 0.94 \\
\mbox{LongTimeScope} & 12 & 16 & 19 & 0.96 & 1.06 \\
\end{xltabular}
\endgroup
\endgroup

\begingroup
\begingroup
\setlength{\LTcapwidth}{\linewidth}
\setlength{\tabcolsep}{4pt}
\begin{xltabular}{\linewidth}{>{\raggedright\arraybackslash}p{101.66bp}>{\raggedleft\arraybackslash}p{26.16bp}>{\raggedleft\arraybackslash}p{45.76bp}>{\raggedleft\arraybackslash}p{24.34bp}>{\raggedleft\arraybackslash}p{29.80bp}>{\raggedleft\arraybackslash}p{29.80bp}>{\raggedleft\arraybackslash}p{29.80bp}>{\raggedleft\arraybackslash}p{71.00bp}>{\raggedleft\arraybackslash}p{27.27bp}}
\caption{Long-video accuracy (\%) at 32, 128, 512, and 600 frames with Claude Opus 5.}\label{tab:lvbudget}\\
\toprule
\rowcolor{SkillPaleBlue} Benchmark & $n$ & Chance & 32f & 128f & 512f & 600f & Med. frames & Sat. \\
\midrule
\endfirsthead
\multicolumn{9}{l}{\tablename~\thetable{} (continued)}\\
\toprule
\rowcolor{SkillPaleBlue} Benchmark & $n$ & Chance & 32f & 128f & 512f & 600f & Med. frames & Sat. \\
\midrule
\endhead
\bottomrule
\endfoot
\mbox{HourVideo} & 300 & 20 & 25 & 26 & 26 & 23 & 600 & 32 \\
\mbox{LEMONADE} & 300 & 25 & 38 & 38 & 39 & 36 & 600 & 32 \\
\mbox{LongTimeScope} & 300 & 17 & 28 & 29 & 40 & 45 & 600 & 512 \\
\mbox{LongVideoBench} & 300 & 21 & 75 & 74 & 79 & 80 & 600 & 32 \\
\mbox{LVBench} & 300 & 25 & 57 & 64 & 74 & 75 & 600 & 512 \\
\mbox{MLVU} & 300 & 25 & 75 & 86 & 87 & 87 & 600 & 128 \\
\mbox{MMR-V} & 300 & 10 & 62 & 64 & 65 & 67 & 394 & 128 \\
\mbox{SciVideoBench} & 300 & 10 & 65 & 69 & 65 & 67 & 600 & 32 \\
\mbox{StreamingBench} & 300 & 25 & 69 & 75 & 75 & 76 & 600 & 128 \\
\mbox{TimeScope} & 300 & 17 & 64 & 78 & 89 & 91 & 600 & 512 \\
\mbox{V2P-Bench} & 300 & 25 & 57 & 64 & 70 & 70 & 360 & 512 \\
\mbox{Video-MME} & 300 & 25 & 81 & 83 & 85 & 85 & 600 & 32 \\
\mbox{Video-MME-v2} & 300 & 13 & 46 & 58 & 61 & 60 & 600 & 128 \\
\mbox{Video-MMMU} & 299 & 12 & 81 & 80 & 81 & 83 & 600 & 32 \\
\mbox{VideoEval-Pro} & 300 & 25 & 55 & 67 & 76 & 76 & 600 & 512 \\
\mbox{VideoKR} & 185 & 14 & 65 & 68 & 71 & 70 & 600 & 128 \\
\mbox{VideoOdyssey} & 320 & 25 & 49 & 54 & 61 & 58 & 600 & 512 \\
\mbox{VSI-Super-Wild} & 252 & 25 & 36 & 39 & 38 & 39 & 600 & 32 \\
\mbox{VURB} & 209 & 27 & 65 & 69 & 74 & 78 & 209 & 512 \\
\mbox{ALLVB} & 300 & 20 & 73 & 76 & 73 & 76 & 600 & 32 \\
\mbox{Causal2Needles} & 300 & 25 & 64 & 74 & 80 & 81 & 600 & 512 \\
\mbox{FineBench} & 300 & 25 & 63 & 63 & 56 & 56 & 600 & 32 \\
\mbox{ForeSeaQA} & 199 & 25 & 70 & 76 & 76 & 74 & 463 & 128 \\
\mbox{HERBench} & 300 & 20 & 58 & 63 & 63 & 65 & 360 & 128 \\
\mbox{HLV-1K} & 300 & 25 & 68 & 76 & 79 & 76 & 600 & 128 \\
\mbox{KFS-Bench} & 300 & 21 & 65 & 71 & 76 & 78 & 600 & 512 \\
\mbox{LongVideo-Reason} & 300 & 27 & 78 & 78 & 80 & 80 & 600 & 32 \\
\mbox{OVO-S-Bench} & 275 & 30 & 49 & 54 & 56 & 54 & 600 & 128 \\
\mbox{S-EMBER} & 299 & 20 & 39 & 48 & 54 & 49 & 600 & 512 \\
\mbox{ScaleLong} & 300 & 25 & 63 & 69 & 77 & 75 & 600 & 512 \\
\end{xltabular}
\endgroup
\endgroup

\section{Breaking levels by fine capability}
\label{app:findings}
\begin{figure}[!htbp]
\centering
\includegraphics[width=\linewidth]{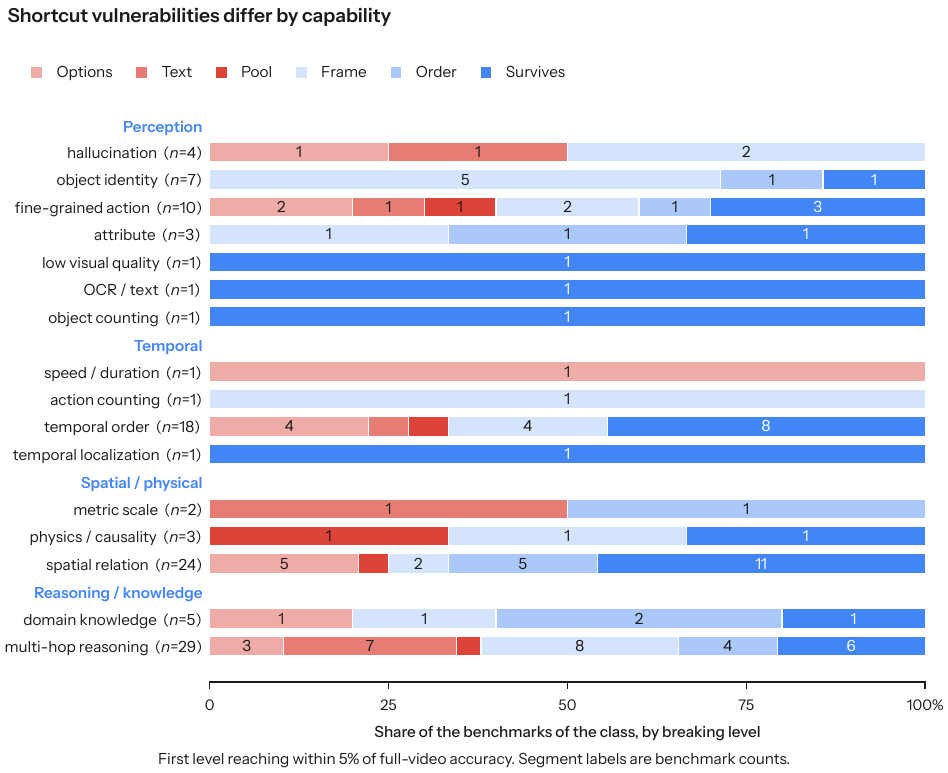} %
\caption{First breaking levels by dominant fine capability, with survivors in dark blue.}
\label{fig:f3claimsfine}
\end{figure}
We resolve the first breaking level by the dominant fine capability of each benchmark within the four groups. As shown in Figure~\ref{fig:f3claimsfine}, the level at which a claim breaks follows the evidence the claim needs rather than the group it belongs to. Object identity breaks at the frame level on five of seven benchmarks, because a single frame already shows what the question asks about. Multi-hop reasoning breaks on text or on a frame for 15 of 29 benchmarks and survives on 6. Most reasoning questions are therefore answered by language priors or by one frame. Temporal order survives on 8 of 18 benchmarks, and four others, namely ForeSeaQA \citep{park2026foresea}, LongTimeScope \citep{zohar2025apollo2}, TempCompass \citep{liu2024tempcompass}, and VSI-Super-Wild \citep{gu2026vsisuperwild}, break on the options alone. Spatial relation breaks at the order level on 5 of 24 benchmarks and survives on 11, the highest share of any large class. A shuffled video still supports those answers, a perturbation that a claim about static relations need not resist. Fine-grained action spreads over every level and survives on 3 of 10.

\section{Claim-matched breaking levels}
\label{app:claimmatched}
We recount the breaking levels with only the attacks that bear on the claim of each benchmark. Order attacks count only for the 18 benchmarks in the temporal group, because the other groups do not claim that answers depend on frame order. Blind readers do not count for the 32 reasoning and knowledge benchmarks, whose claims include knowledge that a reader brings without the video. Every other attacker, the reference, the tolerance $\delta$, the 128-frame rescreen, and the cumulative maximum over levels stay as in the main audit, and a level left without attackers stays unmeasured. The recount reuses the item-level results of the main audit and first reproduces all 115 published breaking levels.

\begin{table}[!htbp]
\centering
\caption{Benchmarks by first breaking level under all attacks and under claim-matched attacks. All counts every attack as in the main audit. Claim drops the order level outside the temporal group and the blind readers for reasoning and knowledge. Groups hold 45 perception, 32 reasoning and knowledge, 20 spatial and physical, and 18 temporal benchmarks. Before video sums the option, text, and pool levels.}
\label{tab:claimmatched}
\begin{tabular}{@{}lrrrrrrrrrr@{}}
\toprule
 & \multicolumn{2}{c}{Perception} & \multicolumn{2}{c}{Reasoning} & \multicolumn{2}{c}{Spatial} & \multicolumn{2}{c}{Temporal} & \multicolumn{2}{c}{Total} \\
\cmidrule(lr){2-3}\cmidrule(lr){4-5}\cmidrule(lr){6-7}\cmidrule(lr){8-9}\cmidrule(l){10-11}
Level & All & Claim & All & Claim & All & Claim & All & Claim & All & Claim \\
\midrule
Option & 3 & 3 & 4 & 4 & 4 & 4 & 6 & 6 & 17 & 17 \\
Text & 5 & 5 & 7 & 0 & 1 & 1 & 0 & 0 & 13 & 6 \\
Pool & 1 & 1 & 1 & 2 & 2 & 2 & 1 & 1 & 5 & 6 \\
Frame & 17 & 17 & 7 & 13 & 2 & 2 & 1 & 1 & 27 & 33 \\
Order & 6 & 0 & 5 & 0 & 4 & 0 & 0 & 0 & 15 & 0 \\
Unbroken & 13 & 19 & 8 & 13 & 7 & 11 & 10 & 10 & 38 & 53 \\
\midrule
Before video & 9 & 9 & 12 & 6 & 7 & 7 & 7 & 7 & 35 & 29 \\
\bottomrule
\end{tabular}
\end{table}

As shown in Table~\ref{tab:claimmatched}, the conclusions of the audit hold under claim-matched attacks. Newer benchmarks still break before visual input more often, 28\% of releases from 2025 and 2026 against 15\% of releases through 2024, and releases from 2025 and 2026 hold all 12 text and pool breaks. The option level is unchanged. Every order break of the main audit falls on a benchmark outside the temporal group, and these 15 benchmarks become unbroken, which raises the survivors from 38 to 53. The temporal group keeps its profile, with 6 of 18 benchmarks breaking on options and 10 surviving. Six reasoning and knowledge benchmarks move from the text level to the frame level, where frame captions alone still break them, namely COVER \citep{zhou2025cover}, LongVideo-Reason \citep{chen2025longvideoreason}, MMR-V \citep{zhu2026mmrv}, MMWorld \citep{he2024mmworld}, VAU-Bench \citep{zhu2025vaur1}, and VideoEspresso \citep{han2025videoespresso}. VideoSIAH-Eval \citep{yang2026longvt} moves from text to pool because its reference stays near chance, where the rule assigns the first measured level.

\section{Robustness to tolerance and reference model}
\label{app:tolerance}
We recount the breaking levels with the tolerance $\delta$ at 3\% and 8\% around the 5\% of the main audit, and every other setting stays fixed. As shown in Table~\ref{tab:tolerance}, newer benchmarks break before visual input more often at every tolerance, and releases from 2025 and 2026 hold most of the text and pool breaks. A tighter tolerance leaves more benchmarks unbroken, and a looser one moves breaks to earlier levels.
\begin{table}[!htbp]
\centering
\caption{Breaking levels at three tolerances. Before video counts the option, text, and pool breaks, and the last four columns compare releases through 2024 with releases from 2025 and 2026. The 5\% row is the main audit.}
\label{tab:tolerance}
\small
\begin{tabular}{@{}ccccccc@{}}
\toprule
 & & & \multicolumn{2}{c}{Before video, share} & \multicolumn{2}{c}{Text and pool breaks} \\
\cmidrule(lr){4-5}\cmidrule(l){6-7}
$\delta$ & Before video & Unbroken & Through 2024 & 2025--2026 & Through 2024 & 2025--2026 \\
\midrule
3\% & 33 & 45 & 19\% & 31\% & 1/26 & 19/89 \\
5\% & 35 & 38 & 19\% & 34\% & 1/26 & 17/89 \\
8\% & 49 & 19 & 31\% & 46\% & 4/26 & 21/89 \\
\bottomrule
\end{tabular}
\end{table}

We also recount the breaking levels with Claude Opus 5 \citep{anthropic2026opus5} at 32 frames, the strongest attacker model, as the reference for every benchmark. As shown in Table~\ref{tab:reference}, this reference leaves 48 benchmarks unbroken, 31 break before visual input, and the text and pool breaks fall on the same benchmarks as in the main audit. Newer benchmarks still break before visual input more often, 29\% against 19\%.
\begin{table}[!htbp]
\centering
\caption{Breaking levels under the two reference models. The Gemini 3.5 Flash row is the main audit. The last two columns give the share breaking before video among releases through 2024 and from 2025 and 2026.}
\label{tab:reference}
\small
\begin{tabular}{@{}lcccccccc@{}}
\toprule
Reference & Option & Text & Pool & Frame & Order & Unbroken & Through 2024 & 2025--2026 \\
\midrule
Gemini 3.5 Flash & 17 & 13 & 5 & 27 & 15 & 38 & 19\% & 34\% \\
Claude Opus 5 & 13 & 12 & 6 & 22 & 14 & 48 & 19\% & 29\% \\
\bottomrule
\end{tabular}
\end{table}

We also compare the order in which benchmarks break. For every pair of benchmarks that the main audit breaks at different levels, we check whether another setting keeps their order, ties them, or reverses them. As shown in Table~\ref{tab:breakorder}, the tolerance barely changes this order. With Claude Opus 5 as the reference, four in five pairs keep their order. With Qwen3-VL-8B \citep{bai2025qwen3} at 32 frames as the reference, most pairs keep their order, and the weaker reference moves most breaks to earlier levels.
\begin{table}[!htbp]
\centering
\caption{Order of breaking levels under other settings, as shares of the benchmark pairs that the main audit breaks at different levels.}
\label{tab:breakorder}
\small
\begin{tabular}{@{}lccc@{}}
\toprule
Setting & Same order & Tied & Reversed \\
\midrule
Tolerance $\delta=3\%$ & 89.9\% & 7.8\% & 2.3\% \\
Tolerance $\delta=8\%$ & 80.4\% & 13.2\% & 6.5\% \\
Claude Opus 5 reference at 32 frames & 78.6\% & 13.3\% & 8.2\% \\
Qwen3-VL-8B reference at 32 frames & 57.9\% & 23.1\% & 18.9\% \\
\bottomrule
\end{tabular}
\end{table}

\section{Resolution, frame rate, and frame budget}
\label{app:framespixels}
\begin{figure}[!htbp]
\centering
\includegraphics[width=\linewidth]{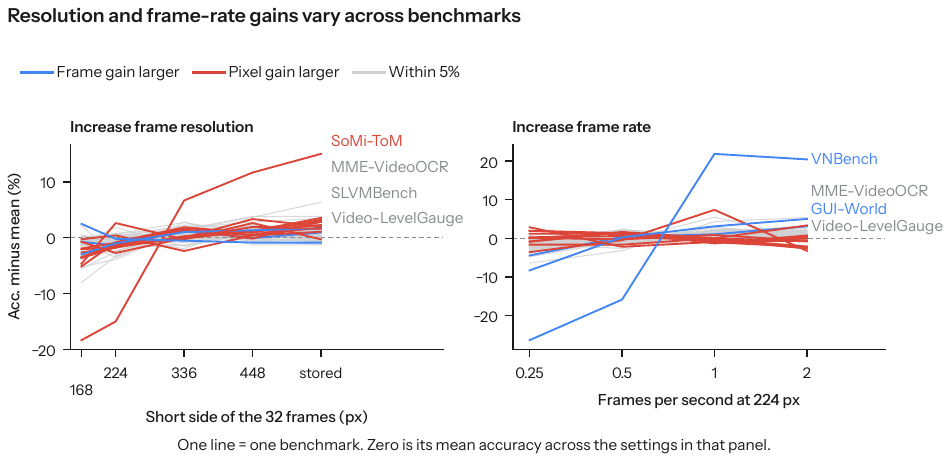} %
\caption{Resolution and frame-rate effects, centered on each benchmark's mean accuracy.}
\label{fig:f9fppanels}
\end{figure}
Qwen3-VL-8B \citep{bai2025qwen3} evaluates 300 items per benchmark passing the pool level. The resolution ladder uses 32 frames with short sides of 168, 224, 336, and 448 pixels, plus stored resolution. The temporal ladder uses 224 pixels at 0.25, 0.5, 1, and 2 frames per second, within context limits. Pixel gain is best-resolution accuracy minus accuracy at 168 pixels. Frame gain is accuracy at 2 fps minus accuracy at 0.25 fps. As shown in Figure~\ref{fig:f9fppanels}, median gains are 3.0\% for pixels and 2.1\% for frames. Both medians sit below the 5\% tolerance. Most benchmarks respond little to either ladder, and pixels help slightly more than frames. Among the 80 benchmarks passing the pool level, 57 have both ladders, and the frame-rate ladder does not apply to 17 benchmarks of short clips.

\section{Errors introduced by video input}
\label{sec:f2}
\begin{figure}[!htbp]
\centering
\includegraphics[width=\linewidth]{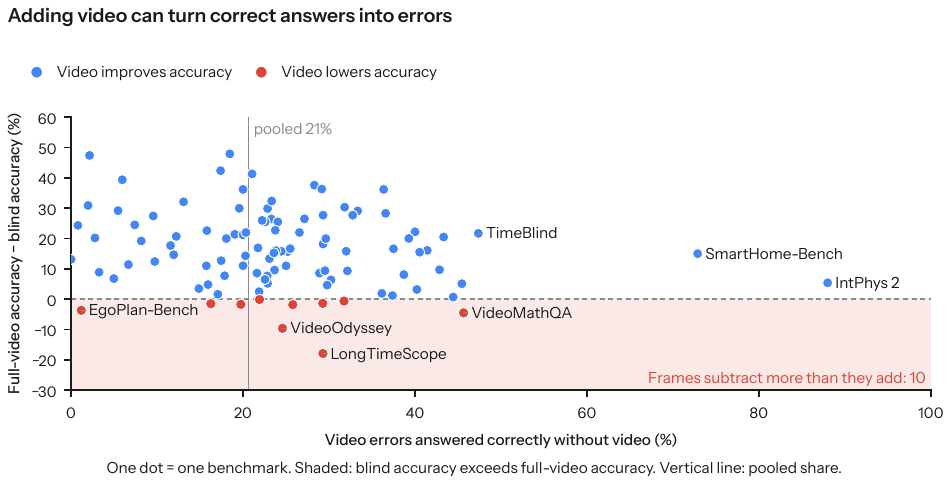} %
\caption{Blind-correct video errors versus video gain. Shading marks net losses from adding video.}
\label{fig:f2subtract}
\end{figure}
Video input can overturn a correct text-only answer. The blind reader answers 21\% of the 9,651 attributed reference errors correctly. Across the middle half of benchmarks, this share ranges from 17 to 30\%. It reaches 88\% on IntPhys 2 \citep{bordes2025intphys2}, where both conditions remain near chance but succeed on different items. Video lowers total accuracy on 10 benchmarks, including drops above 9\% on LongTimeScope \citep{zohar2025apollo2} and VideoOdyssey \citep{he2026videoodyssey}. As shown in Figure~\ref{fig:f2subtract}, video introduces individual errors on most benchmarks but lowers total accuracy on only these 10. Shortcut auditing and error attribution thus address complementary failure modes.

\section{Attribution of reference errors}
\label{sec:f5}
\begin{figure}[!htbp]
\centering
\includegraphics[width=\linewidth]{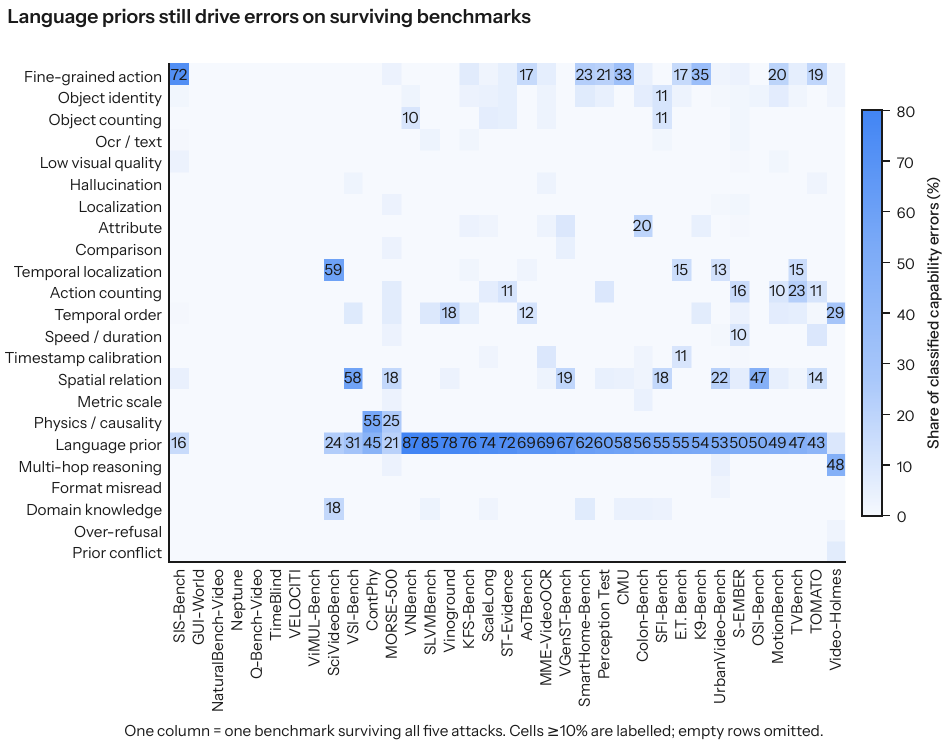} %
\caption{Fine error categories among pyramid survivors. Color encodes capability-error shares.}
\label{fig:attribution}
\end{figure}
The gap to 100\% includes errors that better models cannot resolve. Of 6,863 attributed failures, 5\% arise from labels. Label failures include ambiguous answers and mismatched videos. Evidence missed by sparse sampling counts as a model failure. The median label-error share is 2.1\%, exceeding a quarter on 1 benchmark. Its unrecoverable score gap has median 0.8\% and exceeds 10\% on TempCompass \citep{liu2024tempcompass} and UCF101-AD \citep{abdullah2026ucf101ad}. Among the 38 surviving benchmarks, 30 have label-error attribution, with a median share of 0\%. Of the 35 with fine attribution, language priors are the largest fine category on 30, with a median share of 54.6\%. VSI-Bench \citep{yang2025vsibench} instead assigns 58\% to spatial relation and metric scale. As shown in Figure~\ref{fig:attribution}, language priors still drive most remaining errors on the survivors.

\begin{figure}[!htbp]\centering
\includegraphics[height=0.44\textheight]{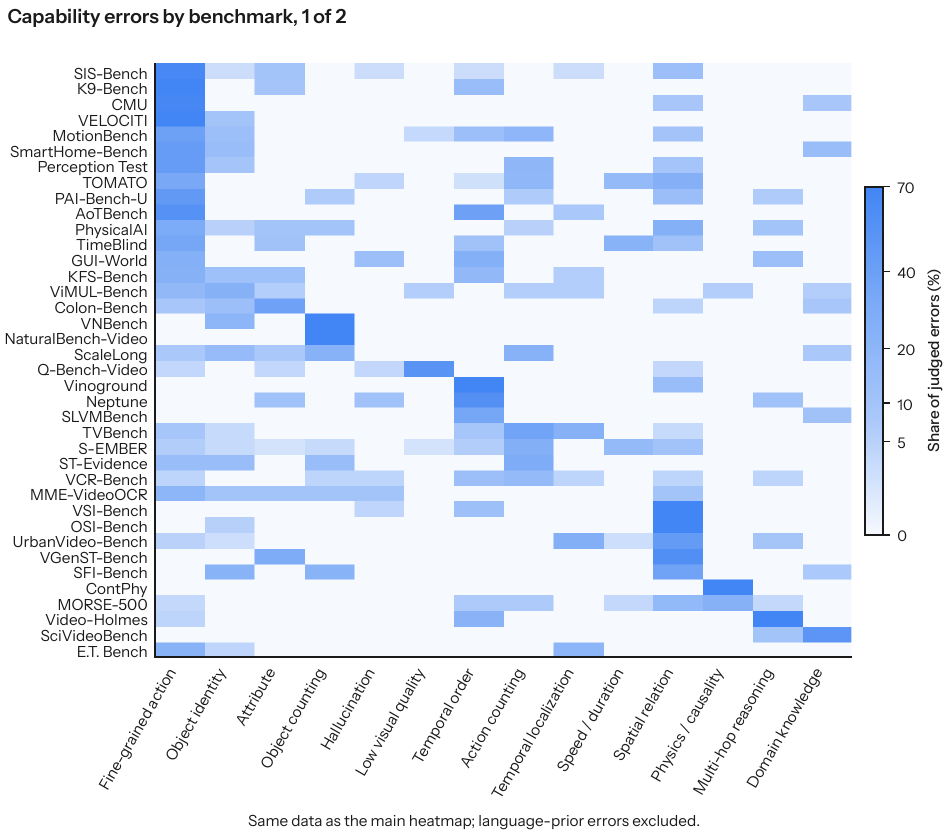}
\par\vspace{2pt}
\includegraphics[height=0.44\textheight]{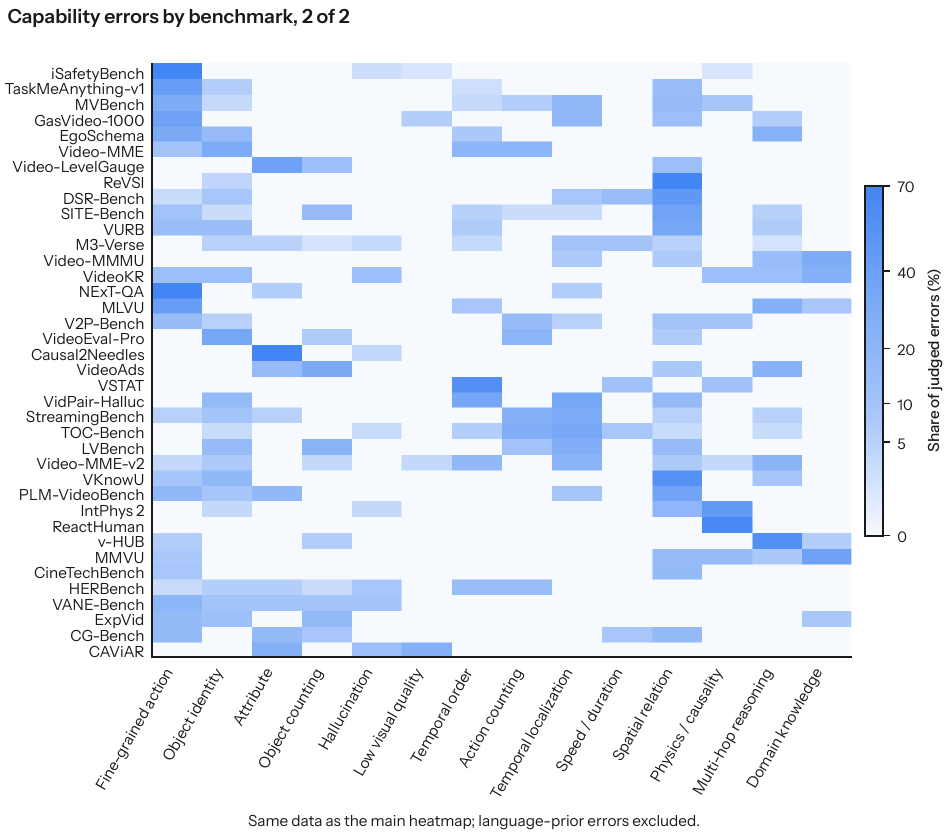}
\caption{Named capability-error profiles, first half above and second half below. Percentages of judged non-prior errors.}
\label{fig:abilitydetail}\panellabel{a}{fig:abilitydetail1}\panellabel{b}{fig:abilitydetail2}
\end{figure}

As shown in Figure~\ref{fig:abilitydetail}, most named profiles concentrate their non-prior errors in two categories.
Tables~\ref{tab:attrfine}--\ref{tab:attrfine_cols2} give perception and temporal error shares. Tables~\ref{tab:attrfine_b}--\ref{tab:attrfine_b_cols2} cover spatial, reasoning, and unresolved categories. Table~\ref{tab:abilitytop} lists the ten benchmarks with the largest share in each category. In Tables~\ref{tab:attrfine} and~\ref{tab:attrfine_b}, a blank cell marks a category with no instance. Benchmarks with fewer than 20 capability instances show no shares, and a dash there marks a category with at least one instance.
\begingroup

\begingroup
\setlength{\LTcapwidth}{\linewidth}
\setlength{\tabcolsep}{4pt}
\begin{xltabular}{\linewidth}{>{\raggedright\arraybackslash}p{170.42bp}>{\raggedleft\arraybackslash}p{25.05bp}>{\raggedleft\arraybackslash}p{25.05bp}>{\raggedleft\arraybackslash}p{24.58bp}>{\raggedleft\arraybackslash}p{25.26bp}>{\raggedleft\arraybackslash}p{25.47bp}>{\raggedleft\arraybackslash}p{25.83bp}>{\raggedleft\arraybackslash}p{35.06bp}>{\raggedleft\arraybackslash}p{28.87bp}}
\caption{Perception and temporal capability-error shares (\%) across benchmarks (1/2).}\label{tab:attrfine}\\
\toprule
\rowcolor{SkillPaleBlue} Benchmark & $n_{\mathrm{cap}}$ & $n_{\mathrm{det}}$ & P-act & P-obj & P-cnt & P-ocr & P-lowq & P-hall \\
\midrule
\endfirsthead
\multicolumn{9}{l}{\tablename~\thetable{} (continued)}\\
\toprule
\rowcolor{SkillPaleBlue} Benchmark & $n_{\mathrm{cap}}$ & $n_{\mathrm{det}}$ & P-act & P-obj & P-cnt & P-ocr & P-lowq & P-hall \\
\midrule
\endhead
\bottomrule
\endfoot
\multicolumn{9}{l}{\textbf{Main set} (Gemini 3.5 Flash errors; 67 benchmarks)}\\
\mbox{AoTBench} & 42 & 42 & 16.7 &  &  &  &  &  \\
\mbox{Artifact-Bench} & 94 & 62 & 22.6 & 6.5 &  & 1.6 & 9.7 & 4.8 \\
\mbox{CMU} & 26 & 24 & 33.3 &  &  &  &  &  \\
\mbox{COVER} & 42 & 39 & 20.5 &  &  &  &  & 2.6 \\
\mbox{DSI-Bench} & 104 & 96 & 1.0 &  &  &  &  &  \\
\mbox{DSR-Bench} & 118 & 110 & 1.8 & 5.5 &  &  &  &  \\
\mbox{EgoSchema} & 35 & 32 & 12.5 & 6.2 &  &  &  &  \\
\mbox{ExAct} & 61 & 59 & 23.7 &  &  &  &  &  \\
\mbox{ExpVid} & 38 & 36 & 38.9 & 8.3 &  & 16.7 &  &  \\
\mbox{FAVOR-Bench} & 70 & 66 & 36.4 &  &  &  & 3.0 &  \\
\mbox{GasVideo-1000} & 52 & 49 & 12.2 &  &  &  & 2.0 &  \\
\mbox{HourVideo} & 129 & 126 & 1.6 & 5.6 &  &  & 0.8 &  \\
\mbox{IntPhys 2} & 69 & 53 &  & 3.8 &  &  &  & 3.8 \\
\mbox{LEMONADE} & 91 & 85 &  &  &  &  &  &  \\
\mbox{LongTimeScope} & 26 & 24 &  &  &  & 12.5 &  & 29.2 \\
\mbox{LongVideoBench} & 18 & 17 & -- &  &  &  &  &  \\
\mbox{LVBench} & 60 & 58 &  & 5.2 & 6.9 &  &  &  \\
\mbox{MLVU} & 25 & 24 & 20.8 &  &  & 4.2 &  &  \\
\mbox{MMR-V} & 72 & 72 & 1.4 & 1.4 &  &  &  &  \\
\mbox{MMSI-Video-Bench} & 128 & 125 & 3.2 &  & 2.4 &  &  &  \\
\mbox{MMVU} & 21 & 20 & 5.0 &  &  &  &  &  \\
\mbox{MMWorld} & 34 & 33 & 9.1 & 3.0 &  &  &  & 3.0 \\
\mbox{MORSE-500} & 35 & 28 & 3.6 &  &  &  &  &  \\
\mbox{MotionBench} & 60 & 59 & 20.3 & 6.8 &  &  & 1.7 &  \\
\mbox{MVBench} & 52 & 47 & 19.1 & 2.1 &  &  &  &  \\
\mbox{Neptune} & 3 & 3 &  &  &  &  & -- &  \\
\mbox{OSI-Bench} & 38 & 38 &  & 2.6 &  &  &  &  \\
\mbox{Perception Test} & 46 & 42 & 21.4 & 4.8 &  &  &  &  \\
\mbox{PerceptionComp} & 155 & 140 & 0.7 & 6.4 & 3.6 & 4.3 & 2.1 &  \\
\mbox{Q-Bench-Video} & 9 & 8 &  &  &  &  & -- &  \\
\mbox{ReVSI} & 42 & 42 &  & 2.4 &  &  &  &  \\
\mbox{SAW-Bench} & 134 & 133 &  &  &  &  &  &  \\
\mbox{SciVideoBench} & 35 & 34 &  &  &  &  &  &  \\
\mbox{SFI-Bench} & 58 & 56 &  & 10.7 & 10.7 & 1.8 &  &  \\
\mbox{SIS-Bench} & 116 & 111 & 72.1 & 1.8 &  & 0.9 & 3.6 &  \\
\mbox{SITE-Bench} & 58 & 56 & 7.1 & 1.8 & 10.7 &  &  &  \\
\mbox{SoMi-ToM} & 3 & 3 &  &  &  & -- &  &  \\
\mbox{SpatialTree} & 18 & 18 &  &  &  &  &  &  \\
\mbox{ST-Evidence} & 20 & 18 & 5.6 & 5.6 & 5.6 &  &  &  \\
\mbox{STI-Bench} & 132 & 127 &  & 0.8 &  &  & 0.8 &  \\
\mbox{StreamingBench} & 41 & 38 & 2.6 & 5.3 &  &  &  &  \\
\mbox{TempCompass} & 41 & 41 & 12.2 &  &  &  &  &  \\
\mbox{TemporalBench} & 35 & 34 & 26.5 &  &  &  &  &  \\
\mbox{TimeBlind} & 19 & 18 & -- &  &  &  &  &  \\
\mbox{TimeScope} & 20 & 19 &  & 5.3 &  & 10.5 &  & 10.5 \\
\mbox{TOMATO} & 85 & 84 & 19.0 &  &  &  &  & 2.4 \\
\mbox{TVBench} & 58 & 53 & 5.7 & 1.9 &  &  &  &  \\
\mbox{v-HUB} & 30 & 26 & 3.8 &  & 3.8 &  &  &  \\
\mbox{V2P-Bench} & 55 & 52 & 5.8 & 1.9 &  & 3.8 &  &  \\
\mbox{Video-Holmes} & 47 & 42 & 2.4 & 2.4 &  &  &  &  \\
\mbox{Video-MME} & 31 & 29 & 3.4 & 10.3 &  &  &  &  \\
\mbox{Video-MME-v2} & 78 & 75 & 1.3 & 2.7 & 1.3 &  & 1.3 &  \\
\mbox{Video-MMMU} & 30 & 25 &  &  &  & 4.0 &  &  \\
\mbox{Video-TT} & 12 & 12 & -- & -- &  &  &  &  \\
\mbox{VideoEval-Pro} & 66 & 63 &  & 7.9 & 1.6 &  &  &  \\
\mbox{VideoKR} & 29 & 29 & 3.4 & 3.4 &  &  &  & 3.4 \\
\mbox{VideoMathQA} & 25 & 25 &  &  & 8.0 & 4.0 &  &  \\
\mbox{VideoOdyssey} & 29 & 29 &  & 6.9 &  &  &  &  \\
\mbox{VideoSIAH-Eval} & 31 & 29 & 6.9 & 20.7 &  & 13.8 &  &  \\
\mbox{VidPair-Halluc} & 40 & 40 &  & 2.5 &  &  &  &  \\
\mbox{ViMUL-Bench} & 15 & 13 & -- &  &  &  & -- &  \\
\mbox{Vinoground} & 28 & 27 &  &  &  &  &  &  \\
\mbox{VITATECS} & 12 & 12 & -- &  &  &  &  &  \\
\mbox{VKnowU} & 33 & 33 & 3.0 & 6.1 &  &  &  &  \\
\mbox{VSI-Bench} & 36 & 36 &  &  &  &  &  & 2.8 \\
\mbox{VSI-Super-Wild} & 92 & 83 &  & 7.2 &  &  &  &  \\
\mbox{VURB} & 26 & 23 & 8.7 & 8.7 &  &  &  &  \\
\midrule
\multicolumn{9}{l}{\textbf{Newly onboarded benchmarks (N1)} (claude-opus-5 errors; 43 benchmarks)}\\
\mbox{ALLVB} & 35 & 35 &  &  &  &  & 2.9 &  \\
\mbox{Causal2Needles} & 65 & 62 &  &  &  &  &  & 1.6 \\
\mbox{CAViAR} & 26 & 23 &  &  &  &  & 8.7 & 4.3 \\
\mbox{CG-Bench} & 23 & 19 & 10.5 &  & 5.3 &  &  &  \\
\mbox{CineTechBench} & 22 & 13 & 7.7 &  &  &  &  &  \\
\mbox{Colon-Bench} & 50 & 46 & 4.3 & 6.5 &  &  &  &  \\
\mbox{ContPhy} & 43 & 42 &  &  &  &  &  &  \\
\mbox{E.T. Bench} & 86 & 66 & 16.7 & 3.0 &  &  &  &  \\
\mbox{EgoPlan-Bench} & 75 & 66 & 21.2 & 7.6 &  &  &  &  \\
\mbox{FineBench} & 30 & 30 & 13.3 &  &  &  &  &  \\
\mbox{ForeSeaQA} & 35 & 32 & 18.8 &  & 9.4 &  & 3.1 &  \\
\mbox{GUI-World} & 18 & 17 & -- &  &  &  &  &  \\
\mbox{HERBench} & 75 & 63 & 1.6 & 3.2 & 1.6 &  &  & 4.8 \\
\mbox{HLV-1K} & 46 & 43 & 2.3 &  & 4.7 &  &  &  \\
\mbox{iSafetyBench} & 122 & 118 & 66.1 &  &  &  & 0.8 & 1.7 \\
\mbox{K9-Bench} & 44 & 43 & 34.9 &  &  &  &  &  \\
\mbox{KFS-Bench} & 58 & 54 & 7.4 & 3.7 &  & 1.9 &  &  \\
\mbox{LongVideo-Reason} & 33 & 33 &  &  &  &  &  &  \\
\mbox{M3-Verse} & 121 & 109 &  & 2.8 & 0.9 &  &  & 1.8 \\
\mbox{MME-VideoOCR} & 32 & 32 & 6.2 & 3.1 & 3.1 &  &  & 3.1 \\
\mbox{NaturalBench-Video} & 19 & 19 &  &  & -- &  &  &  \\
\mbox{OVO-S-Bench} & 87 & 87 &  & 1.1 & 1.1 &  &  &  \\
\mbox{PLM-VideoBench} & 29 & 28 & 7.1 & 3.6 &  &  &  &  \\
\mbox{ReactHuman} & 120 & 102 &  &  &  &  &  &  \\
\mbox{ReXTime} & 11 & 10 & -- &  &  &  &  &  \\
\mbox{S-EMBER} & 121 & 109 & 3.7 & 1.8 & 1.8 & 1.8 & 0.9 &  \\
\mbox{ScaleLong} & 47 & 46 & 2.2 & 4.3 & 6.5 &  &  &  \\
\mbox{SLVMBench} & 38 & 34 &  &  &  & 2.9 &  &  \\
\mbox{SmartHome-Bench} & 30 & 26 & 23.1 & 7.7 &  &  &  &  \\
\mbox{TaskMeAnything-v1} & 87 & 70 & 28.6 & 4.3 &  &  &  &  \\
\mbox{TOC-Bench} & 105 & 102 &  & 1.0 &  &  &  & 1.0 \\
\mbox{UCF101-AD} & 86 & 83 & 84.3 & 4.8 &  &  &  &  \\
\mbox{UrbanVideo-Bench} & 89 & 87 & 2.3 & 1.1 &  &  &  &  \\
\mbox{VANE-Bench} & 26 & 22 & 9.1 & 4.5 & 4.5 &  &  & 4.5 \\
\mbox{VAU-Bench} & 7 & 6 & -- &  &  & -- &  &  \\
\mbox{VELOCITI} & 13 & 12 & -- &  &  &  &  &  \\
\mbox{VF-Eval} & 23 & 18 &  &  &  &  &  & 5.6 \\
\mbox{VGenST-Bench} & 21 & 21 &  &  &  &  &  &  \\
\mbox{Video-LevelGauge} & 21 & 19 &  &  & 5.3 & 5.3 &  &  \\
\mbox{VideoAds} & 27 & 24 &  &  & 16.7 &  &  &  \\
\mbox{VideoEspresso} & 29 & 28 &  & 3.6 &  &  &  &  \\
\mbox{VNBench} & 38 & 38 &  & 2.6 & 10.5 &  &  &  \\
\mbox{VRU-Accident} & 40 & 40 &  &  &  &  &  &  \\
\midrule
\textbf{All, primary attribution protocol} (67 benchmarks) & 3,413 & 3,205 & 9.1 & 3.0 & 0.9 & 0.9 & 0.8 & 0.6 \\
\textbf{All, Claude-role protocol} (43 benchmarks) & 2,170 & 1,993 & 13.1 & 2.0 & 1.4 & 0.3 & 0.3 & 0.7 \\
\end{xltabular}
\endgroup

\begingroup
\setlength{\LTcapwidth}{\linewidth}
\setlength{\tabcolsep}{4pt}
\begin{xltabular}{\linewidth}{>{\raggedright\arraybackslash}p{173.66bp}>{\raggedleft\arraybackslash}p{24.70bp}>{\raggedleft\arraybackslash}p{27.60bp}>{\raggedleft\arraybackslash}p{31.75bp}>{\raggedleft\arraybackslash}p{24.19bp}>{\raggedleft\arraybackslash}p{24.97bp}>{\raggedleft\arraybackslash}p{26.95bp}>{\raggedleft\arraybackslash}p{28.07bp}>{\raggedleft\arraybackslash}p{23.69bp}}
\caption{Perception and temporal capability-error shares (\%) across benchmarks (2/2).}\label{tab:attrfine_cols2}\\
\toprule
\rowcolor{SkillPaleBlue} Benchmark & P-loc & P-attr & P-cmp & T-loc & T-cnt & T-ord & T-spd & T-cal \\
\midrule
\endfirsthead
\multicolumn{9}{l}{\tablename~\thetable{} (continued)}\\
\toprule
\rowcolor{SkillPaleBlue} Benchmark & P-loc & P-attr & P-cmp & T-loc & T-cnt & T-ord & T-spd & T-cal \\
\midrule
\endhead
\bottomrule
\endfoot
\multicolumn{9}{l}{\textbf{Main set} (Gemini 3.5 Flash errors; 67 benchmarks)}\\
\mbox{AoTBench} &  &  &  & 2.4 &  & 11.9 &  &  \\
\mbox{Artifact-Bench} &  &  &  &  &  &  &  &  \\
\mbox{CMU} &  &  &  &  &  &  &  &  \\
\mbox{COVER} &  &  &  &  &  & 5.1 &  &  \\
\mbox{DSI-Bench} &  &  &  &  &  &  &  &  \\
\mbox{DSR-Bench} &  &  &  & 5.5 &  &  & 9.1 & 2.7 \\
\mbox{EgoSchema} &  &  &  &  &  & 3.1 &  &  \\
\mbox{ExAct} &  &  &  &  &  &  &  &  \\
\mbox{ExpVid} &  &  &  &  &  &  &  &  \\
\mbox{FAVOR-Bench} &  &  &  & 3.0 & 1.5 & 6.1 &  &  \\
\mbox{GasVideo-1000} &  &  &  & 6.1 &  &  &  &  \\
\mbox{HourVideo} &  & 0.8 &  &  & 3.2 & 3.2 & 7.1 &  \\
\mbox{IntPhys 2} &  &  &  &  &  &  &  &  \\
\mbox{LEMONADE} &  &  &  & 17.6 & 4.7 & 12.9 & 25.9 &  \\
\mbox{LongTimeScope} &  &  &  &  &  &  &  &  \\
\mbox{LongVideoBench} &  &  &  &  &  & -- &  &  \\
\mbox{LVBench} &  &  &  & 8.6 & 3.4 &  &  &  \\
\mbox{MLVU} &  &  &  &  &  & 4.2 &  &  \\
\mbox{MMR-V} &  &  &  &  &  &  &  & 1.4 \\
\mbox{MMSI-Video-Bench} &  &  & 0.8 &  & 3.2 &  &  &  \\
\mbox{MMVU} &  &  &  &  &  &  &  &  \\
\mbox{MMWorld} &  &  &  & 3.0 & 6.1 &  &  &  \\
\mbox{MORSE-500} & 3.6 &  & 3.6 &  & 7.1 & 7.1 & 3.6 &  \\
\mbox{MotionBench} &  &  &  &  & 10.2 & 6.8 &  &  \\
\mbox{MVBench} &  &  &  & 12.8 & 4.3 & 2.1 &  &  \\
\mbox{Neptune} &  &  &  &  & -- &  &  &  \\
\mbox{OSI-Bench} &  &  &  &  &  &  &  &  \\
\mbox{Perception Test} &  &  &  &  & 9.5 &  &  &  \\
\mbox{PerceptionComp} &  & 3.6 &  &  & 0.7 &  &  &  \\
\mbox{Q-Bench-Video} &  &  &  &  &  &  &  &  \\
\mbox{ReVSI} &  &  &  &  &  &  &  &  \\
\mbox{SAW-Bench} &  &  & 0.8 &  &  &  &  &  \\
\mbox{SciVideoBench} &  &  &  & 58.8 &  &  &  &  \\
\mbox{SFI-Bench} &  &  &  &  &  &  &  &  \\
\mbox{SIS-Bench} &  &  &  &  &  & 0.9 &  &  \\
\mbox{SITE-Bench} & 1.8 &  &  & 1.8 & 1.8 & 3.6 &  &  \\
\mbox{SoMi-ToM} &  &  &  &  &  &  &  &  \\
\mbox{SpatialTree} &  &  &  &  &  &  &  &  \\
\mbox{ST-Evidence} &  &  &  &  & 11.1 &  &  &  \\
\mbox{STI-Bench} &  &  &  & 1.6 &  &  & 3.1 & 0.8 \\
\mbox{StreamingBench} &  & 2.6 &  & 15.8 & 13.2 &  &  &  \\
\mbox{TempCompass} &  &  &  &  &  & 2.4 & 4.9 &  \\
\mbox{TemporalBench} &  &  &  &  & 2.9 &  &  &  \\
\mbox{TimeBlind} &  & -- &  &  &  & -- & -- &  \\
\mbox{TimeScope} &  &  &  &  &  &  &  &  \\
\mbox{TOMATO} &  &  &  &  & 10.7 & 1.2 & 9.5 &  \\
\mbox{TVBench} &  &  &  & 15.1 & 22.6 & 5.7 &  &  \\
\mbox{v-HUB} &  &  &  &  &  &  &  &  \\
\mbox{V2P-Bench} & 1.9 &  &  & 1.9 & 5.8 &  &  &  \\
\mbox{Video-Holmes} &  &  &  &  &  & 28.6 &  &  \\
\mbox{Video-MME} &  &  &  &  & 6.9 & 6.9 &  &  \\
\mbox{Video-MME-v2} &  &  &  & 8.0 &  & 6.7 &  &  \\
\mbox{Video-MMMU} &  &  &  & 4.0 &  &  &  &  \\
\mbox{Video-TT} &  &  &  &  & -- &  & -- &  \\
\mbox{VideoEval-Pro} &  &  &  &  & 4.8 &  &  &  \\
\mbox{VideoKR} &  &  &  &  &  &  &  &  \\
\mbox{VideoMathQA} &  &  &  &  &  &  &  &  \\
\mbox{VideoOdyssey} &  &  &  &  & 3.4 &  &  & 3.4 \\
\mbox{VideoSIAH-Eval} & 3.4 &  &  & 6.9 &  & 3.4 &  &  \\
\mbox{VidPair-Halluc} &  &  &  & 5.0 &  & 5.0 &  &  \\
\mbox{ViMUL-Bench} &  &  &  &  & -- &  &  &  \\
\mbox{Vinoground} &  &  &  &  &  & 18.5 &  &  \\
\mbox{VITATECS} &  &  &  &  &  & -- &  &  \\
\mbox{VKnowU} &  &  & 3.0 &  &  &  &  &  \\
\mbox{VSI-Bench} &  &  &  &  &  & 8.3 &  &  \\
\mbox{VSI-Super-Wild} &  &  &  & 4.8 &  & 19.3 &  &  \\
\mbox{VURB} &  &  &  &  &  & 4.3 &  &  \\
\midrule
\multicolumn{9}{l}{\textbf{Newly onboarded benchmarks (N1)} (claude-opus-5 errors; 43 benchmarks)}\\
\mbox{ALLVB} &  & 8.6 &  &  &  &  &  &  \\
\mbox{Causal2Needles} &  & 35.5 &  &  &  &  &  &  \\
\mbox{CAViAR} &  & 8.7 &  &  &  &  &  &  \\
\mbox{CG-Bench} &  & 10.5 &  &  &  &  & 5.3 &  \\
\mbox{CineTechBench} &  &  &  &  &  &  &  &  \\
\mbox{Colon-Bench} &  & 19.6 &  &  &  &  &  & 2.2 \\
\mbox{ContPhy} &  &  &  &  &  &  &  &  \\
\mbox{E.T. Bench} &  &  &  & 15.2 &  &  &  & 10.6 \\
\mbox{EgoPlan-Bench} &  &  &  & 1.5 &  &  &  &  \\
\mbox{FineBench} &  &  &  &  &  &  &  & 10.0 \\
\mbox{ForeSeaQA} &  &  &  & 3.1 & 3.1 &  &  &  \\
\mbox{GUI-World} &  &  &  &  &  & -- &  &  \\
\mbox{HERBench} & 3.2 & 3.2 & 3.2 &  & 7.9 & 7.9 &  &  \\
\mbox{HLV-1K} & 4.7 &  &  & 2.3 &  & 14.0 &  & 7.0 \\
\mbox{iSafetyBench} &  &  &  &  &  &  &  &  \\
\mbox{K9-Bench} &  & 4.7 &  &  &  & 7.0 &  &  \\
\mbox{KFS-Bench} &  & 3.7 &  & 1.9 &  & 5.6 &  &  \\
\mbox{LongVideo-Reason} &  &  &  &  &  & 3.0 &  & 3.0 \\
\mbox{M3-Verse} &  & 2.8 & 7.3 & 5.5 &  & 1.8 & 5.5 &  \\
\mbox{MME-VideoOCR} &  & 3.1 &  &  &  &  &  & 9.4 \\
\mbox{NaturalBench-Video} &  &  &  &  &  &  &  &  \\
\mbox{OVO-S-Bench} &  &  & 2.3 &  & 1.1 & 4.6 & 4.6 &  \\
\mbox{PLM-VideoBench} &  & 7.1 &  & 3.6 &  &  &  &  \\
\mbox{ReactHuman} &  &  &  &  &  &  &  &  \\
\mbox{ReXTime} &  &  &  &  &  &  &  &  \\
\mbox{S-EMBER} & 1.8 & 0.9 &  &  & 15.6 & 3.7 & 10.1 & 0.9 \\
\mbox{ScaleLong} &  & 2.2 &  &  & 6.5 &  &  & 2.2 \\
\mbox{SLVMBench} &  &  &  &  &  & 8.8 &  &  \\
\mbox{SmartHome-Bench} &  &  &  &  &  &  &  &  \\
\mbox{TaskMeAnything-v1} &  &  &  &  &  & 1.4 &  &  \\
\mbox{TOC-Bench} & 1.0 &  & 1.0 & 10.8 & 8.8 & 2.0 & 2.9 &  \\
\mbox{UCF101-AD} &  &  &  &  &  &  &  &  \\
\mbox{UrbanVideo-Bench} & 1.1 &  &  & 12.6 &  &  & 1.1 &  \\
\mbox{VANE-Bench} &  & 4.5 &  &  &  &  &  &  \\
\mbox{VAU-Bench} &  &  &  &  &  &  &  &  \\
\mbox{VELOCITI} &  &  &  &  &  &  &  &  \\
\mbox{VF-Eval} &  & 11.1 &  &  &  &  &  &  \\
\mbox{VGenST-Bench} &  & 9.5 & 4.8 &  &  &  &  &  \\
\mbox{Video-LevelGauge} &  & 15.8 &  &  &  &  &  &  \\
\mbox{VideoAds} &  & 8.3 &  &  &  &  &  &  \\
\mbox{VideoEspresso} &  &  &  &  &  & 3.6 &  &  \\
\mbox{VNBench} &  &  &  &  &  &  &  &  \\
\mbox{VRU-Accident} &  & 17.5 &  &  &  &  &  &  \\
\midrule
\textbf{All, primary attribution protocol} (67 benchmarks) & 0.1 & 0.2 & 0.1 & 2.9 & 2.3 & 3.0 & 1.8 & 0.2 \\
\textbf{All, Claude-role protocol} (43 benchmarks) & 0.4 & 3.5 & 0.7 & 2.2 & 1.8 & 1.9 & 1.3 & 1.0 \\
\end{xltabular}
\endgroup

\endgroup

\begingroup

\begingroup
\setlength{\LTcapwidth}{\linewidth}
\setlength{\tabcolsep}{4pt}
\begin{xltabular}{\linewidth}{>{\raggedright\arraybackslash}p{188.47bp}>{\raggedleft\arraybackslash}p{25.05bp}>{\raggedleft\arraybackslash}p{25.05bp}>{\raggedleft\arraybackslash}p{22.37bp}>{\raggedleft\arraybackslash}p{33.18bp}>{\raggedleft\arraybackslash}p{33.46bp}>{\raggedleft\arraybackslash}p{35.72bp}>{\raggedleft\arraybackslash}p{30.28bp}}
\caption{Spatial, reasoning, and unresolved capability-error shares (\%) (1/2).}\label{tab:attrfine_b}\\
\toprule
\rowcolor{SkillPaleBlue} Benchmark & $n_{\mathrm{cap}}$ & $n_{\mathrm{det}}$ & S-rel & S-scale & S-phys & R-prior & R-hop \\
\midrule
\endfirsthead
\multicolumn{8}{l}{\tablename~\thetable{} (continued)}\\
\toprule
\rowcolor{SkillPaleBlue} Benchmark & $n_{\mathrm{cap}}$ & $n_{\mathrm{det}}$ & S-rel & S-scale & S-phys & R-prior & R-hop \\
\midrule
\endhead
\bottomrule
\endfoot
\multicolumn{8}{l}{\textbf{Main set} (Gemini 3.5 Flash errors; 67 benchmarks)}\\
\mbox{AoTBench} & 42 & 42 &  &  &  & 69.0 &  \\
\mbox{Artifact-Bench} & 94 & 62 &  &  & 11.3 & 40.3 & 3.2 \\
\mbox{CMU} & 26 & 24 & 4.2 &  &  & 58.3 &  \\
\mbox{COVER} & 42 & 39 &  &  & 2.6 & 53.8 & 7.7 \\
\mbox{DSI-Bench} & 104 & 96 & 68.8 &  & 1.0 & 29.2 &  \\
\mbox{DSR-Bench} & 118 & 110 & 30.0 &  &  & 44.5 &  \\
\mbox{EgoSchema} & 35 & 32 &  &  &  & 68.8 & 9.4 \\
\mbox{ExAct} & 61 & 59 &  &  &  & 69.5 &  \\
\mbox{ExpVid} & 38 & 36 &  &  &  & 30.6 & 2.8 \\
\mbox{FAVOR-Bench} & 70 & 66 &  &  &  & 48.5 & 1.5 \\
\mbox{GasVideo-1000} & 52 & 49 & 4.1 &  &  & 73.5 & 2.0 \\
\mbox{HourVideo} & 129 & 126 & 13.5 &  &  & 62.7 & 0.8 \\
\mbox{IntPhys 2} & 69 & 53 & 20.8 &  & 49.1 & 22.6 &  \\
\mbox{LEMONADE} & 91 & 85 & 3.5 & 10.6 &  &  &  \\
\mbox{LongTimeScope} & 26 & 24 &  &  &  & 45.8 &  \\
\mbox{LongVideoBench} & 18 & 17 & -- &  &  & -- & -- \\
\mbox{LVBench} & 60 & 58 & 5.2 &  &  & 70.7 &  \\
\mbox{MLVU} & 25 & 24 &  &  &  & 54.2 & 12.5 \\
\mbox{MMR-V} & 72 & 72 &  &  &  & 75.0 & 16.7 \\
\mbox{MMSI-Video-Bench} & 128 & 125 & 48.0 & 0.8 &  & 39.2 & 1.6 \\
\mbox{MMVU} & 21 & 20 & 10.0 &  & 10.0 & 40.0 & 5.0 \\
\mbox{MMWorld} & 34 & 33 &  &  &  & 69.7 &  \\
\mbox{MORSE-500} & 35 & 28 & 17.9 & 3.6 & 25.0 & 21.4 & 3.6 \\
\mbox{MotionBench} & 60 & 59 & 5.1 &  &  & 49.2 &  \\
\mbox{MVBench} & 52 & 47 & 10.6 &  & 6.4 & 42.6 &  \\
\mbox{Neptune} & 3 & 3 &  &  &  &  & -- \\
\mbox{OSI-Bench} & 38 & 38 & 47.4 &  &  & 50.0 &  \\
\mbox{Perception Test} & 46 & 42 & 4.8 &  &  & 59.5 &  \\
\mbox{PerceptionComp} & 155 & 140 & 43.6 &  &  & 16.4 & 17.9 \\
\mbox{Q-Bench-Video} & 9 & 8 &  &  &  & -- &  \\
\mbox{ReVSI} & 42 & 42 & 57.1 &  &  & 40.5 &  \\
\mbox{SAW-Bench} & 134 & 133 & 60.2 &  &  & 39.1 &  \\
\mbox{SciVideoBench} & 35 & 34 &  &  &  & 23.5 &  \\
\mbox{SFI-Bench} & 58 & 56 & 17.9 &  &  & 55.4 &  \\
\mbox{SIS-Bench} & 116 & 111 & 4.5 &  &  & 16.2 &  \\
\mbox{SITE-Bench} & 58 & 56 & 25.0 & 7.1 &  & 35.7 & 3.6 \\
\mbox{SoMi-ToM} & 3 & 3 &  &  &  & -- & -- \\
\mbox{SpatialTree} & 18 & 18 & -- &  &  & -- &  \\
\mbox{ST-Evidence} & 20 & 18 &  &  &  & 72.2 &  \\
\mbox{STI-Bench} & 132 & 127 & 18.1 & 15.0 &  & 57.5 &  \\
\mbox{StreamingBench} & 41 & 38 & 2.6 &  &  & 55.3 & 2.6 \\
\mbox{TempCompass} & 41 & 41 & 4.9 &  &  & 73.2 &  \\
\mbox{TemporalBench} & 35 & 34 & 8.8 &  &  & 61.8 &  \\
\mbox{TimeBlind} & 19 & 18 & -- &  &  & -- &  \\
\mbox{TimeScope} & 20 & 19 &  &  &  & 68.4 &  \\
\mbox{TOMATO} & 85 & 84 & 14.3 &  &  & 42.9 &  \\
\mbox{TVBench} & 58 & 53 & 1.9 &  &  & 47.2 &  \\
\mbox{v-HUB} & 30 & 26 &  &  &  & 53.8 & 34.6 \\
\mbox{V2P-Bench} & 55 & 52 & 3.8 &  & 3.8 & 67.3 &  \\
\mbox{Video-Holmes} & 47 & 42 &  &  &  & 9.5 & 47.6 \\
\mbox{Video-MME} & 31 & 29 &  &  &  & 72.4 &  \\
\mbox{Video-MME-v2} & 78 & 75 & 2.7 &  & 1.3 & 66.7 & 8.0 \\
\mbox{Video-MMMU} & 30 & 25 & 4.0 &  &  & 64.0 & 8.0 \\
\mbox{Video-TT} & 12 & 12 &  &  &  & -- &  \\
\mbox{VideoEval-Pro} & 66 & 63 & 1.6 &  &  & 81.0 &  \\
\mbox{VideoKR} & 29 & 29 &  &  & 3.4 & 72.4 & 3.4 \\
\mbox{VideoMathQA} & 25 & 25 & 12.0 &  &  & 48.0 & 20.0 \\
\mbox{VideoOdyssey} & 29 & 29 & 6.9 &  &  & 75.9 &  \\
\mbox{VideoSIAH-Eval} & 31 & 29 &  &  &  & 37.9 & 3.4 \\
\mbox{VidPair-Halluc} & 40 & 40 & 2.5 &  &  & 85.0 &  \\
\mbox{ViMUL-Bench} & 15 & 13 &  &  & -- & -- &  \\
\mbox{Vinoground} & 28 & 27 & 3.7 &  &  & 77.8 &  \\
\mbox{VITATECS} & 12 & 12 &  &  &  & -- &  \\
\mbox{VKnowU} & 33 & 33 & 18.2 &  &  & 66.7 & 3.0 \\
\mbox{VSI-Bench} & 36 & 36 & 58.3 &  &  & 30.6 &  \\
\mbox{VSI-Super-Wild} & 92 & 83 &  &  &  & 62.7 &  \\
\mbox{VURB} & 26 & 23 & 21.7 &  &  & 47.8 & 4.3 \\
\midrule
\multicolumn{8}{l}{\textbf{Newly onboarded benchmarks (N1)} (claude-opus-5 errors; 43 benchmarks)}\\
\mbox{ALLVB} & 35 & 35 &  &  &  & 88.6 &  \\
\mbox{Causal2Needles} & 65 & 62 &  &  &  & 62.9 &  \\
\mbox{CAViAR} & 26 & 23 &  &  &  & 78.3 &  \\
\mbox{CG-Bench} & 23 & 19 & 10.5 &  &  & 57.9 &  \\
\mbox{CineTechBench} & 22 & 13 & 15.4 &  &  & 76.9 &  \\
\mbox{Colon-Bench} & 50 & 46 & 2.2 & 4.3 &  & 56.5 &  \\
\mbox{ContPhy} & 43 & 42 &  &  & 54.8 & 45.2 &  \\
\mbox{E.T. Bench} & 86 & 66 &  &  &  & 54.5 &  \\
\mbox{EgoPlan-Bench} & 75 & 66 &  &  &  & 30.3 & 36.4 \\
\mbox{FineBench} & 30 & 30 &  &  &  & 76.7 &  \\
\mbox{ForeSeaQA} & 35 & 32 & 3.1 &  &  & 59.4 &  \\
\mbox{GUI-World} & 18 & 17 &  &  &  & -- & -- \\
\mbox{HERBench} & 75 & 63 &  &  &  & 63.5 &  \\
\mbox{HLV-1K} & 46 & 43 & 4.7 &  &  & 60.5 &  \\
\mbox{iSafetyBench} & 122 & 118 &  &  & 0.8 & 30.5 &  \\
\mbox{K9-Bench} & 44 & 43 &  &  &  & 53.5 &  \\
\mbox{KFS-Bench} & 58 & 54 &  &  &  & 75.9 &  \\
\mbox{LongVideo-Reason} & 33 & 33 &  &  &  & 84.8 & 9.1 \\
\mbox{M3-Verse} & 121 & 109 & 2.8 & 13.8 &  & 54.1 & 0.9 \\
\mbox{MME-VideoOCR} & 32 & 32 & 3.1 &  &  & 68.8 &  \\
\mbox{NaturalBench-Video} & 19 & 19 &  &  &  & -- &  \\
\mbox{OVO-S-Bench} & 87 & 87 & 9.2 & 2.3 &  & 71.3 & 2.3 \\
\mbox{PLM-VideoBench} & 29 & 28 & 14.3 &  &  & 64.3 &  \\
\mbox{ReactHuman} & 120 & 102 &  &  & 31.4 & 68.6 &  \\
\mbox{ReXTime} & 11 & 10 &  &  & -- & -- & -- \\
\mbox{S-EMBER} & 121 & 109 & 6.4 &  &  & 50.5 &  \\
\mbox{ScaleLong} & 47 & 46 &  &  &  & 73.9 &  \\
\mbox{SLVMBench} & 38 & 34 &  &  &  & 85.3 &  \\
\mbox{SmartHome-Bench} & 30 & 26 &  &  &  & 61.5 &  \\
\mbox{TaskMeAnything-v1} & 87 & 70 & 10.0 &  &  & 55.7 &  \\
\mbox{TOC-Bench} & 105 & 102 & 1.0 &  &  & 69.6 & 1.0 \\
\mbox{UCF101-AD} & 86 & 83 &  &  &  & 9.6 &  \\
\mbox{UrbanVideo-Bench} & 89 & 87 & 21.8 &  &  & 52.9 & 4.6 \\
\mbox{VANE-Bench} & 26 & 22 &  &  &  & 72.7 &  \\
\mbox{VAU-Bench} & 7 & 6 &  &  &  & -- &  \\
\mbox{VELOCITI} & 13 & 12 &  &  &  & -- &  \\
\mbox{VF-Eval} & 23 & 18 &  &  & 5.6 & 77.8 &  \\
\mbox{VGenST-Bench} & 21 & 21 & 19.0 &  &  & 66.7 &  \\
\mbox{Video-LevelGauge} & 21 & 19 & 5.3 &  &  & 68.4 &  \\
\mbox{VideoAds} & 27 & 24 & 4.2 &  &  & 58.3 & 12.5 \\
\mbox{VideoEspresso} & 29 & 28 &  &  &  & 89.3 & 3.6 \\
\mbox{VNBench} & 38 & 38 &  &  &  & 86.8 &  \\
\mbox{VRU-Accident} & 40 & 40 &  &  & 2.5 & 77.5 &  \\
\midrule
\textbf{All, primary attribution protocol} (67 benchmarks) & 3,413 & 3,205 & 16.2 & 1.1 & 1.6 & 48.9 & 3.4 \\
\textbf{All, Claude-role protocol} (43 benchmarks) & 2,170 & 1,993 & 3.2 & 1.0 & 3.0 & 59.8 & 2.1 \\
\end{xltabular}
\endgroup

\begingroup
\setlength{\LTcapwidth}{\linewidth}
\setlength{\tabcolsep}{4pt}
\begin{xltabular}{\linewidth}{>{\raggedright\arraybackslash}p{166.73bp}>{\raggedleft\arraybackslash}p{28.24bp}>{\raggedleft\arraybackslash}p{38.89bp}>{\raggedleft\arraybackslash}p{40.82bp}>{\raggedleft\arraybackslash}p{32.20bp}>{\raggedleft\arraybackslash}p{19.61bp}>{\raggedleft\arraybackslash}p{19.59bp}>{\raggedleft\arraybackslash}p{19.59bp}>{\raggedleft\arraybackslash}p{19.91bp}}
\caption{Spatial, reasoning, and unresolved capability-error shares (\%) (2/2).}\label{tab:attrfine_b_cols2}\\
\toprule
\rowcolor{SkillPaleBlue} Benchmark & R-fmt & R-know & R-refuse & R-conf & U-P & U-T & U-S & U-R \\
\midrule
\endfirsthead
\multicolumn{9}{l}{\tablename~\thetable{} (continued)}\\
\toprule
\rowcolor{SkillPaleBlue} Benchmark & R-fmt & R-know & R-refuse & R-conf & U-P & U-T & U-S & U-R \\
\midrule
\endhead
\bottomrule
\endfoot
\multicolumn{9}{l}{\textbf{Main set} (Gemini 3.5 Flash errors; 67 benchmarks)}\\
\mbox{AoTBench} &  &  &  &  &  &  &  &  \\
\mbox{Artifact-Bench} &  &  &  &  & \textit{30.9} &  & \textit{2.1} & \textit{1.1} \\
\mbox{CMU} &  & 4.2 &  &  & \textit{7.7} &  &  &  \\
\mbox{COVER} &  &  &  & 7.7 & \textit{7.1} &  &  &  \\
\mbox{DSI-Bench} &  &  &  &  & \textit{6.7} &  & \textit{1.0} &  \\
\mbox{DSR-Bench} & 0.9 &  &  &  & \textit{2.5} & \textit{3.4} & \textit{0.8} &  \\
\mbox{EgoSchema} &  &  &  &  & \textit{2.9} &  &  & \textit{5.7} \\
\mbox{ExAct} &  & 6.8 &  &  & \textit{3.3} &  &  &  \\
\mbox{ExpVid} &  & 2.8 &  &  & \textit{2.6} &  &  & \textit{2.6} \\
\mbox{FAVOR-Bench} &  &  &  &  & \textit{4.3} & \textit{1.4} &  &  \\
\mbox{GasVideo-1000} &  &  &  &  & \textit{5.8} &  &  &  \\
\mbox{HourVideo} & 0.8 &  &  &  & \textit{0.8} & \textit{1.6} &  &  \\
\mbox{IntPhys 2} &  &  &  &  & \textit{4.3} &  & \textit{18.8} &  \\
\mbox{LEMONADE} & 3.5 &  & 20.0 & 1.2 &  & \textit{4.4} &  & \textit{2.2} \\
\mbox{LongTimeScope} &  &  & 12.5 &  & \textit{7.7} &  &  &  \\
\mbox{LongVideoBench} &  &  &  &  & -- &  &  &  \\
\mbox{LVBench} &  &  &  &  &  & \textit{1.7} &  & \textit{1.7} \\
\mbox{MLVU} &  & 4.2 &  &  & \textit{4.0} &  &  &  \\
\mbox{MMR-V} &  &  &  & 4.2 &  &  &  &  \\
\mbox{MMSI-Video-Bench} & 0.8 &  &  &  & \textit{0.8} &  & \textit{1.6} &  \\
\mbox{MMVU} &  & 25.0 & 5.0 &  &  &  & \textit{4.8} &  \\
\mbox{MMWorld} &  & 3.0 & 3.0 &  &  &  &  & \textit{2.9} \\
\mbox{MORSE-500} &  &  &  &  & \textit{5.7} &  & \textit{8.6} & \textit{5.7} \\
\mbox{MotionBench} &  &  &  &  & \textit{1.7} &  &  &  \\
\mbox{MVBench} &  &  &  &  & \textit{7.7} & \textit{1.9} &  &  \\
\mbox{Neptune} &  &  &  &  &  &  &  &  \\
\mbox{OSI-Bench} &  &  &  &  &  &  &  &  \\
\mbox{Perception Test} &  &  &  &  & \textit{8.7} &  &  &  \\
\mbox{PerceptionComp} &  & 0.7 &  &  & \textit{8.4} &  &  & \textit{1.3} \\
\mbox{Q-Bench-Video} &  &  &  &  & -- &  &  &  \\
\mbox{ReVSI} &  &  &  &  &  &  &  &  \\
\mbox{SAW-Bench} &  &  &  &  & \textit{0.7} &  &  &  \\
\mbox{SciVideoBench} &  & 17.6 &  &  &  &  &  & \textit{2.9} \\
\mbox{SFI-Bench} &  & 3.6 &  &  & \textit{3.4} &  &  &  \\
\mbox{SIS-Bench} &  &  &  &  & \textit{4.3} &  &  &  \\
\mbox{SITE-Bench} &  &  &  &  & \textit{3.4} &  &  &  \\
\mbox{SoMi-ToM} &  &  &  &  &  &  &  &  \\
\mbox{SpatialTree} & -- &  &  &  &  &  &  &  \\
\mbox{ST-Evidence} &  &  &  &  & \textit{5.0} &  & \textit{5.0} &  \\
\mbox{STI-Bench} &  &  & 2.4 &  &  & \textit{0.8} & \textit{0.8} & \textit{2.3} \\
\mbox{StreamingBench} &  &  &  &  & \textit{4.9} & \textit{2.4} &  &  \\
\mbox{TempCompass} & 2.4 &  &  &  &  &  &  &  \\
\mbox{TemporalBench} &  &  &  &  & \textit{2.9} &  &  &  \\
\mbox{TimeBlind} &  &  &  &  & -- &  &  &  \\
\mbox{TimeScope} &  &  & 5.3 &  & \textit{5.0} &  &  &  \\
\mbox{TOMATO} &  &  &  &  & \textit{1.2} &  &  &  \\
\mbox{TVBench} &  &  &  &  & \textit{5.2} & \textit{1.7} &  & \textit{1.7} \\
\mbox{v-HUB} &  & 3.8 &  &  &  &  &  & \textit{13.3} \\
\mbox{V2P-Bench} & 1.9 &  & 1.9 &  & \textit{5.5} &  &  &  \\
\mbox{Video-Holmes} &  &  & 2.4 & 7.1 &  &  &  & \textit{10.6} \\
\mbox{Video-MME} &  &  &  &  & \textit{6.5} &  &  &  \\
\mbox{Video-MME-v2} &  &  &  &  & \textit{1.3} & \textit{1.3} &  & \textit{1.3} \\
\mbox{Video-MMMU} &  & 16.0 &  &  & \textit{6.7} &  &  & \textit{10.0} \\
\mbox{Video-TT} &  &  &  &  &  &  &  &  \\
\mbox{VideoEval-Pro} &  &  & 1.6 & 1.6 & \textit{3.0} & \textit{1.5} &  &  \\
\mbox{VideoKR} &  & 6.9 & 3.4 &  &  &  &  &  \\
\mbox{VideoMathQA} &  &  & 8.0 &  &  &  &  &  \\
\mbox{VideoOdyssey} &  &  & 3.4 &  &  &  &  &  \\
\mbox{VideoSIAH-Eval} &  & 3.4 &  &  & \textit{6.5} &  &  &  \\
\mbox{VidPair-Halluc} &  &  &  &  &  &  &  &  \\
\mbox{ViMUL-Bench} &  &  &  & -- & -- &  &  &  \\
\mbox{Vinoground} &  &  &  &  & \textit{3.6} &  &  &  \\
\mbox{VITATECS} &  &  &  &  &  &  &  &  \\
\mbox{VKnowU} &  &  &  &  &  &  &  &  \\
\mbox{VSI-Bench} &  &  &  &  &  &  &  &  \\
\mbox{VSI-Super-Wild} & 6.0 &  &  &  & \textit{1.1} & \textit{8.7} &  &  \\
\mbox{VURB} &  &  & 4.3 &  &  & \textit{7.7} &  & \textit{3.8} \\
\midrule
\multicolumn{9}{l}{\textbf{Newly onboarded benchmarks (N1)} (claude-opus-5 errors; 43 benchmarks)}\\
\mbox{ALLVB} &  &  &  &  &  &  &  &  \\
\mbox{Causal2Needles} &  &  &  &  & \textit{4.6} &  &  &  \\
\mbox{CAViAR} &  &  &  &  & \textit{11.5} &  &  &  \\
\mbox{CG-Bench} &  &  &  &  & \textit{13.0} & \textit{4.3} &  &  \\
\mbox{CineTechBench} &  &  &  &  & \textit{40.9} &  &  &  \\
\mbox{Colon-Bench} &  & 4.3 &  &  & \textit{8.0} &  &  &  \\
\mbox{ContPhy} &  &  &  &  &  &  & \textit{2.3} &  \\
\mbox{E.T. Bench} &  &  &  &  & \textit{1.2} & \textit{22.1} &  &  \\
\mbox{EgoPlan-Bench} &  & 3.0 &  &  & \textit{5.3} &  &  & \textit{6.7} \\
\mbox{FineBench} &  &  &  &  &  &  &  &  \\
\mbox{ForeSeaQA} &  &  &  &  & \textit{2.9} & \textit{5.7} &  &  \\
\mbox{GUI-World} &  &  &  &  & -- &  &  &  \\
\mbox{HERBench} &  &  &  &  & \textit{16.0} &  &  &  \\
\mbox{HLV-1K} &  &  &  &  & \textit{2.2} & \textit{4.3} &  &  \\
\mbox{iSafetyBench} &  &  &  &  & \textit{3.3} &  &  &  \\
\mbox{K9-Bench} &  &  &  &  & \textit{2.3} &  &  &  \\
\mbox{KFS-Bench} &  &  &  &  & \textit{6.9} &  &  &  \\
\mbox{LongVideo-Reason} &  &  &  &  &  &  &  &  \\
\mbox{M3-Verse} &  &  &  &  & \textit{5.0} & \textit{4.1} & \textit{0.8} &  \\
\mbox{MME-VideoOCR} &  &  &  &  &  &  &  &  \\
\mbox{NaturalBench-Video} &  &  &  &  &  &  &  &  \\
\mbox{OVO-S-Bench} &  &  &  &  &  &  &  &  \\
\mbox{PLM-VideoBench} &  &  &  &  & \textit{3.4} &  &  &  \\
\mbox{ReactHuman} &  &  &  &  &  &  & \textit{15.0} &  \\
\mbox{ReXTime} &  &  &  &  &  & -- &  &  \\
\mbox{S-EMBER} &  &  &  &  & \textit{5.0} & \textit{5.0} &  &  \\
\mbox{ScaleLong} &  & 2.2 &  &  & \textit{2.1} &  &  &  \\
\mbox{SLVMBench} &  & 2.9 &  &  & \textit{5.3} & \textit{5.3} &  &  \\
\mbox{SmartHome-Bench} &  & 7.7 &  &  & \textit{6.7} &  &  & \textit{6.7} \\
\mbox{TaskMeAnything-v1} &  &  &  &  & \textit{19.5} &  &  &  \\
\mbox{TOC-Bench} &  &  &  &  & \textit{1.9} & \textit{1.0} &  &  \\
\mbox{UCF101-AD} &  & 1.2 &  &  & \textit{3.5} &  &  &  \\
\mbox{UrbanVideo-Bench} & 2.3 &  &  &  & \textit{1.1} & \textit{1.1} &  &  \\
\mbox{VANE-Bench} &  &  &  &  & \textit{15.4} &  &  &  \\
\mbox{VAU-Bench} &  &  &  &  & -- &  &  &  \\
\mbox{VELOCITI} &  &  &  &  & -- &  &  &  \\
\mbox{VF-Eval} &  &  &  &  & \textit{21.7} &  &  &  \\
\mbox{VGenST-Bench} &  &  &  &  &  &  &  &  \\
\mbox{Video-LevelGauge} &  &  &  &  & \textit{4.8} & \textit{4.8} &  &  \\
\mbox{VideoAds} &  &  &  &  & \textit{7.4} & \textit{3.7} &  &  \\
\mbox{VideoEspresso} &  &  &  &  & \textit{3.4} &  &  &  \\
\mbox{VNBench} &  &  &  &  &  &  &  &  \\
\mbox{VRU-Accident} &  & 2.5 &  &  &  &  &  &  \\
\midrule
\textbf{All, primary attribution protocol} (67 benchmarks) & 0.4 & 0.9 & 1.1 & 0.4 & \textit{3.6} & \textit{0.8} & \textit{0.7} & \textit{0.9} \\
\textbf{All, Claude-role protocol} (43 benchmarks) & 0.1 & 0.5 &  &  & \textit{5.0} & \textit{1.9} & \textit{0.9} & \textit{0.3} \\
\end{xltabular}
\endgroup

\endgroup

\begingroup
\begingroup
\setlength{\LTcapwidth}{\linewidth}
\setlength{\tabcolsep}{4pt}
\begin{xltabular}{\linewidth}{>{\raggedright\arraybackslash}p{151.15bp}>{\raggedright\arraybackslash}p{290.43bp}}
\caption{Top ten benchmarks per fine error category (\% share); at least 20 capability errors required.}\label{tab:abilitytop}\\
\toprule
\rowcolor{SkillPaleBlue} Category & Benchmarks \\
\midrule
\endfirsthead
\multicolumn{2}{l}{\tablename~\thetable{} (continued)}\\
\toprule
\rowcolor{SkillPaleBlue} Category & Benchmarks \\
\midrule
\endhead
\bottomrule
\endfoot
\multicolumn{2}{l}{\textit{Perception}} \\
fine-grained action & \mbox{SIS-Bench} 69, \mbox{ExpVid} 37, \mbox{FAVOR-Bench} 34, \mbox{CMU} 31, \mbox{TemporalBench} 26, \mbox{ExAct} 23, \mbox{MLVU} 20, \mbox{MotionBench} 20, \mbox{Perception Test} 20, \mbox{COVER} 19 \\
object identity & \mbox{VideoSIAH-Eval} 19, \mbox{SFI-Bench} 10, \mbox{Video-MME} 10, \mbox{ExpVid} 8, \mbox{VURB} 8, \mbox{VideoEval-Pro} 8, \mbox{VideoOdyssey} 7, \mbox{MotionBench} 7, \mbox{VSI-Super-Wild} 7, \mbox{VKnowU} 6 \\
object counting & \mbox{SFI-Bench} 10, \mbox{SITE-Bench} 10, \mbox{VideoMathQA} 8, \mbox{LVBench} 7, \mbox{ST-Evidence} 5, \mbox{v-HUB} 3, \mbox{PerceptionComp} 3, \mbox{MMSI-Video-Bench} 2, \mbox{VideoEval-Pro} 2, \mbox{Video-MME-v2} 1 \\
text and OCR & \mbox{ExpVid} 16, \mbox{VideoSIAH-Eval} 13, \mbox{LongTimeScope} 12, \mbox{TimeScope} 10, \mbox{MLVU} 4, \mbox{VideoMathQA} 4, \mbox{PerceptionComp} 4, \mbox{V2P-Bench} 4, \mbox{Video-MMMU} 3, \mbox{SFI-Bench} 2 \\
low visual quality & \mbox{Artifact-Bench} 6, \mbox{SIS-Bench} 3, \mbox{FAVOR-Bench} 3, \mbox{PerceptionComp} 2, \mbox{GasVideo-1000} 2, \mbox{MotionBench} 2, \mbox{Video-MME-v2} 1, \mbox{HourVideo} 1, \mbox{STI-Bench} 1 \\
hallucinated content & \mbox{LongTimeScope} 27, \mbox{TimeScope} 10, \mbox{VideoKR} 3, \mbox{Artifact-Bench} 3, \mbox{MMWorld} 3, \mbox{IntPhys 2} 3, \mbox{VSI-Bench} 3, \mbox{COVER} 2, \mbox{TOMATO} 2 \\
localization & \mbox{VideoSIAH-Eval} 3, \mbox{MORSE-500} 3, \mbox{V2P-Bench} 2, \mbox{SITE-Bench} 2 \\
attribute & \mbox{PerceptionComp} 3, \mbox{StreamingBench} 2, \mbox{HourVideo} 1 \\
comparison & \mbox{VKnowU} 3, \mbox{MORSE-500} 3, \mbox{MMSI-Video-Bench} 1, \mbox{SAW-Bench} 1 \\
\multicolumn{2}{l}{\textit{Temporal}} \\
temporal localization & \mbox{SciVideoBench} 57, \mbox{LEMONADE} 16, \mbox{StreamingBench} 15, \mbox{TVBench} 14, \mbox{MVBench} 12, \mbox{LVBench} 8, \mbox{Video-MME-v2} 8, \mbox{VideoSIAH-Eval} 6, \mbox{GasVideo-1000} 6, \mbox{DSR-Bench} 5 \\
action counting & \mbox{TVBench} 21, \mbox{StreamingBench} 12, \mbox{TOMATO} 11, \mbox{ST-Evidence} 10, \mbox{MotionBench} 10, \mbox{Perception Test} 9, \mbox{Video-MME} 6, \mbox{MMWorld} 6, \mbox{MORSE-500} 6, \mbox{V2P-Bench} 5 \\
temporal order & \mbox{Video-Holmes} 26, \mbox{Vinoground} 18, \mbox{VSI-Super-Wild} 17, \mbox{LEMONADE} 12, \mbox{AoTBench} 12, \mbox{VSI-Bench} 8, \mbox{MotionBench} 7, \mbox{Video-MME} 6, \mbox{Video-MME-v2} 6, \mbox{FAVOR-Bench} 6 \\
speed and duration & \mbox{LEMONADE} 24, \mbox{TOMATO} 9, \mbox{DSR-Bench} 8, \mbox{HourVideo} 7, \mbox{TempCompass} 5, \mbox{STI-Bench} 3, \mbox{MORSE-500} 3 \\
frame-to-timestamp calibration & \mbox{VideoOdyssey} 3, \mbox{DSR-Bench} 3, \mbox{MMR-V} 1, \mbox{STI-Bench} 1 \\
\multicolumn{2}{l}{\textit{Spatial and physical}} \\
spatial relation & \mbox{DSI-Bench} 63, \mbox{SAW-Bench} 60, \mbox{VSI-Bench} 58, \mbox{ReVSI} 57, \mbox{OSI-Bench} 47, \mbox{MMSI-Video-Bench} 47, \mbox{PerceptionComp} 39, \mbox{DSR-Bench} 28, \mbox{SITE-Bench} 24, \mbox{VURB} 19 \\
metric scale & \mbox{STI-Bench} 14, \mbox{LEMONADE} 10, \mbox{SITE-Bench} 7, \mbox{MORSE-500} 3, \mbox{MMSI-Video-Bench} 1 \\
physics and causality & \mbox{IntPhys 2} 38, \mbox{MORSE-500} 20, \mbox{MMVU} 10, \mbox{Artifact-Bench} 7, \mbox{MVBench} 6, \mbox{V2P-Bench} 4, \mbox{VideoKR} 3, \mbox{COVER} 2, \mbox{Video-MME-v2} 1, \mbox{DSI-Bench} 1 \\
\multicolumn{2}{l}{\textit{Reasoning and knowledge}} \\
language prior & \mbox{VidPair-Halluc} 85, \mbox{VideoEval-Pro} 77, \mbox{VideoOdyssey} 76, \mbox{MMR-V} 75, \mbox{Vinoground} 75, \mbox{TempCompass} 73, \mbox{VideoKR} 72, \mbox{GasVideo-1000} 69, \mbox{AoTBench} 69, \mbox{LVBench} 68 \\
multi-hop reasoning & \mbox{Video-Holmes} 43, \mbox{v-HUB} 30, \mbox{VideoMathQA} 20, \mbox{MMR-V} 17, \mbox{PerceptionComp} 16, \mbox{MLVU} 12, \mbox{EgoSchema} 9, \mbox{Video-MME-v2} 8, \mbox{COVER} 7, \mbox{Video-MMMU} 7 \\
task format & \mbox{VSI-Super-Wild} 5, \mbox{LEMONADE} 3, \mbox{TempCompass} 2, \mbox{V2P-Bench} 2, \mbox{DSR-Bench} 1, \mbox{HourVideo} 1, \mbox{MMSI-Video-Bench} 1 \\
domain knowledge & \mbox{MMVU} 24, \mbox{SciVideoBench} 17, \mbox{Video-MMMU} 13, \mbox{VideoKR} 7, \mbox{ExAct} 7, \mbox{MLVU} 4, \mbox{CMU} 4, \mbox{SFI-Bench} 3, \mbox{v-HUB} 3, \mbox{VideoSIAH-Eval} 3 \\
over-refusal & \mbox{LEMONADE} 19, \mbox{LongTimeScope} 12, \mbox{VideoMathQA} 8, \mbox{TimeScope} 5, \mbox{MMVU} 5, \mbox{VURB} 4, \mbox{VideoKR} 3, \mbox{VideoOdyssey} 3, \mbox{MMWorld} 3, \mbox{STI-Bench} 2 \\
language prior conflict & \mbox{COVER} 7, \mbox{Video-Holmes} 6, \mbox{MMR-V} 4, \mbox{VideoEval-Pro} 2, \mbox{LEMONADE} 1 \\
\end{xltabular}
\endgroup
\endgroup
Tables~\ref{tab:attrshares}--\ref{tab:attrshares_cols2} report attribution shares and exclusion rates from the error-verification workflow. Tables~\ref{tab:attrshares_n1}--\ref{tab:attrshares_n1_cols2} give the Claude-role analysis on newly onboarded benchmarks. Tables~\ref{tab:attrshares} and~\ref{tab:attrshares_n1} count errors as the attributed 32-frame failures of Gemini 3.5 Flash \citep{google2026gemini35flash} and of Claude Opus 5 \citep{anthropic2026opus5} in turn, and the attributed count removes the disputed, unstable, and unverified instances to form the denominator of the three shares. Per-benchmark rows require at least 20 errors, which keeps 63 of 77 benchmarks in the first table and 40 of 43 in the second. In the second table the caption signal shown to the judges is answered by the erring model itself. Five rows of the first table, SIS-Bench \citep{zou2026sisbench}, Video-Holmes \citep{cheng2025videoholmes}, SciVideoBench \citep{deng2025scivideobench}, STI-Bench \citep{li2025stibench}, and ExpVid \citep{xu2025expvid}, were computed from predictions made with letter-only option lists.
\begingroup
\begingroup
\setlength{\LTcapwidth}{\linewidth}
\setlength{\tabcolsep}{4pt}
\begin{xltabular}{\linewidth}{>{\raggedright\arraybackslash}p{163.39bp}>{\raggedleft\arraybackslash}p{61.89bp}>{\raggedleft\arraybackslash}p{63.75bp}>{\raggedleft\arraybackslash}p{65.33bp}>{\raggedleft\arraybackslash}p{63.22bp}}
\caption{Verified attribution shares and excluded-instance rates (\%) (1/2).}\label{tab:attrshares}\\
\toprule
\rowcolor{SkillPaleBlue} Benchmark & Errors & Attrib. & Annot. & Cover. \\
\midrule
\endfirsthead
\multicolumn{5}{l}{\tablename~\thetable{} (continued)}\\
\toprule
\rowcolor{SkillPaleBlue} Benchmark & Errors & Attrib. & Annot. & Cover. \\
\midrule
\endhead
\bottomrule
\endfoot
\textbf{Pooled, instance-weighted} & 6,114 & 4,618 & 21.4 & 4.6 \\
\textbf{Pooled, item-weighted} & 6,114 & 4,618 & 21.4 & 4.6 \\
\midrule
\mbox{LongTimeScope} & 198 & 114 & 75.4 & 1.8 \\
\mbox{SciVideoBench} & 282 & 151 & 74.2 & 2.6 \\
\mbox{Video-Holmes} & 250 & 159 & 69.2 & 1.3 \\
\mbox{ExpVid} & 191 & 116 & 67.2 & 0.0 \\
\mbox{TempCompass} & 112 & 101 & 59.4 & 0.0 \\
\mbox{VideoSIAH-Eval} & 184 & 117 & 48.7 & 24.8 \\
\mbox{LEMONADE} & 257 & 183 & 47.5 & 2.7 \\
\mbox{TimeScope} & 120 & 73 & 45.2 & 27.4 \\
\mbox{SIS-Bench} & 279 & 176 & 34.1 & 0.0 \\
\mbox{Video-MMMU} & 59 & 45 & 33.3 & 0.0 \\
\mbox{VKnowU} & 54 & 50 & 32.0 & 2.0 \\
\mbox{SpatialTree} & 27 & 24 & 25.0 & 0.0 \\
\mbox{MMWorld} & 49 & 44 & 22.7 & 0.0 \\
\mbox{Q-Bench-Video} & 22 & 14 & 21.4 & 14.3 \\
\mbox{LongVideoBench} & 49 & 33 & 21.2 & 24.2 \\
\mbox{COVER} & 79 & 53 & 20.8 & 0.0 \\
\mbox{V2P-Bench} & 119 & 81 & 19.8 & 12.3 \\
\mbox{MMR-V} & 111 & 93 & 19.4 & 3.2 \\
\mbox{StreamingBench} & 73 & 57 & 17.5 & 10.5 \\
\mbox{HourVideo} & 203 & 159 & 17.0 & 1.9 \\
\mbox{Video-MME} & 58 & 42 & 16.7 & 9.5 \\
\mbox{ViMUL-Bench} & 26 & 20 & 15.0 & 10.0 \\
\mbox{LVBench} & 115 & 87 & 14.9 & 16.1 \\
\mbox{VURB} & 49 & 34 & 14.7 & 8.8 \\
\mbox{VideoOdyssey} & 61 & 48 & 14.6 & 25.0 \\
\mbox{EgoSchema} & 53 & 42 & 14.3 & 2.4 \\
\mbox{STI-Bench} & 207 & 151 & 11.9 & 0.7 \\
\mbox{VideoEval-Pro} & 131 & 93 & 11.8 & 17.2 \\
\mbox{Video-MME-v2} & 121 & 97 & 11.3 & 8.2 \\
\mbox{VidPair-Halluc} & 52 & 45 & 11.1 & 0.0 \\
\mbox{MLVU} & 70 & 46 & 10.9 & 34.8 \\
\mbox{VideoKR} & 45 & 32 & 9.4 & 0.0 \\
\mbox{VideoMathQA} & 46 & 32 & 9.4 & 12.5 \\
\mbox{MMSI-Video-Bench} & 159 & 137 & 6.6 & 0.0 \\
\mbox{SITE-Bench} & 70 & 63 & 6.3 & 1.6 \\
\mbox{v-HUB} & 40 & 32 & 6.2 & 0.0 \\
\mbox{PerceptionComp} & 221 & 178 & 5.6 & 7.3 \\
\mbox{MVBench} & 70 & 55 & 5.5 & 0.0 \\
\mbox{ST-Evidence} & 27 & 21 & 4.8 & 0.0 \\
\mbox{MMVU} & 27 & 22 & 4.5 & 0.0 \\
\mbox{FAVOR-Bench} & 90 & 73 & 4.1 & 0.0 \\
\mbox{GasVideo-1000} & 63 & 55 & 3.6 & 1.8 \\
\mbox{SFI-Bench} & 69 & 60 & 3.3 & 0.0 \\
\mbox{TOMATO} & 126 & 92 & 3.3 & 4.3 \\
\mbox{ExAct} & 70 & 63 & 3.2 & 0.0 \\
\mbox{MORSE-500} & 47 & 36 & 2.8 & 0.0 \\
\mbox{TemporalBench} & 44 & 36 & 2.8 & 0.0 \\
\mbox{Perception Test} & 58 & 47 & 2.1 & 0.0 \\
\mbox{Artifact-Bench} & 107 & 96 & 2.1 & 0.0 \\
\mbox{DSR-Bench} & 146 & 120 & 1.7 & 0.0 \\
\mbox{TVBench} & 79 & 66 & 1.5 & 10.6 \\
\mbox{VSI-Super-Wild} & 110 & 93 & 1.1 & 0.0 \\
\mbox{DSI-Bench} & 107 & 105 & 0.9 & 0.0 \\
\mbox{AoTBench} & 48 & 42 & 0.0 & 0.0 \\
\mbox{CMU} & 34 & 26 & 0.0 & 0.0 \\
\mbox{IntPhys 2} & 100 & 76 & 0.0 & 9.2 \\
\mbox{MotionBench} & 70 & 62 & 0.0 & 3.2 \\
\mbox{OSI-Bench} & 39 & 38 & 0.0 & 0.0 \\
\mbox{ReVSI} & 45 & 43 & 0.0 & 2.3 \\
\mbox{SAW-Bench} & 140 & 134 & 0.0 & 0.0 \\
\mbox{TimeBlind} & 23 & 19 & 0.0 & 0.0 \\
\mbox{Vinoground} & 33 & 28 & 0.0 & 0.0 \\
\mbox{VSI-Bench} & 40 & 36 & 0.0 & 0.0 \\
\end{xltabular}
\endgroup

\begingroup
\setlength{\LTcapwidth}{\linewidth}
\setlength{\tabcolsep}{4pt}
\begin{xltabular}{\linewidth}{>{\raggedright\arraybackslash}p{159.32bp}>{\raggedleft\arraybackslash}p{61.79bp}>{\raggedleft\arraybackslash}p{63.50bp}>{\raggedleft\arraybackslash}p{65.07bp}>{\raggedleft\arraybackslash}p{67.90bp}}
\caption{Verified attribution shares and excluded-instance rates (\%) (2/2).}\label{tab:attrshares_cols2}\\
\toprule
\rowcolor{SkillPaleBlue} Benchmark & Capab. & Disput. & Unstab. & Unverif. \\
\midrule
\endfirsthead
\multicolumn{5}{l}{\tablename~\thetable{} (continued)}\\
\toprule
\rowcolor{SkillPaleBlue} Benchmark & Capab. & Disput. & Unstab. & Unverif. \\
\midrule
\endhead
\bottomrule
\endfoot
\textbf{Pooled, instance-weighted} & 74.0 & 19.4 & 1.2 & 3.9 \\
\textbf{Pooled, item-weighted} & 74.0 & 19.4 & 1.2 & 3.9 \\
\midrule
\mbox{LongTimeScope} & 22.8 & 29.3 & 1.0 & 12.1 \\
\mbox{SciVideoBench} & 23.2 & 41.1 & 0.0 & 5.3 \\
\mbox{Video-Holmes} & 29.6 & 33.6 & 0.0 & 2.8 \\
\mbox{ExpVid} & 32.8 & 35.6 & 0.0 & 3.7 \\
\mbox{TempCompass} & 40.6 & 9.8 & 0.0 & 0.0 \\
\mbox{VideoSIAH-Eval} & 26.5 & 15.8 & 0.0 & 20.7 \\
\mbox{LEMONADE} & 49.7 & 25.7 & 0.0 & 3.1 \\
\mbox{TimeScope} & 27.4 & 27.5 & 3.3 & 8.3 \\
\mbox{SIS-Bench} & 65.9 & 36.6 & 0.4 & 0.0 \\
\mbox{Video-MMMU} & 66.7 & 23.7 & 0.0 & 0.0 \\
\mbox{VKnowU} & 66.0 & 7.4 & 0.0 & 0.0 \\
\mbox{SpatialTree} & 75.0 & 7.4 & 3.7 & 0.0 \\
\mbox{MMWorld} & 77.3 & 6.1 & 4.1 & 0.0 \\
\mbox{Q-Bench-Video} & 64.3 & 31.8 & 0.0 & 4.5 \\
\mbox{LongVideoBench} & 54.5 & 20.4 & 4.1 & 8.2 \\
\mbox{COVER} & 79.2 & 26.6 & 0.0 & 6.3 \\
\mbox{V2P-Bench} & 67.9 & 22.7 & 0.8 & 8.4 \\
\mbox{MMR-V} & 77.4 & 14.4 & 0.0 & 1.8 \\
\mbox{StreamingBench} & 71.9 & 17.8 & 2.7 & 1.4 \\
\mbox{HourVideo} & 81.1 & 14.3 & 1.5 & 5.9 \\
\mbox{Video-MME} & 73.8 & 19.0 & 5.2 & 3.4 \\
\mbox{ViMUL-Bench} & 75.0 & 23.1 & 0.0 & 0.0 \\
\mbox{LVBench} & 69.0 & 13.9 & 1.7 & 8.7 \\
\mbox{VURB} & 76.5 & 20.4 & 4.1 & 6.1 \\
\mbox{VideoOdyssey} & 60.4 & 11.5 & 0.0 & 9.8 \\
\mbox{EgoSchema} & 83.3 & 18.9 & 0.0 & 1.9 \\
\mbox{STI-Bench} & 87.4 & 27.1 & 0.0 & 0.0 \\
\mbox{VideoEval-Pro} & 71.0 & 8.4 & 3.1 & 17.6 \\
\mbox{Video-MME-v2} & 80.4 & 12.4 & 0.8 & 6.6 \\
\mbox{VidPair-Halluc} & 88.9 & 9.6 & 0.0 & 3.8 \\
\mbox{MLVU} & 54.4 & 25.7 & 1.4 & 7.1 \\
\mbox{VideoKR} & 90.6 & 26.7 & 0.0 & 2.2 \\
\mbox{VideoMathQA} & 78.1 & 19.6 & 4.3 & 6.5 \\
\mbox{MMSI-Video-Bench} & 93.4 & 8.8 & 1.3 & 3.8 \\
\mbox{SITE-Bench} & 92.1 & 7.1 & 1.4 & 1.4 \\
\mbox{v-HUB} & 93.8 & 15.0 & 5.0 & 0.0 \\
\mbox{PerceptionComp} & 87.1 & 18.1 & 0.0 & 1.4 \\
\mbox{MVBench} & 94.5 & 17.1 & 4.3 & 0.0 \\
\mbox{ST-Evidence} & 95.2 & 14.8 & 0.0 & 7.4 \\
\mbox{MMVU} & 95.5 & 18.5 & 0.0 & 0.0 \\
\mbox{FAVOR-Bench} & 95.9 & 13.3 & 3.3 & 2.2 \\
\mbox{GasVideo-1000} & 94.5 & 7.9 & 1.6 & 3.2 \\
\mbox{SFI-Bench} & 96.7 & 10.1 & 2.9 & 0.0 \\
\mbox{TOMATO} & 92.4 & 21.4 & 4.0 & 1.6 \\
\mbox{ExAct} & 96.8 & 10.0 & 0.0 & 0.0 \\
\mbox{MORSE-500} & 97.2 & 23.4 & 0.0 & 0.0 \\
\mbox{TemporalBench} & 97.2 & 9.1 & 6.8 & 2.3 \\
\mbox{Perception Test} & 97.9 & 15.5 & 3.4 & 0.0 \\
\mbox{Artifact-Bench} & 97.9 & 10.3 & 0.0 & 0.0 \\
\mbox{DSR-Bench} & 98.3 & 16.4 & 1.4 & 0.0 \\
\mbox{TVBench} & 87.9 & 13.9 & 1.3 & 1.3 \\
\mbox{VSI-Super-Wild} & 98.9 & 15.5 & 0.0 & 0.0 \\
\mbox{DSI-Bench} & 99.1 & 1.9 & 0.0 & 0.0 \\
\mbox{AoTBench} & 100.0 & 6.2 & 6.2 & 0.0 \\
\mbox{CMU} & 100.0 & 23.5 & 0.0 & 0.0 \\
\mbox{IntPhys 2} & 90.8 & 16.0 & 2.0 & 6.0 \\
\mbox{MotionBench} & 96.8 & 5.7 & 0.0 & 5.7 \\
\mbox{OSI-Bench} & 100.0 & 2.6 & 0.0 & 0.0 \\
\mbox{ReVSI} & 97.7 & 0.0 & 4.4 & 0.0 \\
\mbox{SAW-Bench} & 100.0 & 3.6 & 0.7 & 0.0 \\
\mbox{TimeBlind} & 100.0 & 8.7 & 8.7 & 0.0 \\
\mbox{Vinoground} & 100.0 & 12.1 & 3.0 & 0.0 \\
\mbox{VSI-Bench} & 100.0 & 7.5 & 2.5 & 0.0 \\
\end{xltabular}
\endgroup
\endgroup

\begingroup
\begingroup
\setlength{\LTcapwidth}{\linewidth}
\setlength{\tabcolsep}{4pt}
\begin{xltabular}{\linewidth}{>{\raggedright\arraybackslash}p{163.39bp}>{\raggedleft\arraybackslash}p{61.89bp}>{\raggedleft\arraybackslash}p{63.75bp}>{\raggedleft\arraybackslash}p{65.33bp}>{\raggedleft\arraybackslash}p{63.22bp}}
\caption{Claude-role attribution and exclusion rates on new benchmarks (\%) (1/2).}\label{tab:attrshares_n1}\\
\toprule
\rowcolor{SkillPaleBlue} Benchmark & Errors & Attrib. & Annot. & Cover. \\
\midrule
\endfirsthead
\multicolumn{5}{l}{\tablename~\thetable{} (continued)}\\
\toprule
\rowcolor{SkillPaleBlue} Benchmark & Errors & Attrib. & Annot. & Cover. \\
\midrule
\endhead
\bottomrule
\endfoot
\textbf{Pooled, instance-weighted} & 3,058 & 2,322 & 5.0 & 2.2 \\
\textbf{Pooled, item-weighted} & 3,058 & 2,322 & 5.0 & 2.2 \\
\midrule
\mbox{CAViAR} & 60 & 42 & 38.1 & 0.0 \\
\mbox{SmartHome-Bench} & 59 & 42 & 28.6 & 0.0 \\
\mbox{VideoAds} & 44 & 35 & 22.9 & 0.0 \\
\mbox{UCF101-AD} & 151 & 109 & 21.1 & 0.0 \\
\mbox{VideoEspresso} & 47 & 33 & 12.1 & 0.0 \\
\mbox{EgoPlan-Bench} & 194 & 85 & 11.8 & 0.0 \\
\mbox{ScaleLong} & 89 & 58 & 10.3 & 8.6 \\
\mbox{ALLVB} & 63 & 43 & 9.3 & 9.3 \\
\mbox{VRU-Accident} & 58 & 44 & 9.1 & 0.0 \\
\mbox{LongVideo-Reason} & 57 & 36 & 8.3 & 0.0 \\
\mbox{CG-Bench} & 43 & 26 & 7.7 & 3.9 \\
\mbox{KFS-Bench} & 84 & 67 & 7.5 & 6.0 \\
\mbox{SLVMBench} & 51 & 41 & 7.3 & 0.0 \\
\mbox{GUI-World} & 24 & 19 & 5.3 & 0.0 \\
\mbox{iSafetyBench} & 157 & 127 & 3.9 & 0.0 \\
\mbox{VNBench} & 65 & 51 & 3.9 & 21.6 \\
\mbox{HLV-1K} & 83 & 53 & 3.8 & 9.4 \\
\mbox{MME-VideoOCR} & 38 & 33 & 3.0 & 0.0 \\
\mbox{OVO-S-Bench} & 123 & 93 & 2.1 & 4.3 \\
\mbox{UrbanVideo-Bench} & 126 & 90 & 1.1 & 0.0 \\
\mbox{TOC-Bench} & 119 & 106 & 0.9 & 0.0 \\
\mbox{M3-Verse} & 148 & 122 & 0.8 & 0.0 \\
\mbox{Causal2Needles} & 86 & 65 & 0.0 & 0.0 \\
\mbox{CineTechBench} & 27 & 22 & 0.0 & 0.0 \\
\mbox{Colon-Bench} & 56 & 50 & 0.0 & 0.0 \\
\mbox{ContPhy} & 50 & 43 & 0.0 & 0.0 \\
\mbox{E.T. Bench} & 99 & 86 & 0.0 & 0.0 \\
\mbox{FineBench} & 41 & 32 & 0.0 & 6.2 \\
\mbox{ForeSeaQA} & 44 & 35 & 0.0 & 0.0 \\
\mbox{HERBench} & 114 & 79 & 0.0 & 5.1 \\
\mbox{K9-Bench} & 56 & 44 & 0.0 & 0.0 \\
\mbox{NaturalBench-Video} & 22 & 21 & 0.0 & 9.5 \\
\mbox{PLM-VideoBench} & 31 & 29 & 0.0 & 0.0 \\
\mbox{ReactHuman} & 126 & 120 & 0.0 & 0.0 \\
\mbox{S-EMBER} & 168 & 129 & 0.0 & 6.2 \\
\mbox{TaskMeAnything-v1} & 98 & 87 & 0.0 & 0.0 \\
\mbox{VANE-Bench} & 30 & 26 & 0.0 & 0.0 \\
\mbox{VF-Eval} & 35 & 25 & 0.0 & 8.0 \\
\mbox{VGenST-Bench} & 23 & 21 & 0.0 & 0.0 \\
\mbox{Video-LevelGauge} & 24 & 21 & 0.0 & 0.0 \\
\end{xltabular}
\endgroup

\begingroup
\setlength{\LTcapwidth}{\linewidth}
\setlength{\tabcolsep}{4pt}
\begin{xltabular}{\linewidth}{>{\raggedright\arraybackslash}p{159.32bp}>{\raggedleft\arraybackslash}p{61.79bp}>{\raggedleft\arraybackslash}p{63.50bp}>{\raggedleft\arraybackslash}p{65.07bp}>{\raggedleft\arraybackslash}p{67.90bp}}
\caption{Claude-role attribution and exclusion rates on new benchmarks (\%) (2/2).}\label{tab:attrshares_n1_cols2}\\
\toprule
\rowcolor{SkillPaleBlue} Benchmark & Capab. & Disput. & Unstab. & Unverif. \\
\midrule
\endfirsthead
\multicolumn{5}{l}{\tablename~\thetable{} (continued)}\\
\toprule
\rowcolor{SkillPaleBlue} Benchmark & Capab. & Disput. & Unstab. & Unverif. \\
\midrule
\endhead
\bottomrule
\endfoot
\textbf{Pooled, instance-weighted} & 92.7 & 19.9 & 1.1 & 3.0 \\
\textbf{Pooled, item-weighted} & 92.7 & 19.9 & 1.1 & 3.0 \\
\midrule
\mbox{CAViAR} & 61.9 & 20.0 & 1.7 & 8.3 \\
\mbox{SmartHome-Bench} & 71.4 & 28.8 & 0.0 & 0.0 \\
\mbox{VideoAds} & 77.1 & 18.2 & 2.3 & 0.0 \\
\mbox{UCF101-AD} & 78.9 & 27.8 & 0.0 & 0.0 \\
\mbox{VideoEspresso} & 87.9 & 23.4 & 0.0 & 6.4 \\
\mbox{EgoPlan-Bench} & 88.2 & 34.0 & 0.0 & 22.2 \\
\mbox{ScaleLong} & 81.0 & 27.0 & 1.1 & 6.7 \\
\mbox{ALLVB} & 81.4 & 27.0 & 0.0 & 4.8 \\
\mbox{VRU-Accident} & 90.9 & 22.4 & 0.0 & 1.7 \\
\mbox{LongVideo-Reason} & 91.7 & 29.8 & 3.5 & 3.5 \\
\mbox{CG-Bench} & 88.5 & 32.6 & 0.0 & 7.0 \\
\mbox{KFS-Bench} & 86.6 & 13.1 & 3.6 & 3.6 \\
\mbox{SLVMBench} & 92.7 & 17.6 & 2.0 & 0.0 \\
\mbox{GUI-World} & 94.7 & 16.7 & 0.0 & 4.2 \\
\mbox{iSafetyBench} & 96.1 & 15.3 & 0.6 & 3.2 \\
\mbox{VNBench} & 74.5 & 16.9 & 1.5 & 3.1 \\
\mbox{HLV-1K} & 86.8 & 33.7 & 0.0 & 2.4 \\
\mbox{MME-VideoOCR} & 97.0 & 5.3 & 2.6 & 5.3 \\
\mbox{OVO-S-Bench} & 93.5 & 22.8 & 0.8 & 0.8 \\
\mbox{UrbanVideo-Bench} & 98.9 & 28.6 & 0.0 & 0.0 \\
\mbox{TOC-Bench} & 99.1 & 10.9 & 0.0 & 0.0 \\
\mbox{M3-Verse} & 99.2 & 15.5 & 2.0 & 0.0 \\
\mbox{Causal2Needles} & 100.0 & 22.1 & 1.2 & 1.2 \\
\mbox{CineTechBench} & 100.0 & 18.5 & 0.0 & 0.0 \\
\mbox{Colon-Bench} & 100.0 & 10.7 & 0.0 & 0.0 \\
\mbox{ContPhy} & 100.0 & 14.0 & 0.0 & 0.0 \\
\mbox{E.T. Bench} & 100.0 & 10.1 & 2.0 & 1.0 \\
\mbox{FineBench} & 93.8 & 19.5 & 2.4 & 0.0 \\
\mbox{ForeSeaQA} & 100.0 & 15.9 & 2.3 & 2.3 \\
\mbox{HERBench} & 94.9 & 27.2 & 3.5 & 0.0 \\
\mbox{K9-Bench} & 100.0 & 17.9 & 3.6 & 0.0 \\
\mbox{NaturalBench-Video} & 90.5 & 4.5 & 0.0 & 0.0 \\
\mbox{PLM-VideoBench} & 100.0 & 6.5 & 0.0 & 0.0 \\
\mbox{ReactHuman} & 100.0 & 4.0 & 0.0 & 0.8 \\
\mbox{S-EMBER} & 93.8 & 18.5 & 2.4 & 2.4 \\
\mbox{TaskMeAnything-v1} & 100.0 & 11.2 & 0.0 & 0.0 \\
\mbox{VANE-Bench} & 100.0 & 10.0 & 3.3 & 0.0 \\
\mbox{VF-Eval} & 92.0 & 25.7 & 0.0 & 2.9 \\
\mbox{VGenST-Bench} & 100.0 & 4.3 & 4.3 & 0.0 \\
\mbox{Video-LevelGauge} & 100.0 & 12.5 & 0.0 & 0.0 \\
\end{xltabular}
\endgroup
\endgroup

All 1,826 yes-or-no items in VF-Eval \citep{song2025vfeval} have answer yes, and 282 CAViAR \citep{garg2026caviar} items carry conflicting labels. Blind accuracy falls 11 to 21\% below chance on 4 sources, namely VideoSIAH-Eval \citep{yang2026longvt}, VidPair-Halluc \citep{huang2026vidpairhalluc}, PerceptionComp \citep{li2026perceptioncomp}, and Artifact-Bench \citep{tang2026artifactbench}. These results flag formats whose question-answer relations mislead the blind reader. Chance metadata also needs checking. 33 multiple-choice benchmarks omit it, and sampled option counts differ by at least 5\% on 13 of 15 with declarations. Micro versus macro averaging reverses 79 of 1,344 system pairs. Eight benchmarks are flagged by three of the four independent screens. CAViAR, ExAct \citep{yi2025exact}, LongTimeScope \citep{zohar2025apollo2}, TimeScope \citep{zohar2025timescope}, and UCF101-AD break at or before the frame level, iSafetyBench \citep{abdullah2025isafetybench} breaks at the order level, and Colon-Bench \citep{hamdi2026colonbench} and SmartHome-Bench \citep{zhao2025smarthomebench} survive. Those screens assess protocol errors, effective size, near-duplicates, and text predictability.

\section{Benchmark recommendations by capability}
\label{app:whichbench}
A benchmark that survives every level holds a certificate that none of our attackers can forge. Of the 115 benchmarks with a certificate, 38 survive. For perception, 13 of 45, namely Colon-Bench \citep{hamdi2026colonbench}, K9-Bench \citep{attarde2026k9bench}, KFS-Bench \citep{li2026kfsbench}, MME-VideoOCR \citep{shi2025mmevideoocr}, NaturalBench-Video \citep{li2025naturalbenchvideo}, Perception Test \citep{patraucean2023perceptiontest}, Q-Bench-Video \citep{zhang2025qbenchvideo}, ST-Evidence \citep{wang2026stevidence}, ScaleLong \citep{ma2025scalelong}, TimeBlind \citep{li2026timeblind}, VELOCITI \citep{saravanan2025velociti}, VNBench \citep{zhao2025vnbench}, and ViMUL-Bench \citep{shafique2025vimul}. For temporal understanding, 10 of 18, namely AoTBench \citep{xue2025aotbench}, E.T. Bench \citep{liu2024etbench}, MotionBench \citep{hong2025motionbench}, PhysicalAI \citep{zhou2026paibench}, S-EMBER \citep{wang2026sember}, SLVMBench \citep{yang2026slvmbench}, TOMATO \citep{shangguan2024tomato}, TVBench \citep{cores2024tvbench}, VCR-Bench \citep{qi2025vcrbench}, and Vinoground \citep{zhang2024vinoground}. For spatial and physical understanding, 7 of 20, namely CMU \citep{tan2026acam}, ContPhy \citep{zheng2024contphy}, OSI-Bench \citep{wu2026osibench}, SIS-Bench \citep{zou2026sisbench}, UrbanVideo-Bench \citep{zhao2025urbanvideo}, VGenST-Bench \citep{park2026vgenstbench}, and VSI-Bench \citep{yang2025vsibench}. For reasoning and knowledge, 8 of 32, namely GUI-World \citep{chen2025guiworld}, MORSE-500 \citep{cai2025morse500}, Neptune \citep{nagrani2024neptune}, PAI-Bench-U \citep{zhou2026paibench}, SFI-Bench \citep{zhang2026sfibench}, SciVideoBench \citep{deng2025scivideobench}, SmartHome-Bench \citep{zhao2025smarthomebench}, and Video-Holmes \citep{cheng2025videoholmes}. A survivor resists the attackers we tried and certifies nothing beyond them. Its report card names the level at which it came closest to breaking, and a user who needs the strongest guard for one claim can rank the survivors of a group by that margin. Tables~\ref{tab:rankperception}, \ref{tab:ranktemporal}, \ref{tab:rankspatial}, and~\ref{tab:rankreasoning} give that ranking for the ten strongest benchmarks of each group, where the margin is the breaking threshold $s^{*}-c-\delta$ minus the largest exploitability over the measured levels, in \%.
\begin{table}[!htbp]
\centering
\caption{Perception. Ten strongest of 45 benchmarks, 13 of which survive, ranked by breaking level and then by the margin below the breaking threshold (\%).}\label{tab:rankperception}
\begin{tabularx}{\linewidth}{r>{\raggedright\arraybackslash}Xlr}
\toprule
\rowcolor{SkillPaleBlue} Rank & Benchmark & Level & Margin \\
\midrule
1 & \mbox{VNBench} & Survives & +17.1 \\
2 & \mbox{NaturalBench-Video} & Survives & +15.7 \\
3 & \mbox{MME-VideoOCR} & Survives & +12.3 \\
4 & \mbox{Colon-Bench} & Survives & +5.7 \\
5 & \mbox{ViMUL-Bench} & Survives & +5.5 \\
6 & \mbox{Q-Bench-Video} & Survives & +3.3 \\
7 & \mbox{TimeBlind} & Survives & +3.2 \\
8 & \mbox{K9-Bench} & Survives & +2.8 \\
9 & \mbox{KFS-Bench} & Survives & +2.3 \\
10 & \mbox{ScaleLong} & Survives & +1.4 \\
\bottomrule
\end{tabularx}
\end{table}

\begin{table}[!htbp]
\centering
\caption{Temporal. Ten strongest of 18 benchmarks, 10 of which survive, ranked by breaking level and then by the margin below the breaking threshold (\%).}\label{tab:ranktemporal}
\begin{tabularx}{\linewidth}{r>{\raggedright\arraybackslash}Xlr}
\toprule
\rowcolor{SkillPaleBlue} Rank & Benchmark & Level & Margin \\
\midrule
1 & \mbox{VCR-Bench} & Survives & +9.8 \\
2 & \mbox{TVBench} & Survives & +8.2 \\
3 & \mbox{PhysicalAI} & Survives & +8.1 \\
4 & \mbox{TOMATO} & Survives & +6.4 \\
5 & \mbox{MotionBench} & Survives & +3.8 \\
6 & \mbox{E.T. Bench} & Survives & +1.3 \\
7 & \mbox{SLVMBench} & Survives & +1.3 \\
8 & \mbox{AoTBench} & Survives & +1.2 \\
9 & \mbox{S-EMBER} & Survives & +0.7 \\
10 & \mbox{Vinoground} & Survives & +0.5 \\
\bottomrule
\end{tabularx}
\end{table}

\begin{table}[!htbp]
\centering
\caption{Spatial / physical. Ten strongest of 20 benchmarks, 7 of which survive, ranked by breaking level and then by the margin below the breaking threshold (\%).}\label{tab:rankspatial}
\begin{tabularx}{\linewidth}{r>{\raggedright\arraybackslash}Xlr}
\toprule
\rowcolor{SkillPaleBlue} Rank & Benchmark & Level & Margin \\
\midrule
1 & \mbox{VGenST-Bench} & Survives & +12.0 \\
2 & \mbox{SIS-Bench} & Survives & +5.3 \\
3 & \mbox{CMU} & Survives & +4.1 \\
4 & \mbox{ContPhy} & Survives & +3.6 \\
5 & \mbox{VSI-Bench} & Survives & +2.7 \\
6 & \mbox{OSI-Bench} & Survives & +1.0 \\
7 & \mbox{UrbanVideo-Bench} & Survives & +0.0 \\
8 & \mbox{ReVSI} & Order & -3.3 \\
9 & \mbox{M3-Verse} & Order & -3.4 \\
10 & \mbox{VURB} & Order & -7.3 \\
\bottomrule
\end{tabularx}
\end{table}

\begin{table}[!htbp]
\centering
\caption{Reasoning / knowledge. Ten strongest of 32 benchmarks, 8 of which survive, ranked by breaking level and then by the margin below the breaking threshold (\%).}\label{tab:rankreasoning}
\begin{tabularx}{\linewidth}{r>{\raggedright\arraybackslash}Xlr}
\toprule
\rowcolor{SkillPaleBlue} Rank & Benchmark & Level & Margin \\
\midrule
1 & \mbox{PAI-Bench-U} & Survives & +6.0 \\
2 & \mbox{Video-Holmes} & Survives & +4.7 \\
3 & \mbox{Neptune} & Survives & +4.0 \\
4 & \mbox{SciVideoBench} & Survives & +2.3 \\
5 & \mbox{SFI-Bench} & Survives & +2.1 \\
6 & \mbox{SmartHome-Bench} & Survives & +1.7 \\
7 & \mbox{MORSE-500} & Survives & +1.0 \\
8 & \mbox{GUI-World} & Survives & +0.7 \\
9 & \mbox{EgoSchema} & Order & -5.4 \\
10 & \mbox{Video-MMMU} & Order & -5.8 \\
\bottomrule
\end{tabularx}
\end{table}

\section{Pool construction}
\label{app:pool}
Options-only and blind screens use Qwen3-VL-8B \citep{bai2025qwen3} margins over chance across four option permutations. The single-frame screen uses the correct option's log-probability margin. The 32-frame and shuffled-frame screens use Qwen3-VL-2B answers, followed by cross-source question deduplication. Records retain source, video identifier, duration, task, event, format, option count, license, attack margins, and model outcomes. They also retain BGE-large \citep{xiao2024c} question embeddings and SigLIP2 \citep{tschannen2025siglip} middle-frame embeddings. An item labeler assigns 18 fine capability categories in four groups from the question, options, and gold answer. A scene labeler uses eight uniform frames and the source name, without question text, to assign one of 19 video-content categories, among them daily life, sports, films, scientific experiments, and an ``other'' class. We retain at least one item per available task-event cell and flag cells below 30 items as thin. Caption attacks run on selected candidates because captions depend on the requested frame sample.

\begin{figure}[!htbp]
\centering
\includegraphics[width=\linewidth]{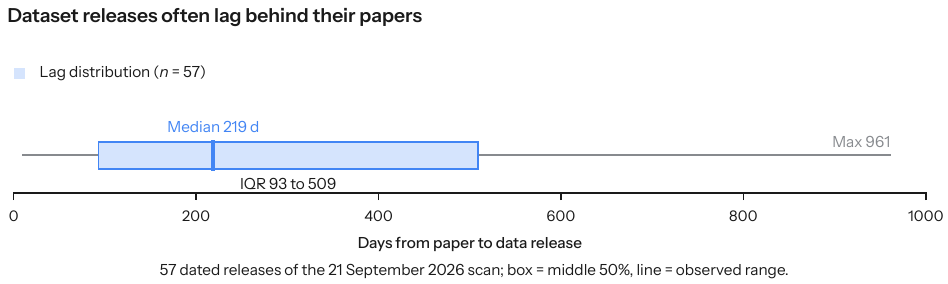} %
\caption{Paper-to-data lag of the 57 dated candidates at the scan of 21 September 2026.}
\label{fig:census}
\end{figure}
The watcher scans registries, technical reports, and release pages every twelve hours, with a retrospective scan over 32 months. At the scan of 21 September 2026, the watcher lists 585 candidates, 69 of them in the pyramid. As shown in Figure~\ref{fig:census}, the median paper-to-data lag over 57 dated releases is 219 days and the middle half spans 93 to 509 days. At that scan, 44 candidates had released only part of their videos or annotations. Each approved source is pinned to a revision and audited before joining a pool snapshot. Compositions can be reissued on later snapshots.
\begin{figure}[!htbp]
\centering
\includegraphics[width=\linewidth]{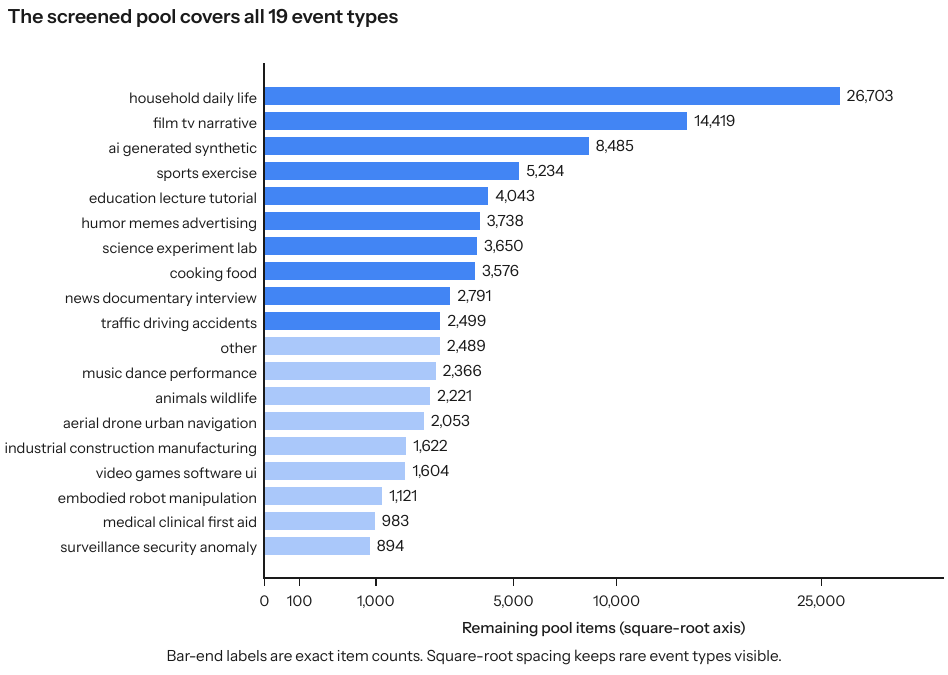} %
\caption{Screened QA counts across nineteen event types. Bar labels give exact counts.}
\label{fig:events}
\end{figure}
As shown in Figure~\ref{fig:events}, household and daily life supplies the most items and surveillance and security anomalies the fewest, 894, and even the rarest class supplies candidates for targeted compositions.

A composition query uses equality and range predicates over task, event, duration, format, option count, and license, with coverage quotas when requested. The lockfile records the round-robin selector's caps and any relaxations. The gate applies each attack's item-level success rule. Options-only and blind checks use margins across four option permutations. Single-frame and caption checks use answer correctness. Order checks use shuffled frames or the first tenth. Refilling stops when no item is dropped. The JSON lockfile records snapshot hash, query, selector version, seed, item identifiers, option permutations, round count, and residual exploitability. Its SHA-256 hash names the released benchmark.

We release Video-Index alongside other meta-benchmarks that prioritize coverage breadth, namely the 10,000-item coverage composition and a separate 834-item coverage census that draws up to eight items from each source without attacker filtering.
Tables~\ref{tab:metahardattrs} to~\ref{tab:sourceslevels_cols2} give, in turn, Video-Index accuracy by duration, viewpoint, and authenticity, the census scores and video gains, the scene clusters by duration, the coverage composition by scene and duration, and each source's contribution against its attack results.

\paragraph{Collecting the meta-benchmark.}
The meta-benchmark is a difficulty-first census of the screened pool. It contains 840 items and uses the following construction procedure.

\paragraph{Candidates and attacker difficulty.}
We start from 63{,}630 multiple-choice items from the 115 audited benchmarks. Every screening step is complete, and no declared attacker solves these items. The attackers receive options only, question text, a single frame, 32 frames with the 2B model, or shuffled frames. For each item, we rank each of its five attacker margins by percentile within the pool. The largest percentile is the item's worst-case attacker percentile. A small value means that even the strongest attacker remains far from the answer. We retain the 3{,}000 lowest-ranked items, with one item per video and ties broken by item identifier.

\paragraph{Capability labels.}
Claude Opus 5 \citep{anthropic2026opus5} reads the question, options, and gold answer without frames. It assigns the primary capability needed to answer from video, choosing among the 18 capability categories of the attribution taxonomy. The five data-and-protocol categories are withheld. Each category belongs to Perception, Temporal, Spatial / physical, or Reasoning / knowledge. The same call flags first-person wording and questions about presence or authenticity. These flags and video duration replace the former long-video, egocentric and hallucination classes. We report them as appendix attributes.

\paragraph{Answer verification.}
Selecting items that defeat every attacker also concentrates questionable gold answers. In the audit of the unverified census, a third were confirmed and a quarter contradicted by the frames. We therefore check every candidate before selection. Claude Opus 5 receives the question, options, marked answer and video frames sampled at 1\,fps, initially capped at 32. Items not confirmed at that budget are reread with the evaluation budget of 512 frames, and the later verdict takes precedence.

The verifier returns one of four verdicts. Supported means that the frames establish the marked answer and rule out the alternatives. Contradicted means that another option is visibly correct. Not verifiable means that the available frames neither confirm nor rule out the marked answer. Definition dependent means that the answer depends on a benchmark-specific definition or subjective judgment.

We keep supported items and reversed clips from AoTBench \citep{xue2025aotbench} RTime and TempCompass \citep{liu2024tempcompass} direction, whose labels describe the reversed action by construction. We also keep not-verifiable items from videos of at least 60\,s when the sampled frames provide no contradictory evidence. We drop contradicted and definition-dependent items, and not-verifiable items from shorter videos. Two format checks precede verification. VidPair-Halluc ordering questions whose choices are individual events and whose stored answer is a single letter cannot be scored as single-choice items. We exclude them. Options stored as label/text records are rendered as text.

Of the 840 unverified census items, verification keeps 468 and drops 372, namely 205 contradicted, 127 not verifiable on short videos, and 40 definition dependent. We then check the next 2{,}095 hardest labeled items on previously unused videos. This step keeps 1{,}208 and drops 887, namely 409 contradicted, 396 not verifiable, and 82 definition dependent. The verified pool contains 1{,}676 items.

\paragraph{Selection by attacker difficulty.}
Within each of the four capability groups, we take verified items in increasing worst-case attacker percentile, with one item per video, until reaching 210. Reference-model accuracy does not enter the selection rule.

The final census contains 840 items from 76 benchmarks, namely 468 retained from the unverified census and 372 replacements. Verification keeps 533 as supported, 219 as not verifiable on long videos, and 88 as reversed clips. The worst-case attacker percentile has median 0.14 and maximum 0.23. Mean chance is 29.7\,\%. The largest sources are S-EMBER \citep{wang2026sember} with 109 items, AoTBench with 92, and ALLVB \citep{tan2025allvb} with 45. The reference answers 477 items correctly, or 56.8\,\%. It answers 131 correctly in the perception group, 118 in the temporal group, 98 in the spatial group, and 130 in the reasoning group. This score measures performance on the selected set. The previous composition imposed a reference-accuracy cap and shares 738 items with this release.

\paragraph{Protocols and scoring.}
The video protocol samples one frame per second, capped at 512 frames, and resizes frames to a short side of 224 pixels. Longer sequences are uniformly subsampled. Because the normalized store holds at most 1{,}024 frames, videos longer than 1{,}024\,s are sampled below 1\,fps. When memory or context limits reject a request, the runner halves the frame count and retries. It records the number of frames actually sent.

The blind protocol supplies the question and options without video. It averages accuracy over four option permutations per item using seed 42. Video gain is video accuracy minus blind accuracy. Every model receives the same introductory line, question, lettered choices and instruction to return an option letter. A shared letter-extraction scorer evaluates the response. Each model applies its released preprocessing to the shared frames.

\paragraph{Model implementations.}
We use the 8B variant when available, otherwise the largest smaller variant, namely VideoChat3-4B \citep{li2026videochat3}, LLaVA-OneVision-2-8B \citep{an2026llava}, Qwen3-VL-8B \citep{bai2025qwen3}, InternVL3.5-8B \citep{wang2025internvl3}, Molmo2-8B \citep{clark2026molmo2}, VideoLLaMA3-7B \citep{zhang2025videollama}, Video-XL-2 \citep{qin2025video} at 7B, InternVideo2.5-Chat-8B \citep{wang2025internvideo2}, and the standard Cambrian-S-7B \citep{yang2026cambrian} checkpoint. Claude Opus 5 is the API reference. Qwen3-VL participates in pool construction and is excluded from ranking-agreement statistics.

Qwen3-VL uses vLLM 0.11 \citep{kwon2023efficient}. The remaining models use their released code. LLaVA-OneVision-2 and VideoChat3 run with transformers \citep{wolf2020transformers} 5.16, Molmo2 and InternVL3.5 with 4.57, VideoLLaMA3, Video-XL-2, and Cambrian-S with 4.46, and InternVideo2.5 with 4.40. Each runs on one B200. InternVideo2.5 groups consecutive frames in fours and rounds its input count down to a multiple of four. VideoLLaMA3 falls back to 64 or 32 frames on 38\% of items after rejecting other counts.

\paragraph{Comparison census.}
Table~\ref{tab:metabenchcov} reports a separate coverage census drawn from audit samples without attacker filtering. It takes up to eight items per benchmark, stratified by the benchmark's task labels, for 834 items across 107 benchmarks. The reported comparison gives reference accuracy 66.7\% and blind accuracy 49.1\% on this census. On the released Video-Index composition, reference accuracy is 56.8\%, mean chance is 29.7\%, and blind accuracy is 21.3\%. Both video and blind accuracy are lower than on the coverage census.

\paragraph{Benchmark capability groups.}
The same labeler assigns groups to a stratified sample of 30 items per benchmark, totaling 3{,}124 items, and to all 720 sampled items of CG-Bench \citep{chen2025cgbench}, EgoPlan-Bench \citep{chen2025egoplan}, and PLM-VideoBench \citep{cho2025perceptionlm}. Each benchmark takes its majority group. Perception covers 45 benchmarks, reasoning and knowledge 32, spatial and physical 20, and temporal 18. Only audited benchmarks contribute to the group totals.
\begingroup
\setlength{\LTcapwidth}{\linewidth}
\setlength{\tabcolsep}{4pt}
\begin{xltabular}{\linewidth}{>{\raggedright\arraybackslash}p{117.36bp}>{\raggedleft\arraybackslash}p{40.32bp}>{\raggedleft\arraybackslash}p{30.27bp}>{\raggedleft\arraybackslash}p{27.55bp}>{\raggedleft\arraybackslash}p{59.19bp}>{\raggedleft\arraybackslash}p{61.72bp}>{\raggedleft\arraybackslash}p{65.16bp}}
\caption{Video-Index accuracy (\%) by duration, viewpoint, and presence/authenticity.}\label{tab:metahardattrs}\\
\toprule
\rowcolor{SkillPaleBlue} Model & Overall & Blind & Gain & Long video & First person & Authenticity \\
\midrule
\endfirsthead
\multicolumn{7}{l}{\tablename~\thetable{} (continued)}\\
\toprule
\rowcolor{SkillPaleBlue} Model & Overall & Blind & Gain & Long video & First person & Authenticity \\
\midrule
\endhead
\bottomrule
\endfoot
claude-opus-5 & 56.8 & 21.3 & 35.5 & 45.0 & 37.6 & 67.6 \\
Molmo2-8B & 19.0 & 4.8 & 14.2 & 9.9 & 10.2 & 26.8 \\
InternVL3.5-8B & 15.2 & 4.4 & 10.8 & 6.0 & 5.7 & 16.9 \\
LLaVA-OneVision-2-8B & 14.2 & 1.8 & 12.4 & 10.3 & 8.9 & 4.2 \\
Video-XL-2 & 12.4 & 7.3 & 5.1 & 9.2 & 6.4 & 14.1 \\
InternVideo2.5-8B & 12.4 & 7.5 & 4.9 & 6.0 & 3.8 & 8.5 \\
VideoLLaMA3-7B & 12.4 & 5.2 & 7.1 & 5.7 & 6.4 & 7.0 \\
Cambrian-S-7B & 11.3 & 5.3 & 6.0 & 7.8 & 3.8 & 8.5 \\
Qwen3-VL-8B$^\dagger$ & 9.5 & 1.4 & 8.1 & 8.5 & 6.4 & 5.6 \\
VideoChat3-4B & 9.4 & 3.8 & 5.7 & 4.6 & 3.8 & 15.5 \\
\midrule
Items & 840 & -- & -- & 282 & 157 & 71 \\
\end{xltabular}
\endgroup
\begingroup
\setlength{\LTcapwidth}{\linewidth}
\setlength{\tabcolsep}{4pt}
\begin{xltabular}{\linewidth}{>{\raggedright\arraybackslash}p{169.88bp}>{\raggedleft\arraybackslash}p{92.84bp}>{\raggedleft\arraybackslash}p{82.79bp}>{\raggedleft\arraybackslash}p{80.07bp}}
\caption{Coverage-census accuracy and video gain (\%) without attacker filtering (1/2).}\label{tab:metabenchcov}\\
\toprule
\rowcolor{SkillPaleBlue} Model & Overall & Blind & Gain \\
\midrule
\endfirsthead
\multicolumn{4}{l}{\tablename~\thetable{} (continued)}\\
\toprule
\rowcolor{SkillPaleBlue} Model & Overall & Blind & Gain \\
\midrule
\endhead
\bottomrule
\endfoot
claude-opus-5 & 66.7 & 48.8 & 17.9 \\
Molmo2-8B & 58.8 & 40.3 & 18.4 \\
InternVL3.5-8B & 56.4 & -- & -- \\
Qwen3-VL-8B$^\dagger$ & 55.1 & 39.1 & 16.1 \\
LLaVA-OneVision-2-8B & 54.5 & 36.5 & 18.1 \\
Video-XL-2 & 53.8 & 39.8 & 14.0 \\
VideoChat3-4B & 51.8 & 38.9 & 12.8 \\
Cambrian-S-7B & 50.4 & 36.7 & 13.7 \\
InternVideo2.5-8B & 50.1 & 39.8 & 10.2 \\
VideoLLaMA3-7B & 49.5 & 16.0 & 33.5 \\
\midrule
Items & 827 & -- & -- \\
\end{xltabular}
\endgroup

\begingroup
\setlength{\LTcapwidth}{\linewidth}
\setlength{\tabcolsep}{4pt}
\begin{xltabular}{\linewidth}{>{\raggedright\arraybackslash}p{137.78bp}>{\raggedleft\arraybackslash}p{76.25bp}>{\raggedleft\arraybackslash}p{70.65bp}>{\raggedleft\arraybackslash}p{57.76bp}>{\raggedleft\arraybackslash}p{75.15bp}}
\caption{Coverage-census accuracy and video gain (\%) without attacker filtering (2/2).}\label{tab:metabenchcov_cols2}\\
\toprule
\rowcolor{SkillPaleBlue} Model & Perception & Temporal & Spatial & Reasoning \\
\midrule
\endfirsthead
\multicolumn{5}{l}{\tablename~\thetable{} (continued)}\\
\toprule
\rowcolor{SkillPaleBlue} Model & Perception & Temporal & Spatial & Reasoning \\
\midrule
\endhead
\bottomrule
\endfoot
claude-opus-5 & 72.1 & 62.6 & 49.5 & 79.0 \\
Molmo2-8B & 66.0 & 51.6 & 47.8 & 63.6 \\
InternVL3.5-8B & 60.6 & 48.8 & 46.0 & 73.1 \\
Qwen3-VL-8B$^\dagger$ & 62.5 & 49.7 & 44.6 & 58.0 \\
LLaVA-OneVision-2-8B & 59.9 & 56.1 & 44.0 & 54.5 \\
Video-XL-2 & 62.5 & 49.7 & 41.3 & 55.1 \\
VideoChat3-4B & 57.4 & 47.7 & 38.6 & 59.1 \\
Cambrian-S-7B & 57.4 & 49.0 & 37.5 & 52.8 \\
InternVideo2.5-8B & 55.8 & 45.2 & 40.2 & 54.5 \\
VideoLLaMA3-7B & 55.4 & 45.2 & 39.7 & 52.8 \\
\midrule
Items & 312 & 155 & 184 & 176 \\
\end{xltabular}
\endgroup

\paragraph{Constructing the coverage composition.}
The screening funnel leaves 62,142 multiple-choice items on 38,426 videos. No declared attacker solves them, every screening step is complete, and each video has a duration and scene label. The attackers use options only, question text, a single frame, 32 frames with the 2B model, or shuffled frames. We select 10,000 items to balance video durations, cover scene and question types, and retain difficulty under these attackers.

\paragraph{Duration groups and visual scenes.}
We use seven duration groups, namely $<$10\,s, 10--30\,s, 30--60\,s, 1--3\,min, 3--10\,min, 10--30\,min, and $>$30\,min. The mean SigLIP2 embedding of 32 uniform frames approaches the global centroid for long videos with many scenes. We therefore detect visual near-duplicates separately within each duration group. Each threshold matches the random-pair false-match rate of cosine similarity at least 0.90 in the shortest group. In increasing duration order, the thresholds are 0.903, 0.853, 0.882, 0.946, 0.901, 0.901, and 0.975. Greedy leader clustering produces 19,263 scene clusters over 38,835 videos, as reported in Table~\ref{tab:scenegroups}.

\paragraph{Coverage and selection constraints.}
Selection cycles through duration groups, then the 19 scene types, then source benchmarks. Budget flows to strata that still contain admissible items. The initial caps allow one item per video, one video per scene cluster, five items per benchmark--question-template pair, and 5\% of the budget per benchmark. A template replaces entities, numbers and nouns in the question with placeholders. Within each benchmark, we first prefer the least-templated item, then the item with the weakest strongest attacker. Attacker strength is the largest percentile rank among the five recorded margins.

A preliminary coverage pass selects one item per benchmark--question-type pair, covering 735 of 736 pairs. This pass ignores cluster and template caps to keep the rare question types of fully templated benchmarks. A duration--scene cell is scarce when its supply falls below an equal share of the budget. Such cells allow up to three items per video and lift cluster and template caps.

If a duration group still cannot fill its share, we raise the cluster cap to four and the template cap to ten, then lift its benchmark cap. These relaxations are needed for the 3--10\,min, 10--30\,min and $>$30\,min groups. Only the 10--30\,min group exceeds the benchmark cap, for S-EMBER \citep{wang2026sember} and Video-MME-v2 \citep{fu2026videommev2}.

\paragraph{Selected composition.}
The result contains 10,000 items from 9,688 videos, 9,045 scene clusters, and 105 benchmarks. Every duration group includes all 19 scene types. The 230 videos with two or three selected items all belong to scarce cells. In increasing duration order, item counts are 1,432, 1,434, 1,432, 1,427, 1,427, 1,422, and 1,426. Median duration is 1.6\,min.

The largest scene types are household daily life with 1,089 items, film and television narrative with 1,077, education and tutorials with 787, news and documentary with 766, and humor and advertising with 749. The smallest are industrial and construction scenes with 285 items, embodied robot manipulation with 270, and medical and first-aid scenes with 194.

A difficulty-first variant restricts selection to items whose worst attacker percentile is at most 0.7. It raises the share hard for all five attackers from 29\% to 50\%, but yields 1,015--1,884 items per duration group and places 12.8\% in one benchmark. We retain the coverage-first composition, detailed in Tables~\ref{tab:videoindexsources} and~\ref{tab:videoindexcomposition}.
\begingroup

\begingroup
\setlength{\LTcapwidth}{\linewidth}
\setlength{\tabcolsep}{4pt}
\begin{xltabular}{\linewidth}{>{\raggedright\arraybackslash}p{76.85bp}>{\raggedleft\arraybackslash}p{60.62bp}>{\raggedleft\arraybackslash}p{57.69bp}>{\raggedleft\arraybackslash}p{76.84bp}>{\raggedleft\arraybackslash}p{59.69bp}>{\raggedleft\arraybackslash}p{77.90bp}}
\caption{Visual-scene clusters by duration at matched false-match rates (1/2).}\label{tab:scenegroups}\\
\toprule
\rowcolor{SkillPaleBlue} Duration & Videos & Items & Threshold & Scenes & Singletons \\
\midrule
\endfirsthead
\multicolumn{6}{l}{\tablename~\thetable{} (continued)}\\
\toprule
\rowcolor{SkillPaleBlue} Duration & Videos & Items & Threshold & Scenes & Singletons \\
\midrule
\endhead
\bottomrule
\endfoot
$<$10 s & 9,445 & 11,043 & 0.903 & 6,003 & 5,243 \\
10--30 s & 11,592 & 14,943 & 0.853 & 4,823 & 3,356 \\
30--60 s & 5,146 & 7,457 & 0.882 & 2,254 & 1,610 \\
1--3 min & 3,122 & 5,313 & 0.946 & 1,787 & 1,396 \\
3--10 min & 5,028 & 9,178 & 0.901 & 2,013 & 1,333 \\
10--30 min & 2,248 & 4,376 & 0.901 & 1,030 & 681 \\
$>$30 min & 2,254 & 10,451 & 0.975 & 1,353 & 1,157 \\
\midrule
All & 38,835 & 62,761 & -- & 19,263 & 14,776 \\
\end{xltabular}
\endgroup

\begingroup
\setlength{\LTcapwidth}{\linewidth}
\setlength{\tabcolsep}{4pt}
\begin{xltabular}{\linewidth}{>{\raggedright\arraybackslash}p{103.83bp}>{\raggedleft\arraybackslash}p{90.38bp}>{\raggedright\arraybackslash}p{114.74bp}>{\raggedleft\arraybackslash}p{116.64bp}}
\caption{Visual-scene clusters by duration at matched false-match rates (2/2).}\label{tab:scenegroups_cols2}\\
\toprule
\rowcolor{SkillPaleBlue} Duration & Largest & Sharing & Scenes @0.90 \\
\midrule
\endfirsthead
\multicolumn{4}{l}{\tablename~\thetable{} (continued)}\\
\toprule
\rowcolor{SkillPaleBlue} Duration & Largest & Sharing & Scenes @0.90 \\
\midrule
\endhead
\bottomrule
\endfoot
$<$10 s & 349 & 4,202 (44\%) & 5,943 \\
10--30 s & 248 & 8,236 (71\%) & 6,348 \\
30--60 s & 123 & 3,536 (69\%) & 2,537 \\
1--3 min & 117 & 1,726 (55\%) & 1,315 \\
3--10 min & 150 & 3,695 (73\%) & 1,999 \\
10--30 min & 79 & 1,567 (70\%) & 1,018 \\
$>$30 min & 136 & 1,097 (49\%) & 685 \\
\midrule
All & -- & 24,059 (62\%) & 19,845 \\
\end{xltabular}
\endgroup

\endgroup

\begingroup

\begingroup
\setlength{\LTcapwidth}{\linewidth}
\setlength{\tabcolsep}{4pt}
\begin{xltabular}{\linewidth}{>{\raggedright\arraybackslash}p{211.97bp}>{\raggedleft\arraybackslash}p{51.40bp}>{\raggedleft\arraybackslash}p{51.40bp}>{\raggedleft\arraybackslash}p{51.40bp}>{\raggedleft\arraybackslash}p{51.40bp}}
\caption{Coverage-pool composition by scene and duration; brackets give eligible counts (1/2).}\label{tab:videoindexcomposition}\\
\toprule
\rowcolor{SkillPaleBlue} Scene type & $<$10 s & 10--30 s & 30--60 s & 1--3 min \\
\midrule
\endfirsthead
\multicolumn{5}{l}{\tablename~\thetable{} (continued)}\\
\toprule
\rowcolor{SkillPaleBlue} Scene type & $<$10 s & 10--30 s & 30--60 s & 1--3 min \\
\midrule
\endhead
\bottomrule
\endfoot
household\ daily\ life & 82 (829) & 92 (5,424) & 177 (3,209) & 158 (1,507) \\
film\ tv\ narrative & 81 (1,338) & 91 (783) & 81 (166) & 144 (751) \\
education\ lecture\ tutorial & 77 (199) & 76 (429) & 122 (346) & 145 (347) \\
news\ documentary\ interview & 71 (71) & 68 (113) & 66 (66) & 91 (150) \\
humor\ memes\ advertising & 80 (420) & 76 (508) & 148 (306) & 141 (335) \\
sports\ exercise & 82 (1,099) & 75 (719) & 97 (228) & 106 (177) \\
cooking\ food & 74 (196) & 76 (467) & 121 (401) & 70 (108) \\
animals\ wildlife & 74 (158) & 78 (341) & 97 (233) & 119 (241) \\
ai\ generated\ synthetic & 102 (2,606) & 101 (1,550) & 56 (665) & 65 (534) \\
video\ games\ software\ ui & 76 (221) & 78 (283) & 43 (109) & 51 (51) \\
science\ experiment\ lab & 75 (961) & 82 (751) & 98 (394) & 72 (72) \\
music\ dance\ performance & 76 (383) & 78 (455) & 50 (76) & 46 (77) \\
other & 75 (436) & 74 (732) & 51 (311) & 34 (102) \\
traffic\ driving\ accidents & 76 (674) & 78 (906) & 45 (199) & 15 (15) \\
surveillance\ security\ anomaly & 47 (202) & 62 (196) & 53 (55) & 62 (68) \\
aerial\ drone\ urban\ navigation & 77 (354) & 66 (480) & 18 (149) & 30 (347) \\
industrial\ construction\ manufacturing & 70 (440) & 71 (172) & 28 (78) & 44 (44) \\
embodied\ robot\ manipulation & 73 (311) & 75 (181) & 64 (117) & 16 (16) \\
medical\ clinical\ first\ aid & 64 (144) & 37 (174) & 17 (127) & 18 (254) \\
\midrule
All & 1,432 & 1,434 & 1,432 & 1,427 \\
\end{xltabular}
\endgroup

\begingroup
\setlength{\LTcapwidth}{\linewidth}
\setlength{\tabcolsep}{4pt}
\begin{xltabular}{\linewidth}{>{\raggedright\arraybackslash}p{198.34bp}>{\raggedleft\arraybackslash}p{51.40bp}>{\raggedleft\arraybackslash}p{51.40bp}>{\raggedleft\arraybackslash}p{51.40bp}>{\raggedleft\arraybackslash}p{65.04bp}}
\caption{Coverage-pool composition by scene and duration; brackets give eligible counts (2/2).}\label{tab:videoindexcomposition_cols2}\\
\toprule
\rowcolor{SkillPaleBlue} Scene type & 3--10 min & 10--30 min & $>$30 min & Total \\
\midrule
\endfirsthead
\multicolumn{5}{l}{\tablename~\thetable{} (continued)}\\
\toprule
\rowcolor{SkillPaleBlue} Scene type & 3--10 min & 10--30 min & $>$30 min & Total \\
\midrule
\endhead
\bottomrule
\endfoot
household\ daily\ life & 131 (3,801) & 366 (1,775) & 83 (629) & 1,089 (17,174) \\
film\ tv\ narrative & 135 (1,331) & 172 (479) & 373 (6,545) & 1,077 (11,393) \\
education\ lecture\ tutorial & 131 (589) & 175 (331) & 61 (291) & 787 (2,532) \\
news\ documentary\ interview & 126 (344) & 88 (198) & 256 (769) & 766 (1,711) \\
humor\ memes\ advertising & 97 (224) & 80 (192) & 127 (360) & 749 (2,345) \\
sports\ exercise & 103 (326) & 93 (222) & 123 (447) & 679 (3,218) \\
cooking\ food & 85 (376) & 91 (394) & 55 (324) & 572 (2,266) \\
animals\ wildlife & 96 (559) & 59 (152) & 40 (100) & 563 (1,784) \\
ai\ generated\ synthetic & 64 (326) & 40 (41) & 39 (170) & 467 (5,892) \\
video\ games\ software\ ui & 59 (164) & 65 (181) & 41 (201) & 413 (1,210) \\
science\ experiment\ lab & 30 (478) & 26 (134) & 23 (23) & 406 (2,813) \\
music\ dance\ performance & 48 (127) & 46 (51) & 55 (266) & 399 (1,435) \\
other & 26 (152) & 28 (118) & 63 (73) & 351 (1,924) \\
traffic\ driving\ accidents & 62 (64) & 27 (27) & 30 (37) & 333 (1,922) \\
surveillance\ security\ anomaly & 61 (68) & 10 (13) & 9 (21) & 304 (623) \\
aerial\ drone\ urban\ navigation & 66 (71) & 21 (21) & 18 (19) & 296 (1,441) \\
industrial\ construction\ manufacturing & 26 (95) & 26 (38) & 20 (160) & 285 (1,027) \\
embodied\ robot\ manipulation & 32 (34) & 6 (6) & 4 (10) & 270 (675) \\
medical\ clinical\ first\ aid & 49 (49) & 3 (3) & 6 (6) & 194 (757) \\
\midrule
All & 1,427 & 1,422 & 1,426 & 10,000 \\
\end{xltabular}
\endgroup

\endgroup
\begingroup

\begingroup
\setlength{\LTcapwidth}{\linewidth}
\setlength{\tabcolsep}{4pt}
\begin{xltabular}{\linewidth}{>{\raggedright\arraybackslash}p{99.52bp}>{\raggedright\arraybackslash}p{129.92bp}>{\raggedleft\arraybackslash}p{30.50bp}>{\raggedleft\arraybackslash}p{40.35bp}>{\raggedleft\arraybackslash}p{62.77bp}>{\raggedleft\arraybackslash}p{46.53bp}}
\caption{Benchmark contributions to the coverage pool and Video-Index, with attack results (1/2).}\label{tab:sourceslevels}\label{tab:videoindexsources}\\
\toprule
\rowcolor{SkillPaleBlue} Benchmark & Class & Cand. & Remain. & Coverage set & Share (\%) \\
\midrule
\endfirsthead
\multicolumn{6}{l}{\tablename~\thetable{} (continued)}\\
\toprule
\rowcolor{SkillPaleBlue} Benchmark & Class & Cand. & Remain. & Coverage set & Share (\%) \\
\midrule
\endhead
\bottomrule
\endfoot
\mbox{ALLVB} & Perception & 24,955 & 5,008 & 139 & 1.4 \\
\mbox{Perception Test} & Perception & 17,803 & 3,145 & 101 & 1.0 \\
\mbox{VITATECS} & Perception & 11,203 & 623 & 7 & 0.1 \\
\mbox{HLV-1K} & Perception & 10,472 & 1,494 & 421 & 4.2 \\
\mbox{NExT-QA} & Perception & 8,550 & 213 & 84 & 0.8 \\
\mbox{FAVOR-Bench} & Perception & 7,985 & 2,162 & 77 & 0.8 \\
\mbox{ExpVid} & Perception & 5,521 & 1,768 & 177 & 1.8 \\
\mbox{VF-Eval} & Perception & 4,943 & 824 & 139 & 1.4 \\
\mbox{K9-Bench} & Perception & 4,687 & 1,086 & 305 & 3.0 \\
\mbox{ViMUL-Bench} & Perception & 4,483 & 1,046 & 244 & 2.4 \\
\mbox{UCF101-AD} & Perception & 4,224 & 7 & 5 & 0.1 \\
\mbox{VELOCITI} & Perception & 4,131 & 176 & 16 & 0.2 \\
\mbox{TemporalBench} & Perception & 3,748 & 630 & 179 & 1.8 \\
\mbox{MVBench} & Perception & 3,569 & 563 & 68 & 0.7 \\
\mbox{SITE-Bench} & Perception & 3,255 & 629 & 184 & 1.8 \\
\mbox{TaskMeAnything-v1} & Perception & 3,194 & 1,180 & 51 & 0.5 \\
\mbox{CAViAR} & Perception & 2,996 & 140 & 10 & 0.1 \\
\mbox{ST-Evidence} & Perception & 2,694 & 327 & 62 & 0.6 \\
\mbox{StreamingBench} & Perception & 2,380 & 419 & 203 & 2.0 \\
\mbox{TOC-Bench} & Perception & 2,211 & 1,030 & 97 & 1.0 \\
\mbox{MLVU} & Perception & 2,112 & 459 & 111 & 1.1 \\
\mbox{ScaleLong} & Perception & 1,707 & 381 & 161 & 1.6 \\
\mbox{LVBench} & Perception & 1,527 & 584 & 70 & 0.7 \\
\mbox{Colon-Bench} & Perception & 1,485 & 460 & 45 & 0.5 \\
\mbox{MME-VideoOCR} & Perception & 1,416 & 356 & 184 & 1.8 \\
\mbox{LongVideoBench} & Perception & 1,326 & 331 & 221 & 2.2 \\
\mbox{VideoEval-Pro} & Perception & 1,268 & 94 & 51 & 0.5 \\
\mbox{iSafetyBench} & Perception & 1,204 & 431 & 13 & 0.1 \\
\mbox{V2P-Bench} & Perception & 1,164 & 295 & 161 & 1.6 \\
\mbox{Video-LevelGauge} & Perception & 1,162 & 117 & 74 & 0.7 \\
\mbox{Causal2Needles} & Perception & 747 & 175 & 26 & 0.3 \\
\mbox{HERBench} & Perception & 628 & 156 & 48 & 0.5 \\
\mbox{Q-Bench-Video} & Perception & 622 & 84 & 69 & 0.7 \\
\mbox{TimeBlind} & Perception & 583 & 122 & 33 & 0.3 \\
\mbox{KFS-Bench} & Perception & 574 & 17 & 13 & 0.1 \\
\mbox{Artifact-Bench} & Perception & 565 & 30 & 5 & 0.1 \\
\mbox{VANE-Bench} & Perception & 556 & 64 & 39 & 0.4 \\
\mbox{NaturalBench-Video} & Perception & 458 & 90 & 38 & 0.4 \\
\mbox{GasVideo-1000} & Perception & 376 & 46 & 32 & 0.3 \\
\mbox{VSTAT} & Perception & 343 & 127 & 32 & 0.3 \\
\mbox{VNBench} & Perception & 320 & 95 & 37 & 0.4 \\
\mbox{FineBench} & Perception & 164 & 15 & 4 & 0.0 \\
\mbox{CG-Bench} & Perception & -- & -- & -- & -- \\
\mbox{PLM-VideoBench} & Perception & -- & -- & -- & 0.0 \\
\mbox{PerceptionComp} & Perception & -- & -- & -- & 0.0 \\
\mbox{S-EMBER} & Temporal & 9,438 & 5,557 & 505 & 5.0 \\
\mbox{AoTBench} & Temporal & 5,294 & 830 & 262 & 2.6 \\
\mbox{MotionBench} & Temporal & 3,971 & 1,042 & 219 & 2.2 \\
\mbox{SLVMBench} & Temporal & 2,247 & 596 & 257 & 2.6 \\
\mbox{TVBench} & Temporal & 2,037 & 507 & 35 & 0.3 \\
\mbox{VidPair-Halluc} & Temporal & 1,907 & 388 & 37 & 0.4 \\
\mbox{TempCompass} & Temporal & 1,522 & 246 & 75 & 0.8 \\
\mbox{E.T. Bench} & Temporal & 1,500 & 503 & 26 & 0.3 \\
\mbox{TimeScope} & Temporal & 1,496 & 386 & 96 & 1.0 \\
\mbox{TOMATO} & Temporal & 1,421 & 741 & 142 & 1.4 \\
\mbox{VSI-Super-Wild} & Temporal & 1,193 & 600 & 116 & 1.2 \\
\mbox{Vinoground} & Temporal & 998 & 201 & 5 & 0.1 \\
\mbox{HourVideo} & Temporal & 993 & 589 & 53 & 0.5 \\
\mbox{ForeSeaQA} & Temporal & 532 & 64 & 56 & 0.6 \\
\mbox{VCR-Bench} & Temporal & 510 & 125 & 81 & 0.8 \\
\mbox{LongTimeScope} & Temporal & 450 & 182 & 28 & 0.3 \\
\mbox{PhysicalAI} & Temporal & 2 & -- & -- & 0.0 \\
\mbox{LEMONADE} & Temporal & -- & -- & -- & 0.0 \\
\mbox{VGenST-Bench} & Spatial / physical & 9,470 & 1,374 & 216 & 2.2 \\
\mbox{VRU-Accident} & Spatial / physical & 5,988 & 834 & 30 & 0.3 \\
\mbox{SIS-Bench} & Spatial / physical & 4,052 & 715 & 89 & 0.9 \\
\mbox{UrbanVideo-Bench} & Spatial / physical & 3,917 & 1,396 & 85 & 0.8 \\
\mbox{ReVSI} & Spatial / physical & 3,381 & 1,126 & 51 & 0.5 \\
\mbox{CMU} & Spatial / physical & 3,228 & 485 & 38 & 0.4 \\
\mbox{M3-Verse} & Spatial / physical & 2,716 & 974 & 24 & 0.2 \\
\mbox{OSI-Bench} & Spatial / physical & 2,516 & 1,125 & 22 & 0.2 \\
\mbox{STI-Bench} & Spatial / physical & 1,885 & 820 & 128 & 1.3 \\
\mbox{DSI-Bench} & Spatial / physical & 1,695 & 414 & 26 & 0.3 \\
\mbox{SAW-Bench} & Spatial / physical & 1,561 & 676 & 28 & 0.3 \\
\mbox{ContPhy} & Spatial / physical & 1,379 & 102 & 11 & 0.1 \\
\mbox{DSR-Bench} & Spatial / physical & 1,334 & 703 & 246 & 2.5 \\
\mbox{OVO-S-Bench} & Spatial / physical & 1,172 & 348 & 125 & 1.2 \\
\mbox{MMSI-Video-Bench} & Spatial / physical & 1,085 & 495 & 144 & 1.4 \\
\mbox{IntPhys 2} & Spatial / physical & 1,064 & 145 & 6 & 0.1 \\
\mbox{VSI-Bench} & Spatial / physical & 694 & 125 & 21 & 0.2 \\
\mbox{VURB} & Spatial / physical & 407 & 39 & 17 & 0.2 \\
\mbox{SpatialTree} & Spatial / physical & 351 & 50 & 9 & 0.1 \\
\mbox{CineTechBench} & Spatial / physical & 127 & 36 & 5 & 0.1 \\
\mbox{GUI-World} & Reasoning / knowledge & 3,635 & 326 & 79 & 0.8 \\
\mbox{ExAct} & Reasoning / knowledge & 3,521 & 1,131 & 12 & 0.1 \\
\mbox{Video-Holmes} & Reasoning / knowledge & 3,384 & 1,038 & 69 & 0.7 \\
\mbox{COVER} & Reasoning / knowledge & 3,323 & 284 & 108 & 1.1 \\
\mbox{Video-MME-v2} & Reasoning / knowledge & 2,910 & 1,662 & 501 & 5.0 \\
\mbox{Video-MME} & Reasoning / knowledge & 2,698 & 530 & 308 & 3.1 \\
\mbox{VideoOdyssey} & Reasoning / knowledge & 2,447 & 1,091 & 105 & 1.1 \\
\mbox{VKnowU} & Reasoning / knowledge & 1,583 & 243 & 124 & 1.2 \\
\mbox{SFI-Bench} & Reasoning / knowledge & 1,530 & 366 & 20 & 0.2 \\
\mbox{MMWorld} & Reasoning / knowledge & 1,398 & 202 & 90 & 0.9 \\
\mbox{VideoEspresso} & Reasoning / knowledge & 1,359 & 396 & 116 & 1.2 \\
\mbox{VideoKR} & Reasoning / knowledge & 1,293 & 118 & 59 & 0.6 \\
\mbox{MMR-V} & Reasoning / knowledge & 1,218 & 536 & 156 & 1.6 \\
\mbox{VideoAds} & Reasoning / knowledge & 1,099 & 323 & 101 & 1.0 \\
\mbox{ReactHuman} & Reasoning / knowledge & 1,054 & 408 & 17 & 0.2 \\
\mbox{SmartHome-Bench} & Reasoning / knowledge & 1,023 & 77 & 18 & 0.2 \\
\mbox{Video-TT} & Reasoning / knowledge & 1,000 & 363 & 172 & 1.7 \\
\mbox{v-HUB} & Reasoning / knowledge & 951 & 219 & 74 & 0.7 \\
\mbox{SciVideoBench} & Reasoning / knowledge & 943 & 507 & 33 & 0.3 \\
\mbox{ReXTime} & Reasoning / knowledge & 918 & 108 & 78 & 0.8 \\
\mbox{Video-MMMU} & Reasoning / knowledge & 873 & 267 & 95 & 0.9 \\
\mbox{LongVideo-Reason} & Reasoning / knowledge & 827 & 89 & 87 & 0.9 \\
\mbox{MMVU} & Reasoning / knowledge & 613 & 112 & 89 & 0.9 \\
\mbox{EgoSchema} & Reasoning / knowledge & 500 & 95 & 60 & 0.6 \\
\mbox{VideoMathQA} & Reasoning / knowledge & 419 & 110 & 76 & 0.8 \\
\mbox{VAU-Bench} & Reasoning / knowledge & 251 & 2 & 2 & 0.0 \\
\mbox{PAI-Bench-U} & Reasoning / knowledge & 2 & 1 & 1 & 0.0 \\
\mbox{EgoPlan-Bench} & Reasoning / knowledge & -- & -- & -- & 0.0 \\
\mbox{MORSE-500} & Reasoning / knowledge & -- & -- & -- & 0.0 \\
\mbox{Neptune} & Reasoning / knowledge & -- & -- & -- & 0.0 \\
\mbox{SoMi-ToM} & Reasoning / knowledge & -- & -- & -- & 0.0 \\
\mbox{VideoSIAH-Eval} & Reasoning / knowledge & -- & -- & -- & 0.0 \\
\end{xltabular}
\endgroup

\begingroup
\setlength{\LTcapwidth}{\linewidth}
\setlength{\tabcolsep}{4pt}
\begin{xltabular}{\linewidth}{>{\raggedright\arraybackslash}p{109.88bp}>{\raggedleft\arraybackslash}p{70.34bp}>{\raggedleft\arraybackslash}p{27.96bp}>{\raggedleft\arraybackslash}p{29.35bp}>{\raggedleft\arraybackslash}p{32.04bp}>{\raggedleft\arraybackslash}p{36.87bp}>{\raggedleft\arraybackslash}p{35.42bp}>{\raggedright\arraybackslash}p{51.73bp}}
\caption{Benchmark contributions to the coverage pool and Video-Index, with attack results (2/2).}\label{tab:sourceslevels_cols2}\\
\toprule
\rowcolor{SkillPaleBlue} Benchmark & Video-Index & $\varepsilon_{\mathrm{opt}}$ & $\varepsilon_{\mathrm{text}}$ & $\varepsilon_{\mathrm{pool}}$ & $\varepsilon_{\mathrm{frame}}$ & $\varepsilon_{\mathrm{order}}$ & Break \\
\midrule
\endfirsthead
\multicolumn{8}{l}{\tablename~\thetable{} (continued)}\\
\toprule
\rowcolor{SkillPaleBlue} Benchmark & Video-Index & $\varepsilon_{\mathrm{opt}}$ & $\varepsilon_{\mathrm{text}}$ & $\varepsilon_{\mathrm{pool}}$ & $\varepsilon_{\mathrm{frame}}$ & $\varepsilon_{\mathrm{order}}$ & Break \\
\midrule
\endhead
\bottomrule
\endfoot
\mbox{ALLVB} & 45 & 40 & 53 & 53 & 53 & 53 & text \\
\mbox{Perception Test} & 19 & 13 & 14 & 14 & 32 & 35 & survives \\
\mbox{VITATECS} & 24 & 26 & 29 & 29 & 39 & 41 & frame \\
\mbox{HLV-1K} & 7 & 33 & 42 & 42 & 45 & 45 & text \\
\mbox{NExT-QA} & 2 & 9 & 39 & 39 & 58 & 64 & frame \\
\mbox{FAVOR-Bench} & 19 & 26 & 34 & 34 & 34 & 34 & text \\
\mbox{ExpVid} & 1 & 25 & 28 & 28 & 40 & 40 & frame \\
\mbox{VF-Eval} & 18 & 52 & 58 & 58 & 58 & 58 & text \\
\mbox{K9-Bench} & 21 & 41 & 46 & 46 & 47 & 47 & survives \\
\mbox{ViMUL-Bench} & 7 & 16 & 20 & 20 & 39 & 44 & survives \\
\mbox{UCF101-AD} & -- & 4 & 4 & 75 & 75 & 75 & pool \\
\mbox{VELOCITI} & 11 & 14 & 15 & 15 & 38 & 38 & survives \\
\mbox{TemporalBench} & 12 & 26 & 26 & 26 & 28 & 28 & option \\
\mbox{MVBench} & 1 & 8 & 18 & 18 & 34 & 37 & order \\
\mbox{SITE-Bench} & 2 & 17 & 24 & 24 & 34 & 39 & order \\
\mbox{TaskMeAnything-v1} & 1 & 4 & 7 & 18 & 19 & 40 & order \\
\mbox{CAViAR} & -- & 23 & 23 & 29 & 41 & 41 & frame \\
\mbox{ST-Evidence} & 3 & 13 & 19 & 19 & 29 & 29 & survives \\
\mbox{StreamingBench} & 9 & 25 & 31 & 31 & 35 & 37 & frame \\
\mbox{TOC-Bench} & 38 & 4 & 4 & 8 & 15 & 15 & frame \\
\mbox{MLVU} & 6 & 22 & 34 & 34 & 41 & 41 & frame \\
\mbox{ScaleLong} & 4 & 21 & 24 & 24 & 33 & 33 & survives \\
\mbox{LVBench} & 8 & 8 & 12 & 12 & 22 & 22 & frame \\
\mbox{Colon-Bench} & 1 & 30 & 34 & 34 & 44 & 44 & survives \\
\mbox{MME-VideoOCR} & 5 & 14 & 18 & 18 & 33 & 42 & survives \\
\mbox{LongVideoBench} & 2 & 21 & 31 & 31 & 43 & 43 & frame \\
\mbox{VideoEval-Pro} & 3 & 7 & 10 & 10 & 23 & 23 & frame \\
\mbox{iSafetyBench} & -- & 9 & 9 & 9 & 30 & 34 & order \\
\mbox{V2P-Bench} & 2 & 16 & 16 & 16 & 30 & 30 & frame \\
\mbox{Video-LevelGauge} & 2 & 14 & 28 & 28 & 43 & 64 & order \\
\mbox{Causal2Needles} & -- & 21 & 32 & 32 & 35 & 35 & frame \\
\mbox{HERBench} & -- & 16 & 28 & 28 & 34 & 34 & frame \\
\mbox{Q-Bench-Video} & -- & 15 & 21 & 37 & 37 & 37 & survives \\
\mbox{TimeBlind} & 9 & 6 & 6 & 6 & 13 & 13 & survives \\
\mbox{KFS-Bench} & -- & 19 & 30 & 30 & 39 & 39 & survives \\
\mbox{Artifact-Bench} & -- & -12 & -9 & 30 & 30 & 30 & option \\
\mbox{VANE-Bench} & 1 & 51 & 52 & 53 & 62 & 62 & frame \\
\mbox{NaturalBench-Video} & 7 & 3 & 3 & 3 & 10 & 10 & survives \\
\mbox{GasVideo-1000} & -- & 10 & 17 & 17 & 32 & 57 & order \\
\mbox{VSTAT} & -- & 10 & 15 & 15 & 18 & 18 & frame \\
\mbox{VNBench} & -- & 2 & 2 & 3 & 27 & 27 & survives \\
\mbox{FineBench} & -- & 36 & 36 & 60 & 60 & 60 & option \\
\mbox{CG-Bench} & -- & 30 & 33 & 33 & 38 & -- & frame \\
\mbox{PLM-VideoBench} & -- & 2 & 22 & 22 & 32 & -- & frame \\
\mbox{PerceptionComp} & -- & -- & -13 & -13 & -11 & -5 & text \\
\mbox{S-EMBER} & 109 & 11 & 11 & 11 & 11 & 11 & survives \\
\mbox{AoTBench} & 92 & 2 & 2 & 3 & 23 & 23 & survives \\
\mbox{MotionBench} & 11 & 21 & 25 & 25 & 31 & 31 & survives \\
\mbox{SLVMBench} & 21 & 30 & 43 & 43 & 53 & 53 & survives \\
\mbox{TVBench} & 3 & 2 & 7 & 10 & 16 & 16 & survives \\
\mbox{VidPair-Halluc} & 2 & 3 & 3 & 3 & 8 & 9 & frame \\
\mbox{TempCompass} & 13 & 15 & 15 & 18 & 18 & 18 & option \\
\mbox{E.T. Bench} & 2 & 3 & 5 & 6 & 25 & 25 & survives \\
\mbox{TimeScope} & -- & 9 & 24 & 72 & 72 & 72 & pool \\
\mbox{TOMATO} & 4 & 3 & 3 & 12 & 12 & 12 & survives \\
\mbox{VSI-Super-Wild} & -- & 4 & 5 & 5 & 5 & 5 & option \\
\mbox{Vinoground} & 18 & 1 & 1 & 1 & 27 & 27 & survives \\
\mbox{HourVideo} & 2 & 3 & 3 & 3 & 3 & 3 & option \\
\mbox{ForeSeaQA} & 1 & 44 & 46 & 46 & 46 & 46 & option \\
\mbox{VCR-Bench} & 1 & 18 & 21 & 21 & 27 & 27 & survives \\
\mbox{LongTimeScope} & -- & 8 & 17 & 76 & 76 & 76 & option \\
\mbox{PhysicalAI} & -- & 7 & 9 & 13 & 23 & -- & survives \\
\mbox{LEMONADE} & -- & 15 & 17 & 17 & 17 & 17 & option \\
\mbox{VGenST-Bench} & 6 & 12 & 19 & 28 & 44 & 44 & survives \\
\mbox{VRU-Accident} & 11 & 30 & 30 & 50 & 50 & 50 & pool \\
\mbox{SIS-Bench} & 5 & 16 & 25 & 25 & 37 & 37 & survives \\
\mbox{UrbanVideo-Bench} & 34 & 21 & 21 & 21 & 28 & 28 & survives \\
\mbox{ReVSI} & 2 & 3 & 13 & 13 & 13 & 23 & order \\
\mbox{CMU} & 33 & 17 & 22 & 22 & 34 & 34 & survives \\
\mbox{M3-Verse} & 1 & 2 & 3 & 10 & 10 & 15 & order \\
\mbox{OSI-Bench} & 1 & 5 & 14 & 14 & 14 & 17 & survives \\
\mbox{STI-Bench} & -- & 22 & 27 & 27 & 29 & 29 & text \\
\mbox{DSI-Bench} & 6 & 27 & 27 & 29 & 30 & 42 & option \\
\mbox{SAW-Bench} & 3 & 4 & 4 & 36 & 36 & 36 & pool \\
\mbox{ContPhy} & -- & 3 & 3 & 14 & 16 & 16 & survives \\
\mbox{DSR-Bench} & 1 & 4 & 4 & 8 & 8 & 19 & order \\
\mbox{OVO-S-Bench} & 2 & 13 & 13 & 16 & 16 & 16 & option \\
\mbox{MMSI-Video-Bench} & 1 & 3 & 6 & 6 & 6 & 7 & option \\
\mbox{IntPhys 2} & -- & 2 & 2 & 2 & 2 & 8 & frame \\
\mbox{VSI-Bench} & -- & 7 & 9 & 9 & 23 & 24 & survives \\
\mbox{VURB} & -- & 22 & 22 & 29 & 29 & 38 & order \\
\mbox{SpatialTree} & -- & 54 & 54 & 54 & 54 & 54 & option \\
\mbox{CineTechBench} & -- & 6 & 15 & 15 & 32 & 32 & frame \\
\mbox{GUI-World} & 6 & 36 & 56 & 56 & 61 & 61 & survives \\
\mbox{ExAct} & 1 & 60 & 62 & 62 & 62 & 62 & option \\
\mbox{Video-Holmes} & 21 & 26 & 38 & 38 & 41 & 41 & survives \\
\mbox{COVER} & -- & 16 & 32 & 32 & 35 & 37 & text \\
\mbox{Video-MME-v2} & 28 & 14 & 18 & 18 & 24 & 36 & frame \\
\mbox{Video-MME} & 9 & 21 & 31 & 31 & 40 & 47 & order \\
\mbox{VideoOdyssey} & 26 & 16 & 20 & 20 & 24 & 24 & option \\
\mbox{VKnowU} & 6 & 23 & 26 & 26 & 36 & 36 & frame \\
\mbox{SFI-Bench} & 3 & 23 & 37 & 37 & 37 & 37 & survives \\
\mbox{MMWorld} & 6 & 36 & 52 & 52 & 55 & 57 & text \\
\mbox{VideoEspresso} & 7 & 49 & 51 & 51 & 54 & 54 & text \\
\mbox{VideoKR} & -- & 20 & 23 & 23 & 35 & 50 & order \\
\mbox{MMR-V} & 1 & 39 & 47 & 47 & 48 & 48 & text \\
\mbox{VideoAds} & 31 & 15 & 26 & 26 & 51 & 51 & frame \\
\mbox{ReactHuman} & -- & 4 & 4 & 4 & 12 & 19 & frame \\
\mbox{SmartHome-Bench} & 1 & 13 & 13 & 13 & 21 & 21 & survives \\
\mbox{Video-TT} & 9 & 5 & 12 & 32 & 32 & 33 & pool \\
\mbox{v-HUB} & -- & 5 & 5 & 5 & 48 & 50 & frame \\
\mbox{SciVideoBench} & -- & 40 & 44 & 44 & 48 & 48 & survives \\
\mbox{ReXTime} & 4 & 59 & 60 & 60 & 66 & 66 & frame \\
\mbox{Video-MMMU} & 1 & 30 & 45 & 45 & 45 & 59 & order \\
\mbox{LongVideo-Reason} & 1 & 35 & 49 & 49 & 50 & 50 & text \\
\mbox{MMVU} & 1 & 29 & 38 & 38 & 58 & 58 & frame \\
\mbox{EgoSchema} & 1 & 37 & 38 & 38 & 51 & 57 & order \\
\mbox{VideoMathQA} & 2 & 4 & 12 & 25 & 25 & 25 & option \\
\mbox{VAU-Bench} & -- & 63 & 71 & 71 & 71 & 71 & text \\
\mbox{PAI-Bench-U} & -- & 6 & 9 & 10 & 25 & 25 & survives \\
\mbox{EgoPlan-Bench} & -- & 3 & 8 & 8 & 8 & -- & option \\
\mbox{MORSE-500} & -- & 7 & 7 & 15 & 15 & 15 & survives \\
\mbox{Neptune} & -- & 3 & 52 & 52 & 52 & 52 & survives \\
\mbox{SoMi-ToM} & -- & 7 & 17 & 20 & 37 & 61 & order \\
\mbox{VideoSIAH-Eval} & -- & -- & -18 & -18 & -13 & -13 & text \\
\end{xltabular}
\endgroup

\endgroup

We assess five properties of a composition. Validity combines residual attack scores, blind accuracy, and video gain. Reliability uses frame-offset and option-permutation noise, 1,000 item-bootstrap resamples, and the ratio of score dispersion to the noise floor. Ordering checks pairs with an expected direction, including model sizes, frame budgets, and video versus blind inputs. We count correctly ordered pairs with non-overlapping 95\% intervals and report separation per hundred items. Agreement uses Kendall $\tau$ against mean chance-corrected accuracy over 3 to 8 stratified items from each of 106 census benchmarks. Each source is excluded from the census when its own agreement is evaluated. We repeat the ranking comparison using video gain to distinguish dependence on video from shortcut-driven accuracy. Hardness combines reference accuracy with the share of failures attributed to labels. Models used in screening or gating are excluded from ordering and agreement statistics.
\begingroup
\begingroup
\setlength{\LTcapwidth}{\linewidth}
\setlength{\tabcolsep}{4pt}
\begin{xltabular}{\linewidth}{>{\raggedright\arraybackslash}p{136.16bp}>{\raggedleft\arraybackslash}p{16.86bp}>{\raggedleft\arraybackslash}p{24.43bp}>{\raggedleft\arraybackslash}p{57.75bp}>{\raggedleft\arraybackslash}p{42.96bp}>{\raggedleft\arraybackslash}p{22.58bp}>{\raggedleft\arraybackslash}p{43.81bp}>{\raggedleft\arraybackslash}p{19.59bp}>{\raggedleft\arraybackslash}p{21.43bp}}
\caption{Selection at 840 items: coverage, shortcut exposure, video gain, and rank agreement.}\label{tab:selections}\\
\toprule
\rowcolor{SkillPaleBlue} Selection & $n$ & Cells & Exploit. (\%) & Gain (\%) & SNR & Pairs (\%) & $\tau_{\mathrm{acc}}$ & $\tau_{\mathrm{gain}}$ \\
\midrule
\endfirsthead
\multicolumn{9}{l}{\tablename~\thetable{} (continued)}\\
\toprule
\rowcolor{SkillPaleBlue} Selection & $n$ & Cells & Exploit. (\%) & Gain (\%) & SNR & Pairs (\%) & $\tau_{\mathrm{acc}}$ & $\tau_{\mathrm{gain}}$ \\
\midrule
\endhead
\bottomrule
\endfoot
Video-Index (meta-benchmark) & 840 & 354 & 13.1 & 10.8 & 10.8 & 77 & 0.33 & 0.56 \\
Random draw from the pool (mean of 5) & 840 & 374 & 19.3 & 11.7 & 9.9 & 86 & 0.36 & 0.43 \\
Top benchmarks by breaking level & 840 & 234 & 19.1 & 12.1 & 9.1 & 77 & 0.22 & 0.50 \\
Diversity-aware draw, no gate & 840 & 578 & 19.4 & 11.3 & 11.2 & 77 & 0.28 & 0.59 \\
Coverage census ($\leq 8$ per benchmark) & 834 & -- & -- & 15.5 & 2.6 & 90 & 0.94 & 1.00 \\
\midrule
Median source benchmark & 54 & -- & 19.0 & 9.4 & 3.7 & 38 & 0.30 & 0.22 \\
\end{xltabular}
\endgroup
\endgroup
\begin{figure}[!htbp]
\centering
\includegraphics[width=\linewidth]{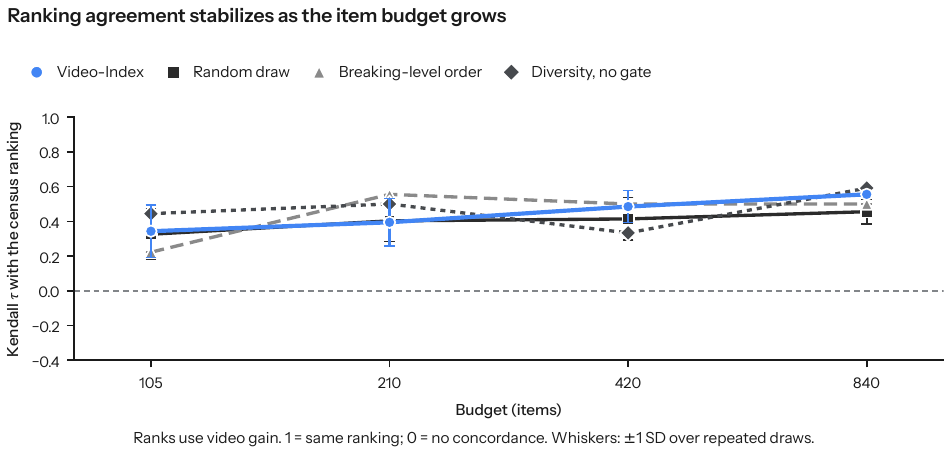} %
\caption{Kendall $\tau$ agreement of model ranks with the census across evaluation budgets.}
\label{fig:budget}
\end{figure}
Video-Index separates all ten video-versus-blind pairs and 77\% of the known-order pairs, the same share as selection by breaking level, against 86\% for the random draws. Table~\ref{tab:selections} reports Kendall $\tau=0.56$ with the census ranking by video gain, comparable to the alternatives. Agreement by accuracy is 0.33. The gate provides the clearest distinction. Strongest-attacker accuracy is 13\%, versus 19\% without gating. As shown in Figure~\ref{fig:budget}, gain-based agreement rises with the budget, from $\tau=0.34$ at 105 items to 0.49 at 420 and 0.56 at 840.

\section{Agent behavior on Video-Index}
\label{sec:agentbehaviour}
Claude Opus 5 \citep{anthropic2026opus5} is evaluated through the same type of tool-enabled CLI protocol, with maximum effort, model-directed sampling, and failures counted as incorrect. It scores 70.6\% and 24.9\% blind over four option permutations. The fixed 1 fps / 512-frame protocol gives 56.8\% with video and 21.3\% blind. 

Figure~\ref{fig:agentcomparison} takes accuracy from the leaderboard and effort from the session logs. Shell calls, elapsed time, and image counts are medians per session. The horizontal runtime intervals in Figure~\ref{fig:agentcomparison} span the first and third quartiles of recorded session times. Runtime quartiles use the sessions with a recorded time. Each operation bar measures the percentage of that agent's logged answers using the operation. The full background track represents 100\%. The four percentages are independent because operations can co-occur. Visual operations are counted once per session. Regular sampling includes fixed-rate, fixed-count, and every-$n$th-frame extraction. Seeking selects timestamps. Tiling assembles frame grids. Cropping or zooming isolates details. Mixed strategies can contribute to multiple operations. Images produced include decoded frames, crops, and contact sheets, whether or not the model viewed them. Image views are counted where the logs record them. The duration panel uses the exported per-agent medians. Tables~\ref{tab:agentbehaviour} and~\ref{tab:agentstrategy} report the remaining tool and sampling details.

Sampling choices differ substantially. Fable decodes every stored frame in 72.2\% of sessions. Gemini Pro uses an explicit frame-rate filter in 47.6\%. Astra and Opus distribute their sessions across full decoding, periodic sampling, and mixed strategies. Contact sheets appear in 99--100\% of sessions for Astra, Fable, and Opus, compared with 69\% for Gemini Pro and 35\% for Flash-Lite. Flash-Lite retains native extraction resolution in every logged session, while Astra downscales in 60.8\%. These choices describe different inspection policies.

Among Astra, Fable, and Opus, pairwise answer agreement is 76.8--80.6\%, while strategy agreement is 33.0--43.8\%. Similar answers can therefore follow different inspection paths. Long videos remain difficult. Beyond ten minutes, accuracy is 67.1\% for Astra, 50.9\% for Fable, and 46.6\% for Opus, down from 88.7\%, 84.4\%, and 86.2\% below 30 seconds in the answer export. Table~\ref{tab:agentduration} gives the tool-log summaries by duration, including both Gemini agents.
Inspection volume also varies with item difficulty. From the lowest to highest image-production quartile, accuracy changes from 85.4\% to 74.1\% for Astra, from 85.4\% to 51.6\% for Fable, and from 73.0\% to 62.2\% for Opus. These associations do not show that producing more images causes worse answers. Separate vendor usage records report median input/output tokens of 31k/0.9k for Astra and 128k/10k for Fable. Input counts include cache reads.

\begingroup
\setlength{\LTcapwidth}{\linewidth}
\setlength{\tabcolsep}{4pt}
\begin{xltabular}{\linewidth}{>{\raggedright\arraybackslash}p{184.26bp}>{\raggedleft\arraybackslash}p{26.93bp}>{\raggedleft\arraybackslash}p{42.82bp}>{\raggedleft\arraybackslash}p{72.13bp}>{\raggedleft\arraybackslash}p{48.42bp}>{\raggedleft\arraybackslash}p{35.02bp}}
\caption{Agent tool use on Video-Index, as medians per item unless indicated otherwise.}\label{tab:agentbehaviour}\\
\toprule
\rowcolor{SkillPaleBlue} Measure & Astra & Fable 5.1 & Gemini 3.1 Pro & Flash-Lite & Opus 5 \\
\midrule
\endfirsthead
\multicolumn{6}{l}{\tablename~\thetable{} (continued)}\\
\toprule
\rowcolor{SkillPaleBlue} Measure & Astra & Fable 5.1 & Gemini 3.1 Pro & Flash-Lite & Opus 5 \\
\midrule
\endhead
\bottomrule
\endfoot
Accuracy (\%) & 79.3 & 71.6 & 55.3 & 21.7 & 70.6 \\
Shell commands per item & 3 & 5 & 4 & 6 & 12 \\
Sessions using ffmpeg (\%) & 50 & 100 & 100 & 100 & 100 \\
Sessions using OpenCV (\%) & 54 & 16 & 26 & 96 & 48 \\
Sessions using Pillow (\%) & 81 & 81 & 51 & 61 & 90 \\
Sessions with a contact sheet (\%) & 99 & 100 & 69 & 35 & 99 \\
Sessions with crops or zooms (\%) & 11 & 74 & 24 & 9 & 84 \\
Images produced per item & 8 & 79 & 39 & 62 & 25 \\
Images viewed per item & -- & 12 & 6 & 6 & 13 \\
Images produced per video second & 0.10 & 2.19 & 1.63 & 2.00 & 0.73 \\
Seconds per item & 57 & 200 & 140 & 95 & 205 \\
Output tokens per item & 824 & 10239 & -- & -- & 13293 \\
\end{xltabular}
\endgroup

\begingroup
\setlength{\LTcapwidth}{\linewidth}
\setlength{\tabcolsep}{4pt}
\begin{xltabular}{\linewidth}{>{\raggedright\arraybackslash}p{236.38bp}>{\raggedleft\arraybackslash}p{26.93bp}>{\raggedleft\arraybackslash}p{26.45bp}>{\raggedleft\arraybackslash}p{36.38bp}>{\raggedleft\arraybackslash}p{48.42bp}>{\raggedleft\arraybackslash}p{35.02bp}}
\caption{Agent sampling strategies (\% of sessions); videos store at most 2 fps and 1,024 frames.}\label{tab:agentstrategy}\\
\toprule
\rowcolor{SkillPaleBlue} Strategy & \shortstack{Astra} & \shortstack{Fable\\5.1} & \shortstack{Gemini\\3.1 Pro} & \shortstack{Flash-Lite} & \shortstack{Opus 5} \\
\midrule
\endfirsthead
\multicolumn{6}{l}{\tablename~\thetable{} (continued)}\\
\toprule
\rowcolor{SkillPaleBlue} Strategy & \shortstack{Astra} & \shortstack{Fable\\5.1} & \shortstack{Gemini\\3.1 Pro} & \shortstack{Flash-Lite} & \shortstack{Opus 5} \\
\midrule
\endhead
\bottomrule
\endfoot
Decode every stored frame (2 fps) & 35.0 & 72.2 & 19.4 & 39.1 & 26.3 \\
Every $n$-th frame & 13.4 & 22.5 & 14.6 & 8.7 & 33.5 \\
Explicit fps filter & 6.5 & 0.7 & 47.6 & 8.7 & 6.2 \\
Fixed number of frames & 1.5 & 1.4 & -- & -- & 0.8 \\
Seeks to chosen timestamps & 7.6 & 0.5 & -- & -- & 1.6 \\
Two or more of the above & 36.0 & 2.7 & 18.4 & 43.5 & 31.4 \\
No decoding command & -- & -- & -- & -- & 0.1 \\
\end{xltabular}
\endgroup
\begingroup
\setlength{\LTcapwidth}{\linewidth}
\setlength{\tabcolsep}{4pt}
\begin{xltabular}{\linewidth}{>{\raggedright\arraybackslash}p{82.46bp}>{\raggedright\arraybackslash}p{85.15bp}>{\raggedright\arraybackslash}p{69.85bp}>{\raggedleft\arraybackslash}p{56.76bp}>{\raggedleft\arraybackslash}p{64.57bp}>{\raggedleft\arraybackslash}p{50.79bp}}
\caption{Agent sampling strategies, image counts, and accuracy by video duration.}\label{tab:agentduration}\\
\toprule
\rowcolor{SkillPaleBlue} \shortstack{Video\\duration} & Agent & Strategy & \shortstack{Share\\(\%)} & Images & \shortstack{Acc.\\(\%)} \\
\midrule
\endfirsthead
\multicolumn{6}{l}{\tablename~\thetable{} (continued)}\\
\toprule
\rowcolor{SkillPaleBlue} \shortstack{Video\\duration} & Agent & Strategy & \shortstack{Share\\(\%)} & Images & \shortstack{Acc.\\(\%)} \\
\midrule
\endhead
\bottomrule
\endfoot
\shortstack[l]{$<$30 s\\(335 items)} & Astra & all & 70 & 3 & 89 \\
 & Fable 5.1 & all & 79 & 34 & 85 \\
 & Gemini Pro & fps & 52 & 20 & 68 \\
 & Flash-Lite & all & 82 & 16 & 27 \\
 & Opus 5 & nth & 44 & 30 & 86 \\
\shortstack[l]{30-120 s\\(114 items)} & Astra & mix & 48 & 3 & 76 \\
 & Fable 5.1 & all & 77 & 114 & 67 \\
 & Gemini Pro & fps & 54 & 63 & 64 \\
 & Flash-Lite & nth & 100 & 62 & 100 \\
 & Opus 5 & mix & 44 & 24 & 66 \\
\shortstack[l]{120-600 s\\(141 items)} & Astra & mix & 88 & 7 & 74 \\
 & Fable 5.1 & all & 51 & 636 & 69 \\
 & Gemini Pro & fps & 56 & 184 & 56 \\
 & Flash-Lite & mix & 83 & 614 & 0 \\
 & Opus 5 & mix & 76 & 18 & 65 \\
\shortstack[l]{$>$600 s\\(164 items)} & Astra & mix & 52 & 53 & 67 \\
 & Fable 5.1 & all & 73 & 1060 & 51 \\
 & Gemini Pro & nth & 40 & 156 & 15 \\
 & Flash-Lite & mix & 100 & 900 & 20 \\
 & Opus 5 & mix & 38 & 24 & 47 \\
\end{xltabular}
\endgroup

\section{Taxonomy and full results}
\label{app:fulltables}
Table~\ref{tab:categories} summarizes capability groups and their audit statistics. Tables~\ref{tab:master}--\ref{tab:master_cols2} collect benchmark-level results, ranked by the accuracy gap between 32-frame and blind inputs. In Table~\ref{tab:master}, the caption column uses Claude Opus 5 \citep{anthropic2026opus5} for every row. Lift is blind accuracy minus chance, where chance is the mean item chance over the multiple-choice items taken from the option list, the options embedded in the question, the declared option count, or an inferred count, in that order. The reasoning and protocol columns give the share of attributed errors in reasoning and knowledge and in data protocol. The two attribution protocols use different judge pairs, as Tables~\ref{tab:attrshares} and~\ref{tab:attrshares_n1} list. Diversity is the Vendi score per item, and the membership column is the min-20\% score under Qwen2.5-7B \citep{qwen2024qwen25}, where a lower value marks more corpus exposure. Diversity, near-duplicate, and membership cells are blank where not computed.
\begingroup
\begingroup
\setlength{\LTcapwidth}{\linewidth}
\setlength{\tabcolsep}{4pt}
\begin{xltabular}{\linewidth}{>{\raggedright\arraybackslash}p{193.96bp}>{\raggedleft\arraybackslash}p{11.41bp}>{\raggedleft\arraybackslash}p{49.18bp}>{\raggedleft\arraybackslash}p{69.28bp}>{\raggedright\arraybackslash}p{93.75bp}}
\caption{Capability groups: counts, video gain, protocol issues, and largest-gain sources.}\label{tab:categories}\\
\toprule
\rowcolor{SkillPaleBlue} Category & $n$ & Median $m$ & Median proto. & Highest-$m$ member \\
\midrule
\endfirsthead
\multicolumn{5}{l}{\tablename~\thetable{} (continued)}\\
\toprule
\rowcolor{SkillPaleBlue} Category & $n$ & Median $m$ & Median proto. & Highest-$m$ member \\
\midrule
\endhead
\bottomrule
\endfoot
Perception & 45 & 16.0 & 6 & \mbox{VNBench} \\
Temporal & 18 & 16.7 & 9 & \mbox{Vinoground} \\
Spatial / physical & 20 & 14.0 & 2 & \mbox{VGenST-Bench} \\
Reasoning / knowledge & 32 & 15.0 & 22 & \mbox{v-HUB} \\
\end{xltabular}
\endgroup
\endgroup

\begingroup

\begingroup
\setlength{\LTcapwidth}{\linewidth}
\setlength{\tabcolsep}{4pt}
\begin{xltabular}{\linewidth}{>{\raggedleft\arraybackslash}p{29.22bp}>{\raggedright\arraybackslash}p{111.88bp}>{\raggedleft\arraybackslash}p{34.68bp}>{\raggedleft\arraybackslash}p{48.82bp}>{\raggedleft\arraybackslash}p{38.89bp}>{\raggedleft\arraybackslash}p{23.77bp}>{\raggedleft\arraybackslash}p{27.40bp}>{\raggedleft\arraybackslash}p{35.33bp}>{\raggedleft\arraybackslash}p{35.58bp}}
\caption{Audit results ranked by 32-frame accuracy minus blind accuracy (1/2).}\label{tab:master}\\
\toprule
\rowcolor{SkillPaleBlue} \# & Benchmark & Yr & Chance & Blind & 1f & 32f & Cap. & $m$ \\
\midrule
\endfirsthead
\multicolumn{9}{l}{\tablename~\thetable{} (continued)}\\
\toprule
\rowcolor{SkillPaleBlue} \# & Benchmark & Yr & Chance & Blind & 1f & 32f & Cap. & $m$ \\
\midrule
\endhead
\bottomrule
\endfoot
2 & \mbox{VNBench} & 2025 & 25 & 26 & 25 & 74 & 60 & 47.9 \\
3 & \mbox{v-HUB} & 2025 & 20 & 17 & 52 & 64 & 71 & 47.4 \\
4 & \mbox{SoMi-ToM} & 2025 & 33 & 51 & 47 & 93 & 85 & 42.7 \\
5 & \mbox{VGenST-Bench} & 2026 & 27 & 45 & 61 & 88 & 81 & 42.3 \\
6 & \mbox{MME-VideoOCR} & 2025 & 25 & 43 & 58 & 84 & 68 & 41.3 \\
7 & \mbox{UCF101-AD} & 2026 & 9 & 7 & 36 & 47 & 49 & 39.3 \\
8 & \mbox{EgoSchema} & 2023 & 20 & 39 & 50 & 77 & 80 & 37.6 \\
9 & \mbox{Video-LevelGauge} & 2026 & 25 & 53 & 65 & 89 & 74 & 36.3 \\
10 & \mbox{Vinoground} & 2024 & 50 & 46 & 56 & 82 & 79 & 36.2 \\
11 & \mbox{ViMUL-Bench} & 2025 & 26 & 44 & 62 & 80 & 75 & 36.2 \\
12 & \mbox{Video-TT} & 2025 & 25 & 28 & 27 & 61 & 45 & 32.4 \\
13 & \mbox{CAViAR} & 2026 & 32 & 45 & 73 & 78 & 75 & 32.3 \\
14 & \mbox{VideoKR} & 2026 & 14 & 23 & 30 & 55 & 57 & 32.1 \\
15 & \mbox{MORSE-500} & 2025 & 20 & 10 & 1 & 41 & 24 & 30.9 \\
16 & \mbox{VideoAds} & 2025 & 25 & 51 & 57 & 81 & 75 & 30.3 \\
17 & \mbox{NaturalBench-Video} & 2025 & 50 & 51 & 55 & 81 & 67 & 30.0 \\
18 & \mbox{MVBench} & 2023 & 29 & 39 & 42 & 69 & 66 & 29.9 \\
19 & \mbox{TaskMeAnything-v1} & 2024 & 25 & 32 & 41 & 62 & 50 & 29.7 \\
20 & \mbox{GasVideo-1000} & 2026 & 30 & 46 & 48 & 75 & 73 & 29.2 \\
21 & \mbox{AoTBench} & 2025 & 47 & 47 & 46 & 77 & 72 & 29.1 \\
22 & \mbox{VELOCITI} & 2025 & 50 & 65 & 75 & 94 & 93 & 28.3 \\
23 & \mbox{Q-Bench-Video} & 2024 & 27 & 44 & 59 & 73 & 46 & 28.3 \\
24 & \mbox{VRU-Accident} & 2025 & 25 & 48 & 67 & 76 & 75 & 27.7 \\
25 & \mbox{Perception Test} & 2023 & 33 & 47 & 54 & 75 & 71 & 27.7 \\
26 & \mbox{iSafetyBench} & 2025 & 6 & 15 & 36 & 42 & 43 & 27.4 \\
27 & \mbox{MotionBench} & 2025 & 25 & 38 & 48 & 65 & 64 & 26.5 \\
28 & \mbox{PAI-Bench-U} & -- & 25 & 34 & 50 & 61 & 56 & 26.5 \\
29 & \mbox{E.T. Bench} & 2024 & 23 & 29 & 33 & 55 & 52 & 26.0 \\
30 & \mbox{TVBench} & 2024 & 33 & 37 & 30 & 63 & 56 & 25.5 \\
31 & \mbox{VKnowU} & 2025 & 36 & 46 & 42 & 71 & 77 & 25.5 \\
32 & \mbox{VSI-Bench} & 2024 & 29 & 36 & 17 & 61 & 45 & 24.5 \\
33 & \mbox{VITATECS} & 2023 & 50 & 68 & 82 & 92 & 89 & 23.6 \\
34 & \mbox{ContPhy} & 2024 & 31 & 33 & 47 & 56 & 34 & 22.7 \\
35 & \mbox{VURB} & 2026 & 27 & 40 & 37 & 62 & 58 & 22.6 \\
36 & \mbox{SITE-Bench} & 2025 & 26 & 43 & 38 & 65 & 65 & 22.2 \\
37 & \mbox{Video-MMMU} & 2025 & 12 & 48 & 44 & 70 & 70 & 22.0 \\
38 & \mbox{SIS-Bench} & -- & 25 & 50 & 58 & 72 & 69 & 22.0 \\
39 & \mbox{TimeBlind} & 2026 & 50 & 49 & 25 & 71 & 66 & 21.7 \\
40 & \mbox{TOMATO} & 2024 & 20 & 22 & 13 & 43 & 30 & 21.4 \\
41 & \mbox{VCR-Bench} & 2025 & 24 & 45 & 47 & 66 & 58 & 21.2 \\
42 & \mbox{ReVSI} & 2026 & 27 & 31 & 9 & 52 & 40 & 20.7 \\
43 & \mbox{ST-Evidence} & 2026 & 47 & 62 & 53 & 82 & 81 & 20.5 \\
44 & \mbox{TempCompass} & 2024 & 27 & 22 & 28 & 42 & 40 & 20.2 \\
45 & \mbox{CineTechBench} & 2025 & 25 & 40 & 50 & 60 & 55 & 20.0 \\
46 & \mbox{MMVU} & 2025 & 20 & 57 & 72 & 77 & 84 & 20.0 \\
47 & \mbox{Colon-Bench} & 2026 & 20 & 54 & 64 & 74 & 62 & 20.0 \\
48 & \mbox{OSI-Bench} & -- & 25 & 29 & 30 & 48 & 42 & 19.2 \\
49 & \mbox{ReactHuman} & 2026 & 33 & 31 & 40 & 50 & 51 & 18.9 \\
50 & \mbox{Video-MME} & 2024 & 25 & 52 & 41 & 70 & 71 & 18.2 \\
51 & \mbox{Video-MME-v2} & 2026 & 13 & 19 & 17 & 36 & 39 & 17.7 \\
52 & \mbox{LVBench} & 2024 & 25 & 29 & 29 & 46 & 47 & 16.9 \\
53 & \mbox{SLVMBench} & 2026 & 20 & 63 & 65 & 80 & 73 & 16.7 \\
54 & \mbox{TimeScope} & 2025 & 17 & 36 & 16 & 53 & 58 & 16.6 \\
55 & \mbox{MLVU} & 2024 & 25 & 48 & 33 & 65 & 69 & 16.1 \\
56 & \mbox{KFS-Bench} & 2026 & 21 & 51 & 52 & 67 & 62 & 16.0 \\
57 & \mbox{MMWorld} & 2024 & 26 & 59 & 61 & 74 & 81 & 15.8 \\
58 & \mbox{VidPair-Halluc} & 2026 & 41 & 27 & 42 & 43 & 42 & 15.8 \\
59 & \mbox{V2P-Bench} & 2025 & 25 & 37 & 43 & 53 & 53 & 15.7 \\
60 & \mbox{StreamingBench} & 2024 & 25 & 46 & 35 & 62 & 62 & 15.5 \\
61 & \mbox{ScaleLong} & 2025 & 25 & 49 & 53 & 64 & 60 & 15.3 \\
62 & \mbox{SmartHome-Bench} & 2025 & 50 & 63 & 71 & 78 & 71 & 15.0 \\
63 & \mbox{TOC-Bench} & 2026 & 32 & 35 & 41 & 50 & 47 & 14.6 \\
64 & \mbox{SFI-Bench} & 2026 & 25 & 55 & 21 & 69 & 61 & 14.3 \\
65 & \mbox{M3-Verse} & 2025 & 24 & 26 & 34 & 40 & 34 & 14.0 \\
66 & \mbox{Video-Holmes} & 2025 & 17 & 54 & 58 & 68 & 63 & 13.3 \\
67 & \mbox{PLM-VideoBench} & 2025 & 50 & 72 & 82 & 85 & 80 & 13.2 \\
68 & \mbox{UrbanVideo-Bench} & 2025 & 21 & 41 & 41 & 54 & 44 & 12.7 \\
69 & \mbox{ExpVid} & 2025 & 24 & 52 & 65 & 65 & 63 & 12.4 \\
70 & \mbox{SpatialTree} & 2025 & 17 & 51 & 23 & 62 & 59 & 11.4 \\
71 & \mbox{GUI-World} & 2025 & 25 & 81 & 83 & 92 & 88 & 11.0 \\
72 & \mbox{VANE-Bench} & 2025 & 25 & 77 & 87 & 88 & 82 & 11.0 \\
73 & \mbox{SciVideoBench} & 2025 & 10 & 54 & 52 & 65 & 62 & 11.0 \\
74 & \mbox{LongVideoBench} & 2024 & 21 & 51 & 26 & 60 & 70 & 9.7 \\
75 & \mbox{VideoEval-Pro} & 2025 & 25 & 35 & 25 & 45 & 46 & 9.6 \\
76 & \mbox{SAW-Bench} & 2026 & 26 & 29 & 17 & 38 & 32 & 9.4 \\
77 & \mbox{K9-Bench} & 2026 & 20 & 66 & 64 & 75 & 69 & 9.3 \\
78 & \mbox{CG-Bench} & 2024 & 15 & 48 & 52 & 57 & 48 & 9.2 \\
79 & \mbox{VideoSIAH-Eval} & 2025 & 25 & 4 & 4 & 13 & 11 & 8.9 \\
80 & \mbox{FAVOR-Bench} & 2025 & 20 & 47 & 28 & 55 & 54 & 8.6 \\
81 & \mbox{MMR-V} & 2025 & 10 & 47 & 40 & 56 & 58 & 8.6 \\
82 & \mbox{COVER} & 2025 & 25 & 51 & 45 & 59 & 60 & 8.1 \\
83 & \mbox{S-EMBER} & 2026 & 20 & 29 & 22 & 37 & 24 & 7.7 \\
84 & \mbox{HERBench} & 2026 & 20 & 48 & 50 & 56 & 55 & 7.7 \\
85 & \mbox{ReXTime} & 2024 & 25 & 84 & 88 & 92 & 90 & 7.5 \\
86 & \mbox{PerceptionComp} & 2026 & 20 & 7 & 5 & 13 & 9 & 6.8 \\
87 & \mbox{OVO-S-Bench} & 2026 & 30 & 40 & 42 & 47 & 48 & 6.6 \\
88 & \mbox{Causal2Needles} & 2025 & 25 & 57 & 59 & 64 & 60 & 6.3 \\
89 & \mbox{IntPhys 2} & 2025 & 50 & 52 & 50 & 57 & 55 & 5.4 \\
90 & \mbox{ExAct} & 2025 & 20 & 64 & 61 & 69 & 79 & 5.2 \\
91 & \mbox{TemporalBench} & 2024 & 50 & 72 & 68 & 77 & 76 & 5.1 \\
92 & \mbox{VideoEspresso} & 2025 & 25 & 75 & 79 & 80 & 77 & 4.6 \\
93 & \mbox{VSI-Super-Wild} & 2026 & 25 & 29 & 13 & 32 & 31 & 3.5 \\
94 & \mbox{Artifact-Bench} & 2026 & 17 & 6 & 8 & 9 & 11 & 3.2 \\
95 & \mbox{ForeSeaQA} & 2026 & 25 & 71 & 71 & 73 & 69 & 2.5 \\
96 & \mbox{HLV-1K} & 2025 & 25 & 67 & 66 & 69 & 70 & 1.9 \\
97 & \mbox{STI-Bench} & 2025 & 20 & 47 & 49 & 49 & 48 & 1.7 \\
98 & \mbox{DSI-Bench} & 2025 & 25 & 52 & 44 & 53 & 60 & 1.2 \\
99 & \mbox{LongVideo-Reason} & 2025 & 27 & 76 & 73 & 77 & 79 & 1.0 \\
100 & \mbox{VF-Eval} & 2025 & 25 & 83 & 81 & 84 & 82 & 0.7 \\
102 & \mbox{DSR-Bench} & -- & 25 & 24 & 26 & 24 & 34 & -0.1 \\
103 & \mbox{ALLVB} & 2025 & 20 & 73 & 70 & 73 & 71 & -0.7 \\
104 & \mbox{FineBench} & 2026 & 25 & 63 & 57 & 62 & 63 & -1.4 \\
105 & \mbox{HourVideo} & 2024 & 20 & 23 & 10 & 21 & 24 & -1.5 \\
106 & \mbox{VAU-Bench} & 2025 & 25 & 96 & 95 & 94 & 95 & -1.6 \\
107 & \mbox{LEMONADE} & 2025 & 25 & 42 & 32 & 40 & 38 & -1.7 \\
108 & \mbox{MMSI-Video-Bench} & 2025 & 24 & 30 & 17 & 28 & 35 & -1.8 \\
109 & \mbox{EgoPlan-Bench} & 2023 & 25 & 33 & 28 & 29 & 30 & -3.7 \\
110 & \mbox{VideoMathQA} & 2025 & 44 & 56 & 26 & 52 & 69 & -4.5 \\
111 & \mbox{VideoOdyssey} & 2026 & 25 & 44 & 23 & 35 & 49 & -9.6 \\
112 & \mbox{LongTimeScope} & -- & 17 & 34 & 4 & 16 & 27 & -17.9 \\
113 & \mbox{NExT-QA} & 2021 & 20 & -- & -- & -- & 79 & -- \\
114 & \mbox{Neptune} & 2024 & 20 & 72 & 71 & 81 & -- & -- \\
115 & \mbox{VSTAT} & 2026 & 26 & -- & -- & -- & 50 & -- \\
\end{xltabular}
\endgroup

\begingroup
\setlength{\LTcapwidth}{\linewidth}
\setlength{\tabcolsep}{4pt}
\begin{xltabular}{\linewidth}{>{\raggedleft\arraybackslash}p{28.71bp}>{\raggedright\arraybackslash}p{111.37bp}>{\raggedleft\arraybackslash}p{35.07bp}>{\raggedleft\arraybackslash}p{49.97bp}>{\raggedleft\arraybackslash}p{41.25bp}>{\raggedleft\arraybackslash}p{31.65bp}>{\raggedleft\arraybackslash}p{57.98bp}>{\raggedleft\arraybackslash}p{37.57bp}}
\caption{Audit results ranked by 32-frame accuracy minus blind accuracy (2/2).}\label{tab:master_cols2}\\
\toprule
\rowcolor{SkillPaleBlue} \# & Benchmark & Lift & Reason. & Proto. & Div. & NearDup & mink \\
\midrule
\endfirsthead
\multicolumn{8}{l}{\tablename~\thetable{} (continued)}\\
\toprule
\rowcolor{SkillPaleBlue} \# & Benchmark & Lift & Reason. & Proto. & Div. & NearDup & mink \\
\midrule
\endhead
\bottomrule
\endfoot
2 & \mbox{VNBench} & 1.0 & 87 & 4 & 0.12 & 58 & 7.0 \\
3 & \mbox{v-HUB} & -3.1 & 39 & 50 & 0.21 & 22 & 9.8 \\
4 & \mbox{SoMi-ToM} & 17.3 & 67 & 0 & 0.05 & 100 & 9.7 \\
5 & \mbox{VGenST-Bench} & 18.7 & 67 & 0 & 0.25 & 28 & 8.0 \\
6 & \mbox{MME-VideoOCR} & 17.7 & 69 & 3 & 0.35 & 10 & 7.3 \\
7 & \mbox{UCF101-AD} & -1.8 & 10 & 21 & 0.06 & 92 & 6.4 \\
8 & \mbox{EgoSchema} & 19.3 & 77 & 22 & 0.12 & 66 & 6.7 \\
9 & \mbox{Video-LevelGauge} & 27.7 & 62 & 0 & 0.34 & 22 & 8.0 \\
10 & \mbox{Vinoground} & -4.0 & 72 & 0 & 0.06 & 81 & 7.5 \\
11 & \mbox{ViMUL-Bench} & 18.3 & 52 & 49 & 0.28 & 10 & 7.3 \\
12 & \mbox{Video-TT} & 3.3 & 58 & 14 & 0.31 & 13 & 7.7 \\
13 & \mbox{CAViAR} & 13.2 & 69 & 38 & 0.02 & 100 & 6.2 \\
14 & \mbox{VideoKR} & 9.1 & 82 & 52 & 0.25 & 26 & 8.0 \\
15 & \mbox{MORSE-500} & -9.8 & 29 & 16 & 0.02 & 100 & 7.4 \\
16 & \mbox{VideoAds} & 26.0 & 63 & 23 & 0.20 & 27 & 6.3 \\
17 & \mbox{NaturalBench-Video} & 0.7 & 89 & 0 & 0.22 & 67 & 8.3 \\
18 & \mbox{MVBench} & 10.3 & 38 & 8 & 0.15 & 67 & 9.2 \\
19 & \mbox{TaskMeAnything-v1} & 6.7 & 45 & 0 & 0.08 & 84 & 8.5 \\
20 & \mbox{GasVideo-1000} & 16.1 & 67 & 71 & 0.27 & 39 & 11.0 \\
21 & \mbox{AoTBench} & 0.7 & 71 & 0 & 0.16 & 35 & 8.8 \\
22 & \mbox{VELOCITI} & 15.3 & 38 & 0 & 0.02 & 99 & 7.2 \\
23 & \mbox{Q-Bench-Video} & 17.0 & 52 & 38 & 0.19 & 12 & 7.8 \\
24 & \mbox{VRU-Accident} & 23.3 & 80 & 9 & 0.02 & 100 & 7.8 \\
25 & \mbox{Perception Test} & 13.7 & 55 & 4 & 0.16 & 77 & 9.0 \\
26 & \mbox{iSafetyBench} & 8.5 & 30 & 4 & 0.05 & 99 & 6.8 \\
27 & \mbox{MotionBench} & 13.0 & 48 & 9 & 0.22 & 29 & 6.9 \\
28 & \mbox{PAI-Bench-U} & 9.0 & 66 & 0 & 0.22 & 46 & 9.1 \\
29 & \mbox{E.T. Bench} & 5.3 & 42 & 0 & 0.06 & 97 & 8.1 \\
30 & \mbox{TVBench} & 4.0 & 44 & 13 & 0.09 & 90 & 8.2 \\
31 & \mbox{VKnowU} & 9.9 & 76 & 29 & 0.23 & 42 & 6.9 \\
32 & \mbox{VSI-Bench} & 7.5 & 20 & 8 & 0.07 & 98 & 7.9 \\
33 & \mbox{VITATECS} & 18.5 & 67 & 14 & 0.06 & 79 & 7.1 \\
34 & \mbox{ContPhy} & 2.3 & 44 & 0 & 0.05 & 100 & 8.6 \\
35 & \mbox{VURB} & 13.2 & 50 & 42 & 0.22 & 61 & 8.0 \\
36 & \mbox{SITE-Bench} & 17.3 & 39 & 9 & 0.21 & 47 & 9.1 \\
37 & \mbox{Video-MMMU} & 36.6 & 83 & 38 & 0.30 & 23 & 6.5 \\
38 & \mbox{SIS-Bench} & 25.0 & 33 & 0 & 0.06 & 85 & 9.2 \\
39 & \mbox{TimeBlind} & -0.8 & 53 & 6 & 0.19 & 80 & 7.9 \\
40 & \mbox{TOMATO} & 2.1 & 42 & 10 & 0.15 & 72 & 9.4 \\
41 & \mbox{VCR-Bench} & 21.1 & 60 & 2 & 0.34 & 9 & 8.9 \\
42 & \mbox{ReVSI} & 4.3 & 21 & 12 & 0.09 & 94 & 8.2 \\
43 & \mbox{ST-Evidence} & 14.2 & 59 & 15 & 0.24 & 35 & 7.9 \\
44 & \mbox{TempCompass} & -4.5 & 75 & 57 & 0.17 & 58 & 8.6 \\
45 & \mbox{CineTechBench} & 15.2 & 45 & 0 & 0.07 & 88 & 8.4 \\
46 & \mbox{MMVU} & 36.9 & 80 & 27 & 0.25 & 25 & 7.3 \\
47 & \mbox{Colon-Bench} & 34.3 & 56 & 0 & 0.04 & 100 & 6.8 \\
48 & \mbox{OSI-Bench} & 3.6 & 20 & 4 & 0.08 & 91 & 8.2 \\
49 & \mbox{ReactHuman} & -2.0 & 58 & 0 & 0.01 & 100 & 8.4 \\
50 & \mbox{Video-MME} & 27.0 & 69 & 40 & 0.27 & 5 & 7.0 \\
51 & \mbox{Video-MME-v2} & 6.0 & 73 & 28 & 0.31 & 15 & 6.5 \\
52 & \mbox{LVBench} & 3.7 & 70 & 40 & 0.15 & 68 & 8.3 \\
53 & \mbox{SLVMBench} & 43.0 & 79 & 7 & 0.20 & 51 & 6.3 \\
54 & \mbox{TimeScope} & 19.3 & 70 & 82 & 0.07 & 100 & 9.7 \\
55 & \mbox{MLVU} & 23.5 & 69 & 59 & 0.11 & 77 & 7.8 \\
56 & \mbox{KFS-Bench} & 30.2 & 71 & 7 & 0.26 & 27 & 7.2 \\
57 & \mbox{MMWorld} & 32.7 & 77 & 26 & 0.23 & 53 & 7.6 \\
58 & \mbox{VidPair-Halluc} & -14.2 & 83 & 31 & 0.14 & 75 & 9.0 \\
59 & \mbox{V2P-Bench} & 11.9 & 67 & 49 & 0.19 & 25 & 9.7 \\
60 & \mbox{StreamingBench} & 21.4 & 51 & 31 & 0.21 & 54 & 7.1 \\
61 & \mbox{ScaleLong} & 23.7 & 74 & 10 & 0.27 & 28 & 6.5 \\
62 & \mbox{SmartHome-Bench} & 13.0 & 67 & 29 & 0.02 & 99 & 5.9 \\
63 & \mbox{TOC-Bench} & 2.9 & 69 & 1 & 0.16 & 42 & 8.7 \\
64 & \mbox{SFI-Bench} & 29.7 & 58 & 3 & 0.14 & 59 & 7.6 \\
65 & \mbox{M3-Verse} & 2.7 & 50 & 1 & 0.02 & 100 & 7.8 \\
66 & \mbox{Video-Holmes} & 37.7 & 88 & 4 & 0.16 & 36 & 8.5 \\
67 & \mbox{PLM-VideoBench} & 21.7 & 62 & 0 & 0.24 & 19 & 6.3 \\
68 & \mbox{UrbanVideo-Bench} & 20.3 & 58 & 1 & 0.07 & 94 & 7.4 \\
69 & \mbox{ExpVid} & 27.8 & 56 & 0 & 0.11 & 69 & 8.7 \\
70 & \mbox{SpatialTree} & 33.3 & 33 & 16 & 0.08 & 79 & 7.4 \\
71 & \mbox{GUI-World} & 56.0 & 72 & 5 & 0.27 & 12 & 7.0 \\
72 & \mbox{VANE-Bench} & 52.3 & 62 & 0 & 0.16 & 72 & 6.9 \\
73 & \mbox{SciVideoBench} & 44.0 & 95 & 2 & 0.08 & 76 & 7.8 \\
74 & \mbox{LongVideoBench} & 29.3 & 68 & 59 & 0.30 & 15 & 8.2 \\
75 & \mbox{VideoEval-Pro} & 10.7 & 79 & 47 & 0.30 & 17 & 7.4 \\
76 & \mbox{SAW-Bench} & 2.9 & 39 & 0 & 0.04 & 97 & 8.8 \\
77 & \mbox{K9-Bench} & 45.5 & 52 & 0 & 0.15 & 41 & 7.5 \\
78 & \mbox{CG-Bench} & 32.9 & 48 & 8 & -- & -- & 6.4 \\
79 & \mbox{VideoSIAH-Eval} & -21.1 & 42 & 82 & 0.32 & 8 & 10.5 \\
80 & \mbox{FAVOR-Bench} & 26.7 & 46 & 10 & 0.19 & 26 & 8.0 \\
81 & \mbox{MMR-V} & 37.4 & 96 & 25 & 0.20 & 24 & 7.5 \\
82 & \mbox{COVER} & 26.1 & 65 & 31 & 0.26 & 8 & 7.4 \\
83 & \mbox{S-EMBER} & 9.4 & 45 & 0 & 0.22 & 12 & 6.5 \\
84 & \mbox{HERBench} & 28.0 & 53 & 0 & 0.06 & 99 & 7.0 \\
85 & \mbox{ReXTime} & 59.3 & 73 & 8 & 0.31 & 6 & 7.0 \\
86 & \mbox{PerceptionComp} & -13.3 & 33 & 17 & 0.17 & 67 & 8.7 \\
87 & \mbox{OVO-S-Bench} & 9.9 & 74 & 2 & 0.08 & 91 & 7.3 \\
88 & \mbox{Causal2Needles} & 32.3 & 60 & 0 & 0.14 & 42 & 9.0 \\
89 & \mbox{IntPhys 2} & 1.7 & 17 & 19 & 0.02 & 100 & 10.6 \\
90 & \mbox{ExAct} & 44.0 & 74 & 6 & 0.02 & 100 & 6.5 \\
91 & \mbox{TemporalBench} & 21.8 & 63 & 5 & 0.14 & 62 & 6.7 \\
92 & \mbox{VideoEspresso} & 50.0 & 90 & 12 & 0.26 & 23 & 6.8 \\
93 & \mbox{VSI-Super-Wild} & 4.0 & 55 & 15 & 0.04 & 97 & 7.6 \\
94 & \mbox{Artifact-Bench} & -11.0 & 47 & 2 & 0.11 & 49 & 10.4 \\
95 & \mbox{ForeSeaQA} & 45.9 & 54 & 0 & 0.13 & 84 & 6.5 \\
96 & \mbox{HLV-1K} & 42.0 & 57 & 4 & 0.16 & 46 & 6.7 \\
97 & \mbox{STI-Bench} & 27.2 & 43 & 1 & 0.10 & 88 & 8.1 \\
98 & \mbox{DSI-Bench} & 26.5 & 27 & 1 & 0.09 & 67 & 6.7 \\
99 & \mbox{LongVideo-Reason} & 49.5 & 94 & 8 & 0.21 & 11 & 8.1 \\
100 & \mbox{VF-Eval} & 58.3 & 61 & 0 & 0.18 & 30 & 7.6 \\
102 & \mbox{DSR-Bench} & -0.9 & 41 & 4 & 0.16 & 49 & 6.4 \\
103 & \mbox{ALLVB} & 53.3 & 89 & 9 & 0.17 & 24 & 6.7 \\
104 & \mbox{FineBench} & 38.3 & 77 & 0 & 0.04 & 100 & 7.5 \\
105 & \mbox{HourVideo} & 2.7 & 62 & 28 & 0.14 & 74 & 7.0 \\
106 & \mbox{VAU-Bench} & 71.0 & 43 & 0 & 0.12 & 70 & 6.3 \\
107 & \mbox{LEMONADE} & 16.7 & 57 & 3 & 0.05 & 100 & 10.7 \\
108 & \mbox{MMSI-Video-Bench} & 5.9 & 41 & 14 & 0.16 & 48 & 7.0 \\
109 & \mbox{EgoPlan-Bench} & 8.0 & 68 & 12 & 0.04 & 97 & 9.0 \\
110 & \mbox{VideoMathQA} & 11.7 & 71 & 43 & 0.20 & 63 & 7.3 \\
111 & \mbox{VideoOdyssey} & 19.4 & 79 & 51 & 0.30 & 27 & 7.1 \\
112 & \mbox{LongTimeScope} & 17.0 & 50 & 88 & 0.05 & 100 & 9.7 \\
113 & \mbox{NExT-QA} & -- & -- & -- & 0.26 & 7 & 9.1 \\
114 & \mbox{Neptune} & -- & 33 & 73 & 0.28 & 14 & 8.8 \\
115 & \mbox{VSTAT} & -- & -- & -- & 0.06 & 100 & 8.7 \\
\end{xltabular}
\endgroup

\endgroup

\clearpage
\section{Benchmark report cards}
\label{app:cards}
One card per audited benchmark, the page released with the pool. Each card gives attack gaps, breaking level, reference and shortcut accuracies, dominant error categories, effective size, near-duplicate share, video dependence, protocol-error share, and pool contribution. The cards regenerate from the same exported tables as the paper.

\noindent\includegraphics[width=1.0\linewidth]{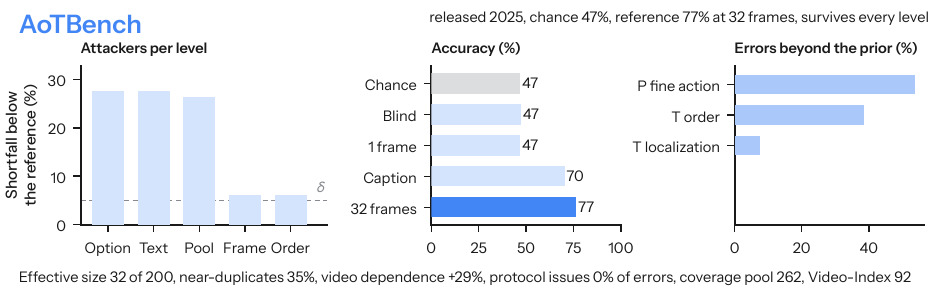}\par\vspace{2pt}
\noindent\includegraphics[width=1.0\linewidth]{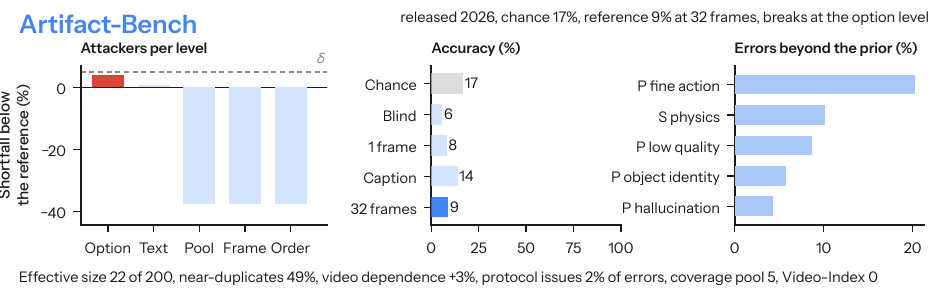}\par\vspace{2pt}
\noindent\includegraphics[width=1.0\linewidth]{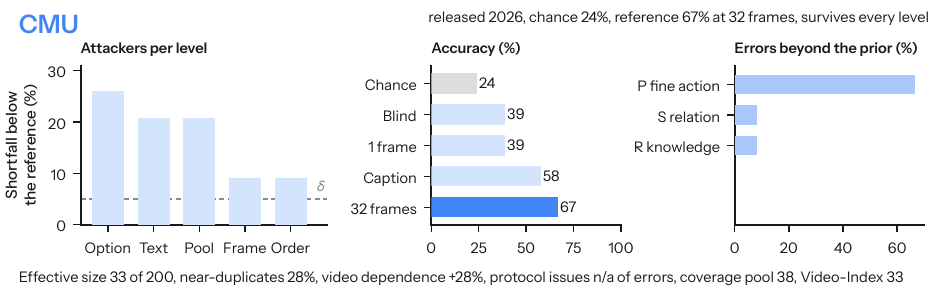}\par\vspace{2pt}
\clearpage
\noindent\includegraphics[width=1.0\linewidth]{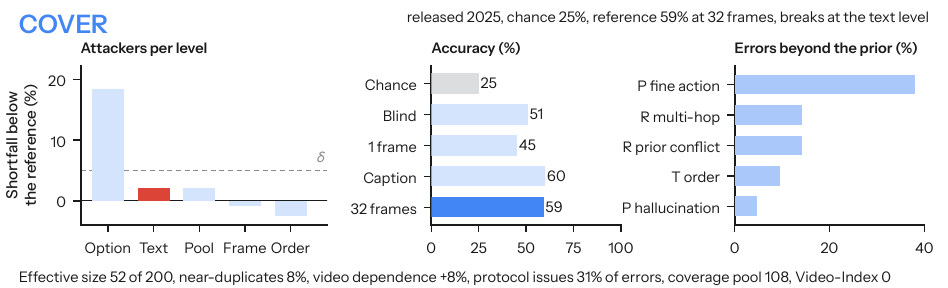}\par\vspace{2pt}
\noindent\includegraphics[width=1.0\linewidth]{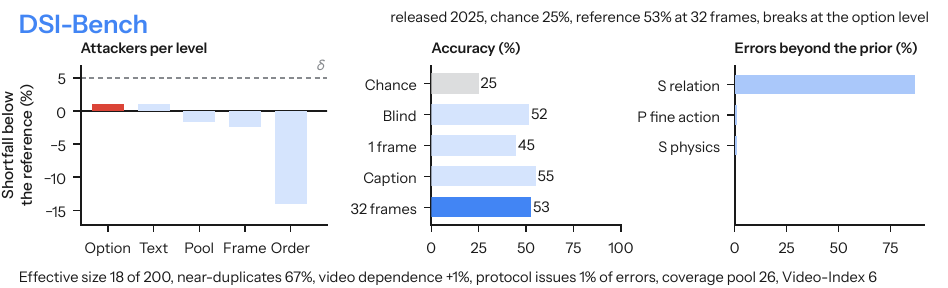}\par\vspace{2pt}
\noindent\includegraphics[width=1.0\linewidth]{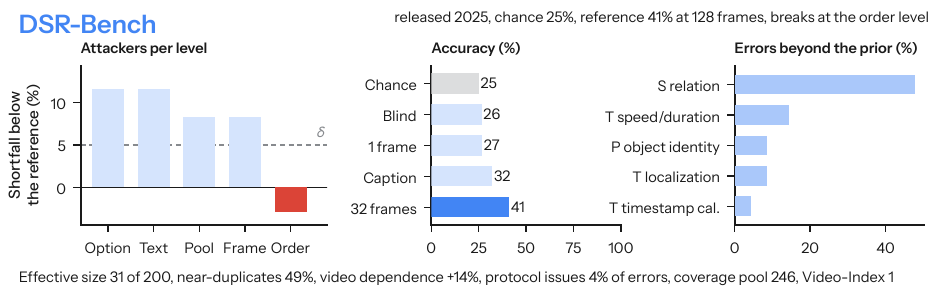}\par\vspace{2pt}
\noindent\includegraphics[width=1.0\linewidth]{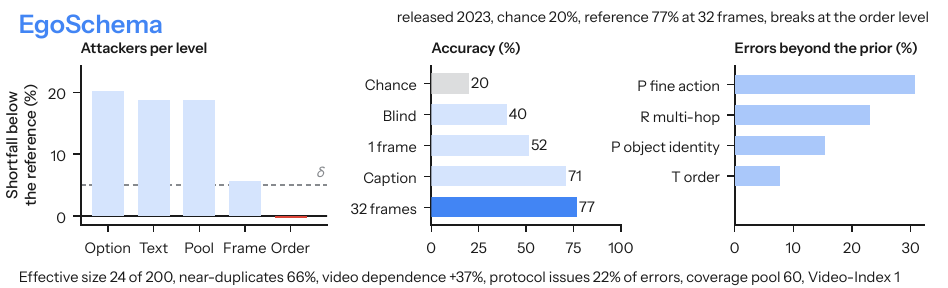}\par\vspace{2pt}
\clearpage
\noindent\includegraphics[width=1.0\linewidth]{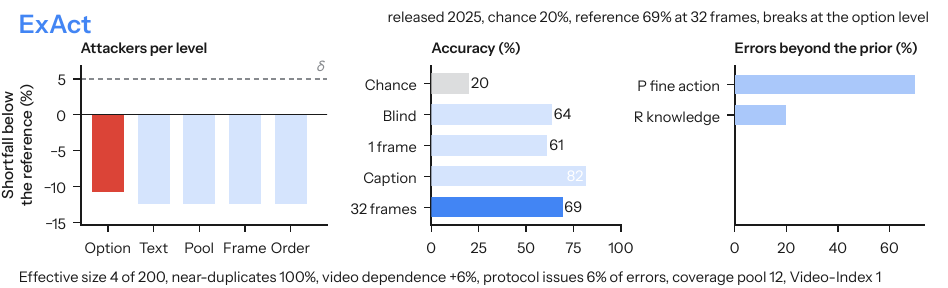}\par\vspace{2pt}
\noindent\includegraphics[width=1.0\linewidth]{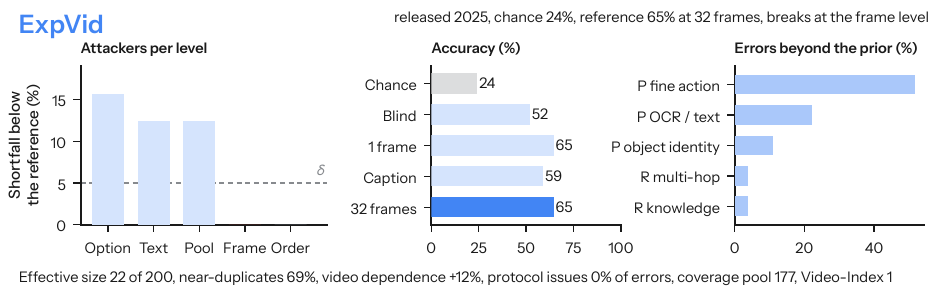}\par\vspace{2pt}
\noindent\includegraphics[width=1.0\linewidth]{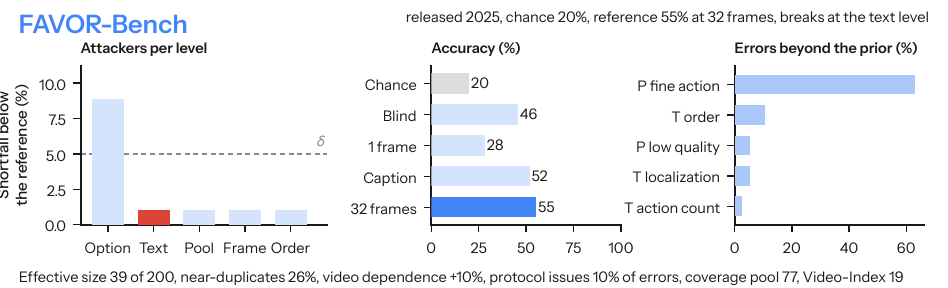}\par\vspace{2pt}
\noindent\includegraphics[width=1.0\linewidth]{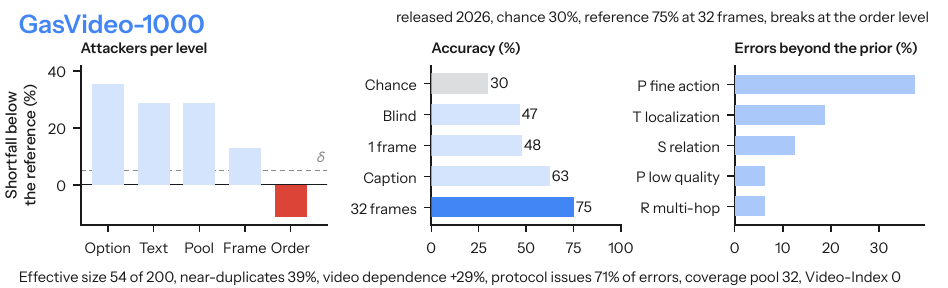}\par\vspace{2pt}
\clearpage
\noindent\includegraphics[width=1.0\linewidth]{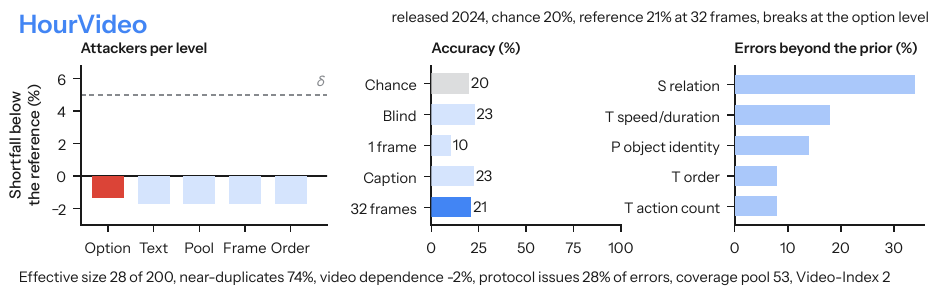}\par\vspace{2pt}
\noindent\includegraphics[width=1.0\linewidth]{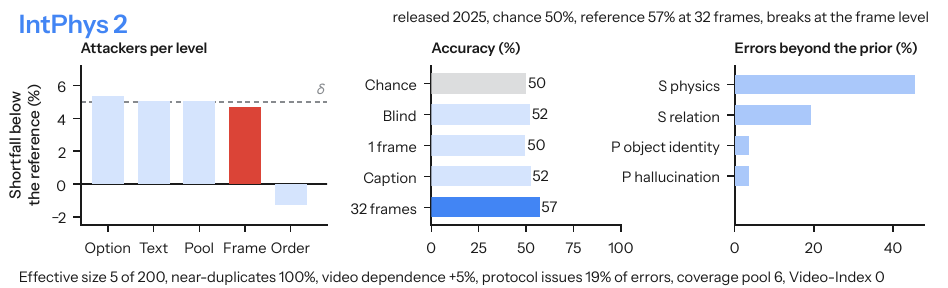}\par\vspace{2pt}
\noindent\includegraphics[width=1.0\linewidth]{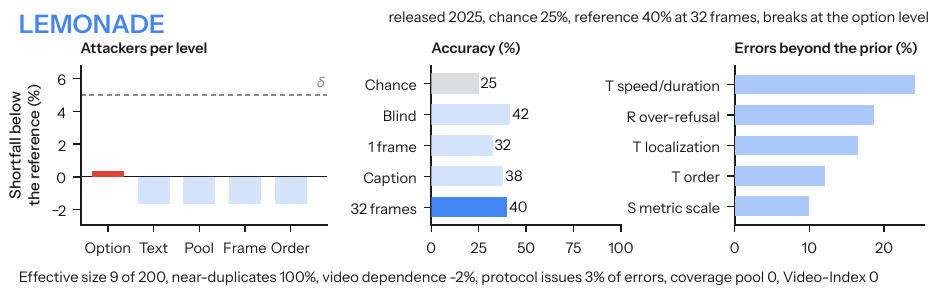}\par\vspace{2pt}
\noindent\includegraphics[width=1.0\linewidth]{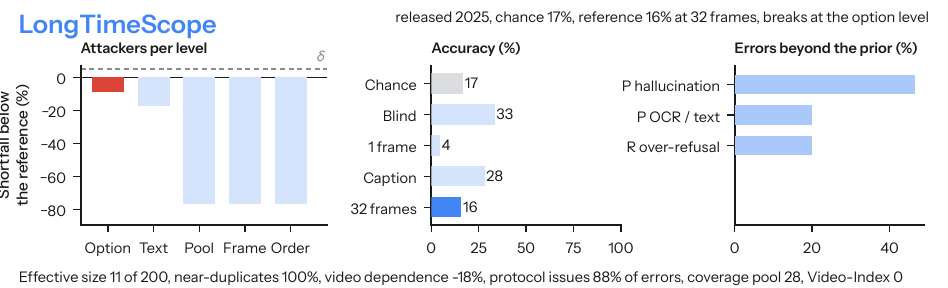}\par\vspace{2pt}
\clearpage
\noindent\includegraphics[width=1.0\linewidth]{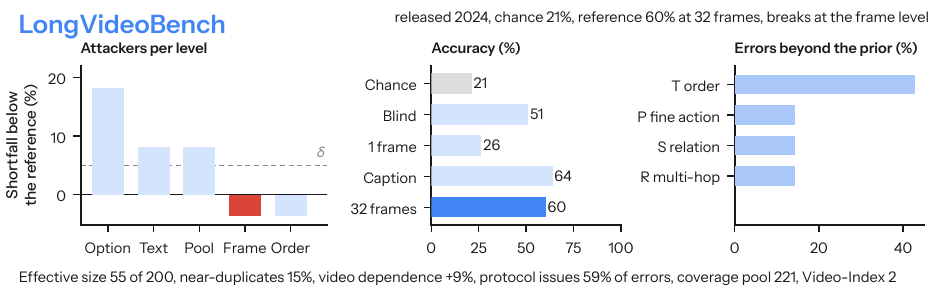}\par\vspace{2pt}
\noindent\includegraphics[width=1.0\linewidth]{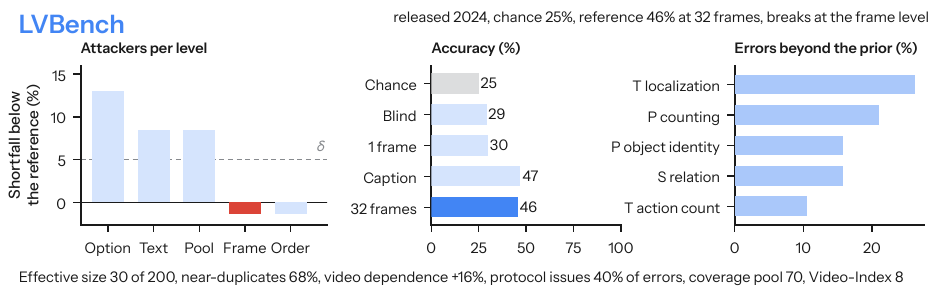}\par\vspace{2pt}
\noindent\includegraphics[width=1.0\linewidth]{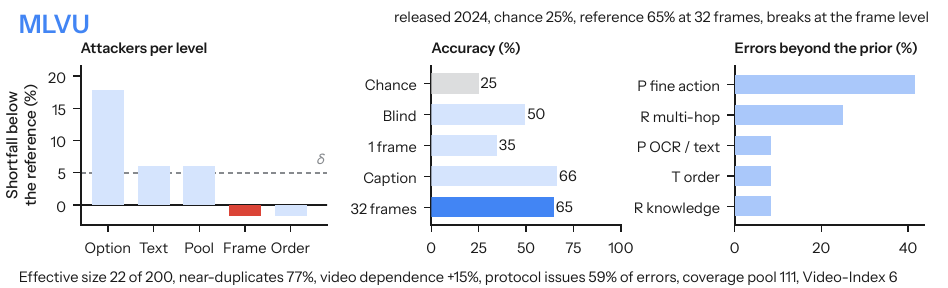}\par\vspace{2pt}
\noindent\includegraphics[width=1.0\linewidth]{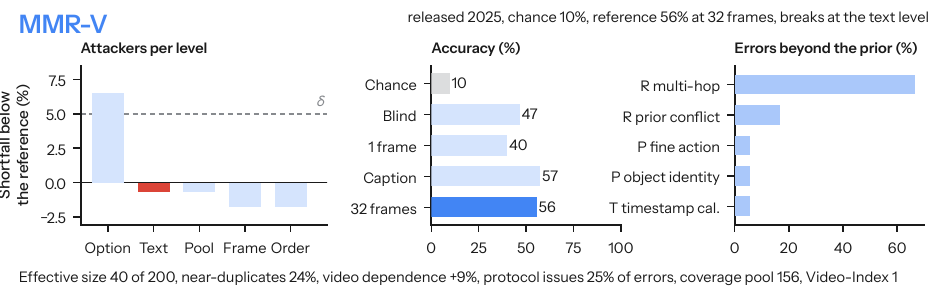}\par\vspace{2pt}
\clearpage
\noindent\includegraphics[width=1.0\linewidth]{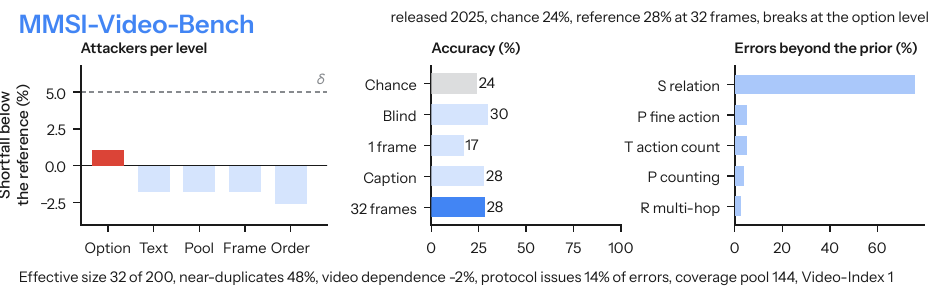}\par\vspace{2pt}
\noindent\includegraphics[width=1.0\linewidth]{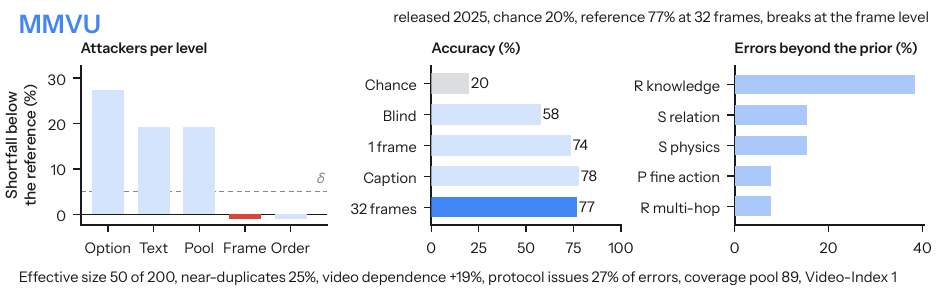}\par\vspace{2pt}
\noindent\includegraphics[width=1.0\linewidth]{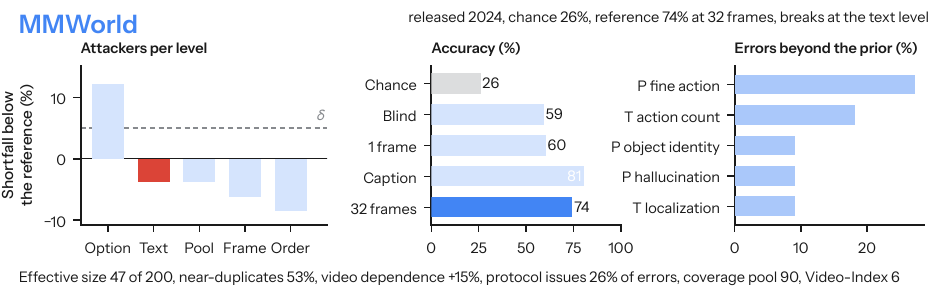}\par\vspace{2pt}
\noindent\includegraphics[width=1.0\linewidth]{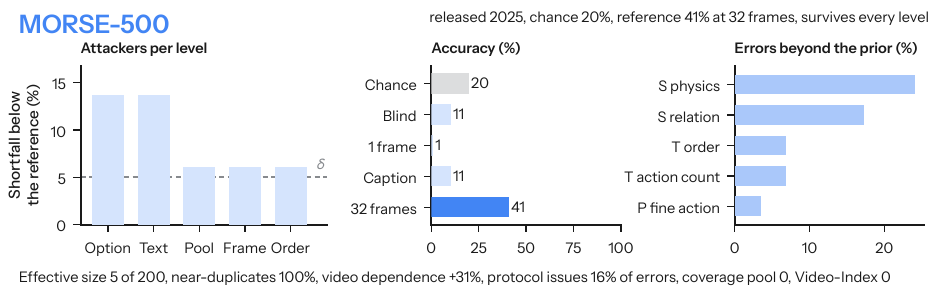}\par\vspace{2pt}
\clearpage
\noindent\includegraphics[width=1.0\linewidth]{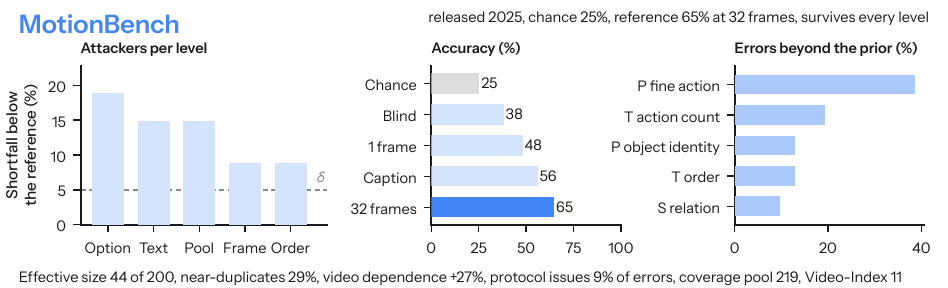}\par\vspace{2pt}
\noindent\includegraphics[width=1.0\linewidth]{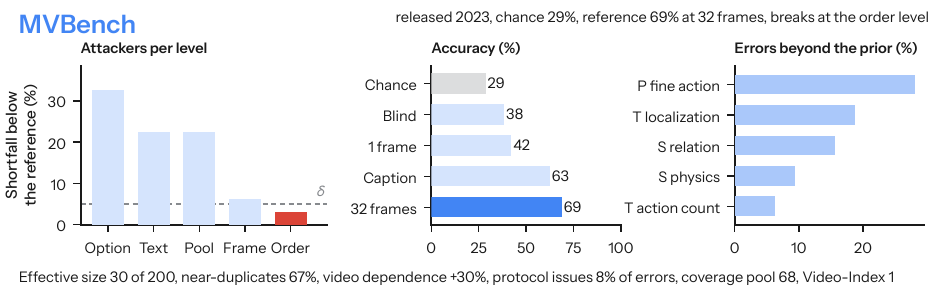}\par\vspace{2pt}
\noindent\includegraphics[width=1.0\linewidth]{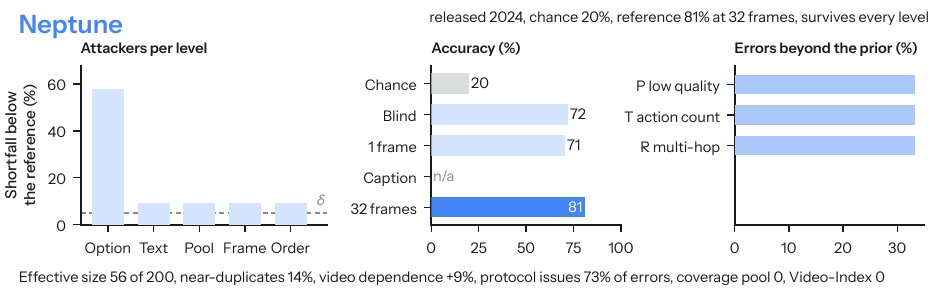}\par\vspace{2pt}
\noindent\includegraphics[width=1.0\linewidth]{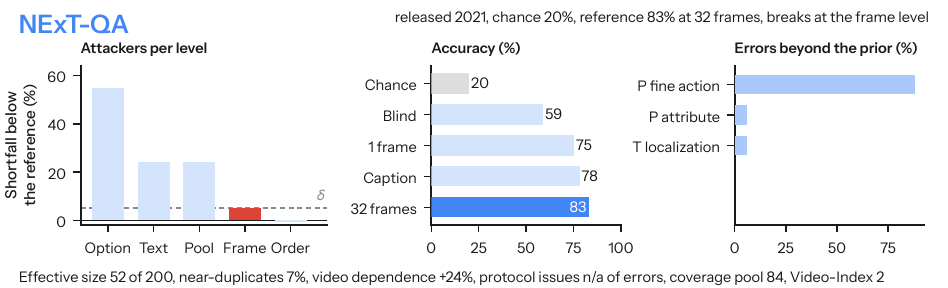}\par\vspace{2pt}
\clearpage
\noindent\includegraphics[width=1.0\linewidth]{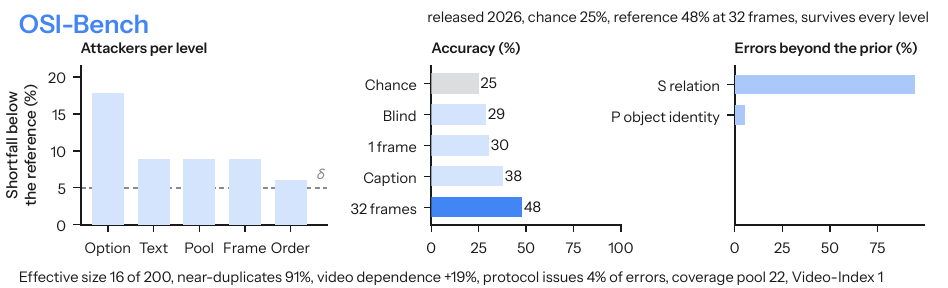}\par\vspace{2pt}
\noindent\includegraphics[width=1.0\linewidth]{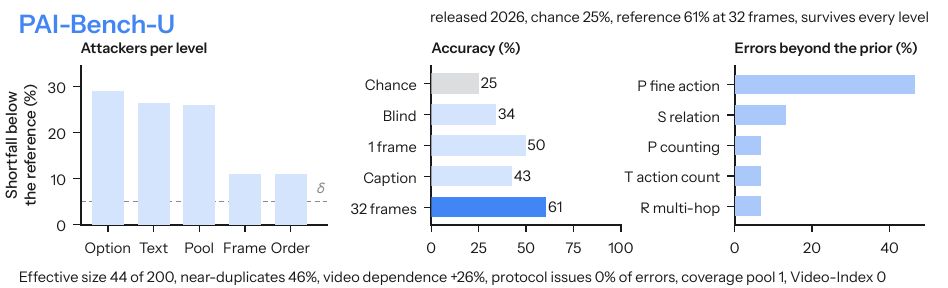}\par\vspace{2pt}
\noindent\includegraphics[width=1.0\linewidth]{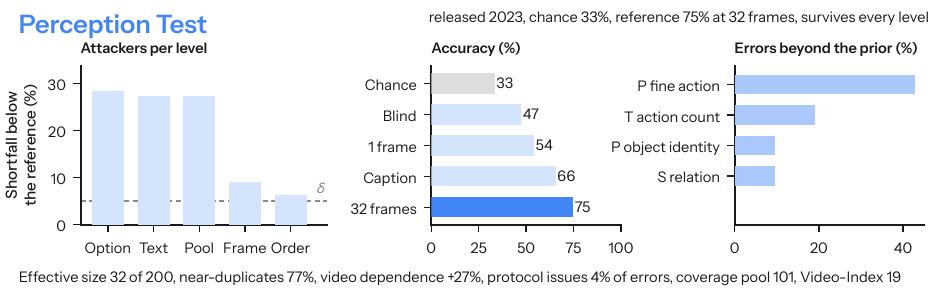}\par\vspace{2pt}
\noindent\includegraphics[width=1.0\linewidth]{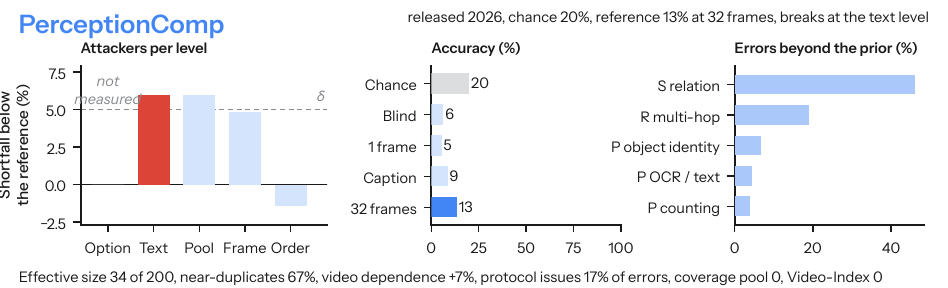}\par\vspace{2pt}
\clearpage
\noindent\includegraphics[width=1.0\linewidth]{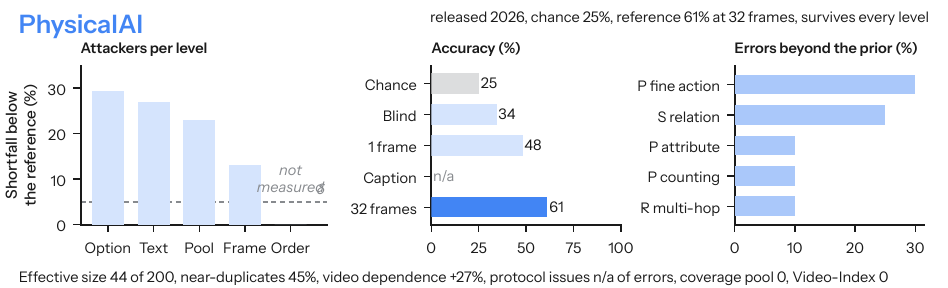}\par\vspace{2pt}
\noindent\includegraphics[width=1.0\linewidth]{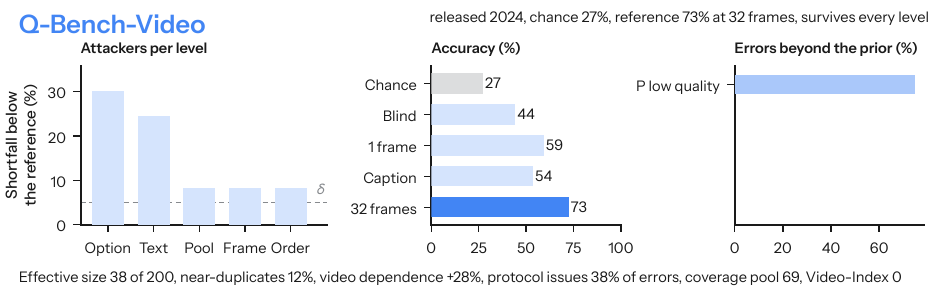}\par\vspace{2pt}
\noindent\includegraphics[width=1.0\linewidth]{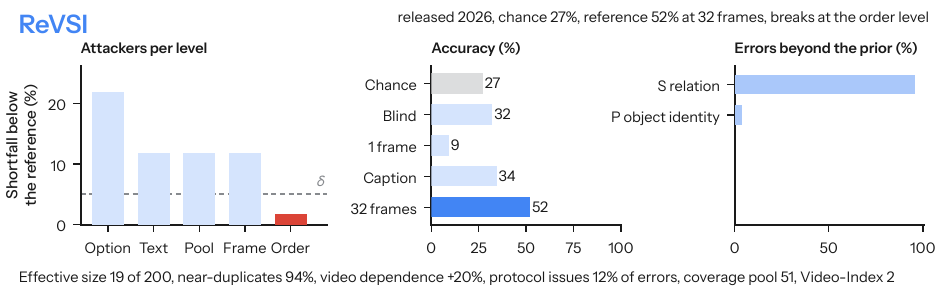}\par\vspace{2pt}
\noindent\includegraphics[width=1.0\linewidth]{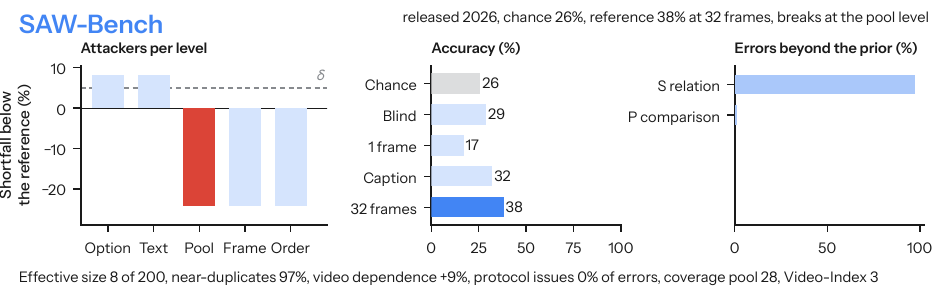}\par\vspace{2pt}
\clearpage
\noindent\includegraphics[width=1.0\linewidth]{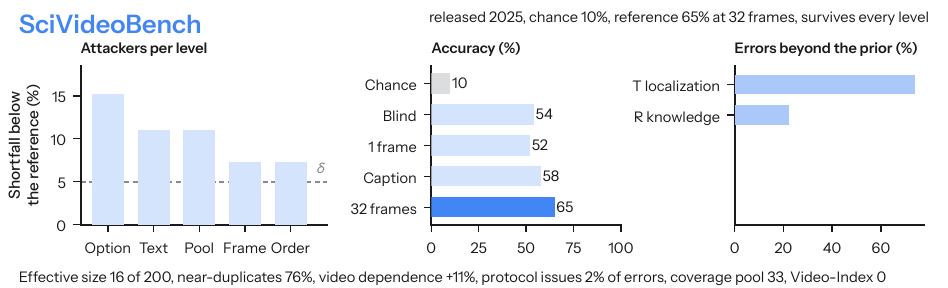}\par\vspace{2pt}
\noindent\includegraphics[width=1.0\linewidth]{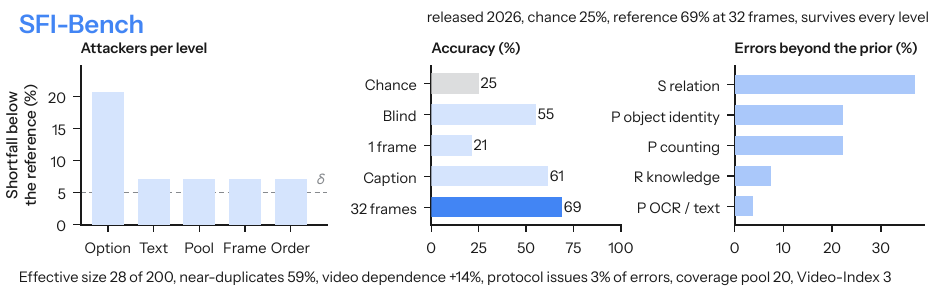}\par\vspace{2pt}
\noindent\includegraphics[width=1.0\linewidth]{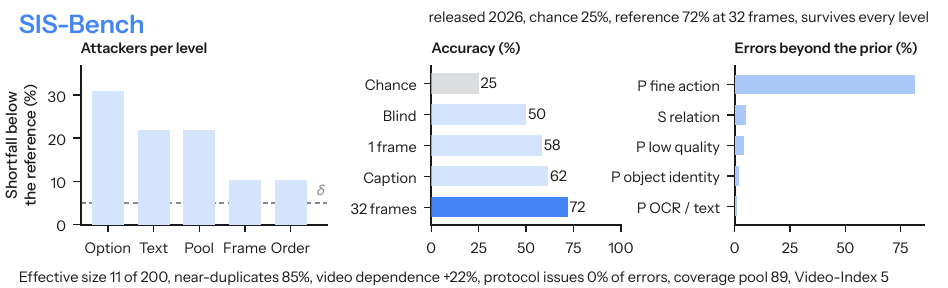}\par\vspace{2pt}
\noindent\includegraphics[width=1.0\linewidth]{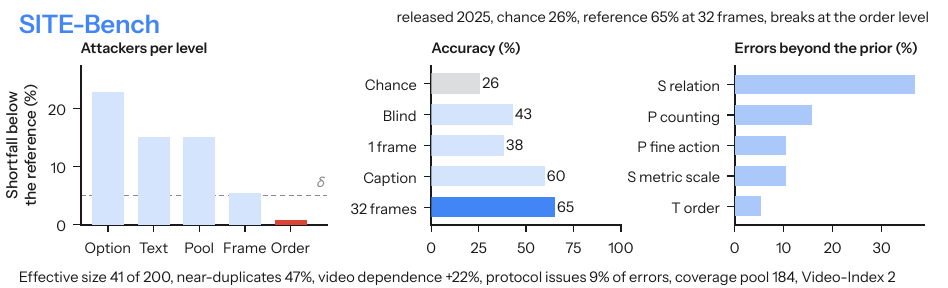}\par\vspace{2pt}
\clearpage
\noindent\includegraphics[width=1.0\linewidth]{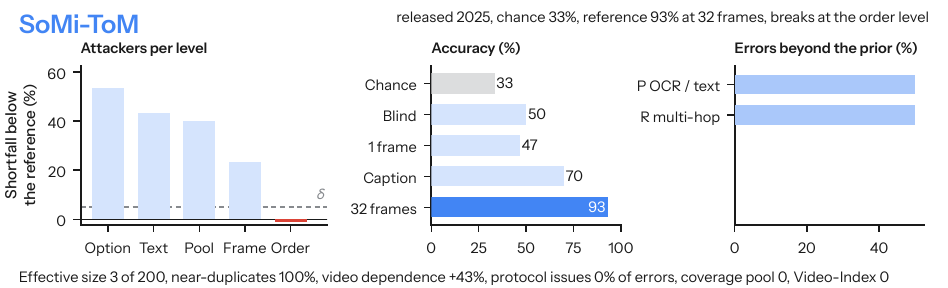}\par\vspace{2pt}
\noindent\includegraphics[width=1.0\linewidth]{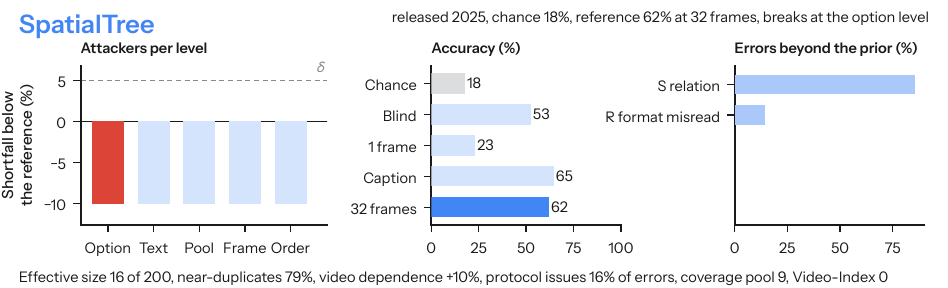}\par\vspace{2pt}
\noindent\includegraphics[width=1.0\linewidth]{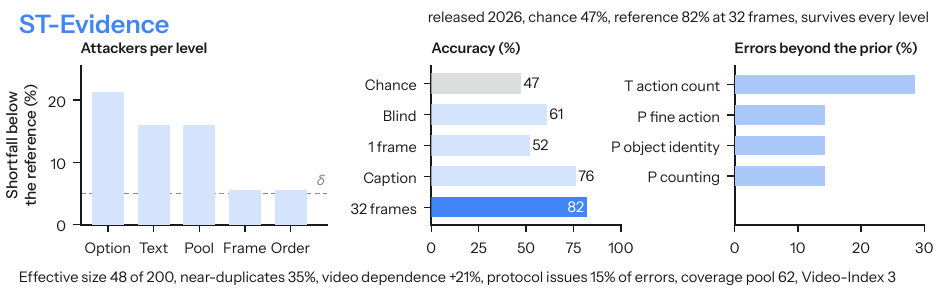}\par\vspace{2pt}
\noindent\includegraphics[width=1.0\linewidth]{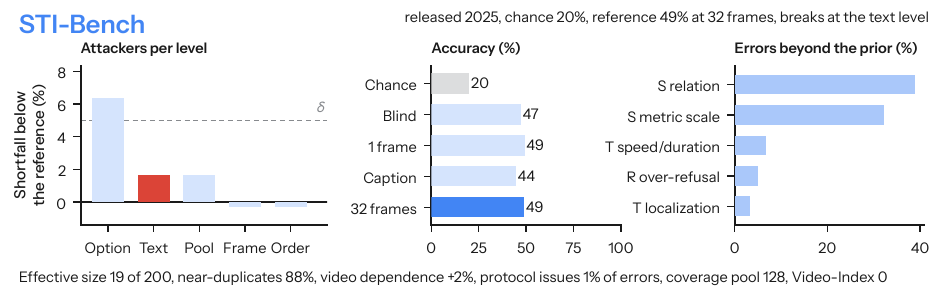}\par\vspace{2pt}
\clearpage
\noindent\includegraphics[width=1.0\linewidth]{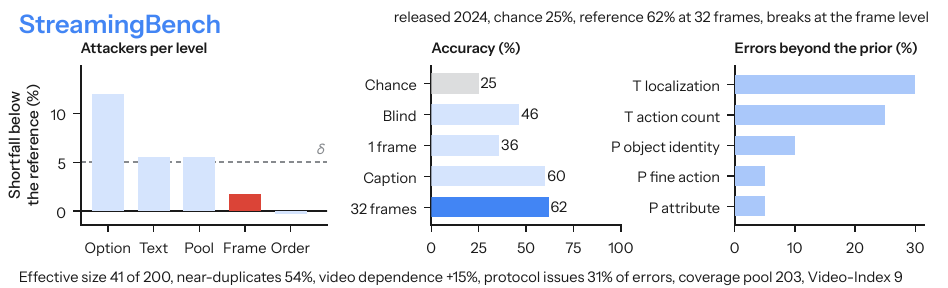}\par\vspace{2pt}
\noindent\includegraphics[width=1.0\linewidth]{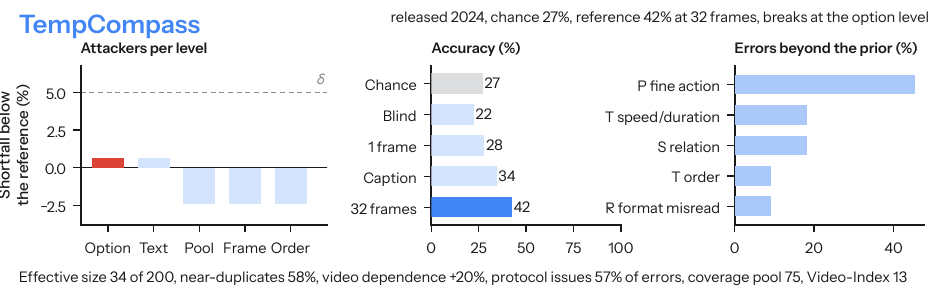}\par\vspace{2pt}
\noindent\includegraphics[width=1.0\linewidth]{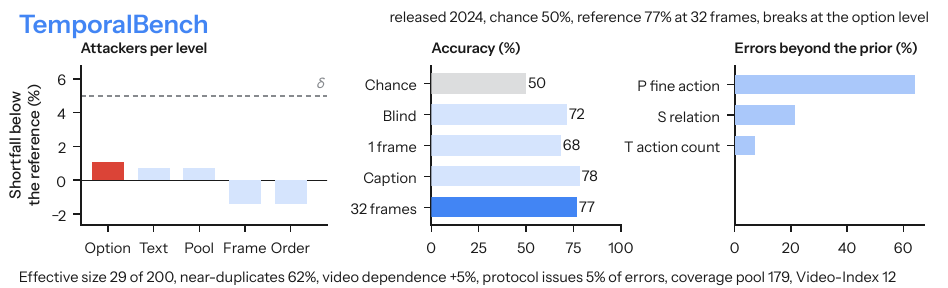}\par\vspace{2pt}
\noindent\includegraphics[width=1.0\linewidth]{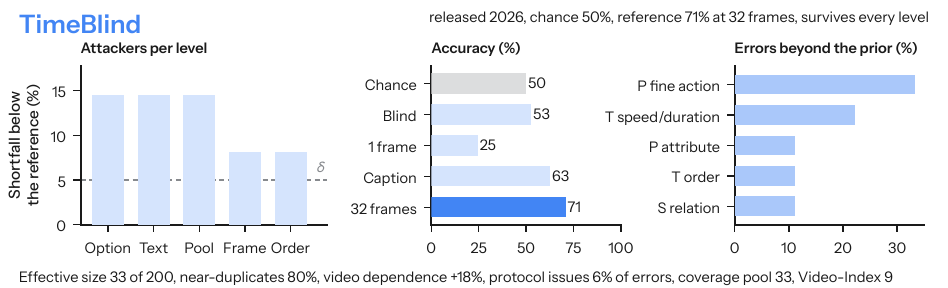}\par\vspace{2pt}
\clearpage
\noindent\includegraphics[width=1.0\linewidth]{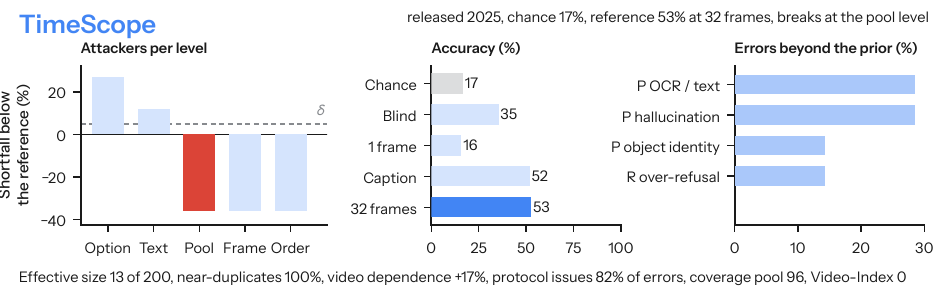}\par\vspace{2pt}
\noindent\includegraphics[width=1.0\linewidth]{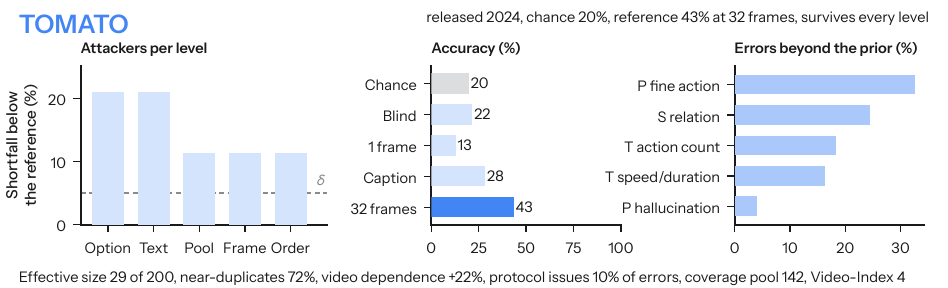}\par\vspace{2pt}
\noindent\includegraphics[width=1.0\linewidth]{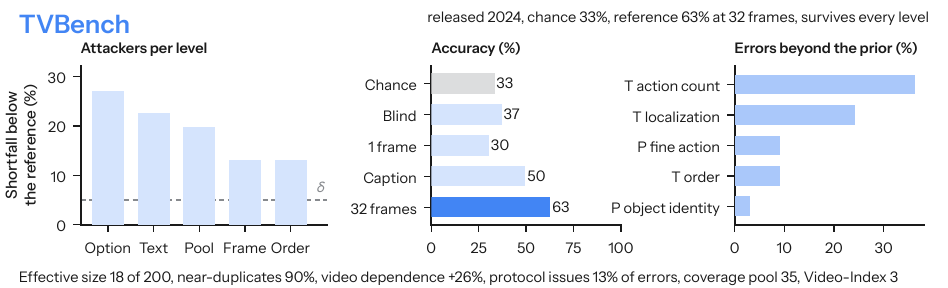}\par\vspace{2pt}
\noindent\includegraphics[width=1.0\linewidth]{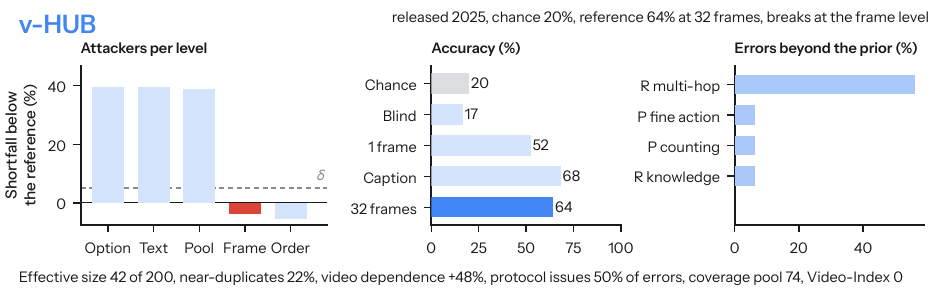}\par\vspace{2pt}
\clearpage
\noindent\includegraphics[width=1.0\linewidth]{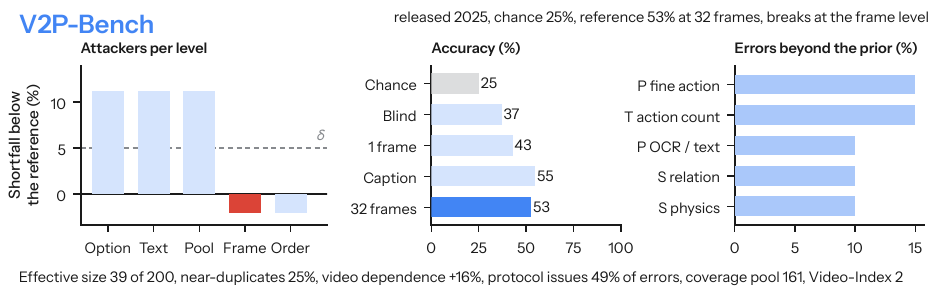}\par\vspace{2pt}
\noindent\includegraphics[width=1.0\linewidth]{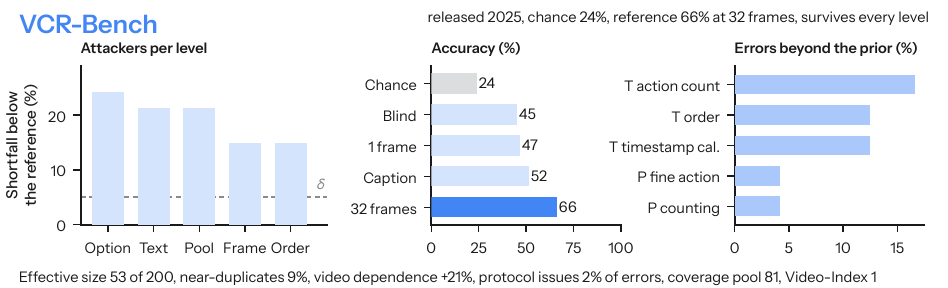}\par\vspace{2pt}
\noindent\includegraphics[width=1.0\linewidth]{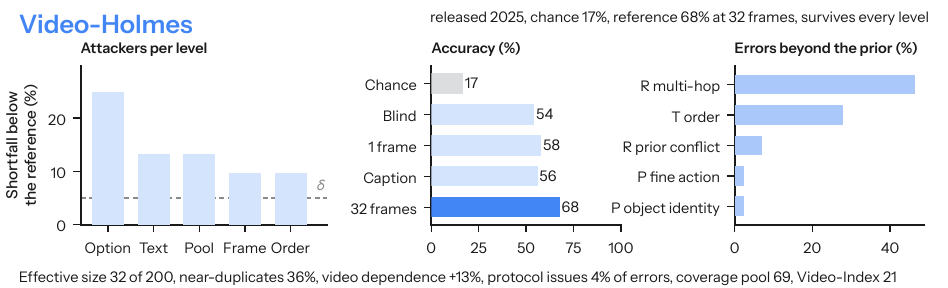}\par\vspace{2pt}
\noindent\includegraphics[width=1.0\linewidth]{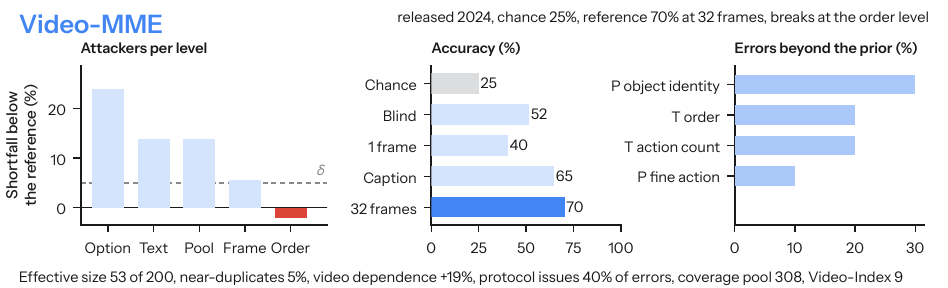}\par\vspace{2pt}
\clearpage
\noindent\includegraphics[width=1.0\linewidth]{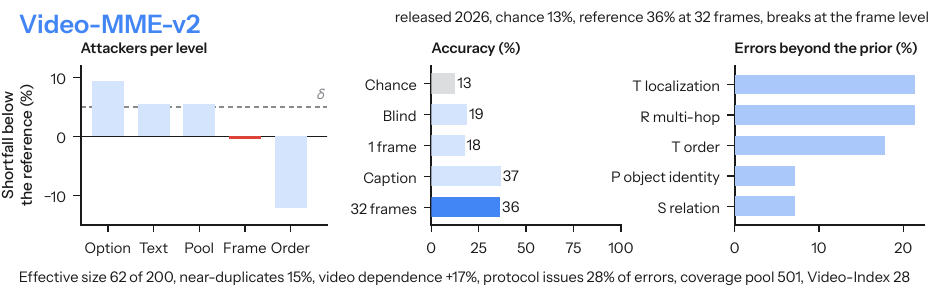}\par\vspace{2pt}
\noindent\includegraphics[width=1.0\linewidth]{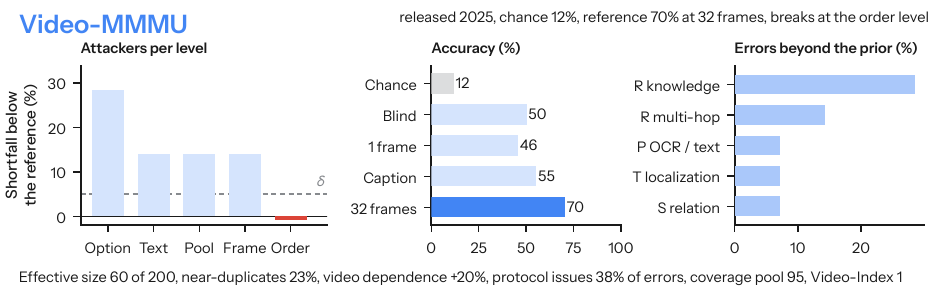}\par\vspace{2pt}
\noindent\includegraphics[width=1.0\linewidth]{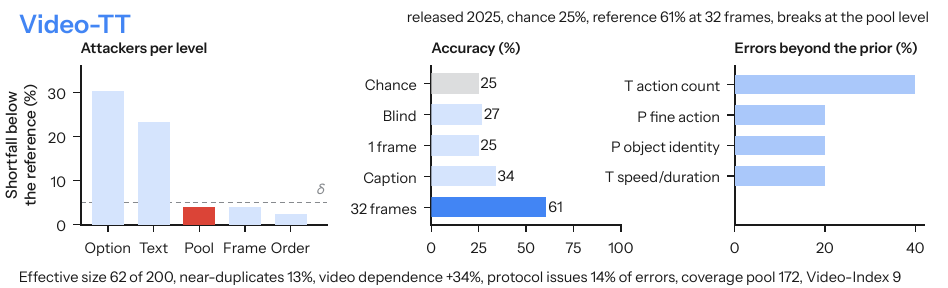}\par\vspace{2pt}
\noindent\includegraphics[width=1.0\linewidth]{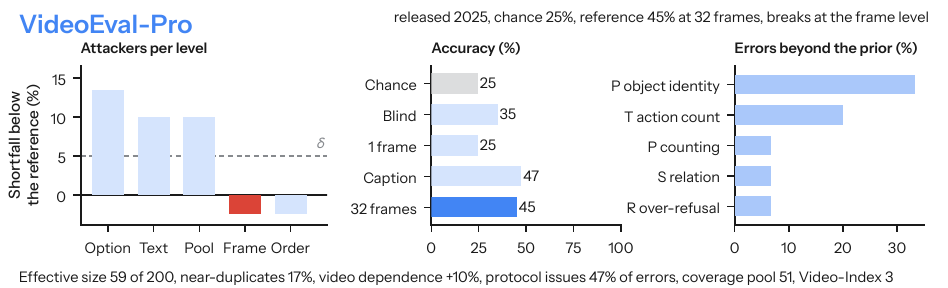}\par\vspace{2pt}
\clearpage
\noindent\includegraphics[width=1.0\linewidth]{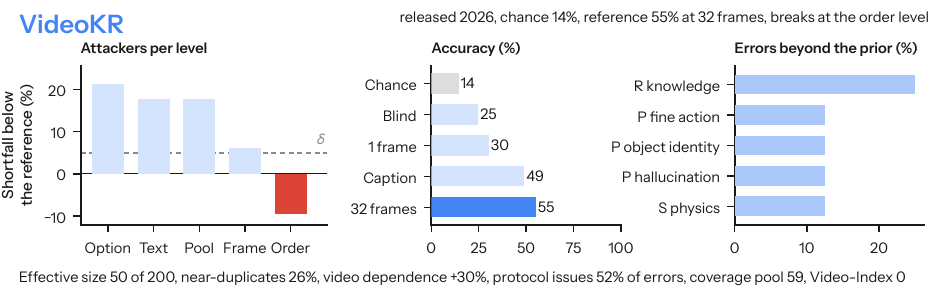}\par\vspace{2pt}
\noindent\includegraphics[width=1.0\linewidth]{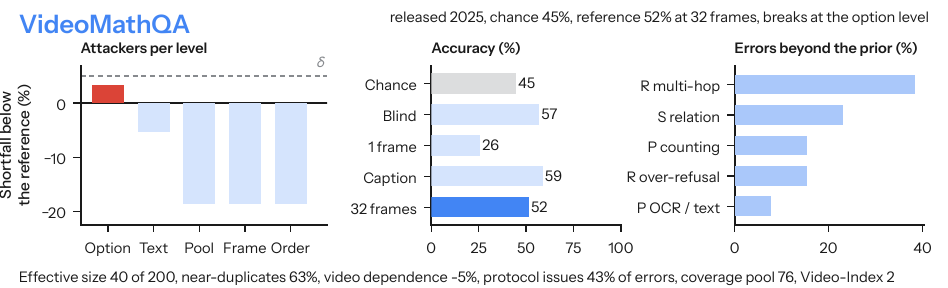}\par\vspace{2pt}
\noindent\includegraphics[width=1.0\linewidth]{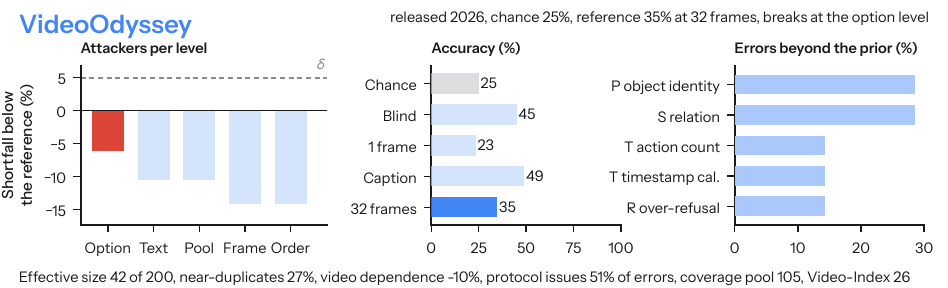}\par\vspace{2pt}
\noindent\includegraphics[width=1.0\linewidth]{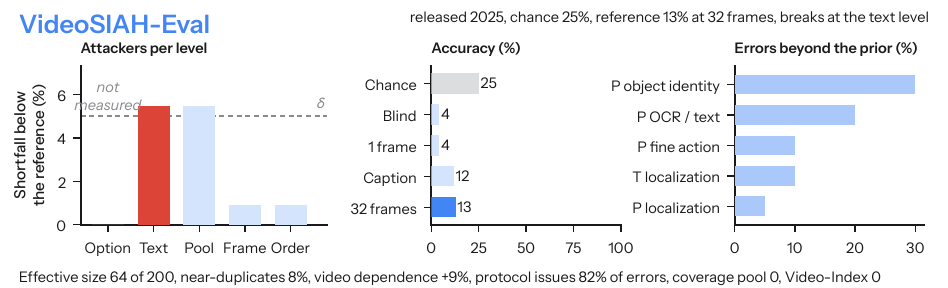}\par\vspace{2pt}
\clearpage
\noindent\includegraphics[width=1.0\linewidth]{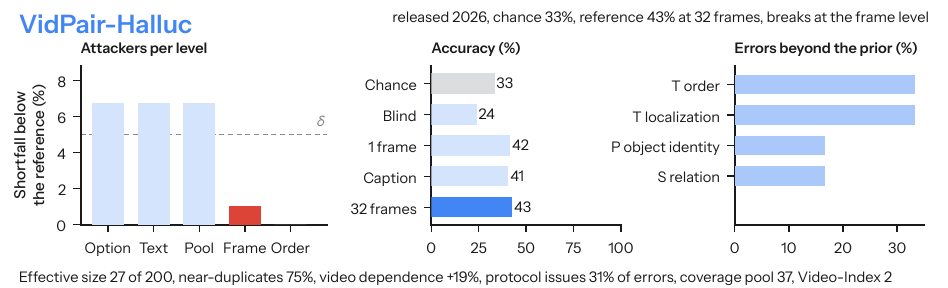}\par\vspace{2pt}
\noindent\includegraphics[width=1.0\linewidth]{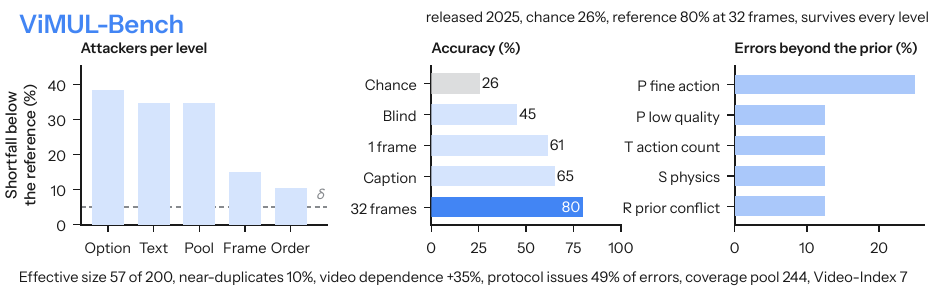}\par\vspace{2pt}
\noindent\includegraphics[width=1.0\linewidth]{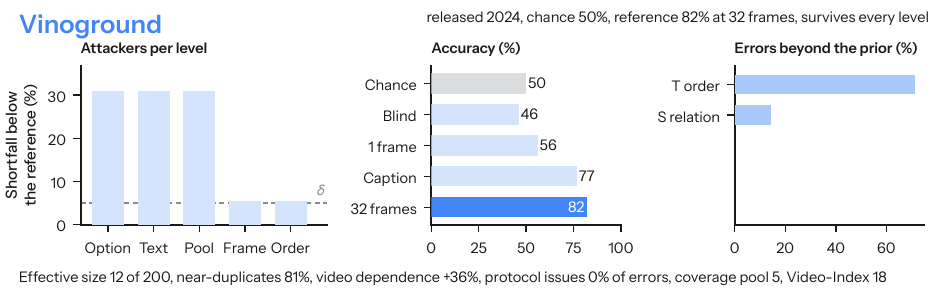}\par\vspace{2pt}
\noindent\includegraphics[width=1.0\linewidth]{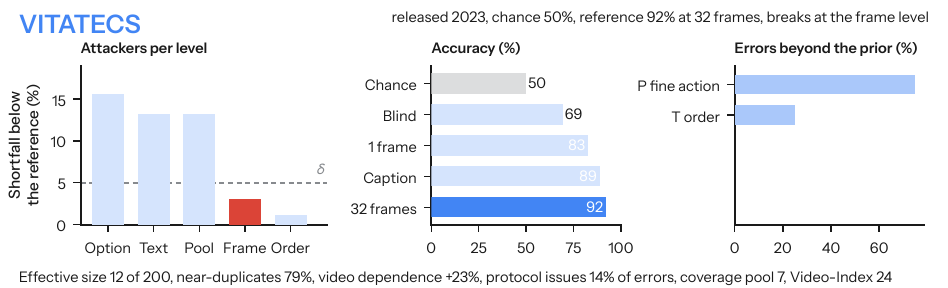}\par\vspace{2pt}
\clearpage
\noindent\includegraphics[width=1.0\linewidth]{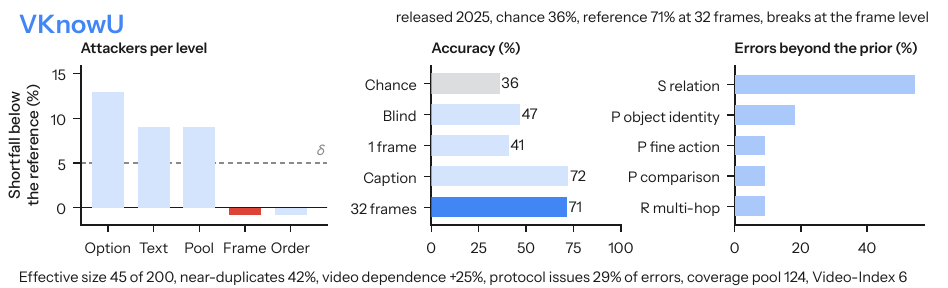}\par\vspace{2pt}
\noindent\includegraphics[width=1.0\linewidth]{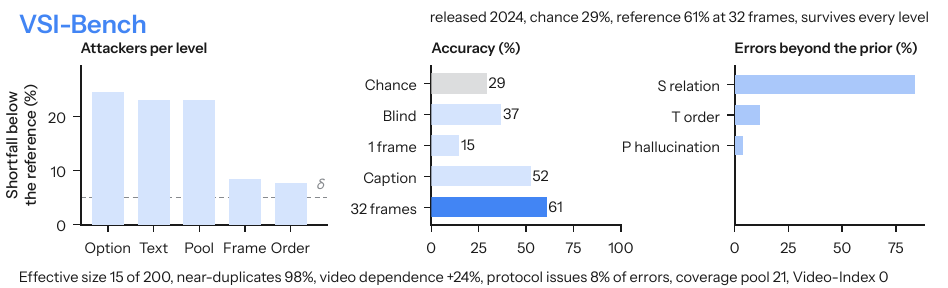}\par\vspace{2pt}
\noindent\includegraphics[width=1.0\linewidth]{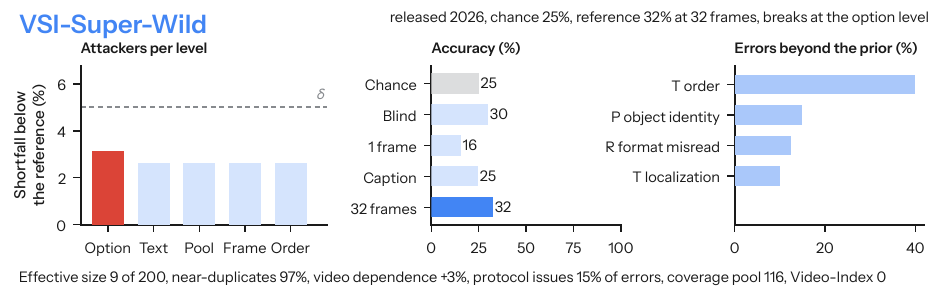}\par\vspace{2pt}
\noindent\includegraphics[width=1.0\linewidth]{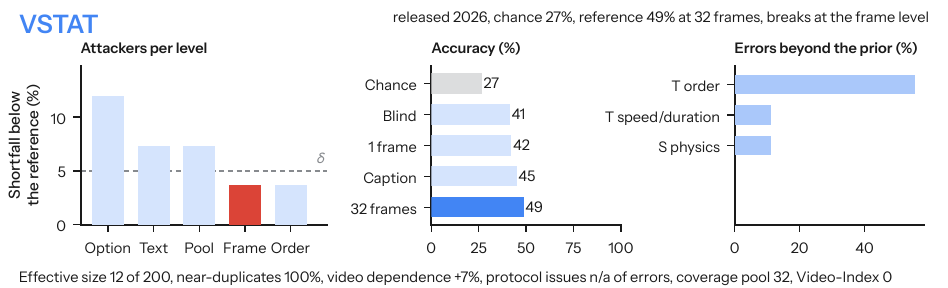}\par\vspace{2pt}
\clearpage
\noindent\includegraphics[width=1.0\linewidth]{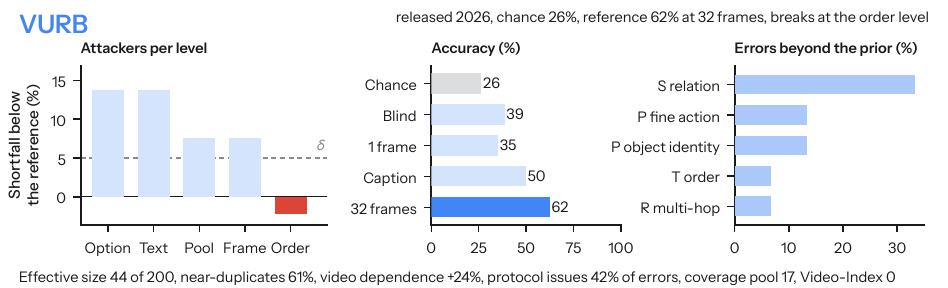}\par\vspace{2pt}
\noindent\includegraphics[width=1.0\linewidth]{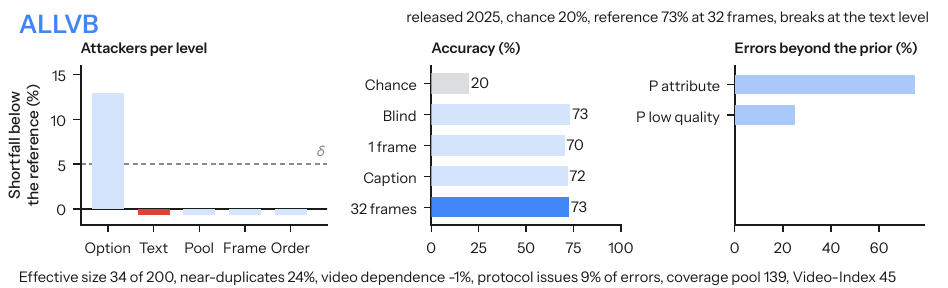}\par\vspace{2pt}
\noindent\includegraphics[width=1.0\linewidth]{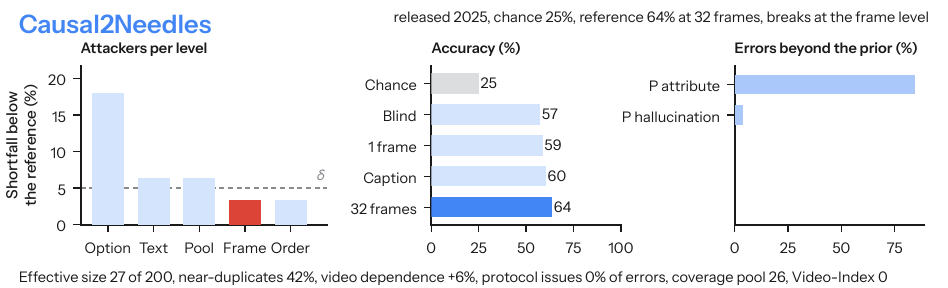}\par\vspace{2pt}
\noindent\includegraphics[width=1.0\linewidth]{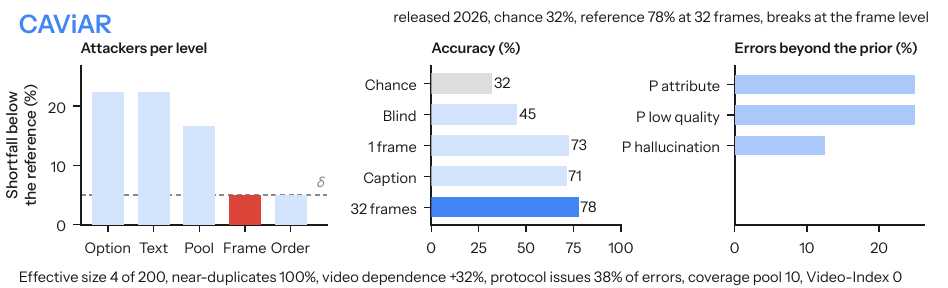}\par\vspace{2pt}
\clearpage
\noindent\includegraphics[width=1.0\linewidth]{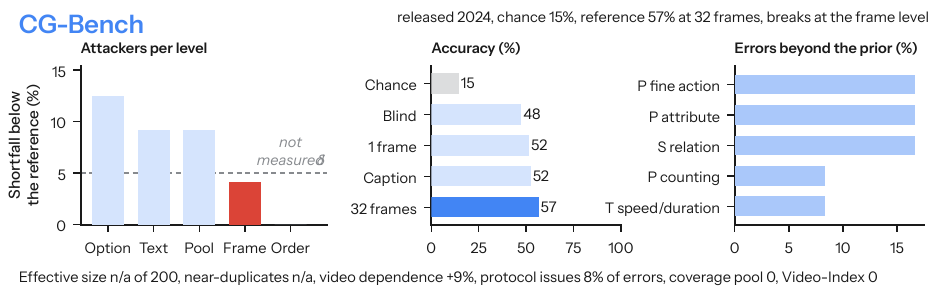}\par\vspace{2pt}
\noindent\includegraphics[width=1.0\linewidth]{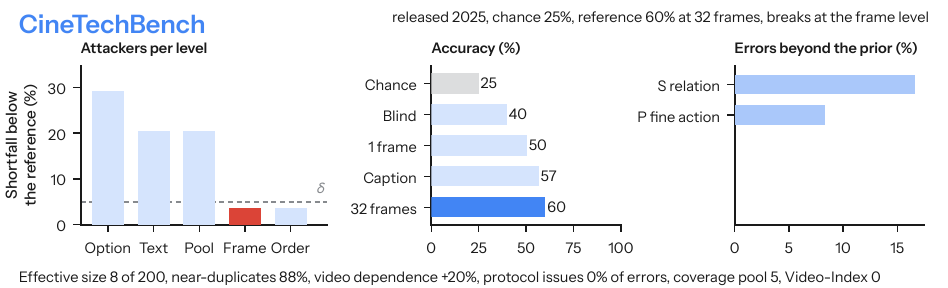}\par\vspace{2pt}
\noindent\includegraphics[width=1.0\linewidth]{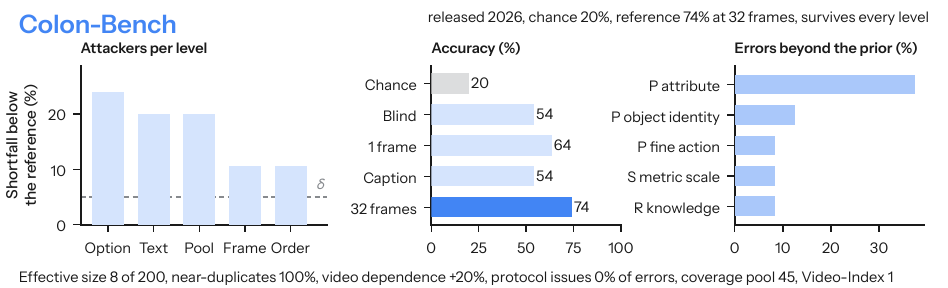}\par\vspace{2pt}
\noindent\includegraphics[width=1.0\linewidth]{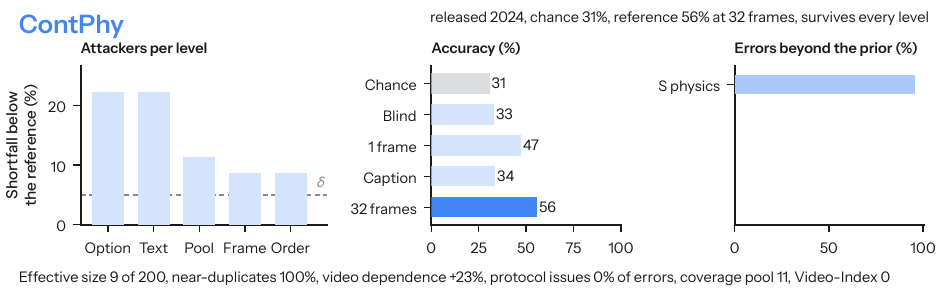}\par\vspace{2pt}
\clearpage
\noindent\includegraphics[width=1.0\linewidth]{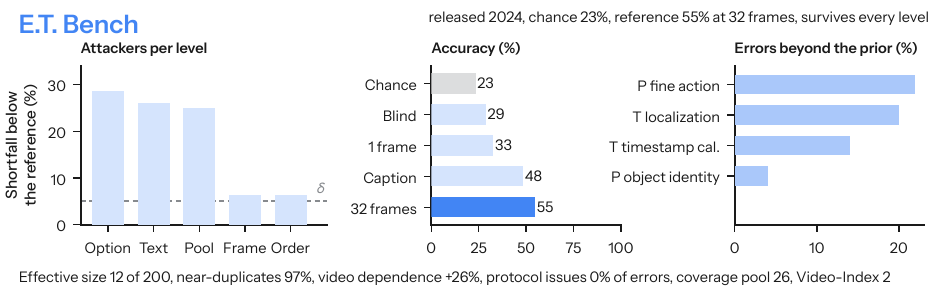}\par\vspace{2pt}
\noindent\includegraphics[width=1.0\linewidth]{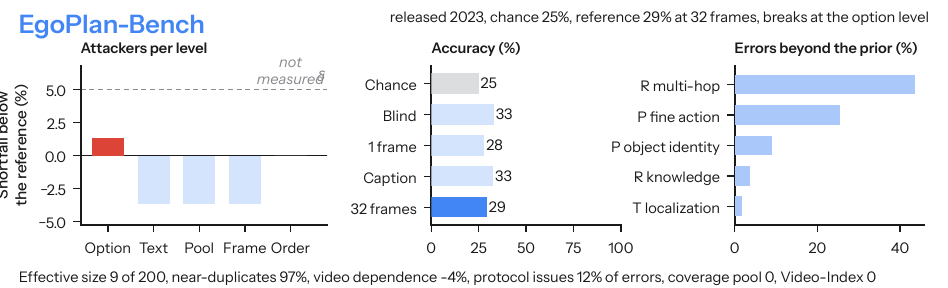}\par\vspace{2pt}
\noindent\includegraphics[width=1.0\linewidth]{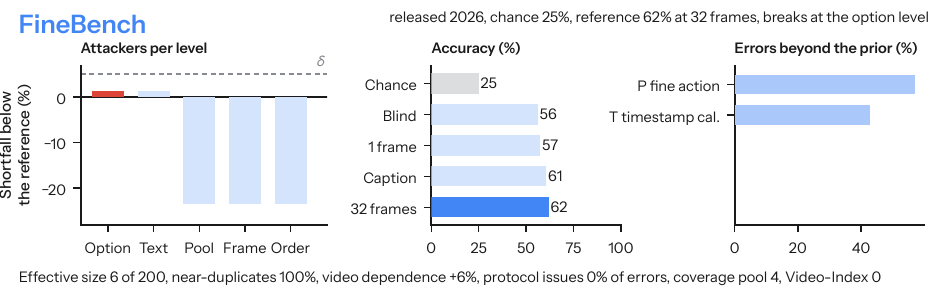}\par\vspace{2pt}
\noindent\includegraphics[width=1.0\linewidth]{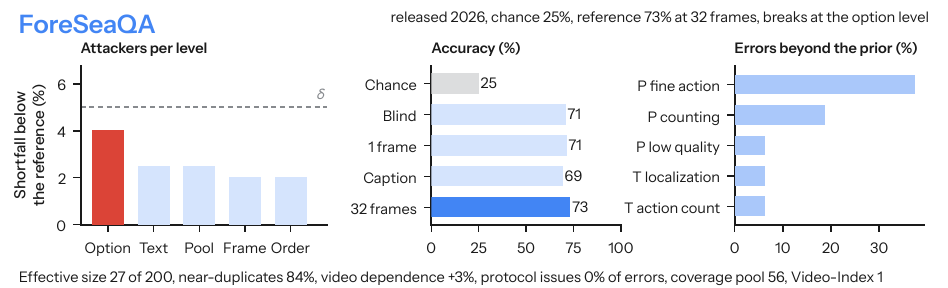}\par\vspace{2pt}
\clearpage
\noindent\includegraphics[width=1.0\linewidth]{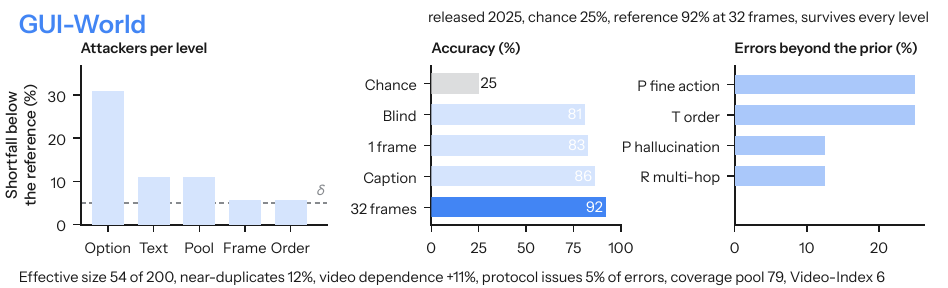}\par\vspace{2pt}
\noindent\includegraphics[width=1.0\linewidth]{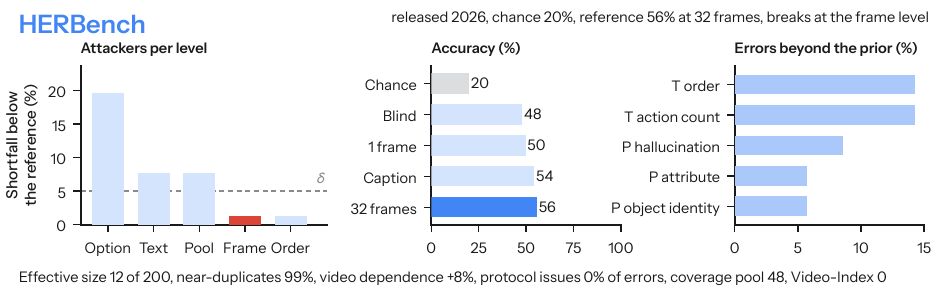}\par\vspace{2pt}
\noindent\includegraphics[width=1.0\linewidth]{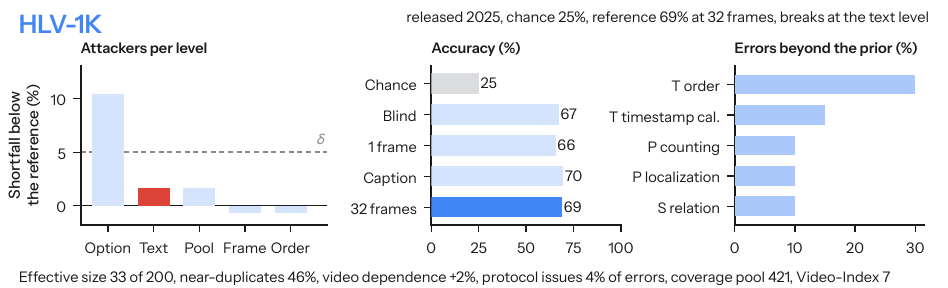}\par\vspace{2pt}
\noindent\includegraphics[width=1.0\linewidth]{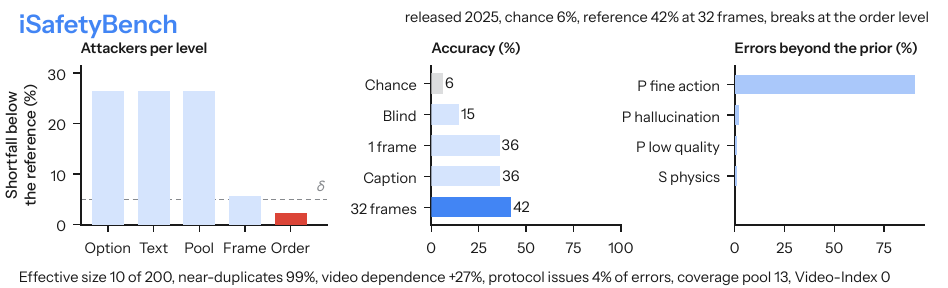}\par\vspace{2pt}
\clearpage
\noindent\includegraphics[width=1.0\linewidth]{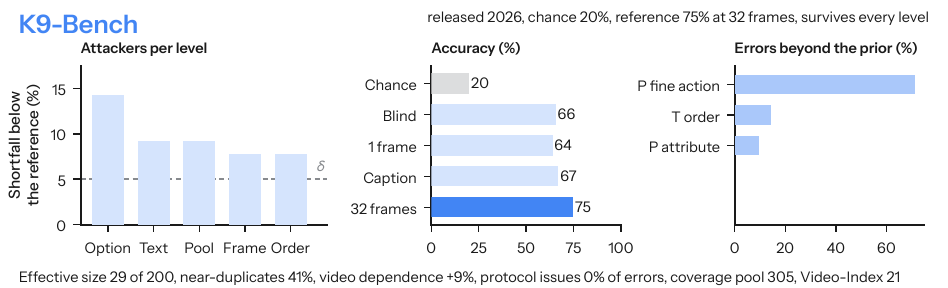}\par\vspace{2pt}
\noindent\includegraphics[width=1.0\linewidth]{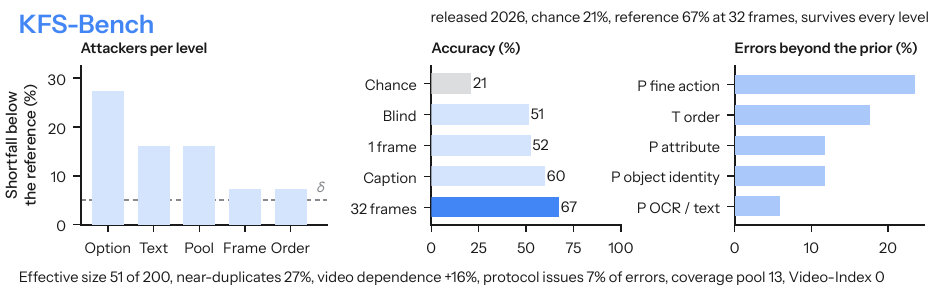}\par\vspace{2pt}
\noindent\includegraphics[width=1.0\linewidth]{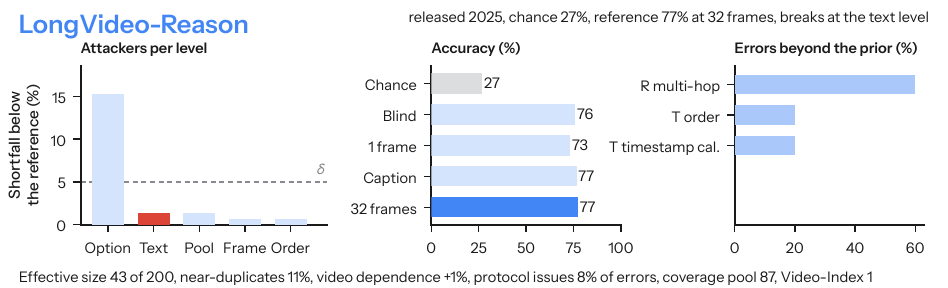}\par\vspace{2pt}
\noindent\includegraphics[width=1.0\linewidth]{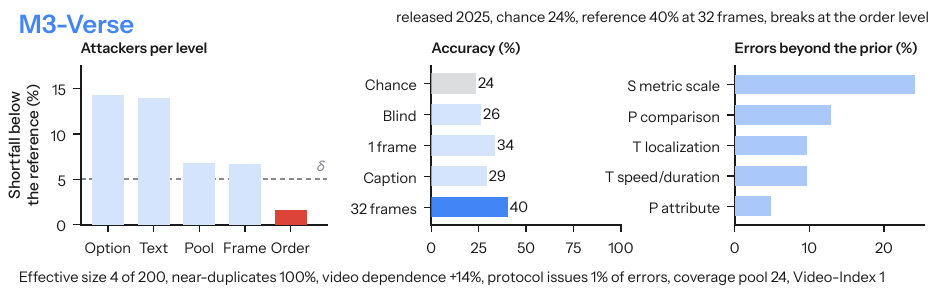}\par\vspace{2pt}
\clearpage
\noindent\includegraphics[width=1.0\linewidth]{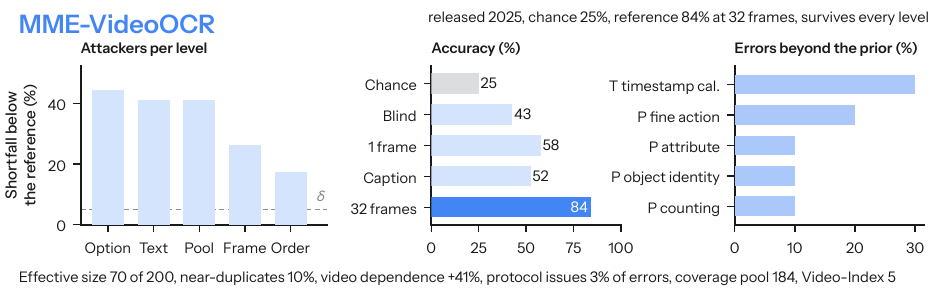}\par\vspace{2pt}
\noindent\includegraphics[width=1.0\linewidth]{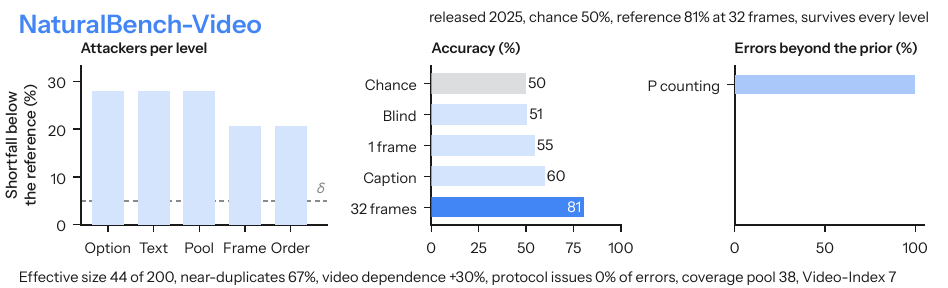}\par\vspace{2pt}
\noindent\includegraphics[width=1.0\linewidth]{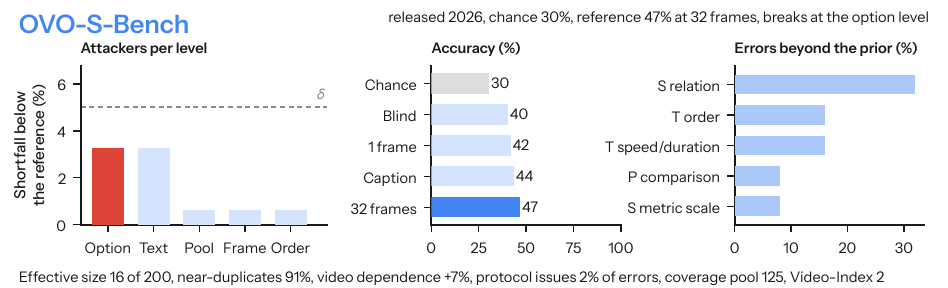}\par\vspace{2pt}
\noindent\includegraphics[width=1.0\linewidth]{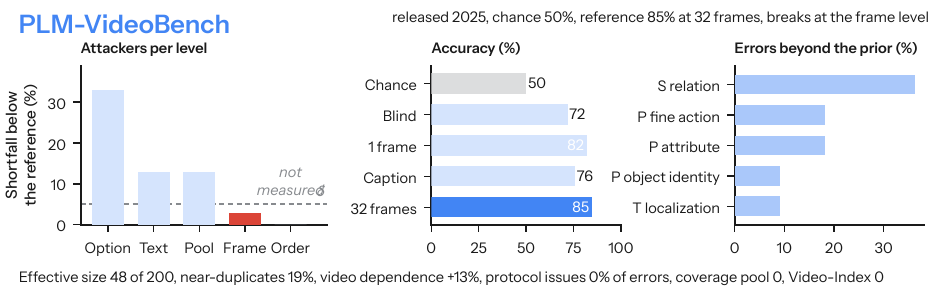}\par\vspace{2pt}
\clearpage
\noindent\includegraphics[width=1.0\linewidth]{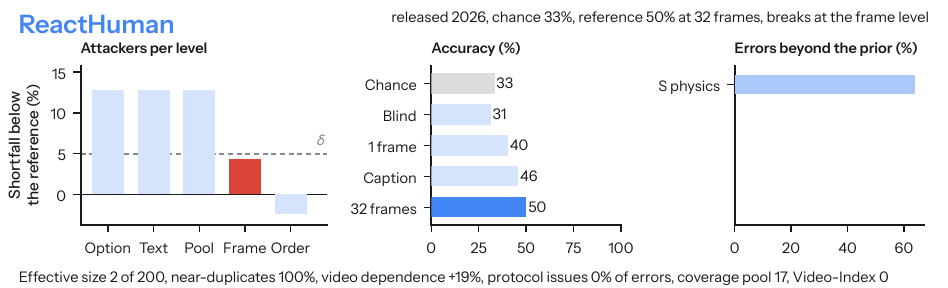}\par\vspace{2pt}
\noindent\includegraphics[width=1.0\linewidth]{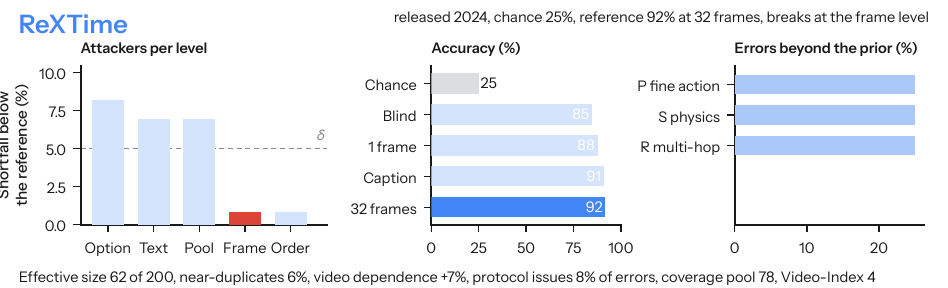}\par\vspace{2pt}
\noindent\includegraphics[width=1.0\linewidth]{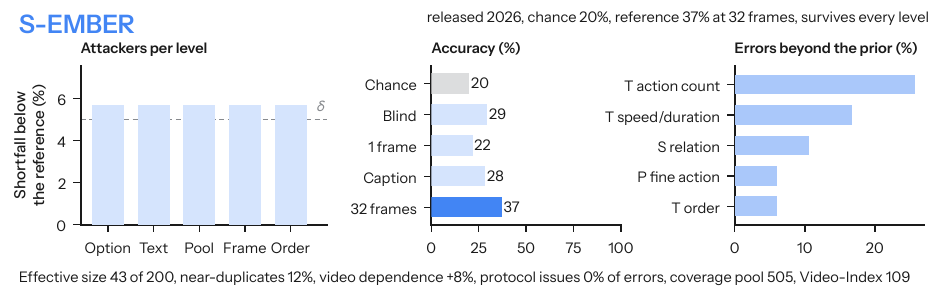}\par\vspace{2pt}
\noindent\includegraphics[width=1.0\linewidth]{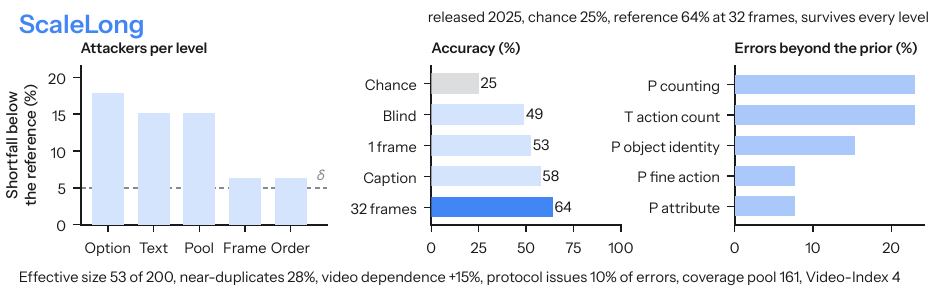}\par\vspace{2pt}
\clearpage
\noindent\includegraphics[width=1.0\linewidth]{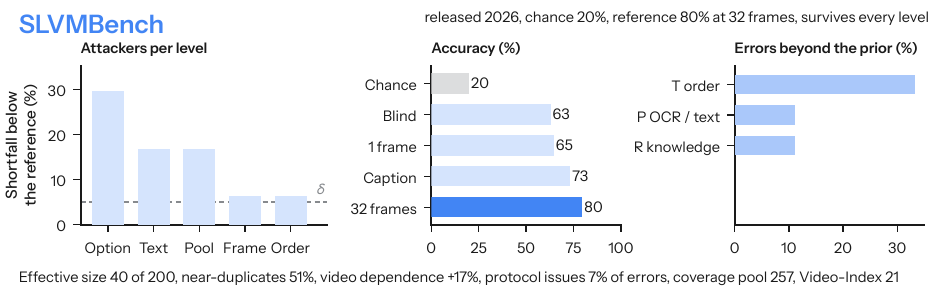}\par\vspace{2pt}
\noindent\includegraphics[width=1.0\linewidth]{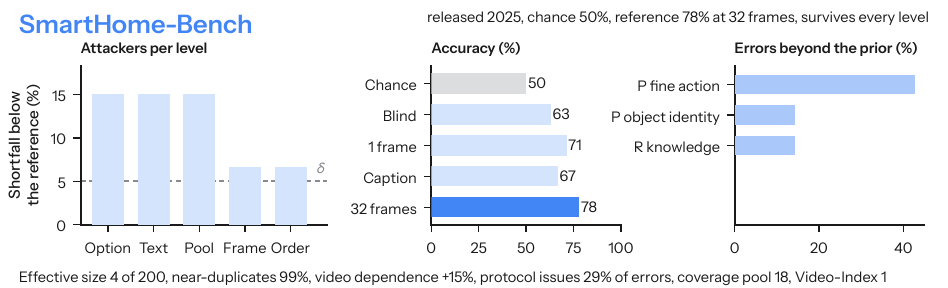}\par\vspace{2pt}
\noindent\includegraphics[width=1.0\linewidth]{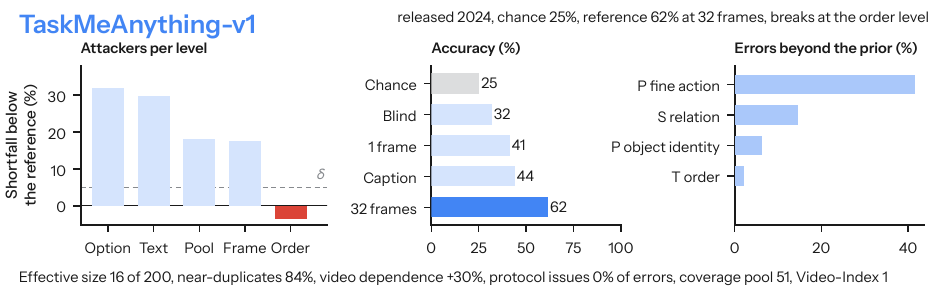}\par\vspace{2pt}
\noindent\includegraphics[width=1.0\linewidth]{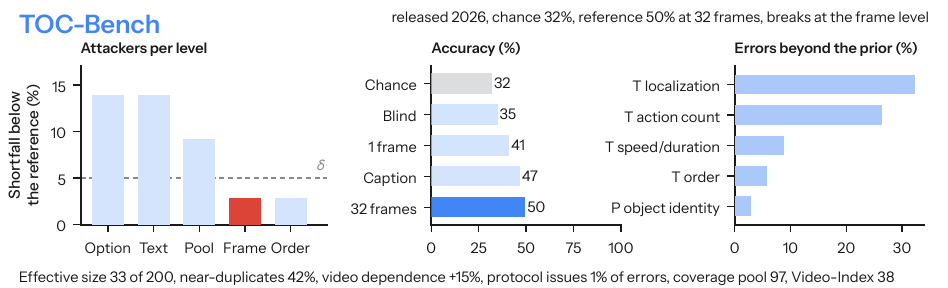}\par\vspace{2pt}
\clearpage
\noindent\includegraphics[width=1.0\linewidth]{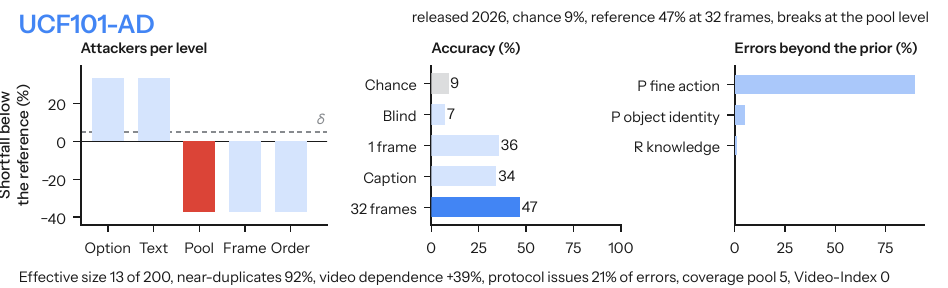}\par\vspace{2pt}
\noindent\includegraphics[width=1.0\linewidth]{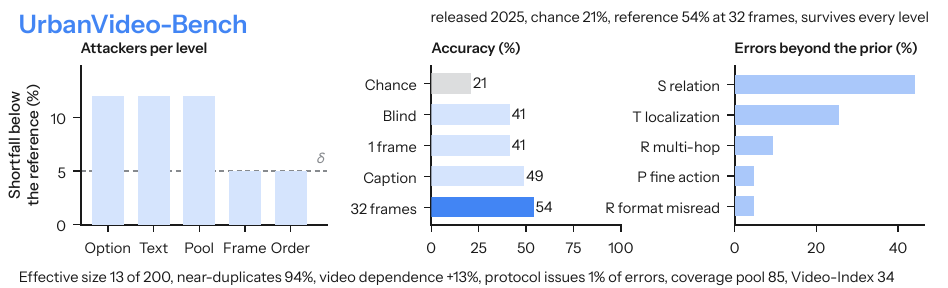}\par\vspace{2pt}
\noindent\includegraphics[width=1.0\linewidth]{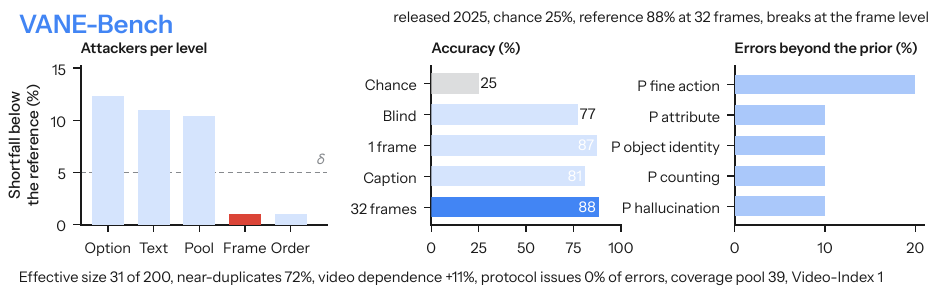}\par\vspace{2pt}
\noindent\includegraphics[width=1.0\linewidth]{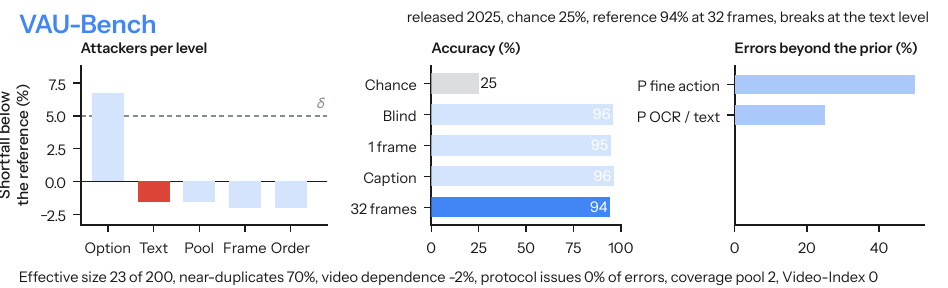}\par\vspace{2pt}
\clearpage
\noindent\includegraphics[width=1.0\linewidth]{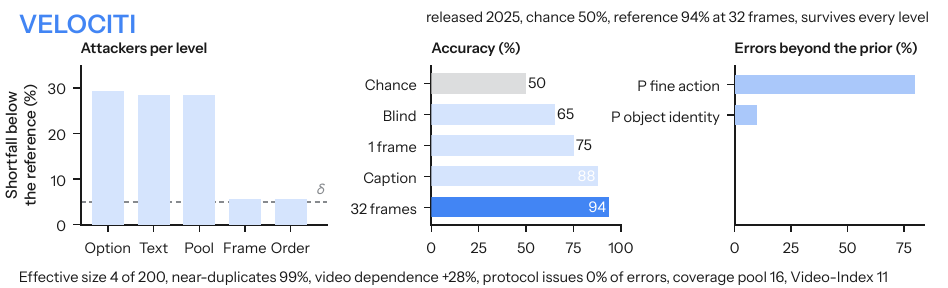}\par\vspace{2pt}
\noindent\includegraphics[width=1.0\linewidth]{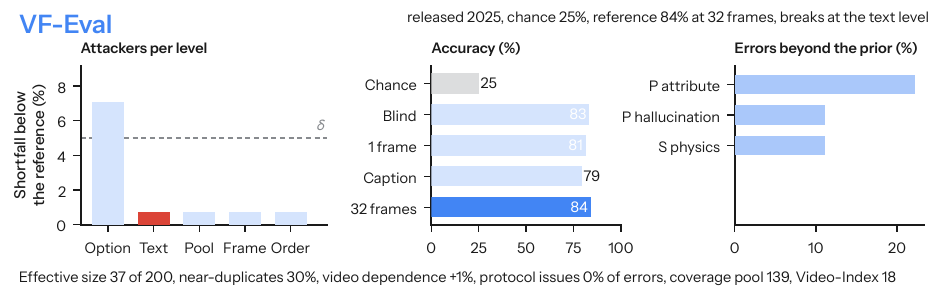}\par\vspace{2pt}
\noindent\includegraphics[width=1.0\linewidth]{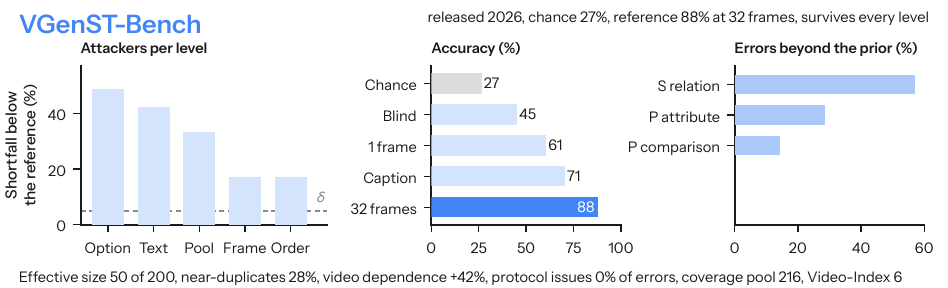}\par\vspace{2pt}
\noindent\includegraphics[width=1.0\linewidth]{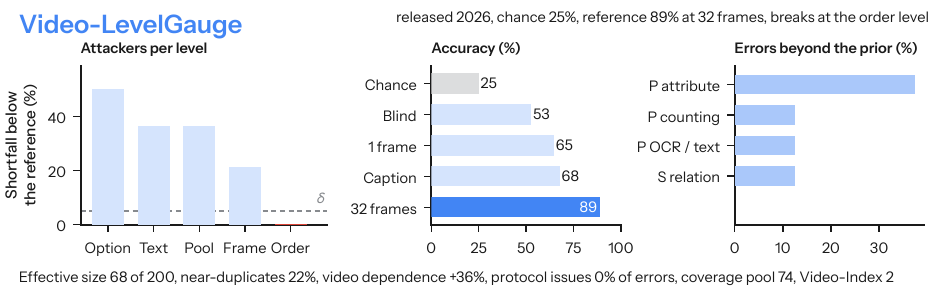}\par\vspace{2pt}
\clearpage
\noindent\includegraphics[width=1.0\linewidth]{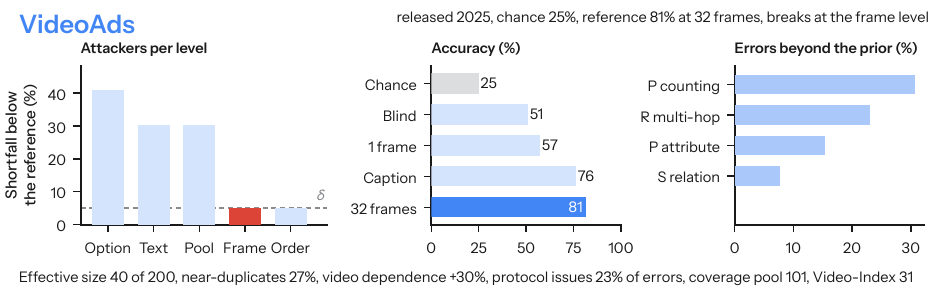}\par\vspace{2pt}
\noindent\includegraphics[width=1.0\linewidth]{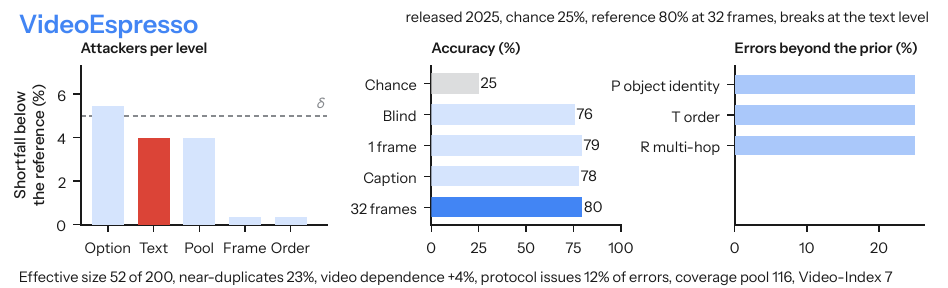}\par\vspace{2pt}
\noindent\includegraphics[width=1.0\linewidth]{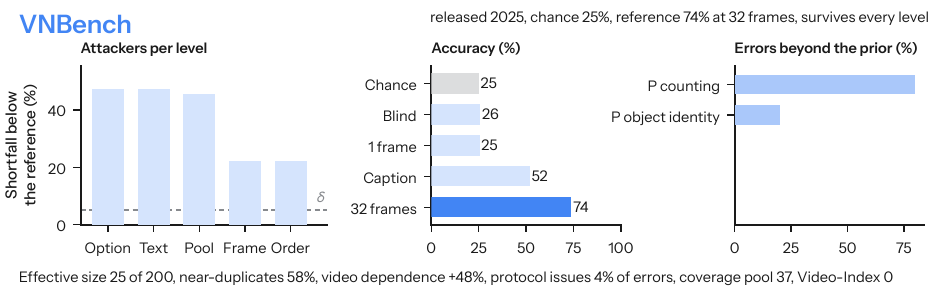}\par\vspace{2pt}
\noindent\includegraphics[width=1.0\linewidth]{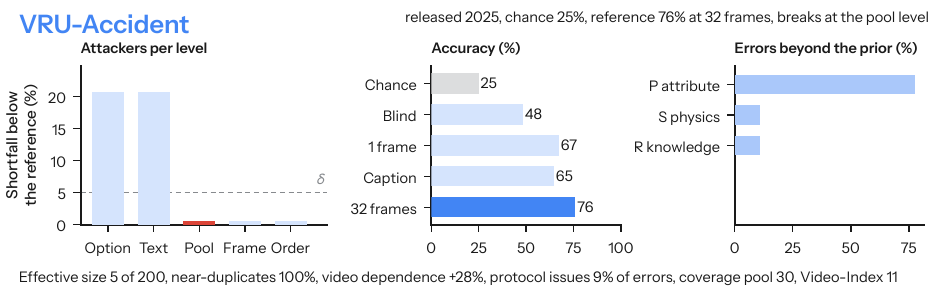}\par\vspace{2pt}
\clearpage

\end{document}